\documentclass[letterpaper]{article} 
\usepackage{aaai24}  
\usepackage{times}  
\usepackage{helvet}  
\usepackage{courier}  
\usepackage[hyphens]{url}  
\usepackage{graphicx} 
\usepackage{natbib}  
\usepackage{caption} 
\setcitestyle{authoryear}

\usepackage[utf8]{inputenc} 
\usepackage{amsfonts}       
\usepackage{nicefrac}       
\usepackage{microtype}      
\usepackage{xcolor}         
\definecolor{bluecite}{HTML}{0875b7}

\usepackage{subcaption}
\usepackage{comment}
\usepackage{amsmath,amssymb}
\usepackage{booktabs}
\usepackage{longtable}
\usepackage{fancyvrb}
\usepackage{algorithm}
\usepackage{algorithmic}
\usepackage[export]{adjustbox}

\usepackage{newfloat}
\usepackage{listings}
\DeclareCaptionStyle{ruled}{labelfont=normalfont,labelsep=colon,strut=off} 
\floatstyle{ruled}
\newfloat{listing}{tb}{lst}{}
\floatname{listing}{Listing}
\title{Efficiently Quantifying Individual Agent\\ Importance in Cooperative MARL}
\author {
    Omayma Mahjoub\textsuperscript{\rm *, \rm 1},
    Ruan de Kock\textsuperscript{\rm *,\rm 1},
    Siddarth Singh\textsuperscript{\rm *,\rm 1},
    Wiem Khlifi\textsuperscript{\rm 1, \rm 2},
    Abidine Vall\textsuperscript{\rm 3},
    Kale-ab Tessera\textsuperscript{\rm 4},
    Rihab Gorsane\textsuperscript{\rm 1},
    Arnu Pretorius\textsuperscript{\rm 1}\\
}

\affiliations {
    \textsuperscript{\rm 1}InstaDeep Ltd\\
    \textsuperscript{\rm 2}National School of Computer Science, Tunisia\\
    \textsuperscript{\rm 3}National School of Engineering of Tunis\\
    \textsuperscript{\rm 4}University of Edinburgh\\
    \textsuperscript{\rm *}Equal Contribution\\
}

\begin{document}
\maketitle

\begin{abstract}

Measuring the contribution of individual agents is challenging in cooperative multi-agent reinforcement learning (MARL). In cooperative MARL, team performance is typically inferred from a single shared global reward. Arguably, among the best current approaches to effectively measure individual agent contributions is to use Shapley values. However, calculating these values is expensive as the computational complexity grows exponentially with respect to the number of agents. In this paper, we adapt difference rewards into an efficient method for quantifying the contribution of individual agents, referred to as Agent Importance, offering a linear computational complexity relative to the number of agents. We show empirically that the computed values are strongly correlated with the true Shapley values, as well as the true underlying individual agent rewards, used as the ground truth in environments where these are available. We demonstrate how Agent Importance can be used to help study MARL systems by diagnosing algorithmic failures discovered in prior MARL benchmarking work. Our analysis illustrates Agent Importance as a valuable explainability component for future MARL benchmarks.
\end{abstract}

\section{Introduction}
In recent years, multi-agent reinforcement learning (MARL) has achieved significant progress, with agents being able to perform similar or better than human players and develop complex coordinated strategies in difficult games such as \textit{Starcraft} \citep{SMAC, Vinyals2019GrandmasterLI}, \textit{Hanabi} \citep{foerester2019bad, Hanabi, hu2021simplified,  wei2021mad} and \textit{Diplomacy} \citep{Bakhtin2022HumanlevelPI}. Furthermore, MARL has also shown promising results in solving real-world problems such as resource allocation, management and sharing, network routing, and traffic signal controls \citep{IntelligentTrafficControl, brittain2019autonomous, DynamicPowerAllocation, AirTrafficManagement, IoTNetworksWithEdgeComputing, TrajectoryDesignandPowerAllocation, pretorius2020game, InternetofControllableThings}. These real-world settings are naturally formulated as cooperative MARL systems, where agents need to coordinate to optimise the same global reward.

One of the critical challenges in cooperative MARL is multi-agent credit assignment \citep{chang2003all}. Since agents typically receive a global reward for their joint actions, this makes determining individual agent contributions challenging. This need for correct attribution becomes especially important as more autonomous systems are deployed in the real world. The inherent complexity of these MARL systems impedes our understanding of decision-making processes and the motivations behind actions, hindering progress in this field. Improved credit assignment could play a vital role in comprehending agent behaviour and system-level decision-making, aiding in accountability, trust, fairness, and facilitating the detection of potential issues such as coordination failures, or unethical behaviour. 

Credit assignment can be considered from a core algorithmic perspective, where components of reinforcement learning (RL) algorithms, such as the value function, are adapted to better decouple the impact of the actions of individual agents. Methods such as VDN \citep{VDN}, COMA \citep{foerster2018counterfactual}, and QMIX \citep{QMIX} fall into this domain. However, since these algorithms are trained end-to-end through the use of function approximators, explainability is difficult, i.e. it is challenging to correlate specific agent actions to reward outcomes over time. Furthermore, since these notions of agent impact are part of the RL algorithms themselves, it is not easy to transfer these between different algorithms.

Accurate credit assignment within a team of agents can also be seen as a form of explainable AI (XAI). XAI consists of machine learning (ML) techniques that are used to provide insights into the workings of models \citep{arrieta2020explainable}. It has been used across various domains in ML, and more recently in single-agent RL\footnote{In this paper, we use the term "RL" to exclusively refer to \emph{single-agent} RL, as opposed to RL as a field of study, of which MARL is a subfield.} \citep{glanois2021survey} and multi-agent systems \citep{heuillet2022collective}. Following from \citep{arrieta2020explainable,glanois2021survey}, we use the notion of explainability to refer to any external post-hoc methodology that is used to gain insights into a trained model. These techniques have the notable advantage of being able to be used across algorithms, often irrespective of their design or formulation.   

Efforts to enhance explainability in RL have resulted in the development of various techniques \citep{juozapaitis2019explainable, madumal2020explainable, puiutta2020explainable, glanois2021survey, heuillet2021explainability, vouros2022explainable,dazeley2023explainable}. In contrast, MARL lacks dedicated explainability tools, with only a limited number of works addressing this topic \citep{Kraus2019AIFE, Boggess2022TowardPE, heuillet2022collective}. One notable approach involves leveraging the Shapley value \citep{shapley1953}, a metric derived from game theory, and adapting it to MARL to quantify agent contributions to the global reward \citep{heuillet2022collective}. Although Shapley values have shown promise in MARL explainability, calculating these values is expensive as the computational complexity grows exponentially with respect to the number of agents.


In this paper, we highlight the need for employing explainable tools to help quantify credit assignment in cooperative MARL systems. We show that an averaged calculation of the difference reward \citep{wolpert2001optimal} across evaluation episodes, can be used as an effective metric for measuring an agent's contribution, which we refer to as the \textit{Agent Importance}. Unlike Shapley values, the Agent Importance has a linear computational complexity (w.r.t. the number of agents) making it more efficient to compute. Through empirical analysis, we demonstrate a strong correlation between the Agent Importance values and the true Shapley values, while also empirically validating the scalability and computational advantage of this approach.  

To showcase the practical use of Agent Importance, we revisit a previous benchmark in cooperative MARL \citep{papoudakis2021benchmarking} and follow the standardised evaluation guideline proposed by \citep{gorsane2022standardised} to reproduce key results from this benchmark under a sound protocol. We then proceed by applying Agent Importance to specific scenarios of interest as highlighted by the authors of this benchmark. This includes investigating: (1) why Multi-Agent Advantage Actor-Critic (MAA2C) \citep{mnih2016asynchronous,papoudakis2021benchmarking} outperforms Multi-Agent Proximal Policy Optimisation (MAPPO) \citep{yu2022surprising} in the Level-Based Foraging (LBF) environment \footnote{A somewhat surprising result since MAPPO uses importance sampling for off-policy correction and is expected to perform at least as well as MAA2C as it incorporates a clipping function based on importance sampling allowing data retraining without divergent policies.} \citep{albrecht2015gametheoretic, albrecht2019reasoning, christianos2020shared}; and (2) why parameter sharing between agents leads to improved performance (3) analyse agents' behaviour in case of heterogeneous settings. Using agent importance, we uncover that for (1) MAA2C achieves a more equal contribution among agents when compared to MAPPO, i.e.\ agents have a more similar importance to the overall team and therefore have a higher degree of cooperation; and that for (2) architectures without parameter sharing exhibit a higher variance in agent importance, leading to credit assignment issues and lower performance compared to architectures with parameter sharing. The source code to reproduce our analysis and compute the agent importance, as well as our raw experiment data is publicly available \footnote{Data and code are accessible at the following links:\\ Data- \url{https://sites.google.com/view/agent-importance/home} \\ Code- \url{https://tinyurl.com/ycx47jz6}}.

\section{Related Work}\label{relatedwork}
\textbf{Explainability in RL} With the surging popularity of Deep RL, which relies on black-box deep neural networks, there has been an increase in literature that attempts to enable human understanding of complex, intelligent RL systems \citep{juozapaitis2019explainable, madumal2020explainable, puiutta2020explainable, glanois2021survey, heuillet2021explainability, vouros2022explainable,dazeley2023explainable}. Additionally, frameworks like ShinRL \citep{ShinRL} and environment suites like bsuite \citep{osband2019behaviour} offer comprehensive debugging tools including state and action space visualizations and reward distributions, and carefully crafted environments for behavioural analysis in RL. 



\textbf{Explainability in MARL} In contrast to explainable RL, there has been a limited amount of work focusing on explainability in MARL \citep{Kraus2019AIFE, Boggess2022TowardPE, heuillet2022collective}. Specifically, we are interested in explainability in the context of cooperative MARL with a shared, global reward and the aim is to effectively quantify credit assignment.


The challenges associated with measuring credit assignment in MARL have motivated researchers to explore the use of the \textbf{Shapley value} \citep{shapley1953}. Originating from game theory, the Shapley value addresses the issue of payoff distribution within a ``grand coalition'' (i.e.\ a cooperative game) and quantifies the contribution of each coalition member toward completing a task. Specifically, consider a cooperative game $\Gamma = (\mathcal{N},v)$, where $\mathcal{N}$ is a set of all players and $v$ is the payoff function used to measure the ``profits'' earned by a given coalition (or subset) $\mathcal{C} \subseteq \mathcal{N}\setminus\{i\}$, such that the marginal contribution of player $i$ is given by $\phi_{i}(\mathcal{C}) = v(\mathcal{C} \cup \{i\}) - v(\mathcal{C})$. The Shapley value of each player $i$ can then be computed as:
\begin{equation}
S_{i}(\Gamma) = \sum_{\mathcal{C} \subseteq \mathcal{N}\setminus\{i\}} \frac{|\mathcal{C}|!(|\mathcal{N}| - |\mathcal{C}|-1)!}{|\mathcal{N}|!} \cdot \phi_{i}(\mathcal{C}).
\end{equation}

Calculating Shapley values in the context of MARL presents two specific challenges: (1) it requires computing $2^{n-1}$ possible coalitions of a potential $n(2^{n-1})$ coalitions (with $|\mathcal{N}|=n$) which is computationally prohibitive and (2) it strictly requires the use of a simulator where agents can be removed from the coalition and the payoff of the same states can be evaluated for each coalition.

Despite its limitations, the Shapley value is able to alleviate the issue of credit assignment and help towards understanding individual agent contributions in MARL. As a result, numerous efforts have been undertaken to incorporate it as a component of an algorithm \citep{wang2020shapley, yang2020q, han2022stable, wang2022}. However, in this work, we focus on the Shapley value as an explainability metric. One such approach is introduced in \citep{heuillet2022collective}, where the authors utilise a Monte Carlo approximation of the Shapley value to estimate the contribution of each agent in a system, which we refer to here as \textbf{MC-Shapley}. This approximate Shapley value is computed as:
\begin{equation}
\hat{S}^{MC}_{i}(\Gamma) = \frac{1}{M} \sum^{M}_{m=1}(r_{\mathcal{C}_m \cup \{i\}} - r_{\mathcal{C}_m}) \approx S_{i}(\Gamma),
\end{equation}
where M is the number of samples (episodes), $C_m$ is a randomly sampled coalition out of all possible coalitions excluding agent $i$, and $r_{\mathcal{C}_m \cup \{i\}}$ and $r_{\mathcal{C}_m}$ are the episode returns obtained with and without agent $i$ included in the coalition.

In essence, \citet{heuillet2022collective} attempts to address the second limitation of the Shapley value, which involves removing agents from the environment. They propose three strategies for proxies of agent removal while computing the return  $r_{\mathcal{C}_m}$. The first hypothesis is to provide the agent $i$ with a no-op (no-operation) action, the second is to assign the agent $i$ with a random action, and the third is to replace the action of agent $i$ with a randomly selected agent's action from the current coalition $C_m$. The paper's findings indicate that using the no-op approach yields the most accurate approximation of the true Shapley value. A primary limitation of this work is the dependence on a significant number of sampled coalitions, with each sample corresponding to a single episode. This characteristic has a notable impact on training speed, especially if the proposed approach is employed as an online metric for detecting the evolution of agents' contributions during system training.

\textbf{Difference Rewards.} Of central relevance to this work is difference rewards \citep{wolpert2001optimal, agogino2004unifying, agogino2008analyzing, devlin2014potential} which presents a method for estimating credit assignment within a system. It can be written as $D_i{(z)} = G(z) - G(z_{-i})$ where $D_i{(z)}$ is the difference reward for agent $i$, $z$ is a state or state-action pair depending on the application, $G(z)$ is the performance of the global system and $G(z_{-i})$ is the performance of a theoretical system that omits agent $i$. Any action taken that increases the difference reward $D_i{(z)}$ also increases $G(z)$ but will have a higher impact on the (typically unknown or hypothetical) individual reward for each agent compared to the global reward. It is from this property that we may determine the relative impact of each agent in a system.

\section{Agent Importance}
We compute the Agent Importance as an average of difference rewards and use it as an efficient estimate of the Shapley value. To ensure accuracy in our estimation, we emphasize the importance of utilizing an adequate number of samples. This is reminiscent of the MC-Shapley approach which uses Monte Carlo approximation over entire episodes \citep{heuillet2022collective}. However, in this work, we show that such an approach to estimation is not necessary and instead, we compute difference returns over samples collected \textit{per step}, rather than per episode, without the need to resample coalitions. We simply compute the difference reward for each agent at each timestep during evaluation and aggregate over all evaluation timesteps. This approach greatly improves the sample efficiency in estimation during online evaluation. Concretely, the \textbf{Agent Importance} is given by

\begin{equation}
\label{eq1}
\hat{S}^{AI}_{i}(\Gamma) = \frac{1}{T} \sum^{T}_{t=1} r^t- r^t_{-i},
\end{equation}

where $T$ is the number of timesteps in a full evaluation interval, $r^t$ is the team reward (i.e.\ the reward of the grand coalition), at timestep $t$ and $r^{t}_{-i}$ is the team reward when agent $i$ performs a no-op action.




Applying Equation \ref{eq1} poses a technical challenge as it requires comparing rewards between agents based on the same exact environment state at a given timestep. In MARL, most simulators are not easily resettable and/or stateless, which makes measuring one reward and undoing that step and then measuring a second reward difficult \footnote{We however do note, that this could easily be achieved with simulators written using pure functions in JAX \citep{brax2021github, gymnax2022github, jumanji2023github}.}. To overcome this limitation, we adopt a simple solution outlined in Algorithm \ref{alg:cap}, where we create a copy of the environment for each agent to be able to compute the Agent Importance.

\begin{algorithm}
\caption{Per timestep difference reward contribution in Agent Importance}
\label{alg:cap}
\begin{algorithmic}[1]
\REQUIRE $t$: evaluation timestep, $marginal\_contribution$: dictionary
\STATE $env\_copies \gets \text{deepcopy}(env, len(agents))$
\STATE $r^{t} \gets \text{env.step}(selected\_actions)$
\FOR{$i = 0 \text{ \textbf{to} } len(agents)$}
    \STATE $actions\_with\_no\_op \mathrel{\gets} \text{disable\_actions}(selected\_actions, i)$
    \STATE $r^{t}_{-i} \gets \text{env\_copies}[i].\text{step}(actions\_with\_no\_op)$
    \STATE $\text{add\_to\_dict}(marginal\_contribution, i, (r^{t}-r^{t}_{-i}))$
\ENDFOR
\end{algorithmic}
\end{algorithm}

\section{Case Study: using Agent Importance to analyse a prior benchmark}
\label{section: experiment_setup}

Our case study setup is based on the work of \citep{papoudakis2021benchmarking}, which made a comparative benchmark of cooperative MARL algorithms. The study conducts evaluations and comparisons of multiple categories of MARL algorithms, covering Q-learning, and policy gradient (PG) methods, across two paradigms: independent learners (ILs), and centralised training with decentralised execution (CTDE). The findings of this study align with those of \citep{gorsane2022standardised}, concluding that current MARL algorithms are most performant on the popular Multi-Particle Environment (MPE) \citep{lowe2017multi} and Starcraft Multi-Agent Challenge (SMAC) \citep{SMAC} environments--with most algorithms achieving comparable performance, in some cases seemingly to the point of overfitting. Consequently, our main analysis focuses on the remaining two environments from this benchmark: LBF, and RWARE. 

\begin{figure}
\centering
\includegraphics[width=0.33\linewidth, height=0.19\linewidth]{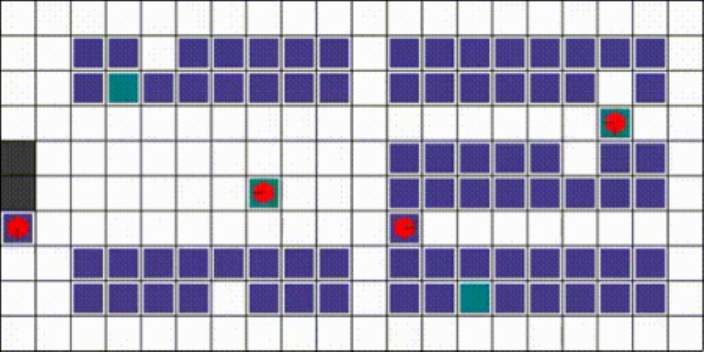}
\includegraphics[width=0.33\linewidth]{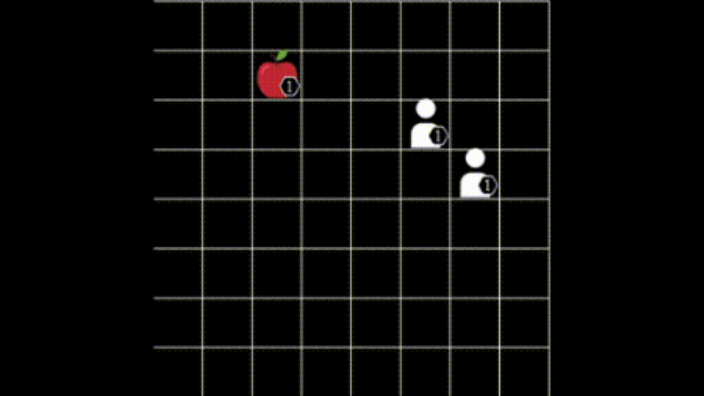}
\includegraphics[width=0.32\linewidth]{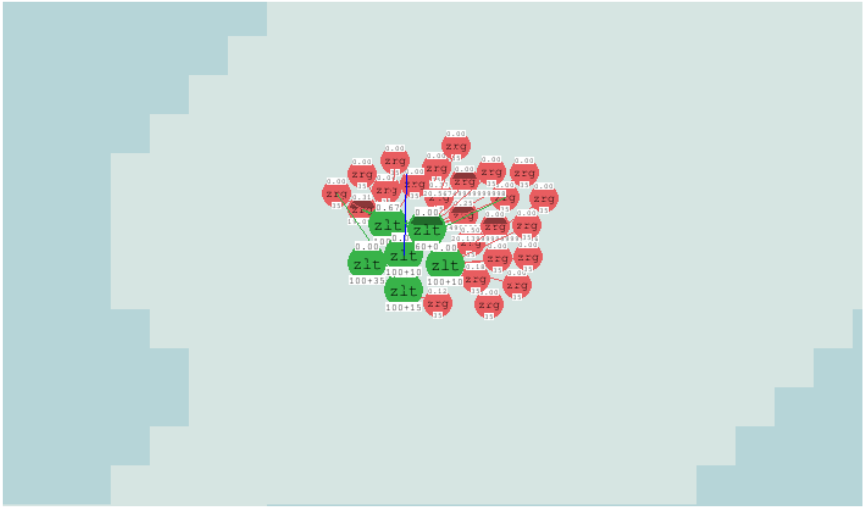}
\caption{\textbf{Left}: Multi-Robot Warehouse (RWARE). \textbf{Middle}: Level-Based Foraging (LBF). \textbf{Right}: SMAClite}
\label{fig:rware_lbf}
\end{figure}

\textbf{Environments. } The Multi-Robot Warehouse (RWARE) \citep{christianos2020shared, papoudakis2021benchmarking} is a multi-agent environment that is designed to represent a simplified setting where robots move goods around a warehouse. The environment requires agents (circles) to move requested shelves (colored squares) to the goal post (dark squares) and back to an empty square as illustrated at the top of Figure \ref{fig:rware_lbf}. Tasks are partially observable with a very sparse reward signal as agents have a limited field of view and are rewarded only upon a successful delivery.

Level-Based Foraging (LBF) \citep{albrecht2015gametheoretic, albrecht2019reasoning, christianos2020shared} is a mixed cooperative-competitive game with a focus on inter-agent coordination illustrated at the bottom of Figure \ref{fig:rware_lbf}. Agents are assigned different levels and navigate a grid world where the goal is to consume food by cooperating with other agents if required. Agents can only consume food if the combined level of the agents adjacent to a given item of food exceeds the level of the food item. Agents are awarded points equal to the level of the collected food divided by their level. LBF has a particularly high level of stochasticity since the spawning position and level assigned to each agent and food are all randomly reset at the start of each episode.

In the original benchmarking work by \cite{papoudakis2021benchmarking}, the authors used the popular Starcraft Multi-Agent Challenge (SMAC) \citep{SMAC} environment. In our case study, we instead use SMAClite \citep{michalski2023smaclite}, an environment designed to replicate SMAC faithfully, in Python. An illustration of SMAClite is given in Figure \ref{fig:rware_lbf}. SMAClite has similar system dynamics to SMAC but does not rely on the StarCraft 2 video game engine as a backend. Due to this SMAClite requires significantly less RAM making it more suitable for utilising parallel processing. This also means it can be used in conjunction with Python methods like \texttt{copy} which makes contribution analysis methods like simpler to implement.


\textbf{Algorithms. }As in the original benchmarking setup of \cite{papoudakis2021benchmarking}, we use the exact same collection of algorithms for our case study. Specifically, we use the value-based algorithms Independent Q-Learning (IQL) \citep{IQL}, Value-Decomposition Network (VDN) \citep{VDN}, and QMIX \citep{QMIX}, alongside two policy-gradient (PG) algorithms, namely Multi-Agent Proximal Policy Optimisation (MAPPO) \citep{yu2022surprising} and Multi-Agent Advantage Actor-Critic (MAA2C) \citep{foerster2018counterfactual}. To investigate the influence of parameter sharing, we conduct experiments with both parameter-sharing and non-parameter-sharing architectures. Further details about the algorithms can be found in the Appendix section \ref{algo_details}.

\textbf{Evaluation Protocol. } We follow the protocol outlined by \cite{gorsane2022standardised}, and apply the evaluation tools from \cite{agarwal2022deep} in the MARL setting as advocated in the protocol. We evaluate agents at 201 equally spaced evaluation intervals for 32 episodes each during training. Following from the recommendations of \cite{papoudakis2021benchmarking} we train off-policy algorithms for a total of 2M timesteps and on-policy algorithms for a total of 20M timesteps summed across all parallel workers. This implies that evaluation occurs at fixed intervals of either $10$k or $100$k total environment steps for off- and on-policy algorithms respectively. For all our experiments, we use the EPyMARL framework \citep{papoudakis2021benchmarking} which is opensourced under the Apache 2.0 licence. This is to ensure we are evaluating all algorithms on the same tasks, using the same codebase as was done by \cite{papoudakis2021benchmarking} for maximal reproducibility. Furthermore, it allows us to use identical hyperparameters as used in their work, which are available in the Appendix section \ref{eval_section}. All results that are presented are aggregated over 10 independent experiment trials. In cases where aggregations are done over multiple tasks within an environment, as opposed to an individual task (e.g. for computing performance profiles), the interquartile mean is reported along with 95\% stratified bootstrap confidence intervals. For all plots except for sample efficiency curves, the absolute metric \citep{colas2018gep, gorsane2022standardised} for a given metric is computed. This metric is the average metric value of the best-performing policy found during training rolled out for 10 times the number of evaluation episodes. 

\textbf{Computational resources}. All experiments were run on an internal cluster using either AMD EPYC 7452 or AMD EPYC 7742 CPUs. Each independent experiment run was assigned 5 CPUs and 5GB of RAM with the exception of the scalability experiments which were exclusively run using AMD EPYC 7742 CPUs and either 5, 15, 30, or 200 GB of memory depending on the number of agents and subsequently the number of environment copies that were required.
\section{Results}

We demonstrate the validity of Agent Importance by considering its correlation to the true Shapley value, its computational scalability and its reliability in quantifying individual agent contributions. We then proceed to illustrate how Agent Importance may be used as an explainability tool. 

\subsection{Validating Agent Importance}

\begin{figure}
  \centering
    \begin{subfigure}[t]{0.008\textwidth}
        \scriptsize
        \textbf{(a)}
    \end{subfigure}
    \begin{subfigure}[t]{0.4\textwidth}
        \begin{subfigure}[t]{1\textwidth}
            \begin{subfigure}[t]{0.4\textwidth}
                \includegraphics[width=\linewidth, valign=t]{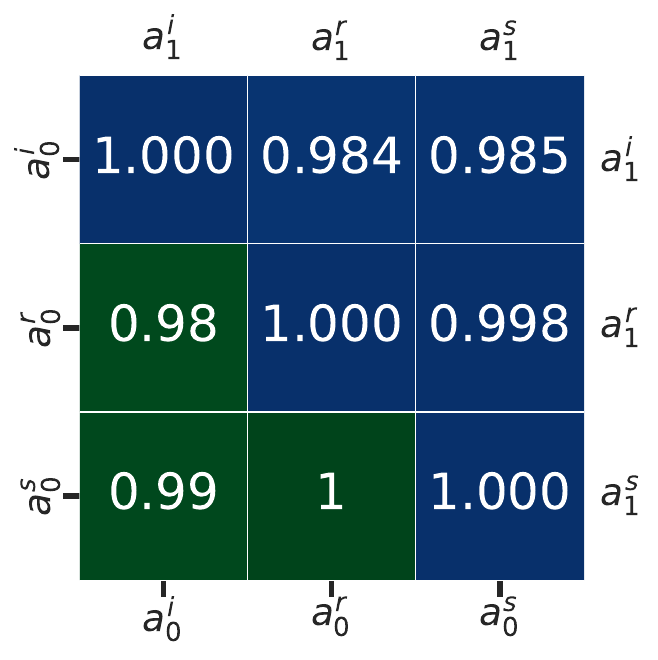}
            \end{subfigure}
            \begin{subfigure}[t]{0.4\textwidth}
                \includegraphics[width=\linewidth, valign=t]{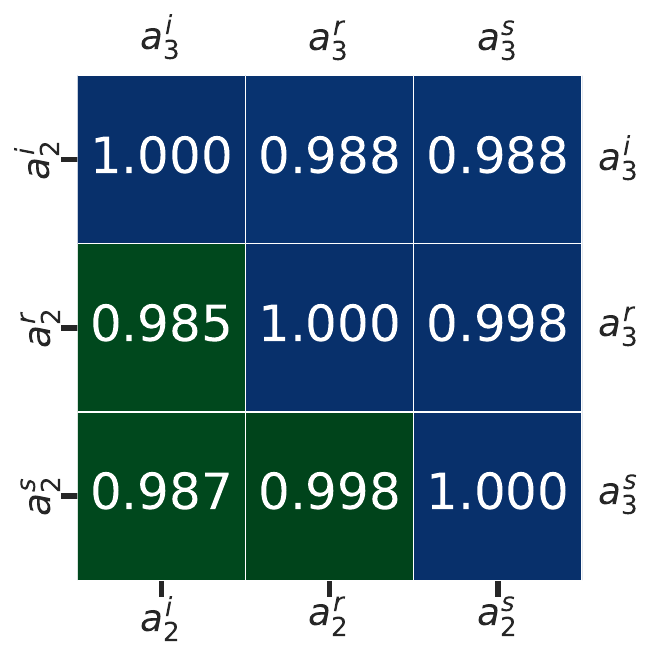}
            \end{subfigure}
    \end{subfigure}
    \begin{subfigure}[t]{1\textwidth}
        \begin{subfigure}[t]{0.4\textwidth}
            \includegraphics[width=\linewidth, valign=t]{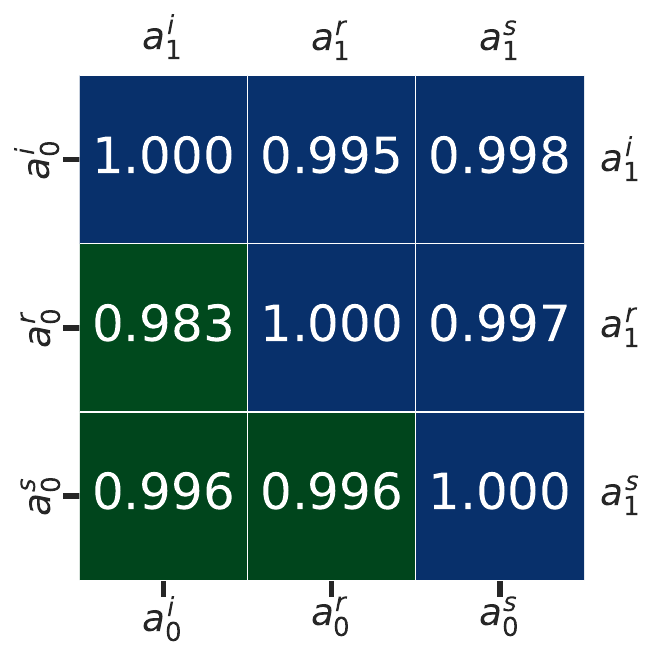}
        \end{subfigure}
        \begin{subfigure}[t]{0.4\textwidth}
            \includegraphics[width=\linewidth, valign=t]{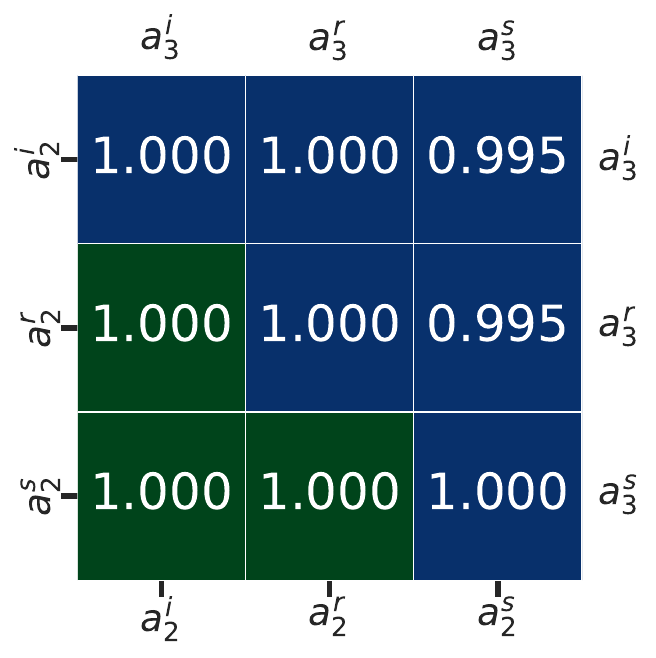}
        \end{subfigure}
    \end{subfigure}
    \label{fig:shapley_importance_heatmap}
  \end{subfigure}

  \begin{subfigure}[t]{0.008\textwidth}
        \scriptsize
        \textbf{(b)}
    \end{subfigure}
    \begin{subfigure}[t]{0.15\textwidth}
      \includegraphics[width=\linewidth, valign=t]{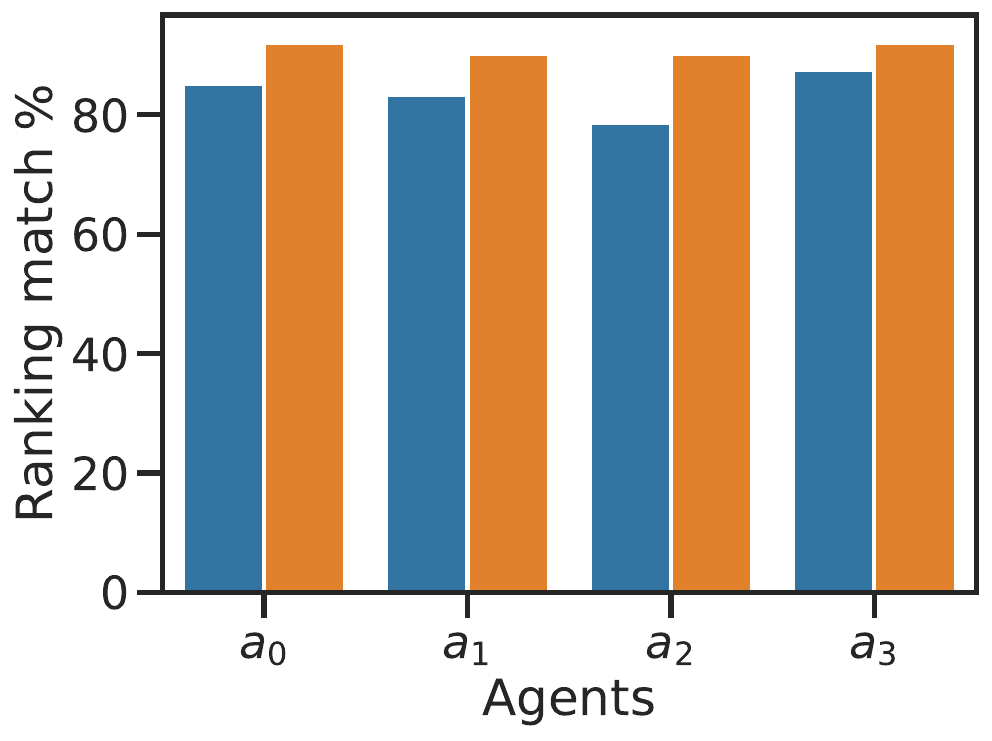}
      \label{fig:ranking_correlation_lbf}
    \end{subfigure}
    \begin{subfigure}[t]{0.008\textwidth}
        \scriptsize
        \textbf{(c)}
    \end{subfigure}
    \begin{subfigure}[t]{0.21\textwidth}
      \includegraphics[width=\linewidth, valign=t]{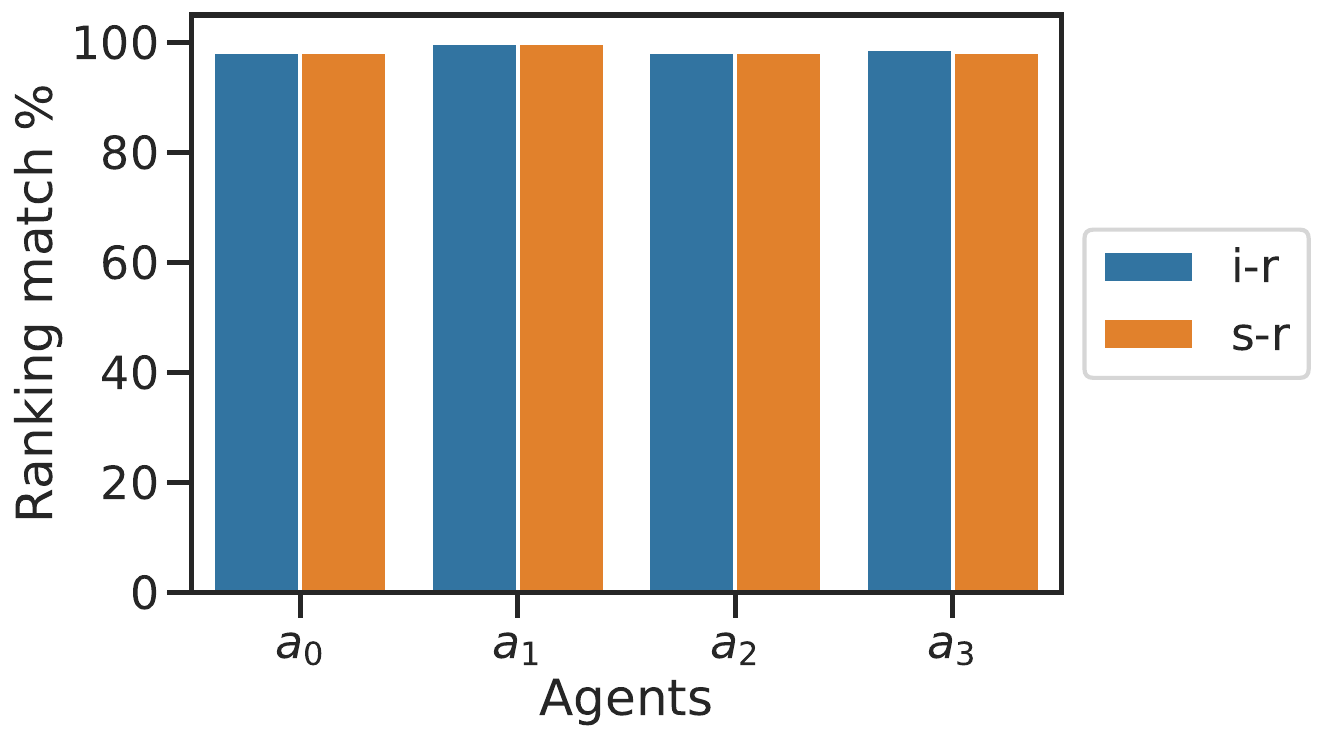}
      \label{fig:ranking_correlation_rware}
    \end{subfigure}
      \caption{\textit{Correlation analysis for agents $\{a_0, a_1, a_2, a_3\}$, for each metric: Agent Importance $i$, Shapley Value $s$, and Individual Reward $r$ using the VDN algorithm.} \textbf{(a)} Heatmap of Correlations among Metrics. \textbf{TOP:} LBF 15x15-4p-5f. \textbf{BOTTOM:} RWARE small-4ag. \textbf{(b)} Matching Rankings Comparison on LBF 15x15-4p-5f. \textbf{(c)} Matching Rankings Comparison on RWARE small-4ag. The legend refers to which metric is being compared to the individual agent rewards.}
  \label{fig:correlation}
\end{figure}

\textbf{Correlation between Agent Importance and the Shapley value. } We note that the Agent Importance metric is not mathematically equivalent to the Shapley value. It focuses on the grand coalition rather than all possible agent coalitions. However, through empirical study, we argue that Agent Importance is sufficient for capturing agents' contributions in the context of cooperative MARL. 

To validate our assertion, we conduct experiments on both LBF and RWARE to empirically assess the correlation between Agent Importance and the Shapley value. We generate a heatmap that describes the correlation between the metrics for the VDN algorithm. Furthermore, we assess the ability of a metric to maintain the relative agent rankings according to each agent's individual rewards (which are not seen by the agents). If a metric gives the same ranking to agents, we count this as a positive result--implying that a higher-ranking match is better. While only results on VDN are displayed, the trend is consistent for all algorithms across various tasks. Further results to this end are given in the Appendix section \ref{validation_section}. 

Figure \ref{fig:correlation} (a) shows that there exists a strong correlation between the Agent Importance, the Shapley value and the individual agent reward as calculated by the Pearson correlation coefficient. This indicates the effectiveness of both the Shapley value and Agent Importance in assessing agents' contributions, making them valuable substitutes for individual agent rewards in environments where such rewards are unavailable. Notably, Agent Importance showcases a promising ability to effectively replace both the Shapley value and individual rewards. While the Shapley value may provide greater consistency in ranking information when compared to the Agent Importance (as illustrated in Figures \ref{fig:correlation} (b,c)) where the frequency of ranking agreement between the individual reward and the contribution estimators is illustrated, it is important to note that Agent Importance is highly correlated with the individual reward and shows a minimal rate of non-matched rankings.

\begin{figure}
\centering
\includegraphics[width=0.8\linewidth]{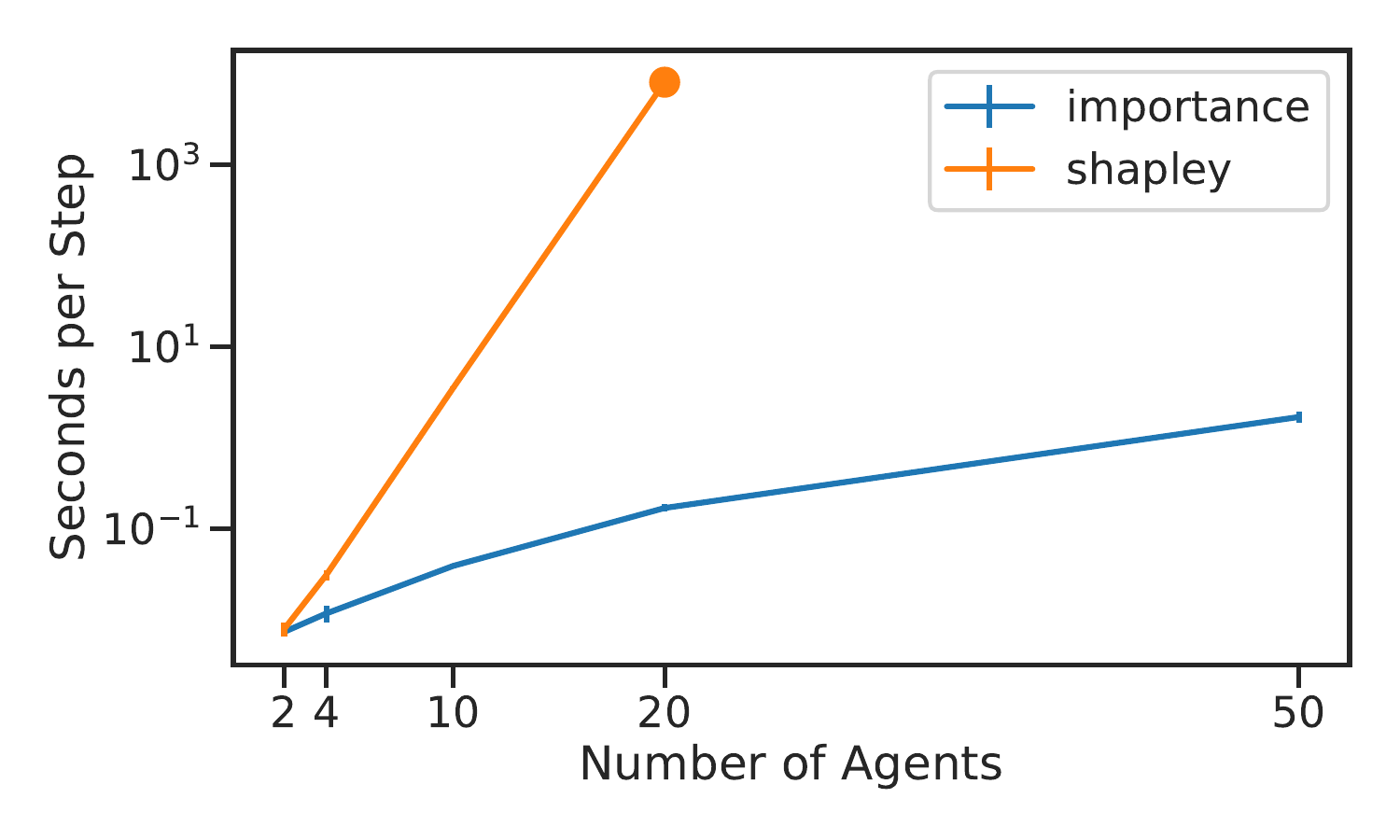}
\caption{\textit{Computational cost of computing the agent importance and the Shapley value.} }
\label{fig: compute_time}
\end{figure}

\textbf{Scalability of Agent Importance. } In order to validate the computational feasibility of the simplified Agent Importance against the full Shapley value we record the run time of both approaches on LBF tasks with $2, 4, 10, 20$, and $50$ agents. We run the algorithm without any training and compute the number of seconds it takes for agents to take a single environment step while computing each metric. The reported results here are the mean and standard deviation over 3 independent runs. The Shapley value became prohibitively slow as the agent number was increased and required approximately $2$ hours to measure a single step within the environment with $20$ agents. Nonetheless, Figure \ref{fig: compute_time} clearly illustrates how the Agent Importance is significantly more computationally efficient than the Shapley value.  

\begin{figure}
     \centering
     \begin{subfigure}{0.18\textwidth}
     \includegraphics[width=\textwidth]{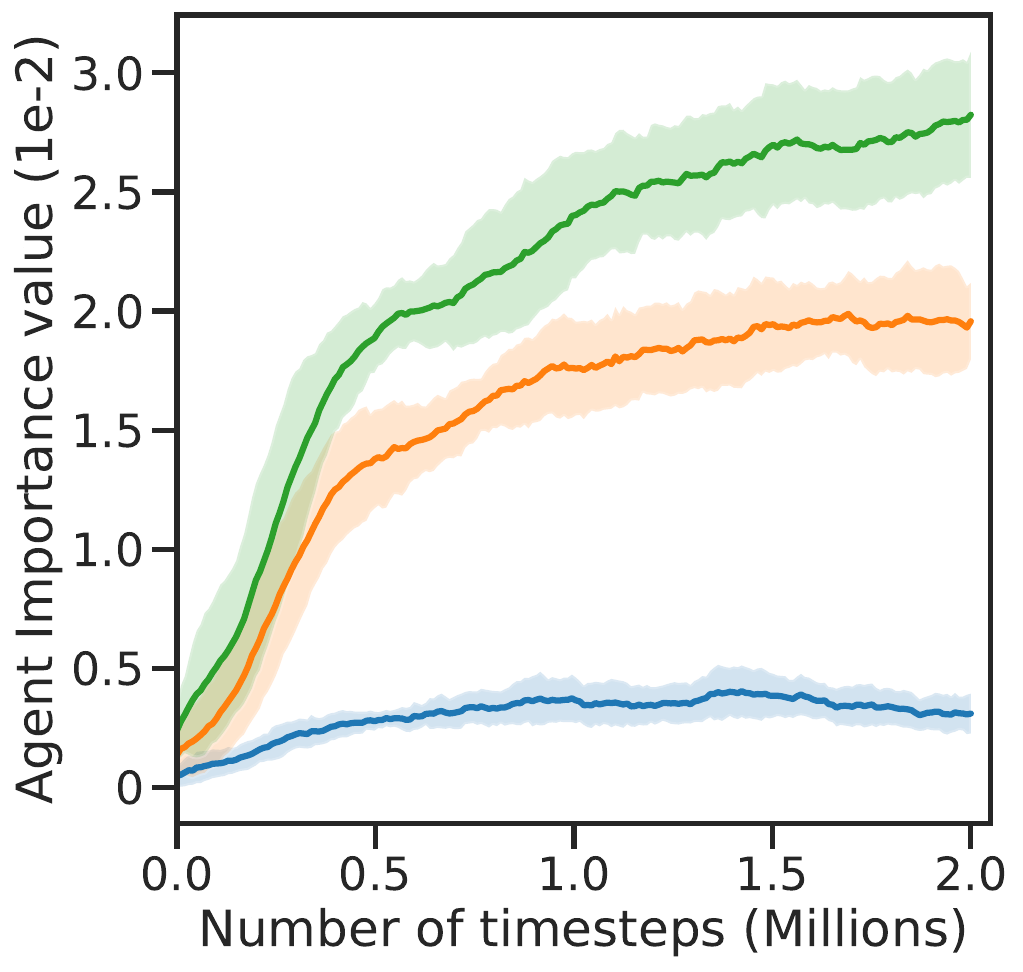}
     \end{subfigure}
     \begin{subfigure}{0.18\textwidth}
         \includegraphics[width=\textwidth]{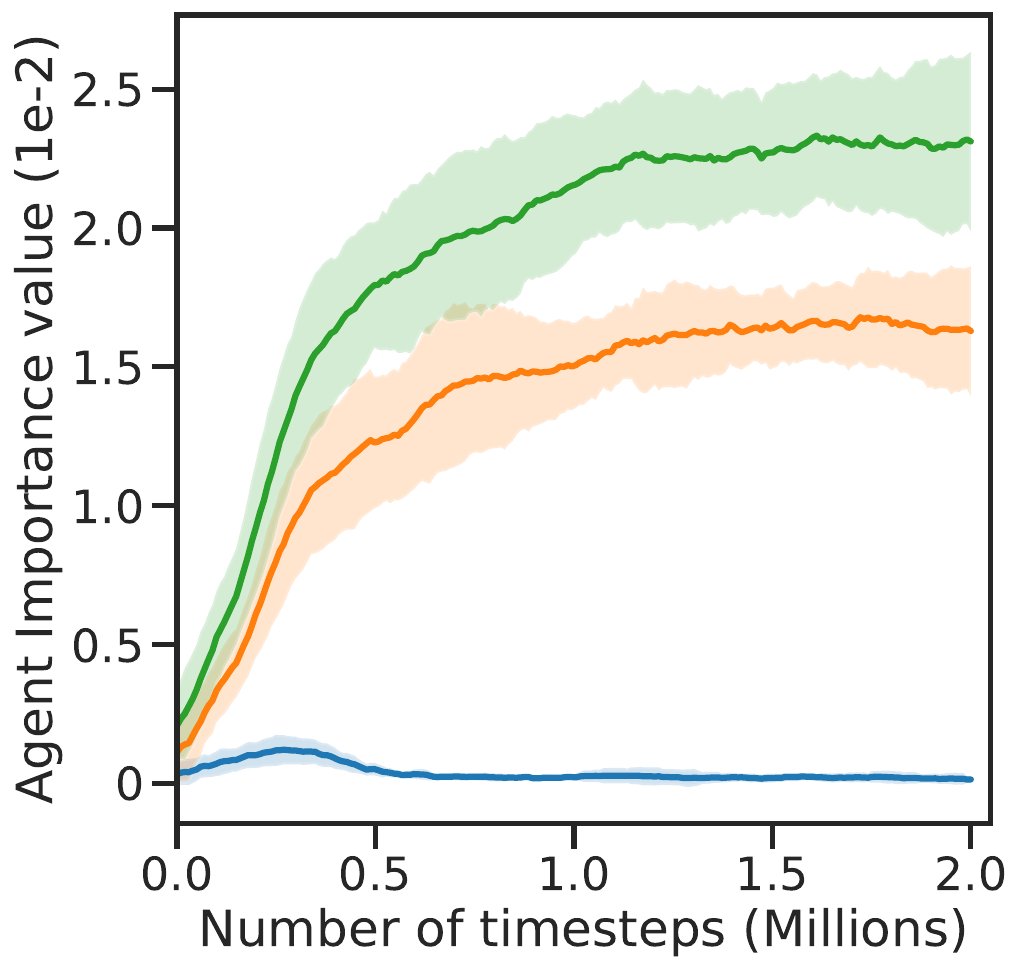}
     \end{subfigure}
     \\
     \begin{subfigure}{0.18\textwidth}
     \includegraphics[width=\textwidth]{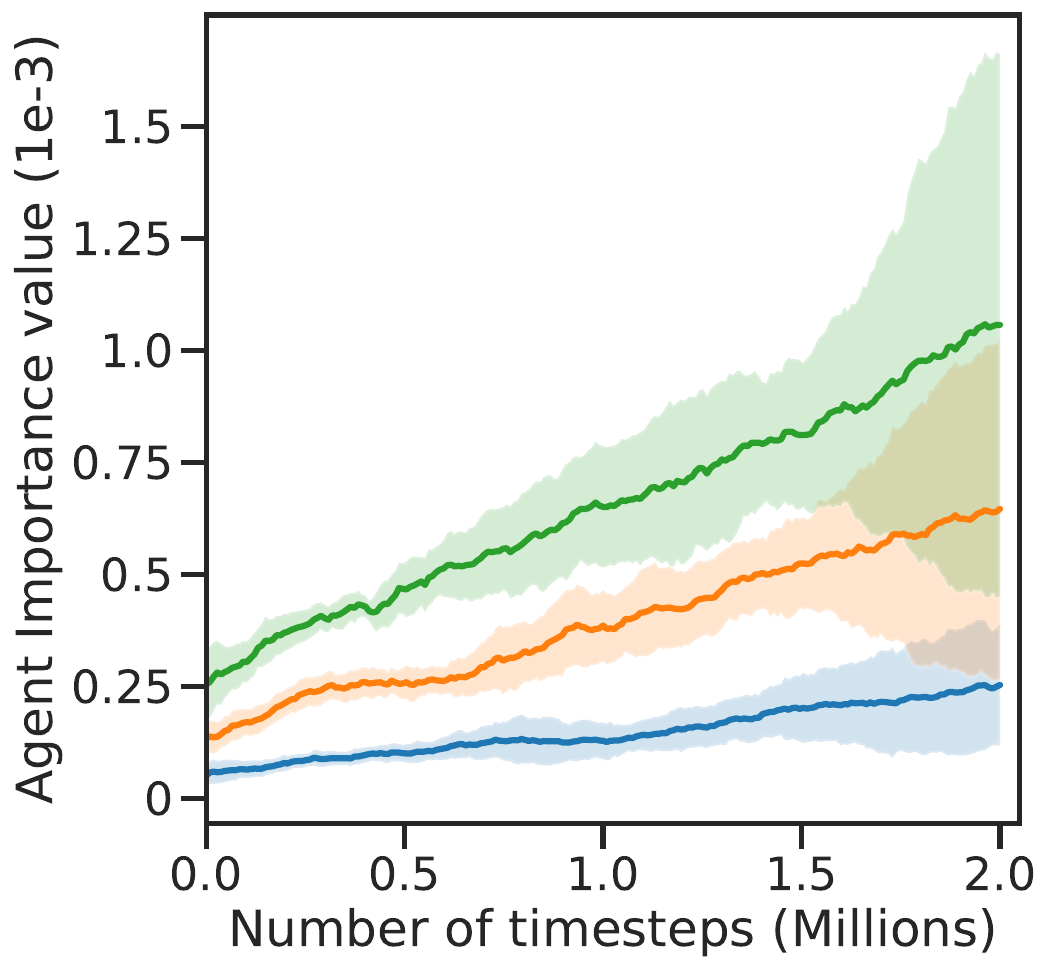}
     \end{subfigure}
     \begin{subfigure}{0.18\textwidth}
         \includegraphics[width=\textwidth]{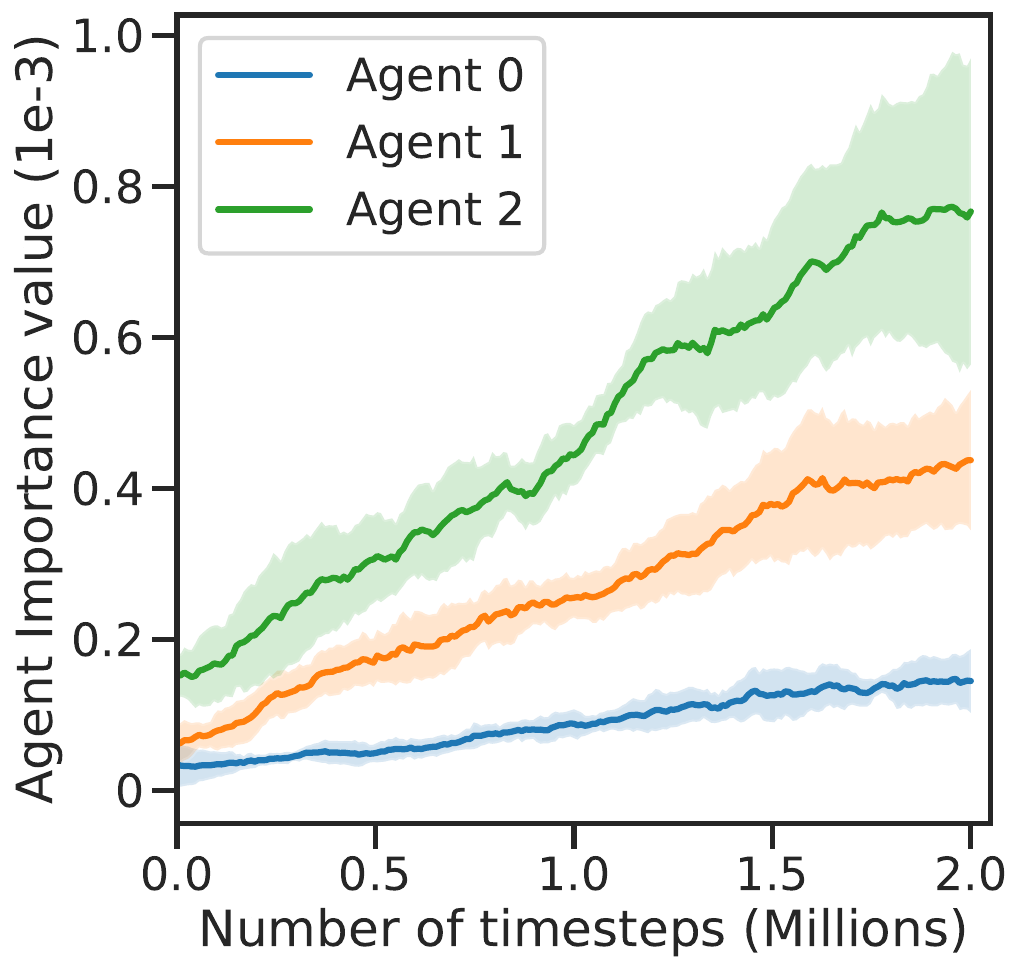}
     \end{subfigure}
     \caption{\textit{Agent importance scores on the deterministic LBF scenario for MAA2C, MAPPO, VDN and QMIX.} Agents $0$, $1$ and $2$ are assigned fixed levels of $1$, $2$ and $3$ respectively--implying that their contributions should be weighted accordingly.}
     \label{fig: det results}
 \end{figure}

\textbf{Reliability of Agent Importance.} In order to validate the ability of the Agent Importance to effectively untangle agent contributions from a shared team reward, we create a deterministic version of LBF where agent levels are always fixed to be $1, 2$, and $3$ respectively, and the maximum level of each food is a random value between 1 and 6. Since agent $2$ is assigned a fixed greater level than its counterparts we should expect it to contribute the most to the team return. Figure \ref{fig: det results} illustrates the ability of Agent Importance to uncover the correct ordering and approximate level of contribution among agents towards the overall team goal. 

\subsection{Applications of Agent Importance}

\begin{figure}
    \centering
    \includegraphics[width=0.4\textwidth]{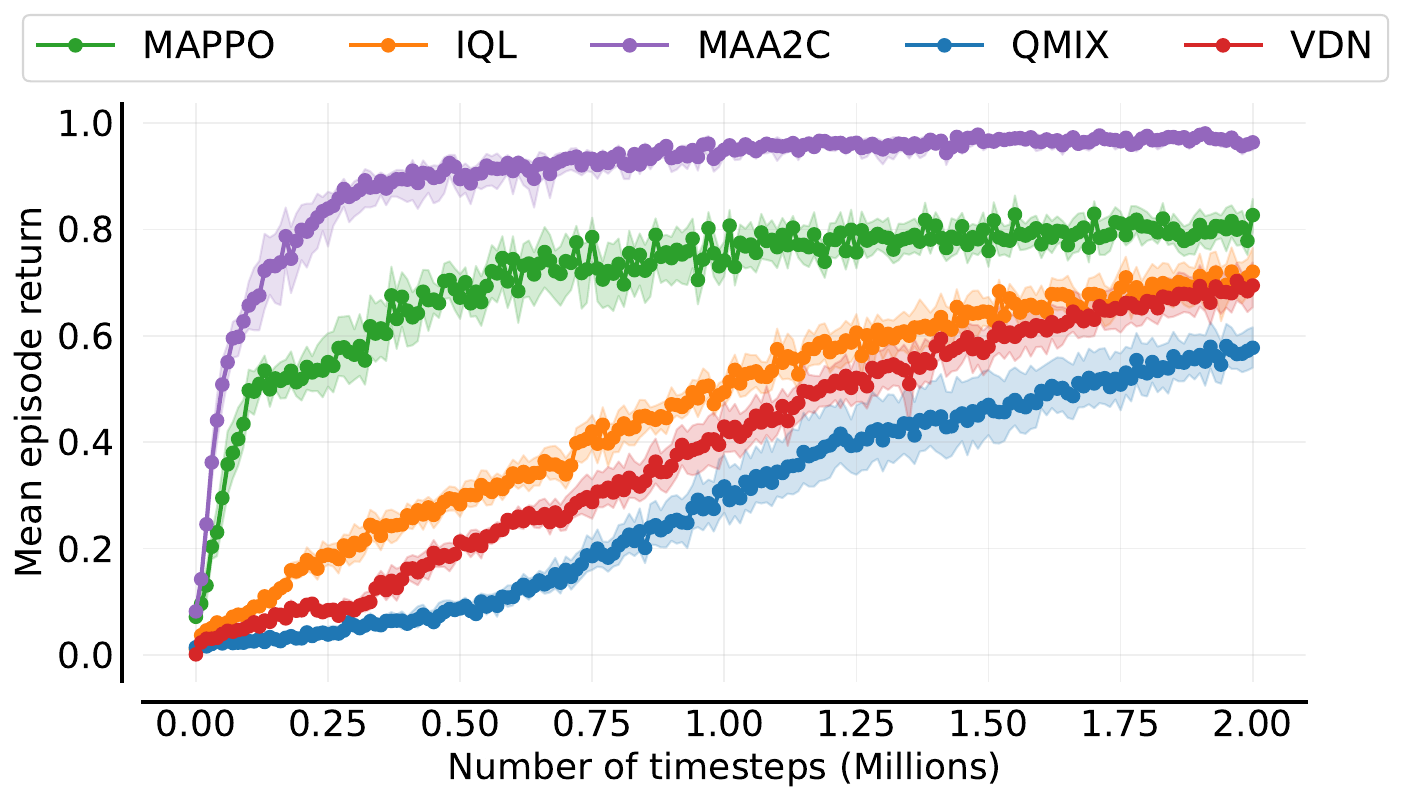}
    \includegraphics[width=0.22\textwidth]{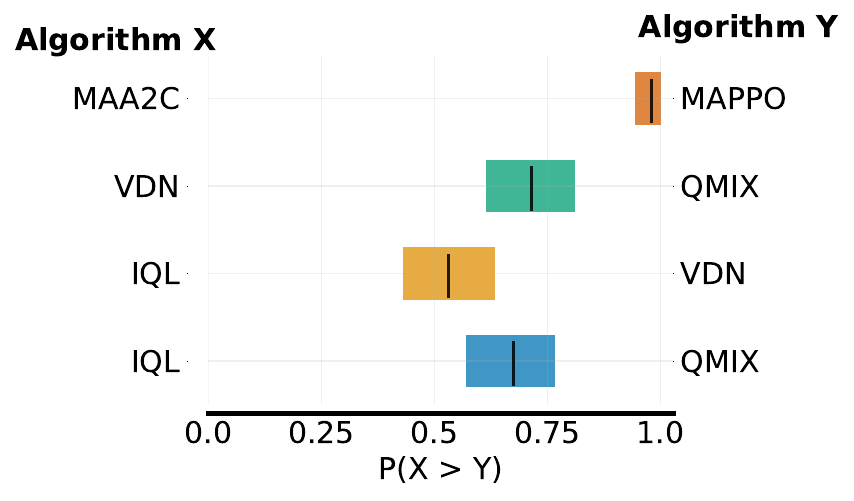}
    \includegraphics[width=0.2\textwidth]{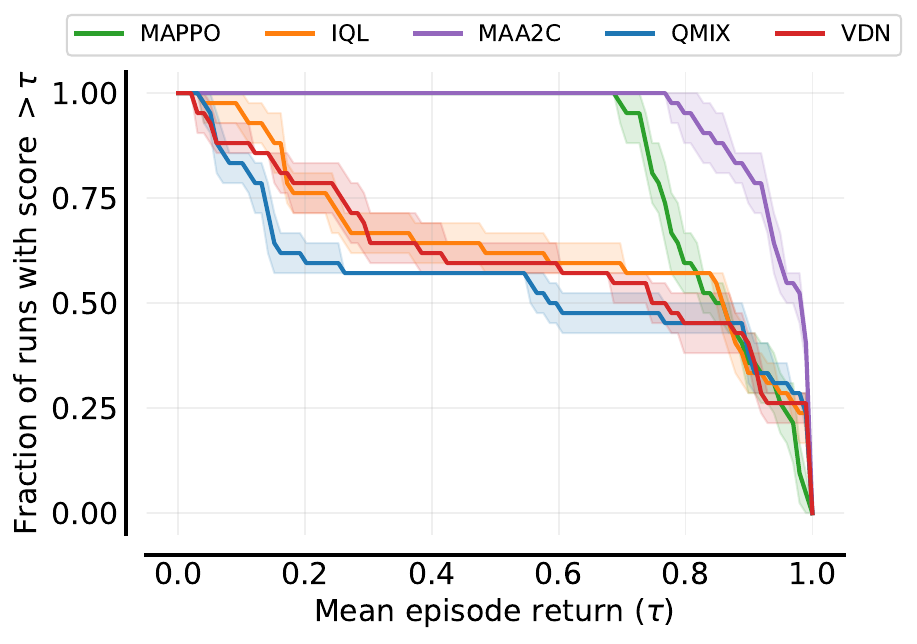}
    \includegraphics[width=0.4\textwidth]{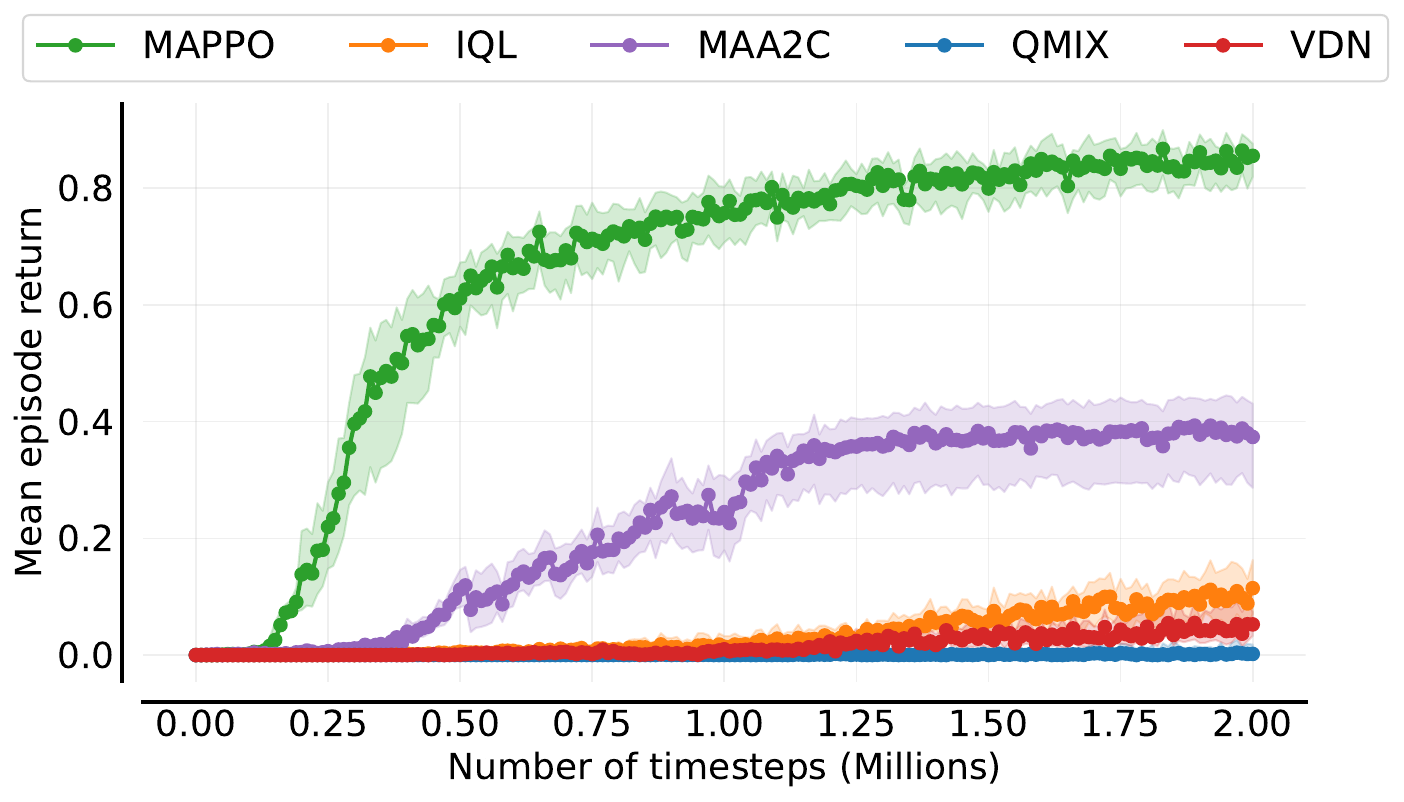}
    \includegraphics[width=0.22\textwidth]{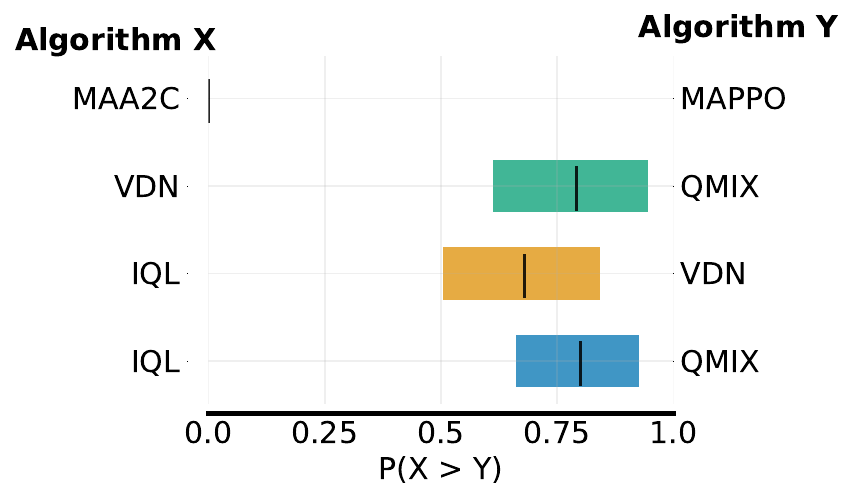}
    \includegraphics[width=0.2\textwidth]{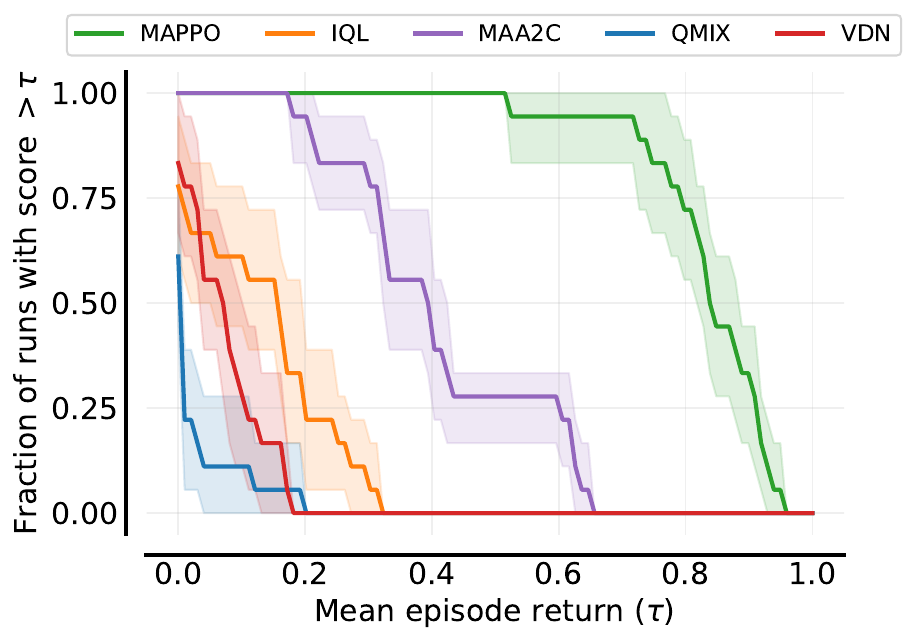}
    \caption{\textit{Algorithm performance on LBF and RWARE including probability of improvement, performance profiles and sample efficiency curves}. \textbf{Top two rows:} Performance of algorithms on 7 LBF tasks. \textbf{Bottom two rows:} Performance of all algorithms on 3 RWARE tasks.}
    \label{fig: benchmarking_paper_redo}
\end{figure}

We replicated the experiments performed by \cite{papoudakis2021benchmarking}, obtaining similar results. However, our work adds value by following a strict protocol \citep{gorsane2022standardised} which includes additional evaluation measurements such as examining the probability of improvement and providing performance profiles \citep{agarwal2022deep}, as shown in Figure \ref{fig: benchmarking_paper_redo}. Additional plots and tabular results for different scenarios and the performance of the algorithms without parameter sharing are included in the Appendix along with more detailed performance plots for SMAClite in Appendix section \ref{smaclite}. Through these experiments, we delve into analysing the agents’ behaviours and previously unexplained results by employing the Agent Importance metric.

\begin{figure}
    \centering
    \begin{subfigure}[t]{0.01\textwidth}
        \scriptsize
        \textbf{(a)}
    \end{subfigure}
    \begin{subfigure}[t]{0.4\textwidth}
    \includegraphics[width=\linewidth, valign=t]{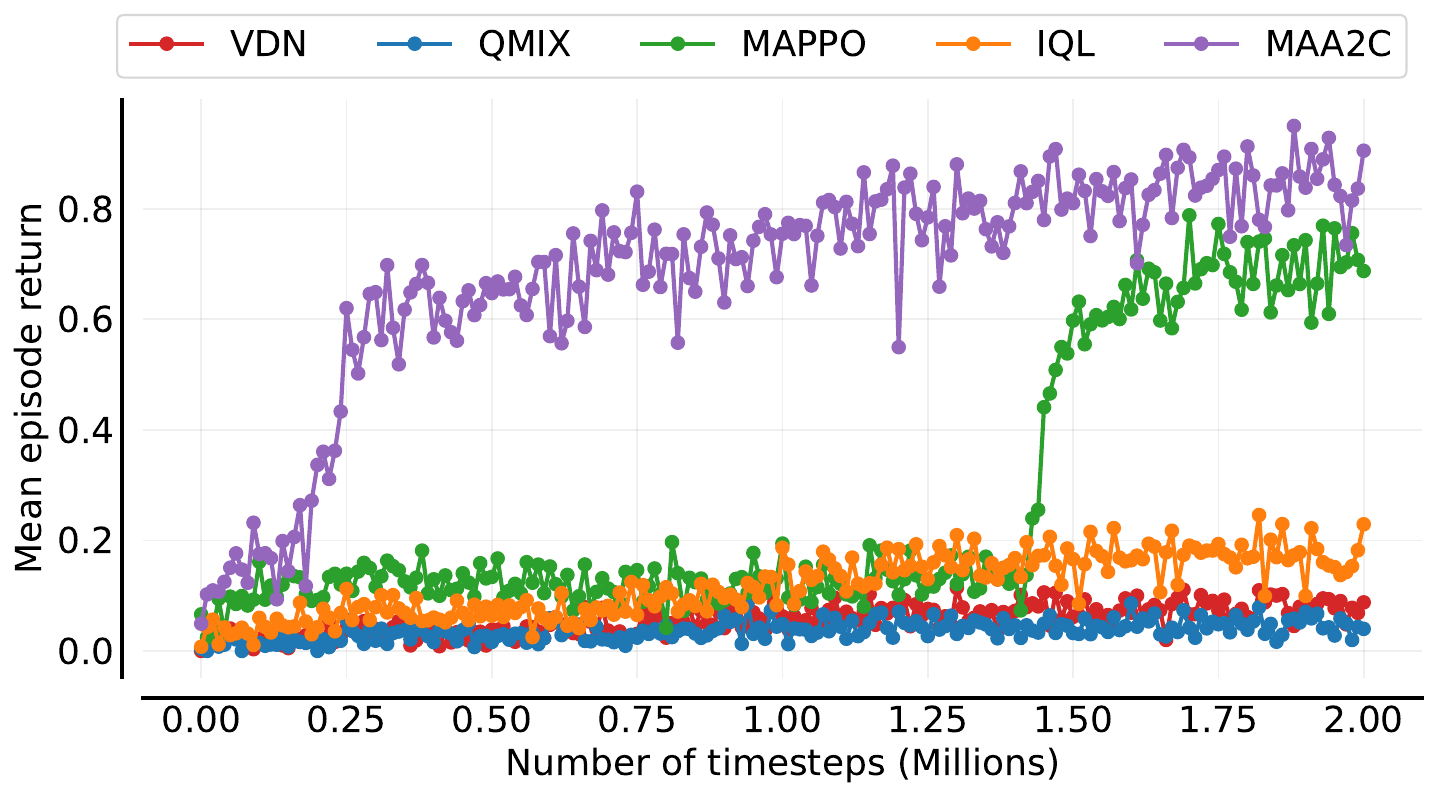}
    \end{subfigure}
    \\
    \begin{subfigure}[t]{0.01\textwidth}
        \scriptsize
        \textbf{(b)}
    \end{subfigure}
    \begin{subfigure}[t]{0.2\textwidth}
        \includegraphics[width=\linewidth, valign=t]{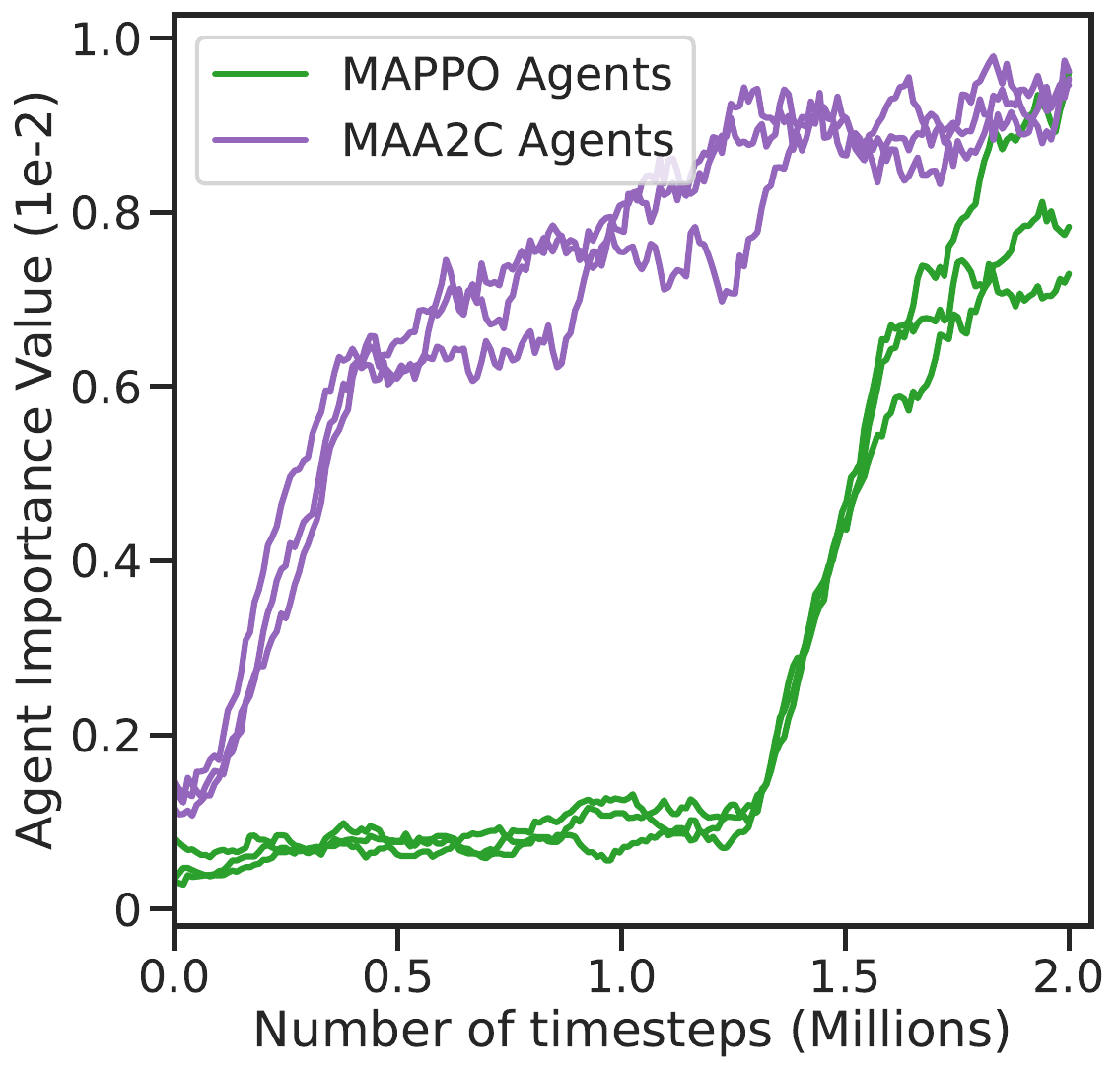}
    \end{subfigure}
    \begin{subfigure}[t]{0.01\textwidth}
        \scriptsize
        \textbf{(c)}
    \end{subfigure}
    \begin{subfigure}[t]{0.2\textwidth}
        \includegraphics[width=\linewidth, valign=t]{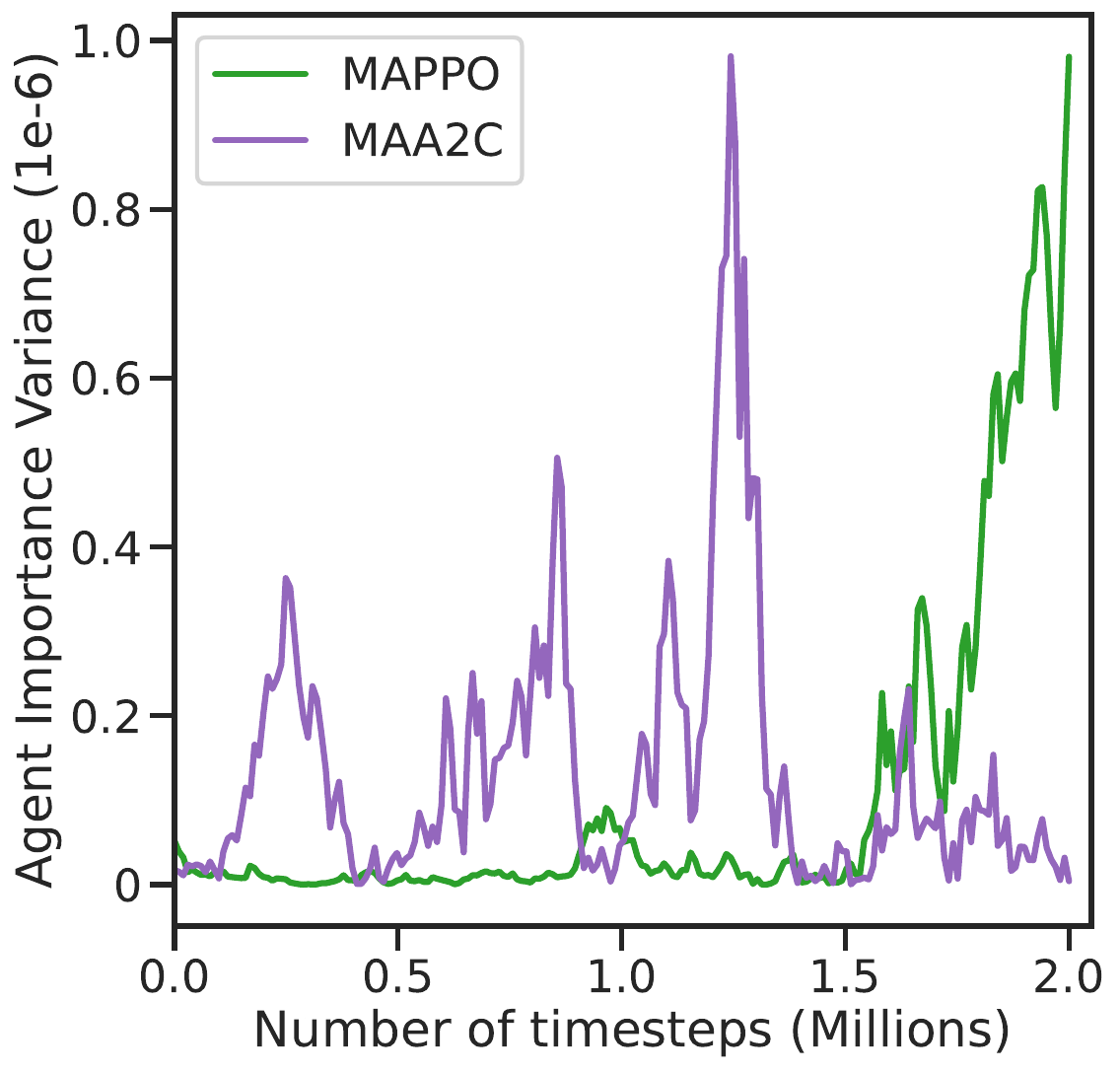}
    \end{subfigure}
    \caption{\textit{MAA2C outperforms MAPPO on the LBF 15x15-3p-5f task.} \textbf{(a)} Sample efficiency curves (one seed). \textbf{(b)} Agent importance for all agents associated with a given algorithm. \textbf{(c)} Variance of Agent Importance. MAA2C has a lower variance in agent importance at convergence.}
    \label{fig: maa2c_vs_mappo}
\end{figure}

\textbf{MAA2C vs MAPPO. } Empirical results in RL consistently demonstrate that PPO tends to outperform A2C \citep{Heess2017EmergenceOL, schulman2017proximal, Henderson2018WhereDM}. This trend naturally leads to the question of whether a similar pattern is observed in the multi-agent setting, i.e.\ between MAPPO and MAA2C. However, when examining the results in Figure \ref{fig: benchmarking_paper_redo}, a conflicting observation arises. In the case of RWARE, we observe the expected behaviour with the probability of improvement aligning with our initial expectations. However, in the case of LBF, the opposite occurs as MAA2C outperforms MAPPO, presenting an unexpected outcome. Figure \ref{fig: maa2c_vs_mappo} (b) highlights a possible reason. By tracking Agent Importance, we may attribute this outcome to a narrowing in the spread of importance values between MAA2C agents at convergence, as compared to MAPPO agents. The assumption of lower variance in Agent Importance leading to improved performance in LBF is due to the stochasticity of the environment. It is reasonable to expect that an algorithm performing well in this environment should have the capability to adapt to the variability in agent and food levels across episodes. From the narrower spread in Agent Importance values in MAA2C we can see it has learnt treat all agents as equally important.
For additional findings on RWARE see section \ref{additional_RWARE} in the supplementary material. 

\begin{figure}[t!]
    \centering
    \begin{subfigure}[b]{0.4\textwidth}
    \includegraphics[width=\textwidth]{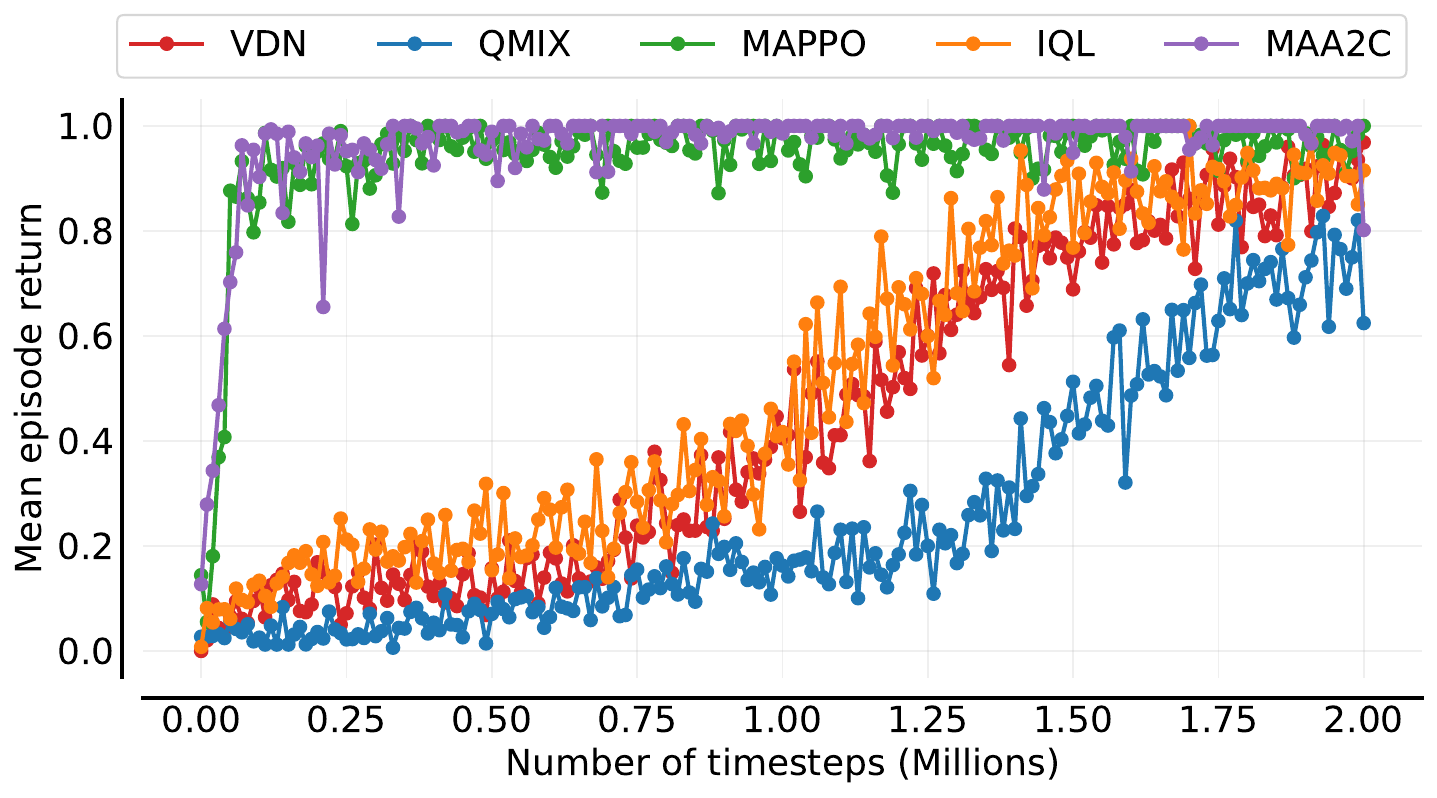}
    \end{subfigure}
    \begin{subfigure}[b]{0.2\textwidth}
        \includegraphics[width=\textwidth]{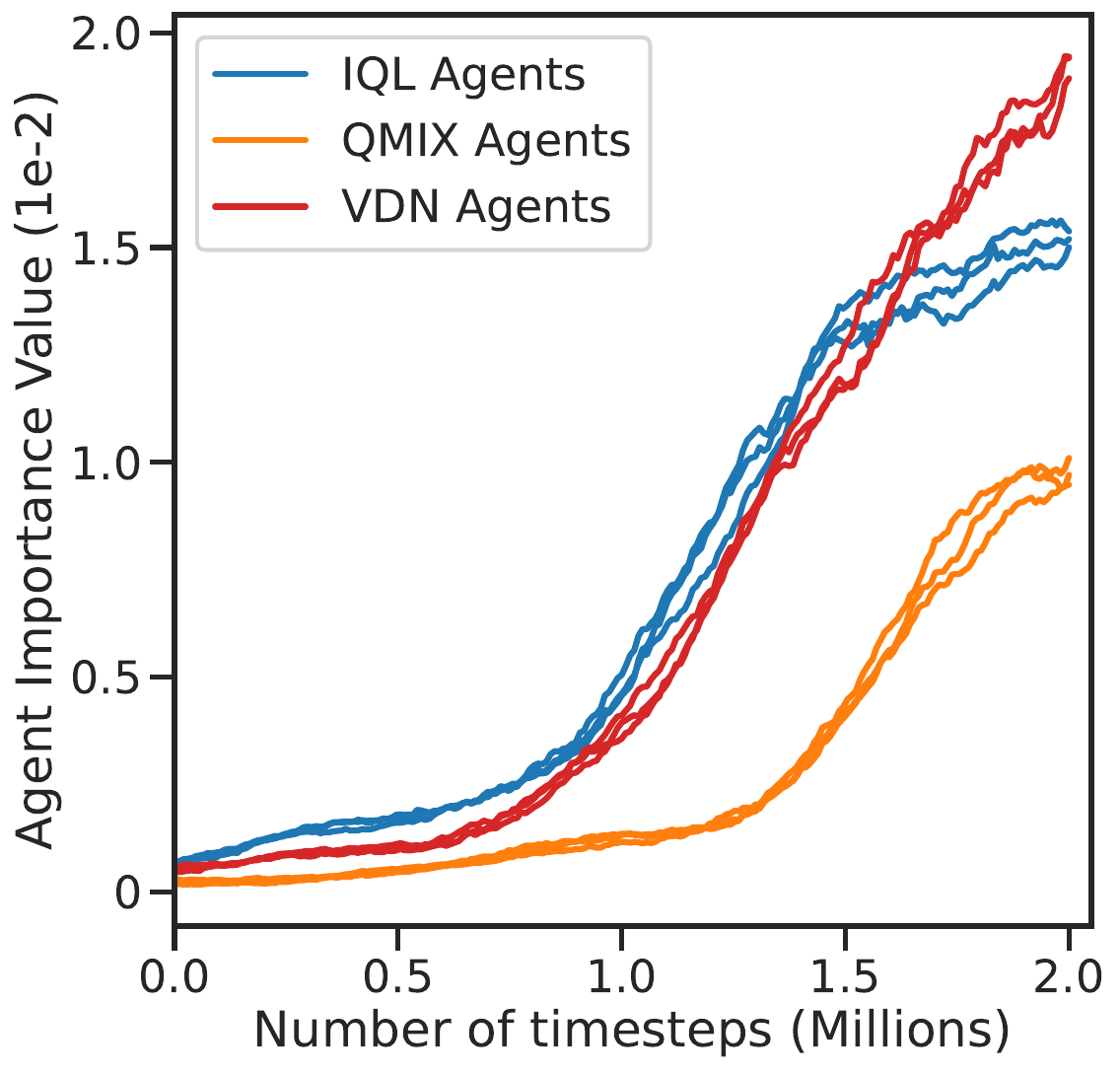}
    \end{subfigure}
    \begin{subfigure}[b]{0.2\textwidth}
        \includegraphics[width=\textwidth]{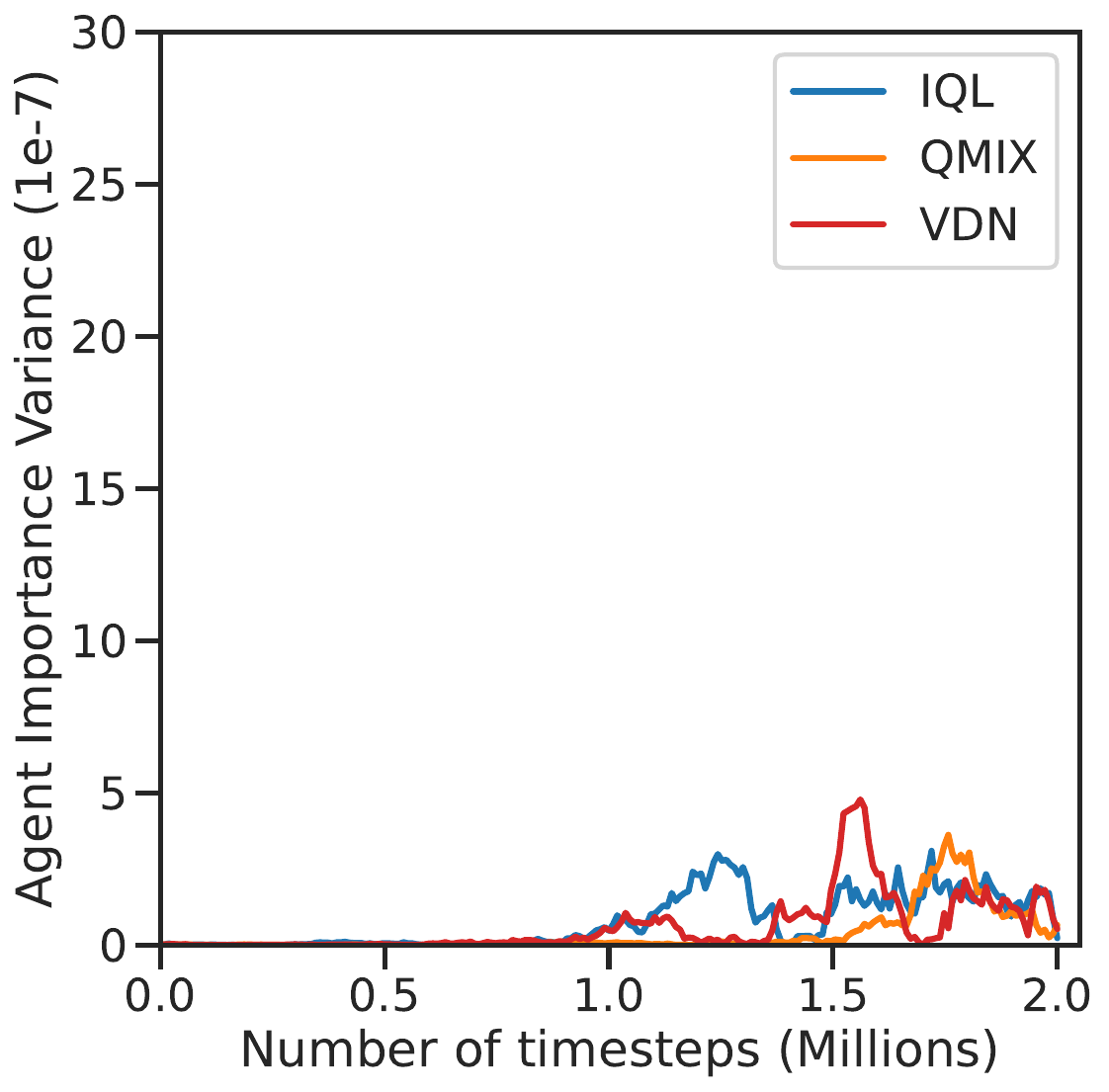}
    \end{subfigure}
    \centering
    \begin{subfigure}[b]{0.4\textwidth}
    \includegraphics[width=\textwidth]{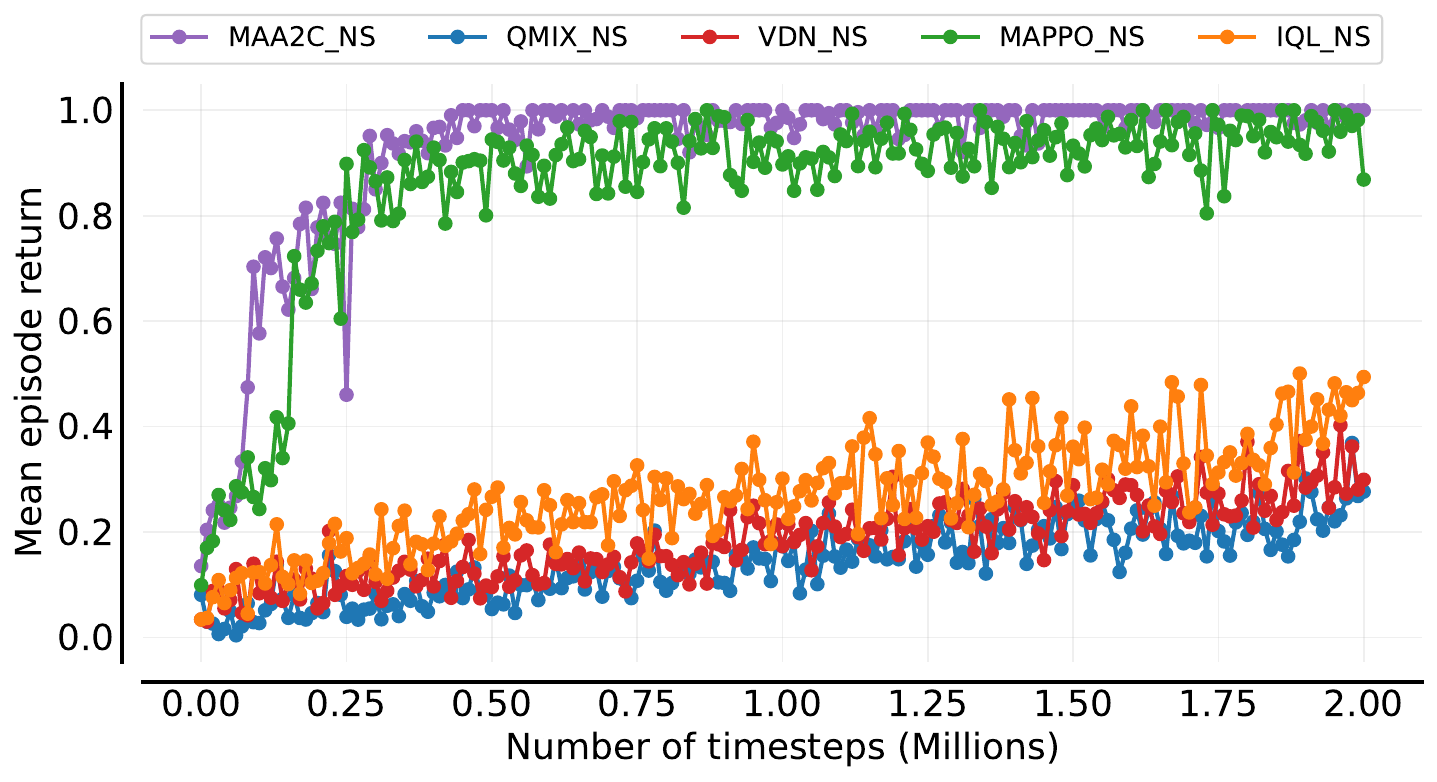}
    \end{subfigure}
    \begin{subfigure}[b]{0.2\textwidth}
        \includegraphics[width=\textwidth]{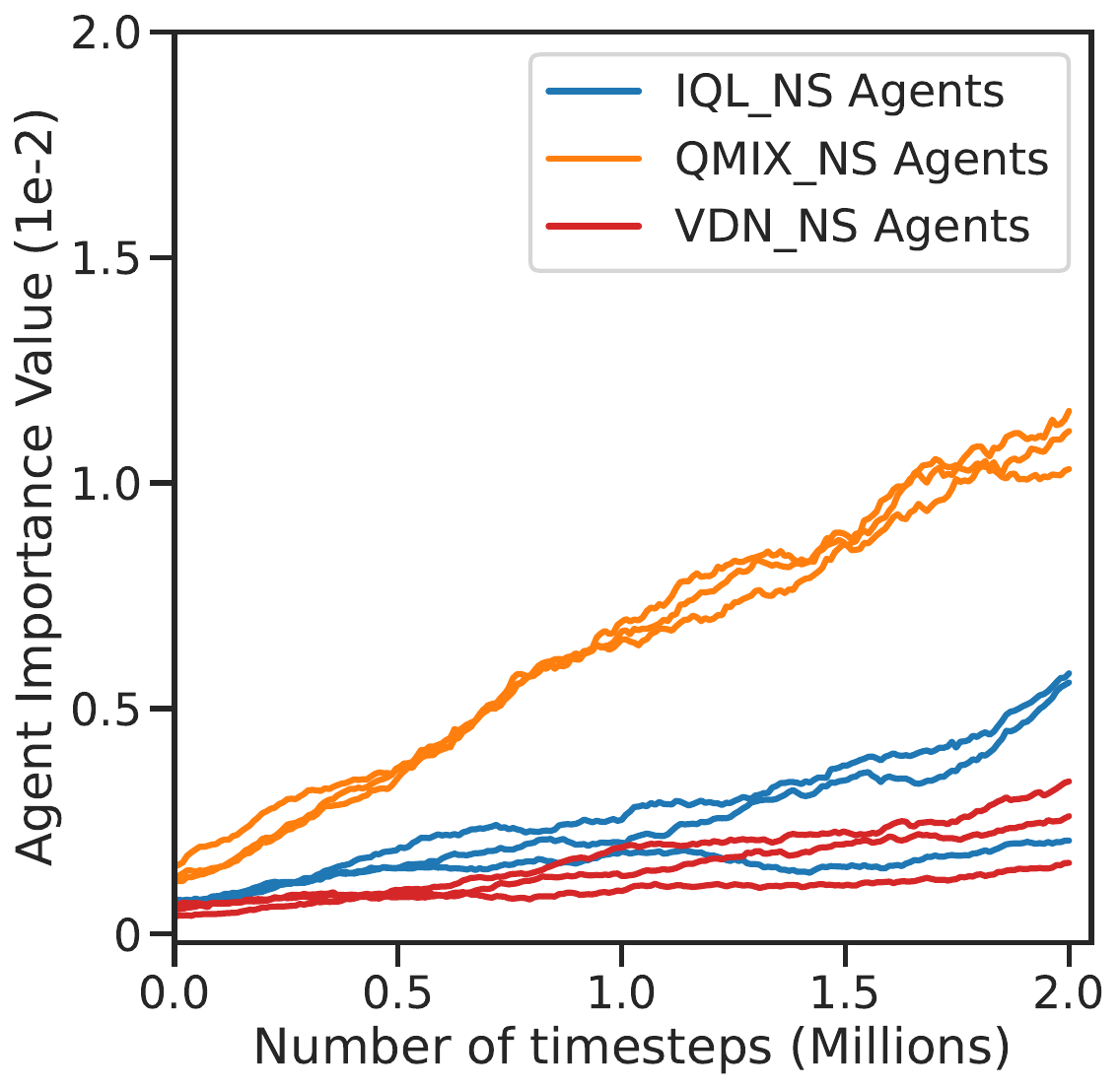}
    \end{subfigure}
    \begin{subfigure}[b]{0.2\textwidth}
        \includegraphics[width=\textwidth]{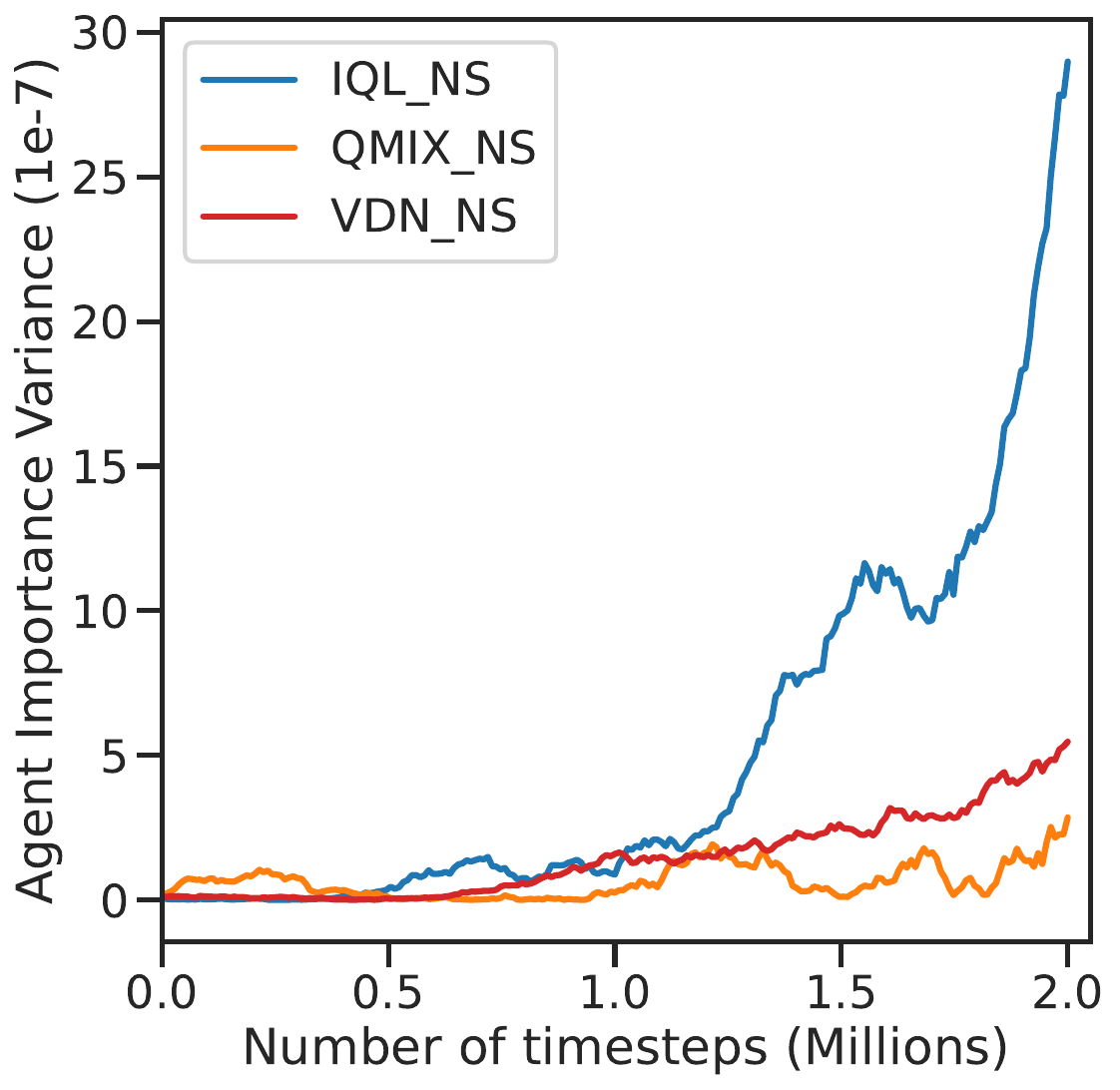}
    \end{subfigure}
    \caption{\textit{Comparison of performance with and without parameter sharing on the LBF 10x10-3p-3f task for one seed including the sample efficiency, Agent Importance, and Agent Importance variance.} \textbf{Top two rows:} Performance with policy parameter sharing. \textbf{Bottom two rows:} Performance without policy parameter sharing. With parameter sharing the agent importance is more evenly distributed.}
    \label{fig: ps_vs_non-ps}
\end{figure}

\textbf{Parameter sharing vs non-parameter sharing. }Consistent with the findings of \cite{papoudakis2021benchmarking}, our experiments demonstrate that algorithms utilizing parameter sharing outperform those without it. As mentioned in the benchmark paper, this outcome is expected as parameter sharing enhances sample efficiency. Additionally, parameter sharing enhances the sharing of learned information across the system. The Agent Importance analysis for IQL, QMIX, and VDN provides clear evidence of the impact of parameter-sharing architectures, as illustrated in Figure \ref{fig: ps_vs_non-ps}. It is apparent that in the absence of parameter sharing, the agents contribute to varying degrees, leading to an uneven distribution of importance. And as mentioned previously, given LBF's characteristics, requiring a high level of coordination in the presence of significant stochasticity, all agents should be expected (on average) to contribute equally. However, in the non-parameter sharing cases, especially for IQL and VDN, we observe that a small number of the agents dominate the contributions, resulting in lower performance compared to when parameter sharing is utilised. 

\begin{figure}
    \centering
    \begin{subfigure}[t]{0.01\textwidth}
        \scriptsize
        \textbf{(a)}
        \phantomcaption
        \label{fig: MMM2_Sample}
    \end{subfigure}
    \begin{subfigure}[t]{0.4\textwidth}
    \includegraphics[width=\linewidth, valign=t]{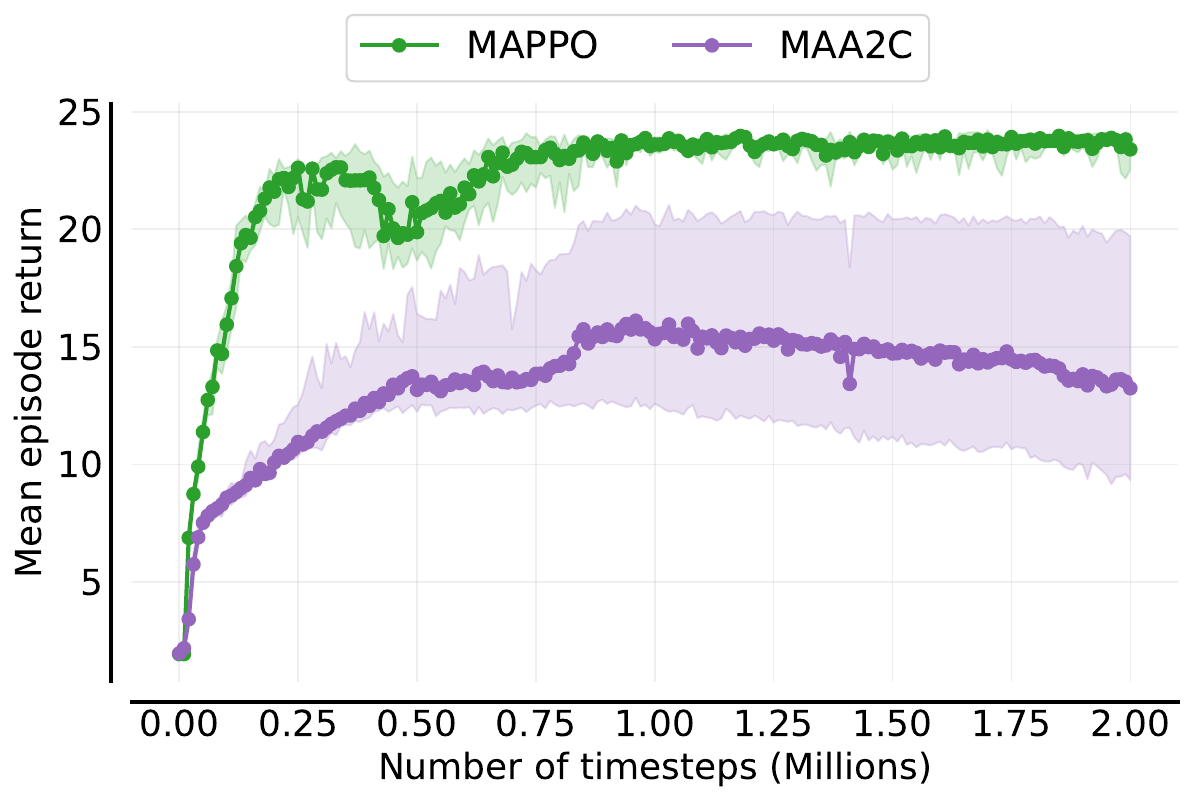}
    \end{subfigure}
    \\
    \begin{subfigure}[t]{0.01\textwidth}
        \scriptsize
        \textbf{(b)}
        \phantomcaption
        \label{fig: maa2c_MMM2}
    \end{subfigure}
    \begin{subfigure}[t]{0.2\textwidth}
    \includegraphics[width=\linewidth, valign=t]{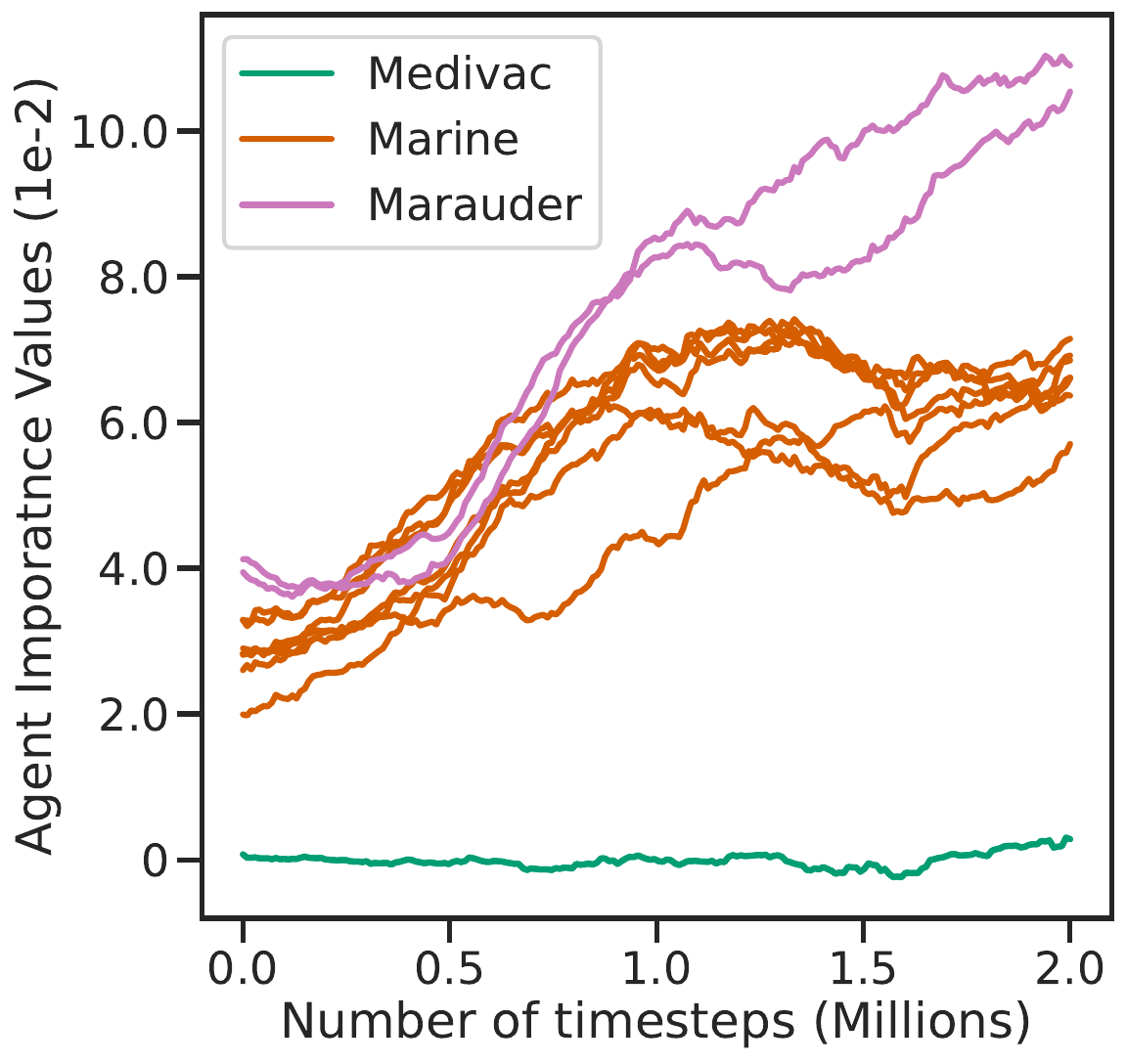}
    \end{subfigure}
    \begin{subfigure}[t]{0.01\textwidth}
        \scriptsize
        \textbf{(c)}
        \phantomcaption
        \label{fig: mappo_MMM2}
    \end{subfigure}
    \begin{subfigure}[t]{0.2\textwidth}
    \includegraphics[width=\linewidth, valign=t]{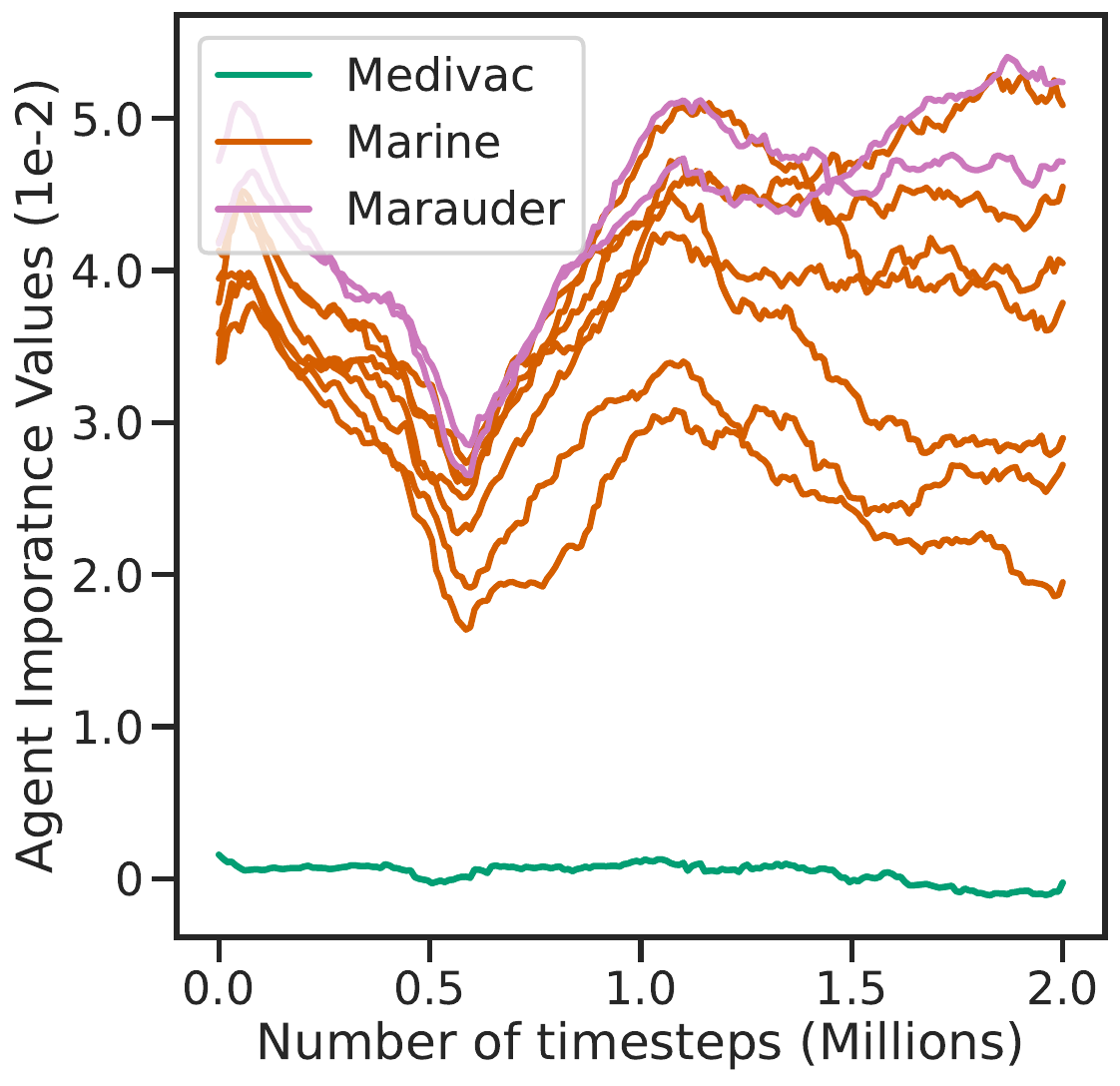}
    \end{subfigure}
    \centering
    \caption{\textit{Comparison between MAA2C and MAPPO in the MMM2 scenario from SMAClite.} \textbf{(a)}: Mean episode returns. \textbf{(b)}: Agent Importance scores MAA2C. \textbf{(c)}: Agent Importance scores MAPPO.} 
    \label{fig: MMM2}
\end{figure}

\textbf{Heterogeneous Agents.} In both LBF and RWARE the importance of each agent and the total reward are highly correlated as all agents have similar capabilities. In the heterogeneous setting of MMM2, rather than converging to similar importance levels over time, agents will instead converge to clear groups of importance levels as seen in figures \ref{fig: maa2c_MMM2} and \ref{fig: mappo_MMM2}. Furthermore, note that agents of the same type can still fall into different levels of importance which is consistent with role decomposition analysis in ROMA \citep{yang2020qatten}. As shown by \cite{yang2020qatten}, the optimal policy in MMM2 requires a subset of marine agents to die early in the episode, who then cannot contribute to the team reward remaining timesteps, whereas a smaller number of marines survive until the end. This is clearly seen in figure \ref{fig: mappo_MMM2} for MAPPO. In the case of MAA2C in figure \ref{fig: maa2c_MMM2} we can see that although clear clusters have formed, it has not learned to assign the correct importance to a subgroup of marines that are required to optimally solve the environment \footnote{An optimal policy for MMM2 can be found in a video by the original SMAC authors in 
\url{https://www.youtube.com/watch?v=VZ7zmQ\_obZ0} and additional information in appendix \ref{param_sharing_MMM2}.} 



\section{Discussion} 


In this work, we illustrate that Agent Importance is an efficient and reliable measure for agent contributions towards the team reward in cooperative MARL. Aside from only quantifying the agent contributions we have also shown how the metric may be used as an explainability tool for uncovering failure modes in existing MARL results.

\textbf{Limitations. } Although Agent Importance is useful, using simulators that allow for an agent's removal during runtime would be highly advantageous. Solely relying on no-op actions could still impact the coalition reward by obstructing other agents' presence and movement in their observations. Unfortunately, agent removal is uncommon in most simulators and some simulators also do not offer the option for a no-op action. Additionally, while popular MARL research environments are fairly low resource, creating multiple parallel instances of the environment during the Agent Importance calculation, makes using more resource-heavy simulators prohibitive from a memory perspective. However, with the growing popularity of the JAX framework, more stateless environments are becoming available where the parallel environments can be replaced with direct access to the environment state \citep{brax2021github, gymnax2022github, jumanji2023github}.

\textbf{Future Work. } It would be useful to investigate the rankings calculated by \textit{agent importance} for simulators which do not have a no-op action. We could consider using random actions or the random actions of specific agents as a proxy for the no-op action or make use of function approximators to learn minimal impact actions for the marginalised agents.

\bibliography{reference}

\appendix
\section{ Experimental details}

\subsection{Environments}

To ensure our experimentation setup is clear and easily reproducible, we make use of the same environment naming conventions used in \citep{papoudakis2021benchmarking}.
 In this section, we provide an overview of the naming conventions employed. Primarily we break down how the naming conventions of each environment correspond to the features of each scenario in the Level-Based Foraging (LBF) and Multi-Robot Warehouse (RWARE) environments.

Figure \ref{fig: all_envs_figures} illustrates a collection of ten scenarios, each corresponding to a specific task in the LBF and RWARE environments. The LBF scenarios are described in detail in Section \ref{LBF}, while the RWARE scenarios are explained in Section \ref{RWARE}.

\subsubsection{Level Based Foraging}  
\label{LBF}
\textbf{Naming Convention.} The scenarios in the LBF environment are named according to the following convention: 

\begin{center}
    $Foraging<obs>-<x\_size>x<y\_size>-<n\_agents>p-<food>f<force\_c>-v1$
\end{center}

Each field in the naming convention has specific options:
\begin{itemize}
\item <obs>: Denotes agent level of partial observability for all agents. If no value is given the agents can see as far as the grid is wide. 
\item $<x\_size>:$ Size of the grid along the horizontal axis.
\item $<y\_size>$: Size of the grid along the vertical axis.
\item $<n\_agents>$: Number of agents in the environment.
\item $<food>$: Number of food items in the environment. This is the total number of food that can spawn per episode. 
\item $<force\_c>$: Optional field indicating a forced cooperative task. It can be empty or set to "-coop" mode. In this mode, the levels of all the food items are intentionally set equal to the sum of the levels of all the agents involved. This implies that the successful acquisition of a food item requires a high degree of cooperation between the agents since no agent will be able to collect a food item by itself. 

As an example, an environment named "Foraging-2s-8x8-2p-2f-coop" has a sight range of "2s" implying that agents can view a 5x5 grid centred on themselves, a grid with horizontal and vertical size 8, contains 2 agents, 2 food objects and is set to a cooperative mode.
\end{itemize}

\textbf{Additional Level Based Foraging Scenarios} 
To gain insights into the interplay between individual agent levels, their impact on team performance, and individual contribution (Agent Importance value), we introduce additional LBF environments. These additional environments serve as a testing ground to study the reliability and scalability of the Agent Importance metric.

With the addition of these new scenarios, we focus on two distinct test features. Firstly, we assess the reliability of agent importance by making tasks that will always have three agents with levels of 1, 2, and 3, respectively. This allows us to compare the agent importance values to the predetermined agent levels to see whether they correspond. We also introduce two versions of the scenarios with fixed agent levels; one where the levels of food items are uniformly random values between 1 and 6 and another where the food levels are always 3. Secondly, we assess the scalability of agent importance w.r.t the number of agents in the setting. To do this we enlarge the grid size of the LBF scenarios to accommodate more agents up to a maximum tested number of 50.

\textbf{Used scenarios.} 
Our research experiments were carried out on a varied set of scenarios, and in all cases, agent positions and food positions as well as agent levels and food levels are randomly generated at each new environment episode. 


\begin{itemize}
  \item \textbf{Main scenarios}
    \begin{itemize}
    \item Figure \ref{fig: all_envs_figures}(a) \textbf{Foraging-2s-8x8-2p-2f-coop}: 8x8 grid, partial observability with sight=2, 2 agents, 2 food items, cooperative mode.
      \item Figure \ref{fig: all_envs_figures}(b) \textbf{Foraging-8x8-2p-2f-coop}: 8x8 grid, full observability, 2 agents, 2 food items, cooperative mode.
      \item Figure \ref{fig: all_envs_figures}(c) \textbf{Foraging-2s-10x10-3p-3f}: 10x10 grid, partial observability, 3 agents, 3 food items.
      \item Figure \ref{fig: all_envs_figures}(d) \textbf{Foraging-10x10-3p-3f}: 10x10 grid, full observability, 3 agents, 3 food items.
      \item Figure \ref{fig: all_envs_figures}(e) \textbf{Foraging-15x15-3p-5f}: 15x15 grid, full observability, 3 agents, 5 food items.
      \item Figure \ref{fig: all_envs_figures}(f) \textbf{Foraging-15x15-4p-3f: 15x15 grid}, full observability, 4 agents, 3 food items.
      \item Figure \ref{fig: all_envs_figures}(g) \textbf{Foraging-15x15-4p-5f: 15x15 grid}, full observability, 4 agents, 5 food items.
    \end{itemize}

  \item \textbf{Reliability Scenarios}
      \begin{itemize}
          \item \textbf{Foraging-15x15-3p-3f-det}: 15x15 grid, full observability, 3 agents, 3 food items. The food levels are fixed to always be 3.
          \item \textbf{Foraging-15x15-3p-3f-det-max-food-sum}: 15x5 grid, full observability, 3 agents, 3 food items. The food levels are uniformly random values between 1 and 6.
      \end{itemize}
      
  \item \textbf{Scalability Scenarios}
      \begin{itemize}
          \item \textbf{Foraging-5x5-2p-2f}: 5x5 grid, full observability, 2 agents, 2 food items.
          \item \textbf{Foraging-10x10-4p-4f}: 10x10 grid, full observability, 4 agents, 4 food items.
          \item \textbf{Foraging-15x15-10p-10f}: 15x15 grid, full observability, 10 agents, 10 food items.
          \item \textbf{Foraging-20x20-20p-20f}: 20x20 grid, full observability, 20 agents, 20 food items.
          \item \textbf{Foraging-25x25-50p-50f}: 25x25 grid, full observability, 50 agents, 50 food items.
      \end{itemize}
      
\end{itemize}

\subsubsection{Multi-Robot Warehouse} 
\label{RWARE}
\textbf{Naming Convention.} 
The scenarios in the RWARE environment are named according to the following convention: 
\begin{center}
    $rware-<size>-<num\_agents>ag<diff>-v1$
\end{center}
Each field in the naming convention has specific options:
\begin{itemize}
    \item $<size>$: Represents the size of the Warehouse (e.g., "tiny", "small", "medium",
    "large").
    \item $<num\_agents>$: Indicates the number of agents (1-20).
    \item $<diff>$: Optional field indicating the difficulty of the task (default: N requests for each of the N agents). 
\end{itemize}

\textbf{Used scenarios}
In this situation, the experiments in our study were carried out using three different scenarios of the RWARE environment.
In each of these scenarios, agents have a 3x3 observation grid centred on themselves, providing information on the location, rotation and surrounding configurations of other agents and shelves.
By default, the number of requested shelves is equal to the number of agents.

\begin{figure}
  \centering
  \begin{subfigure}[t]{0.01\textwidth}
        \scriptsize
        \textbf{(a)}
 \end{subfigure}
  \begin{subfigure}[t]{0.2\linewidth}
    \includegraphics[width=\linewidth, valign=t]{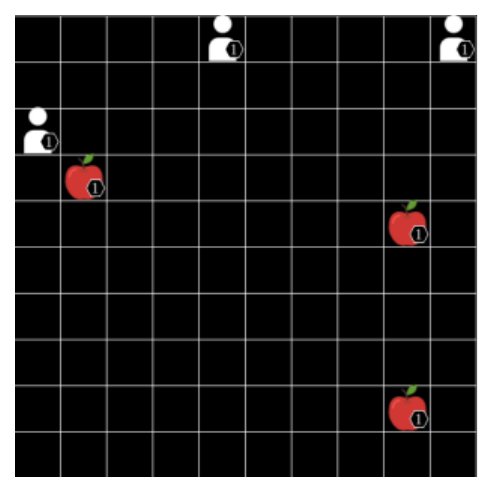}
  \end{subfigure}
  \begin{subfigure}[t]{0.01\textwidth}
        \scriptsize
        \textbf{(b)}
    \end{subfigure}
  \begin{subfigure}[t]{0.2\linewidth}
    \includegraphics[width=\linewidth, valign=t]{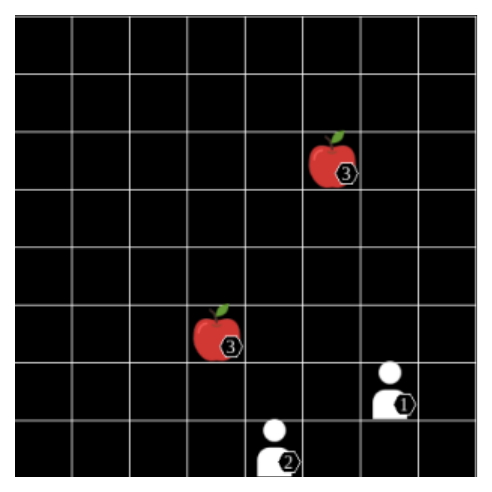}
  \end{subfigure}
  \begin{subfigure}[t]{0.01\textwidth}
        \scriptsize
        \textbf{(c)}
    \end{subfigure}
  \begin{subfigure}[t]{0.2\linewidth}
    \includegraphics[width=\linewidth, valign=t]{figures/appendix/envs_plots/Foraging_2s_10x10_3p_3_v2.pdf}
  \end{subfigure}
  \begin{subfigure}[t]{0.01\textwidth}
        \scriptsize
        \textbf{(d)}
    \end{subfigure}
  \begin{subfigure}[t]{0.2\linewidth}
    \includegraphics[width=\linewidth, valign=t]{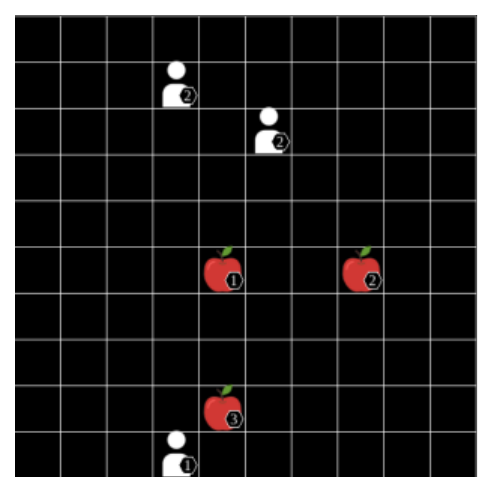}
  \end{subfigure}\\
\begin{subfigure}[t]{0.01\textwidth}
        \scriptsize
        \textbf{(e)}
    \end{subfigure}
  \begin{subfigure}[t]{0.2\linewidth}
    \includegraphics[width=\linewidth, valign=t]{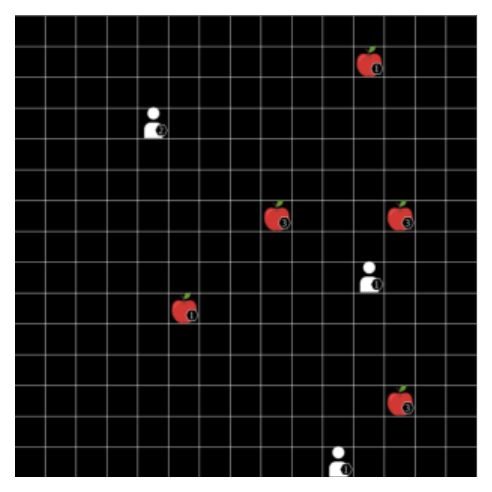}
  \end{subfigure}
  \begin{subfigure}[t]{0.01\textwidth}
        \scriptsize
        \textbf{(f)}
    \end{subfigure}
  \begin{subfigure}[t]{0.2\linewidth}
    \includegraphics[width=\linewidth, valign=t]{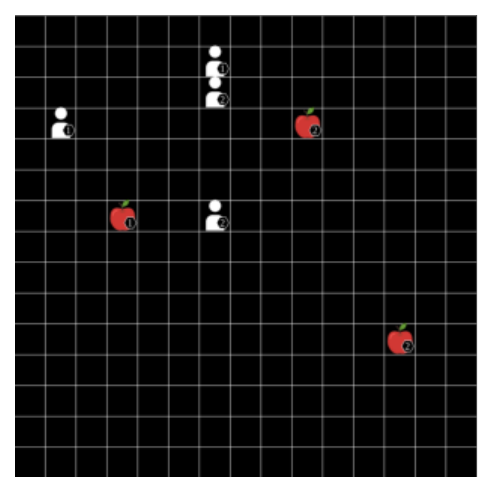}
  \end{subfigure}
    \begin{subfigure}[t]{0.01\textwidth}
        \scriptsize
        \textbf{(g)}
    \end{subfigure}
  \begin{subfigure}[t]{0.2\linewidth}
    \includegraphics[width=\linewidth, valign=t]{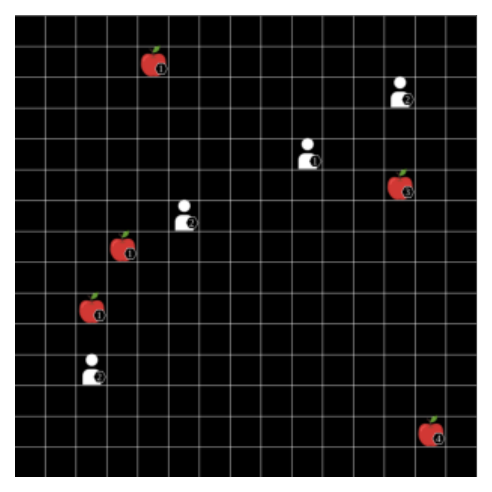}
  \end{subfigure}
  \begin{subfigure}[t]{0.01\textwidth}
        \scriptsize
        \textbf{(h)}
    \end{subfigure}
  \begin{subfigure}[t]{0.2\linewidth}
    \includegraphics[width=\linewidth, valign=t]{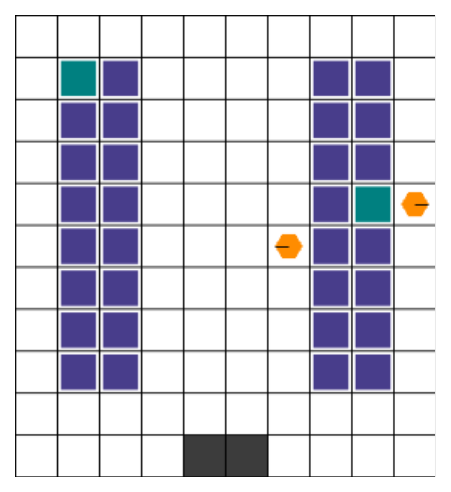}
  \end{subfigure}\\
\begin{subfigure}[t]{0.01\textwidth}
        \scriptsize
        \textbf{(i)}
    \end{subfigure}
  \begin{subfigure}[t]{0.2\linewidth}
    \includegraphics[width=\linewidth, valign=t]{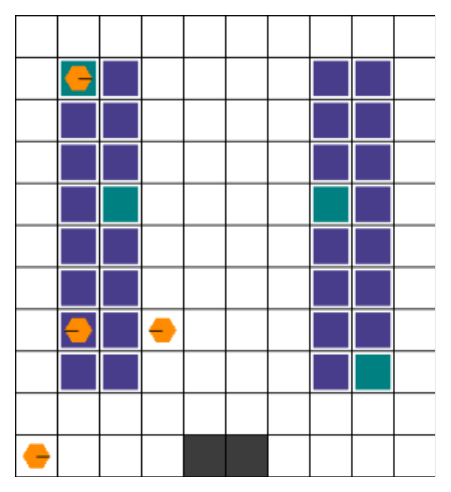}
  \end{subfigure}
  \begin{subfigure}[t]{0.01\textwidth}
        \scriptsize
        \textbf{(j)}
    \end{subfigure}
  \begin{subfigure}[t]{0.2\linewidth}
    \includegraphics[width=\linewidth,height=3.1cm, valign=t]{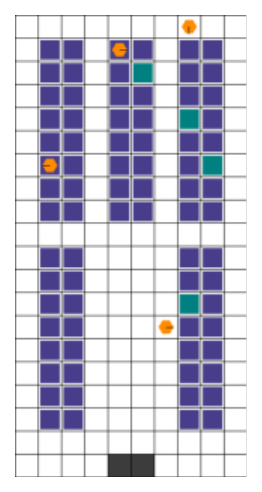}
  \end{subfigure}

  \caption{\textit{Illustration of the seven LBF and three RWARE tasks used for the main experiments.}}
  \label{fig: all_envs_figures}
\end{figure}

\begin{itemize}
    \item Figure \ref{fig: all_envs_figures}(h) \textbf{rware-tiny-2ag}: The tiny map is a grid world of 11x11 squares, partial observability, 2 agents.
    \item Figure \ref{fig: all_envs_figures}(i) \textbf{rware-tiny-4ag}: The tiny map is a grid world of 11x11 squares, partial observability, 4 agents.
    \item Figure \ref{fig: all_envs_figures}(j) \textbf{rware-small-4ag}: The small map is a grid world of 11x20 squares,
    partial observability, 4 agents.
\end{itemize}

\subsection{Algorithms Details}
\label{algo_details}
\textbf{Algorithms Overview}
In our analysis, we restrict ourselves to a limited set of algorithms from MARL literature. Our algorithm selection is done to cover Q-learning and policy gradient (PG) based methods in both the independent learner  (IL) and centralised training decentralised execution (CTDE) paradigms. We also investigate the effect of parameter sharing and non-parameter sharing on the performance of the algorithms.

\subsubsection{Q-learning}

For Q-learning-based methods we have selected, VDN and QMIX which fall into the paradigm of \textbf{CTDE} and IQL which is an \textbf{IL} method.

\textbf{IQL:} For Independent Q-Learning (IQL) \citep{tan1993multi}, each agent learns a policy based purely on their own egocentric experience in the training environment. This policy is parameterised by a Q-value network \citep{mnih2013playing}.

\textbf{VDN:} In Value-Decomposition Network (VDN) \citep{sunehag2017value}, IQL is extended through the use of value decomposition. Rather than learning purely from their own egocentric perspectives with each agent receiving the same reward, VDN formulates the joint Q value of the coalition as a linearly decomposed sum of the individual agent values. Each individual agent then updates its policy using the gradient flow based on a joint additive loss.

\textbf{QMIX}: \citep{rashid2018qmix} then extends VDN by broadening the range of reward functions that can be decomposed. To create a more complex attribution of the Q values, it makes use of a parameterised mixing network to perform the attribution. This mixing network takes the individual agent Q values as input and then policy updates are performed in an end-to-end manner where attribution is done using backpropagation. Qmix also allows the use of data augmentation by accommodating additional data at training time.

\subsubsection{Policy Gradients (PG):}

\textbf{IA2C:} Independent Advantage Actor-Critic (IA2C) is a variant of the A2C algorithm \citep{pmlr-v48-mniha16} applied to the multi-agent setting. IA2C trains agents using their own egocentric experiences in the training environment where each agent has their own critic and actor networks that approximate the optimal policy and state values.

\textbf{IPPO:} Independent Proximal Policy Optimisation (IPPO) is a variant of the PPO algorithm \citep{schulman2017proximal} applied to the multi-agent setting. PPO can be thought of as an improvement to A2C. It uses a surrogate objective which limits the change in the policy at each update step which allows PPO to iterate over the same trajectory of data multiple times without policy divergence. Otherwise, its architecture is the same as A2C.

\textbf{MAPPO \& MAA2C:} Multi-Agent Proximal Policy Optimisation (MAPPO) and Multi-Agent Advantage Actor-Critic (MAA2C) \citep{yu2022surprising} extend IPPO and IA2C to make use of a joint state value function. Instead of multiple per-agent critics, there is a single critic that learns the value of the joint state representation rather than the egocentric individual agent observations. MAA2C is also sometimes referred to as Central-V \citep{foerster2018counterfactual} because of this but, to prevent confusion, we use MAA2C. MAPPO also makes use of the same CTDE type architecture with a centralised critic. 

\textbf{Parameter Sharing vs Non-Parameter Sharing:}
To improve sample efficiency it is common to use \textbf{Parameter Sharing} (PS) in cooperative MARL. When PS is in use, all of the agents on a team share the same set of parameters for their neural networks (NN). In practice, this is equivalent to using a single neural network to represent all members of the team. Typically a one-hot agent ID is added to the local observation of each agent so that the NN can determine which agent to behave as. In some cases, using PS limits performance as agents tend to learn a smaller subset of roles. Alternatively, we can use \textbf{concurrent/non-parameter shared} learning where each agent is represented by a different set of parameters. Under this paradigm, we train each agent's parameters concurrently and maintain separate parameters for each individual agent.

\subsection{Evaluation Protocol}
\label{eval_section}
\subsubsection{Aggregation Metrics:}

\textbf{Median}: The median is the 50th percentile, representing the central point of the sorted raw data. Counts of the datapoints on either side of the median will thus be the same.

\textbf{IQM}: The interquartile mean (IQM) or midmean is a measure of central tendency evaluated based on the truncated mean of the interquartile range. It involves computing the mean over the values that fall within the interquartile range, which is the range between the 25th and 75th percentiles of the data.

\textbf{Optimality Gap}: The optimality gap is the difference between the aggregated value and the optimal value. It provides insight into the performance by quantifying the deviation from the best achievable outcome.

\textbf{Absolute Metric}: The Absolute Metric represents the average performance achieved by the best policy obtained throughout the entire learning process. It's computed by evaluating the algorithm over a number of independent evaluation episodes that is 10 times greater than the original number used during training. 

\subsubsection{Explanation of plots used:}

\textbf{Sample efficiency} The concept of sample efficiency is used to evaluate how effectively an algorithm improves its performance on a specific measure in relation to the amount of data it samples during the training process. These curves are generated by calculating the normalised average performance at each evaluation interval.

\textbf{Performance Profiles}: Performance profiles plot the probability that the normalised return of an algorithm is greater than some fraction of a predetermined value. From these plots, we can see the likelihood of algorithms reaching an optimal score and compare their relative performance at different points.

\textbf{Probability of improvement}: The probability of improvement are plots that indicate the probability that algorithm X has superior performance than algorithm Y with a low score indicating that algorithm Y is likely to be better than algorithm X and vice versa for a high score.

\subsubsection{Experimental Hyperparameters}
In our analysis, we sought to conduct comprehensive experiments in various environments using different algorithms. To ensure reliable and consistent results, it is important carefully select and optimize the hyperparameters for each algorithm in each environment.

To this end, we used the optimized hyperparameters from \cite{papoudakis2021benchmarking}, where the parameters of each algorithm were chosen based on a hyperparameter sweep for a single scenario of each environment and then reused across all other scenarios for the same settings. The choice of the set of hyperparameters is done by selecting the one with the highest evaluation score averaged over three seeds.

Tables \ref{tab: q_learning_hyperparams} and \ref{tab: pg_hyperparams} provide a summary of the shared hyperparameters used in the Q-learning and policy gradient algorithms, respectively. On the other hand, Tables \ref{tab: iql_hyperparams}, \ref{tab: vdn_hyperparams}, and \ref{tab: qmix_hyperparams} specify the algorithm-specific hyperparameters for each Q-learning algorithm, namely IQL, VDN, and QMIX, respectively. Similarly, Tables \ref{tab: mappo_hyperparams} and \ref{tab: maa2c_hyperparams} present the specific hyperparameter settings for the policy-gradient algorithms, namely MAPPO and MAA2C, respectively. 

These aforementioned tables offer an overview of the hyperparameters utilized in each environment, encompassing the applicable algorithms in both parameter-sharing and non-parameter-sharing scenarios.

\begin{table}
  \centering
  \caption{Shared hyperparameters for Q-learning algorithms with and  without parameter sharing}
  \label{tab: q_learning_hyperparams}
  \resizebox{0.5\textwidth}{!}
  {\begin{tabular}{ccccp{2cm}}
    \toprule
    & \multicolumn{2}{c}{\textbf{Parameter Sharing}} & \multicolumn{2}{c}{\textbf{Non-Parameter Sharing}} \\
    \cmidrule(lr){2-3} \cmidrule(l){4-5}
    & \textbf{LBF} & \textbf{RWARE} & \textbf{LBF} & \textbf{RWARE} \\
    \midrule
    Optimizer &Adam & Adam & Adam & Adam \\ 
    Maximum gradient norm & 10 & 10 & 10 & 10\\
    Reward standardisation & True & True  & True & True\\
    Network type & GRU & FC & GRU & FC\\
    Discount factor & 0.99 & 0.99 & 0.99 & 0.99\\
    $\epsilon$ schedule steps & 2e6 & 5e4 & 5e4 & 5e4\\
    $\epsilon$ schedule minimum & 0.05 & 0.05 & 0.05 & 0.05\\
    Batch size & 32 & 32 & 32 & 32\\
    Replay buffer size & 5000 & 5000 & 5000 & 5000\\
    Parallel workers & 1 & 1 & 1 & 1\\
    \bottomrule
  \end{tabular}}
\end{table}

\begin{table}
  \centering
  \resizebox{0.5\textwidth}{!}{
  \begin{tabular}{ccccp{2cm}}
    \toprule
    & \multicolumn{2}{c}{\textbf{Parameter Sharing}} & \multicolumn{2}{c}{\textbf{Non-Parameter Sharing}} \\
    \cmidrule(lr){2-3} \cmidrule(l){4-5}
    & \textbf{LBF} & \textbf{RWARE} & \textbf{LBF} & \textbf{RWARE} \\
    \midrule
    Hidden dimension & 128 & 64 & 64 & 64 \\
    Learning rate & 0.0003 & 0.0005 & 0.0003 & 0.0005 \\
    Reward standardisation & True & True  & True & True\\
    Network type & GRU & FC & GRU & FC\\
    Evaluation epsilon & 0.05 & 0.05 & 0.05 & 0.05 \\
    Target update & 200(hard) & 0.01(soft) & 200(hard) & 0.01 (soft) \\
    \bottomrule
  \end{tabular}}
  \caption{Shared hyperparameters for IQL with and  without parameter sharing}
  \label{tab: iql_hyperparams}
\end{table}

\begin{table}
  \centering
   \resizebox{0.5\textwidth}{!}{
 \begin{tabular}{ccccp{2cm}}
    \toprule
    & \multicolumn{2}{c}{\textbf{Parameter Sharing}} & \multicolumn{2}{c}{\textbf{Non-Parameter Sharing}} \\
    \cmidrule(lr){2-3} \cmidrule(l){4-5}
    & \textbf{LBF} & \textbf{RWARE} & \textbf{LBF} & \textbf{RWARE} \\
    \midrule
    Hidden dimension & 128 & 64 & 64 & 64 \\
    Learning rate & 0.0003 & 0.0005 & 0.0001 & 0.0005 \\
    Reward standardisation & True & True  & True & True\\
    Network type & GRU & FC & GRU & FC\\
    Evaluation epsilon & 0.0 & 0.05 & 0.05 & 0.05 \\
    Target update & 0.01(\textbf{soft}) & 0.01(\textbf{soft}) & 200(\textbf{hard}) & 0.01 (\textbf{soft}) \\
    \bottomrule
  \end{tabular}}
  \caption{Hyperparameters for VDN with and  without parameter sharing}
  \label{tab: vdn_hyperparams}
\end{table}

\begin{table}
  \centering
  \resizebox{0.5\textwidth}{!}{
  \begin{tabular}{ccccp{2cm}}
    \toprule
    & \multicolumn{2}{c}{\textbf{Parameter Sharing}} & \multicolumn{2}{c}{\textbf{Non-Parameter Sharing}} \\
    \cmidrule(lr){2-3} \cmidrule(l){4-5}
    & \textbf{LBF} & \textbf{RWARE} & \textbf{LBF} & \textbf{RWARE} \\
    \midrule
    Hidden dimension & 64 & 64 & 64 & 64 \\
    Network type & GRU & FC & GRU & FC\\
    Mixing network size & 32 & 32& 32 & 32\\
    Mixing network type & FC &FC & FC & FC\\
    Mixing network activation & ReLU & ReLU & ReLU & ReLU\\
    Hypernetwork size & 64 & 64 & 64 & 64\\
    Hypernetwork activation & ReLU & ReLU & ReLU & ReLU\\
    Hypernetworks layers& 2&2&2&2\\
    Learning rate & 0.0003 & 0.0005 & 0.0001 & 0.0003 \\
    Reward standardisation & True & True  & True & True\\
    Evaluation epsilon & 0.05 & 0.05 & 0.05 & 0.05 \\
    Target update & 0.01(\textbf{soft}) & 0.01(\textbf{soft}) & 0.01 (\textbf{soft}) & 0.01 (\textbf{soft}) \\
    \bottomrule
  \end{tabular}}
  \caption{Hyperparameters for QMIX with and  without parameter sharing}
  \label{tab: qmix_hyperparams}
\end{table}


\begin{table}
  \centering
  \caption{Shared hyperparameters for Policy-based algorithms with and without parameter sharing}
  \label{tab: pg_hyperparams}
  \resizebox{0.5\textwidth}{!}{
    \begin{tabular}{ccccp{2cm}}
      \toprule
      & \multicolumn{2}{c}{\textbf{Parameter Sharing}} & \multicolumn{2}{c}{\textbf{Non-Parameter Sharing}} \\
      \cmidrule(lr){2-3} \cmidrule(l){4-5}
      & \textbf{LBF} & \textbf{RWARE} & \textbf{LBF} & \textbf{RWARE} \\
      \midrule
      Optimizer &Adam & Adam & Adam & Adam \\ 
      Maximum gradient norm & 10 & 10 & 10 & 10\\
      Discount factor & 0.99 & 0.99 & 0.99 & 0.99\\
      Entropy coefficient & 0.001 & 0.001 & 0.001 & 0.001\\
      Batch size & 10 & 10 & 10 & 10\\
      Replay buffer size & 10 & 10 & 10 & 10\\
      Parallel workers & 10 & 10 & 10 & 10\\
      \bottomrule
    \end{tabular}
  }
\end{table}

\begin{table}
  \centering
  \resizebox{0.5\textwidth}{!}{
  \begin{tabular}{ccccp{2cm}}
    \toprule
    & \multicolumn{2}{c}{\textbf{Parameter Sharing}} & \multicolumn{2}{c}{\textbf{Non-Parameter Sharing}} \\
    \cmidrule(lr){2-3} \cmidrule(l){4-5}
    & \textbf{LBF} & \textbf{RWARE} & \textbf{LBF} & \textbf{RWARE} \\
    \midrule
    Hidden dimension & 128 & 128 & 128 & 128 \\
    Learning rate & 0.0003 & 0.0005 & 0.0001 & 0.0005 \\
    Reward standardisation & False & False  & False & False\\
    Network type & FC & FC & FC & FC\\
    Evaluation epsilon & 0.05 & 0.05 & 0.05 & 0.05 \\
    Epsilon clip & 0.2 & 0.2 & 0.2 & 0.2 \\
    Epochs & 4 & 4 & 4 & 4 \\
    Target update & 0.01(soft) & 0.01(soft) & 200 (hard) & 0.01 (soft) \\
    n-step & 5 & 10 & 10 & 10\\
    \bottomrule
  \end{tabular}}
  \caption{Hyperparameters for MAPPO with and  without parameter sharing}
  \label{tab: mappo_hyperparams}
\end{table}

\begin{table}
  \centering
  \resizebox{0.5\textwidth}{!}{%
  \begin{tabular}{ccccp{2cm}}
    \toprule
    & \multicolumn{2}{c}{\textbf{Parameter Sharing}} & \multicolumn{2}{c}{\textbf{Non-Parameter Sharing}} \\
    \cmidrule(lr){2-3} \cmidrule(l){4-5}
    & \textbf{LBF} & \textbf{RWARE} & \textbf{LBF} & \textbf{RWARE} \\
    \midrule
    Hidden dimension & 128 & 64 & 128 & 64 \\
    Learning rate & 0.0005 & 0.0005 & 0.0005 & 0.0005 \\
    Reward standardisation & True & True  & True & True\\
    Network type & GRU & FC & GRU & FC\\
    Evaluation epsilon & 0.01 & 0.01 & 0.01 & 0.01 \\
    Target update & 0.01(soft) & 0.01(soft) & 0.01 (soft) & 0.01 (soft) \\
    n-step & 10 & 5 & 5 & 5\\
    \bottomrule
  \end{tabular}%
  }
    \caption{Hyperparameters for MAA2C with and  without parameter sharing}
  \label{tab: maa2c_hyperparams}
\end{table}

\section{Main experiment results}

In this section, we present additional plots that complement the figures presented in the main paper. These plots provide a more comprehensive visualization of the experimental results and support the analysis presented in the paper.

\begin{figure}
  \centering
    \begin{subfigure} {1\linewidth}
    \includegraphics[width=\linewidth]{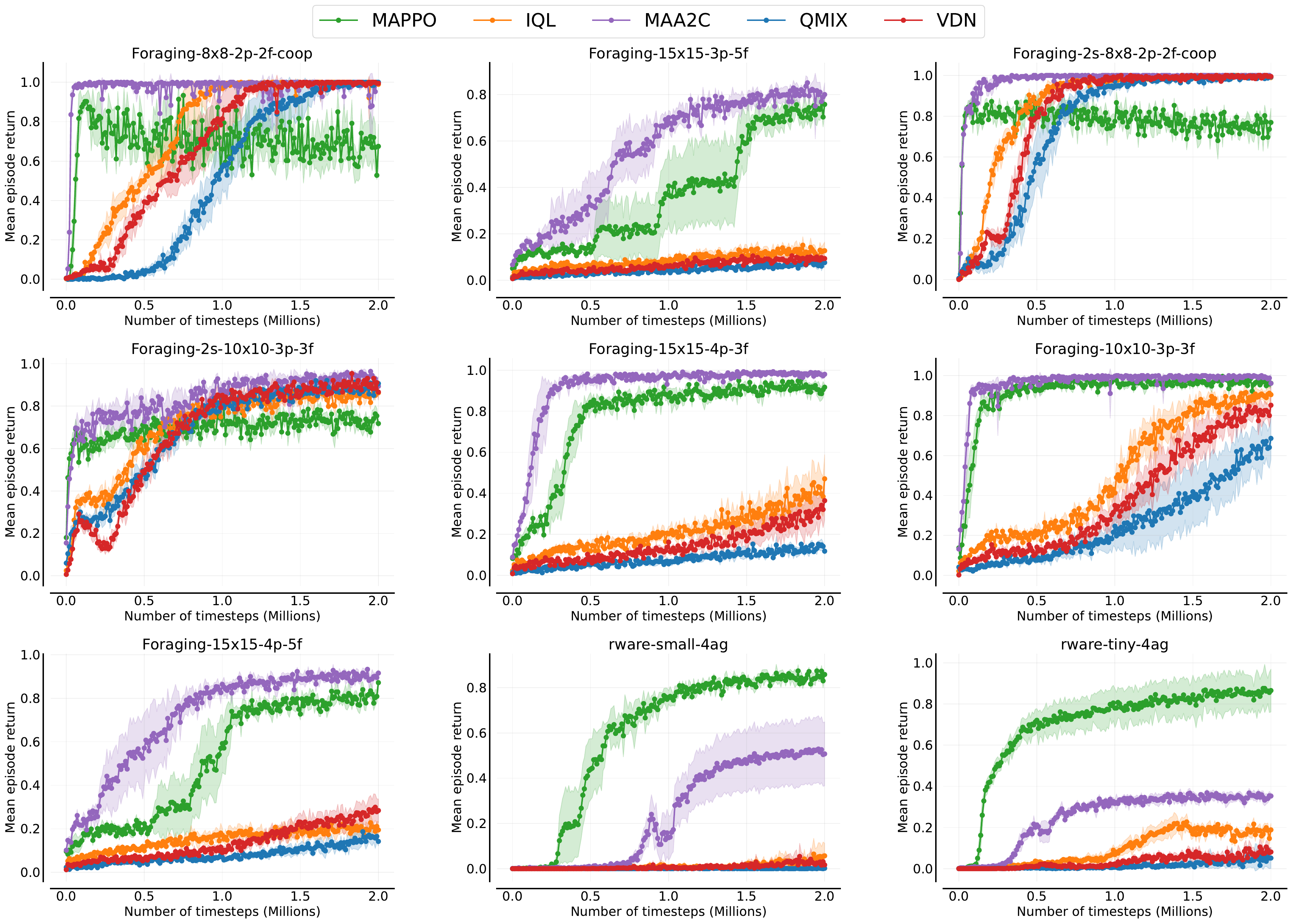} 
    \end{subfigure}
    
    \begin{subfigure} {0.4\linewidth}
    \includegraphics[width=\linewidth]{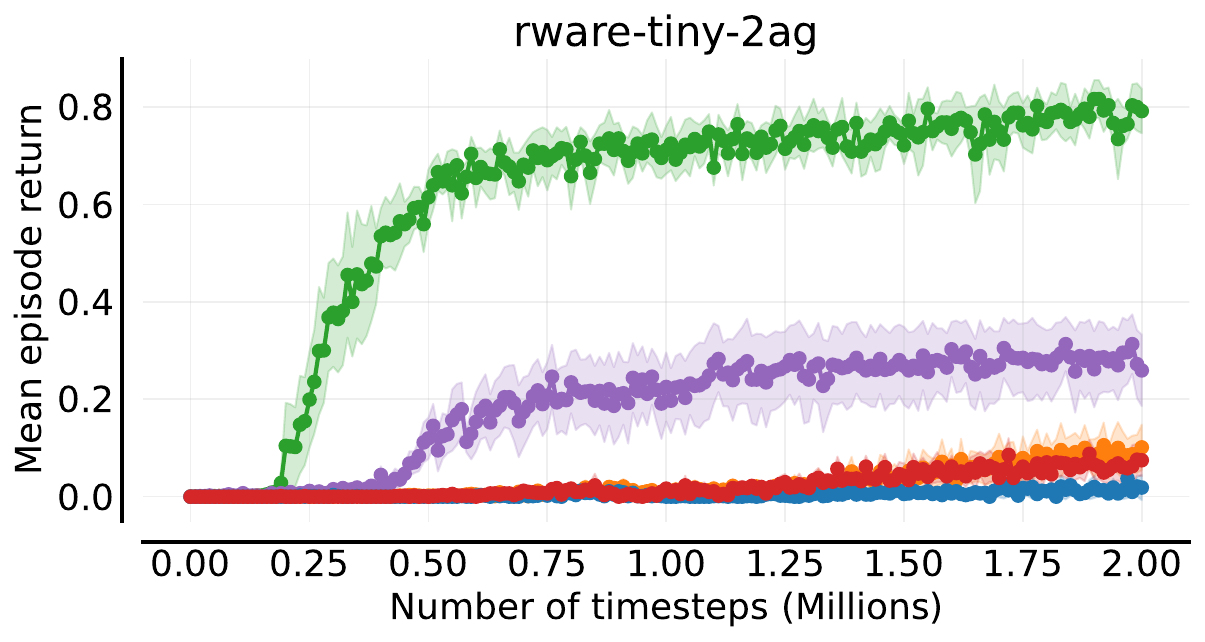}
    \end{subfigure}
  \caption{Mean episode returns for all algorithms with parameter sharing in seven LBF scenarios and three RWARE scenarios, with the mean and 95\% confidence intervals over 10 distinct seeds.}
  \label{fig: shared_all_tasks}
  
\end{figure}

\subsection{Parameter sharing Experiments}
\label{sec: param_share_exp}
In Figure \ref{fig: shared_all_tasks}, we can observe the performance of the aforementioned algorithms in the seven LBF tasks and the 3 RWARE tasks where we recorded the results of Mean episode returns with the mean and 95\% confidence intervals over 10 distinct seeds in the 201 evaluations.

In contrast, Tables \ref{tab: stat_lbf_s} and \ref{tab: stat_rware_s} present the comprehensive tabulated results of algorithms performance in the LBF and RWARE environments. These tables showcase the utilization of various aggregation metrics, including Median, IQM, Mean, and the Optimality gap, along with their corresponding 95\% confidence intervals. The confidence intervals are estimated using the percentile bootstrap with stratified sampling.


\begin{table}
  \centering
  \caption{Normalized Episode Return: Aggregated Scores with 95\% Confidence Intervals in the LBF Environment with Parameter Sharing}
  \label{tab: stat_lbf_s}
  \resizebox{0.5\textwidth}{!}
  {\begin{tabular}{lccccc}
    \toprule
      & \textbf{MAPPO} & \textbf{IQL} & \textbf{MAA2C} & \textbf{QMIX} & \textbf{VDN} \\
    \midrule
    Median & 0.84 $\pm 0.03$ & 0.87 $\pm 0.01$ & 0.98 $\pm 0.01$ & 0.67 $\pm 0.12$ & 0.78 $\pm 0.06$ \\
    IQM & 0.85 $\pm 0.01$ & 0.71 $\pm 0.04$ & 0.97 $\pm 0.01$ & 0.58 $\pm 0.03$ & 0.67 $\pm 0.04$ \\
    Mean & 0.85 $\pm 0.01$ & 0.64 $\pm 0.02$ & 0.95 $\pm 0.01$ & 0.56 $\pm 0.02$ & 0.61 $\pm 0.02$ \\
    Optimality Gap & 0.15 $\pm 0.01$ & 0.36 $\pm 0.02$ & 0.05 $\pm 0.01$ & 0.44 $\pm 0.02$ & 0.39 $\pm 0.02$ \\
    \bottomrule
  \end{tabular}}
\end{table}

\begin{table}
  \centering
  \caption{Normalized Episode Return: Aggregated Scores with 95\% Confidence Intervals in the RWARE Environment with Parameter Sharing}
  \label{tab: stat_rware_s}
 \resizebox{0.5\textwidth}{!}
  {\begin{tabular}{lccccc}
    \toprule
      & \textbf{MAPPO} & \textbf{IQL} & \textbf{MAA2C} & \textbf{QMIX} & \textbf{VDN} \\
    \midrule
    Median & 0.85 $\pm 0.04$ & 0.11 $\pm 0.06$ & 0.35 $\pm 0.03$ & 0.03 $\pm 0.02$ & 0.07 $\pm 0.03$ \\
    IQM & 0.85 $\pm 0.03$ & 0.12 $\pm 0.05$ & 0.39 $\pm 0.05$ & 0.0 $\pm 0.01$ & 0.06 $\pm 0.03$ \\
    Mean & 0.83 $\pm 0.04$ & 0.13 $\pm 0.04$ & 0.41 $\pm 0.04$ & 0.02 $\pm 0.02$ & 0.07 $\pm 0.02$ \\
    Optimality Gap & 0.17 $\pm 0.04$ & 0.87 $\pm 0.04$ & 0.59 $\pm 0.04$ & 0.98 $\pm 0.02$ & 0.93 $\pm 0.02$ \\
    \bottomrule
  \end{tabular}}
\end{table}

\subsection{Non-Parameter sharing Experiments}
\begin{figure}
    \centering
    \includegraphics[width=0.4\textwidth]{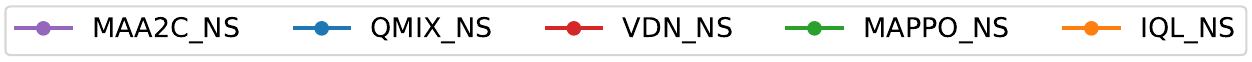}
    \includegraphics[width=0.2\textwidth]{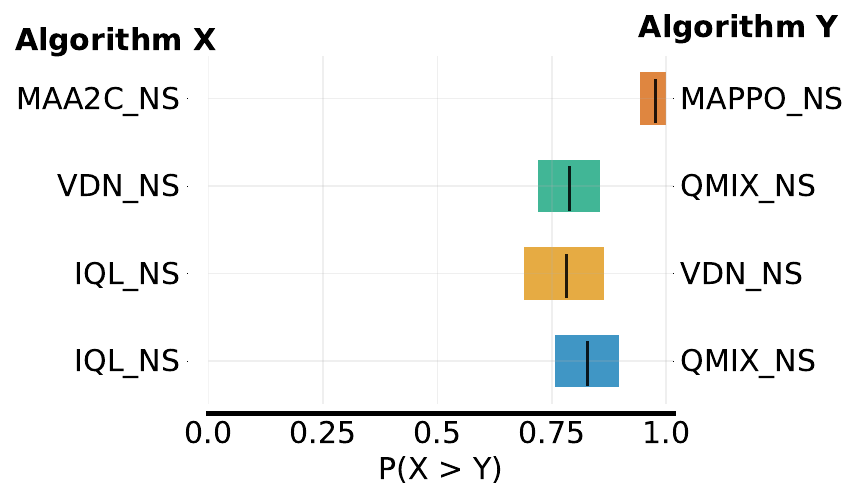}
    \includegraphics[width=0.2\textwidth]{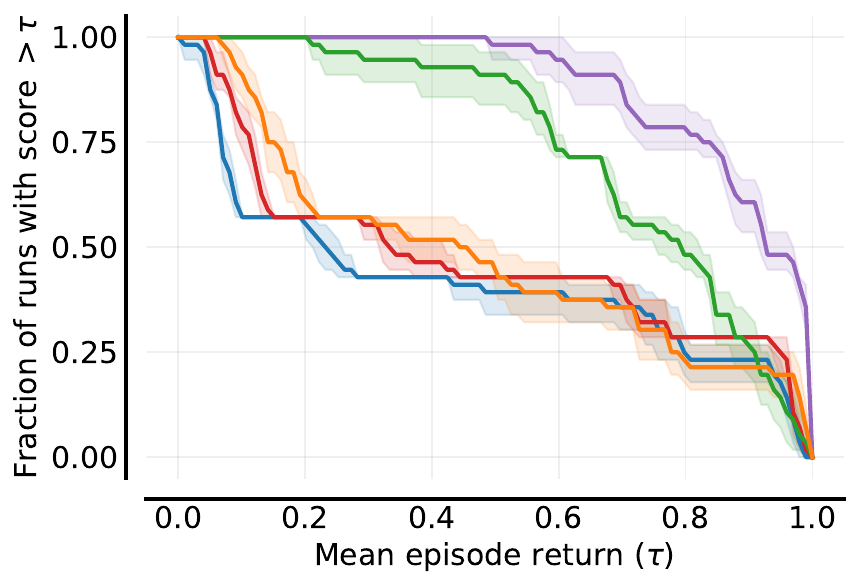}
    \includegraphics[width=0.35\textwidth]{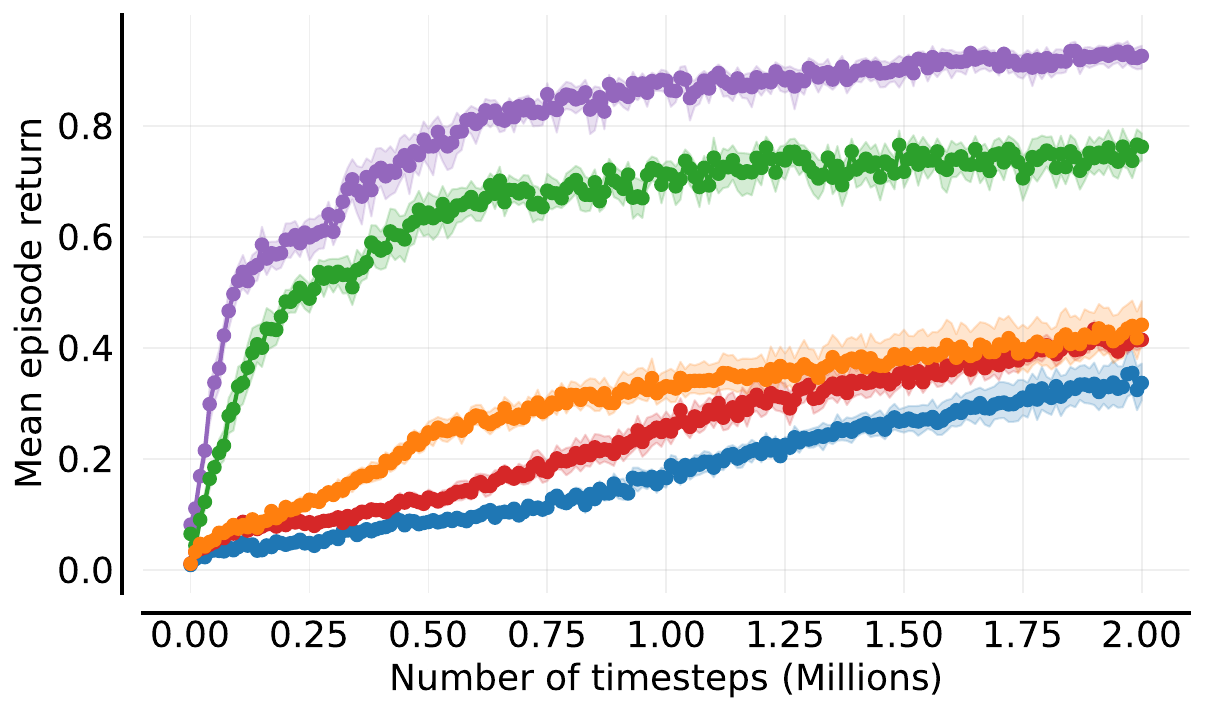}
    \includegraphics[width=0.2\textwidth]{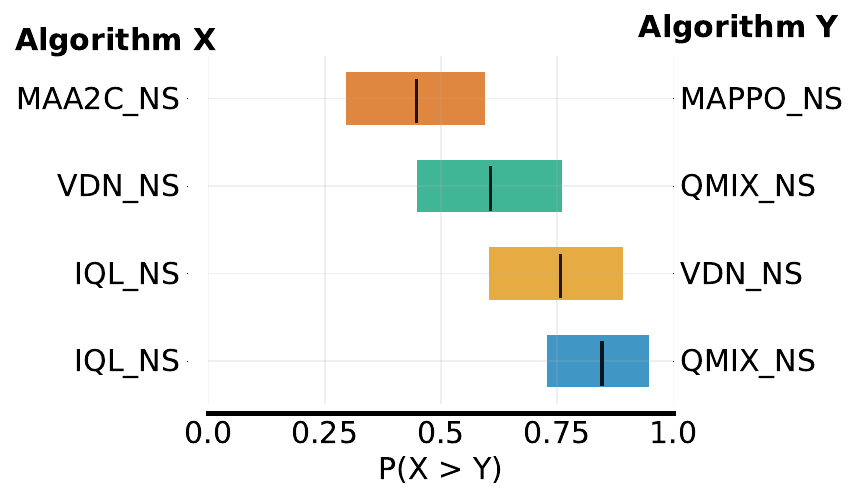}
    \includegraphics[width=0.2\textwidth]{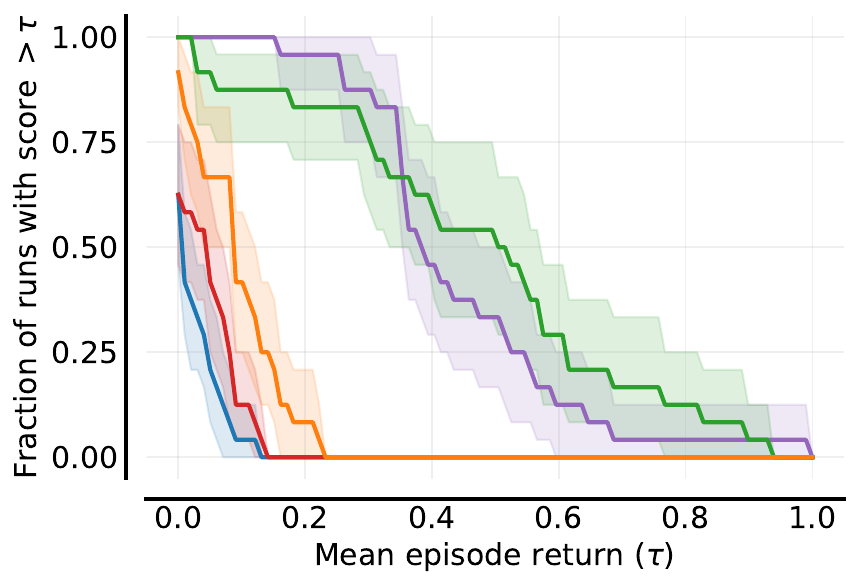}
    \includegraphics[width=0.35\textwidth]{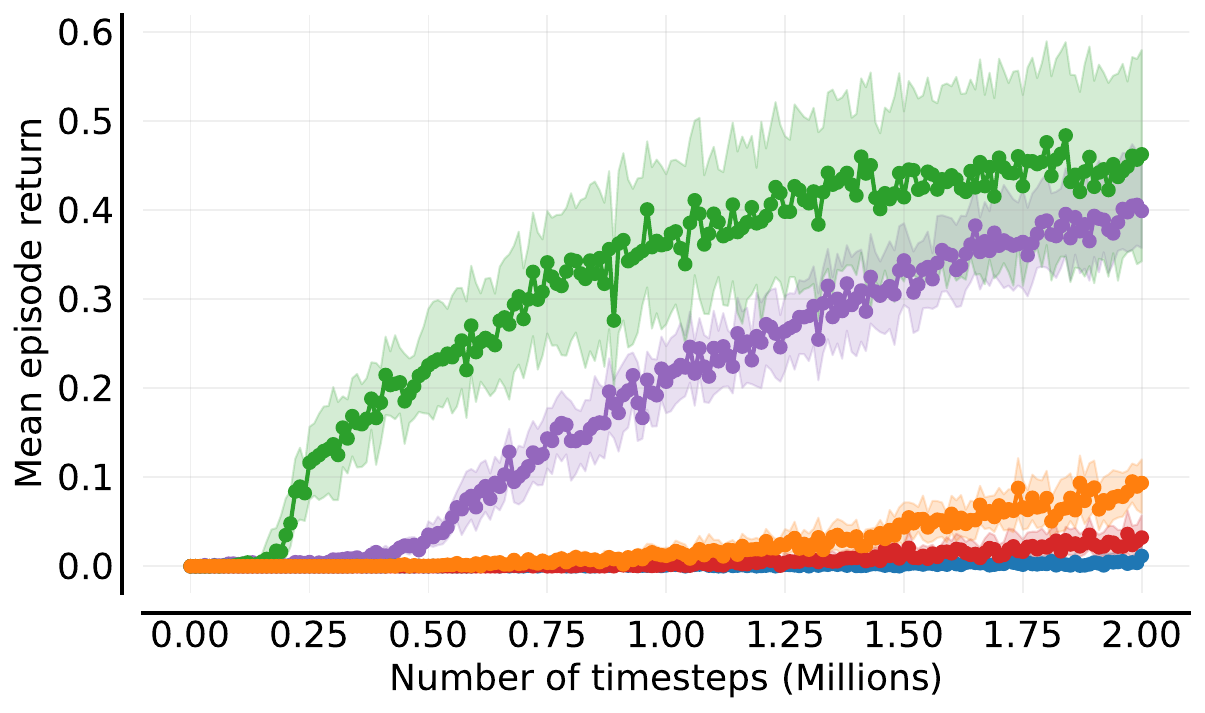}
    \caption{\textit{Results from running the same experimental hyperparameters on the same tasks as \cite{papoudakis2021benchmarking} including the probability of improvement, performance profiles and sample efficiency curves}. \textbf{Top row:} Performance of algorithms on 7 LBF tasks. \textbf{Bottom row:} Performance of all algorithms on 3 RWARE tasks.}
    \label{fig: benchmarking_paper_redo_non_shared}
\end{figure}

Similar to the replication of experiments conducted in the parameter sharing case, we also conducted a replication of the outcomes when agents do not share learning parameters. The outcomes of these experiments are illustrated in Figure \ref{fig: benchmarking_paper_redo_non_shared}. 

By examining the results for both the LBF and RWARE environments, we aimed to compare and contrast the performance of algorithms under these distinct conditions. The replicated experiments serve to validate and complement the findings presented in Figure \ref{fig: benchmarking_paper_redo_non_shared}, offering an understanding of the impact of parameter sharing on algorithm performance.

\begin{figure}
  \centering
  \begin{subfigure} {1\linewidth}
    \includegraphics[width=\linewidth]{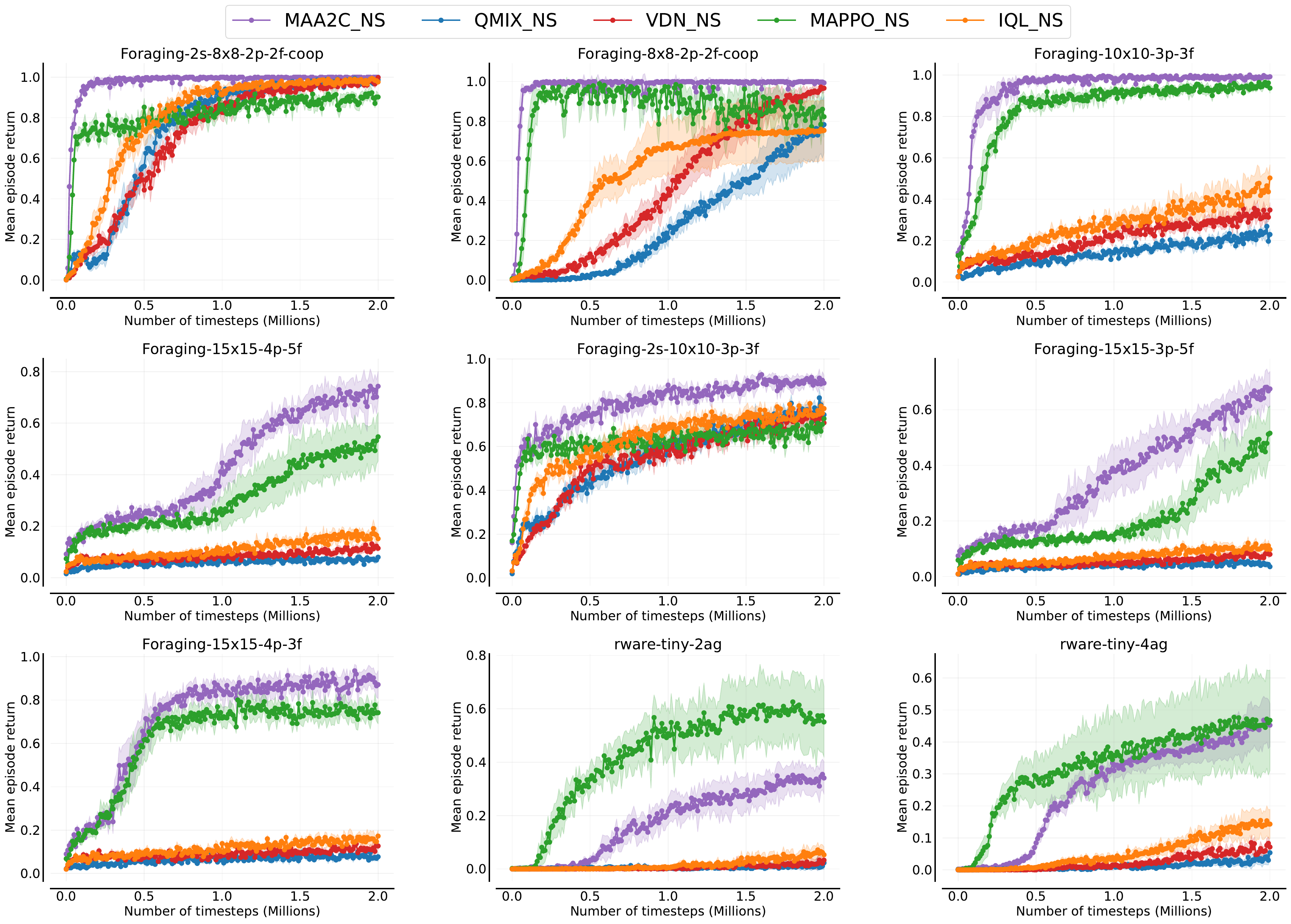} 
    \end{subfigure}
    
    \begin{subfigure} {0.4\linewidth}
    \includegraphics[width=\linewidth]{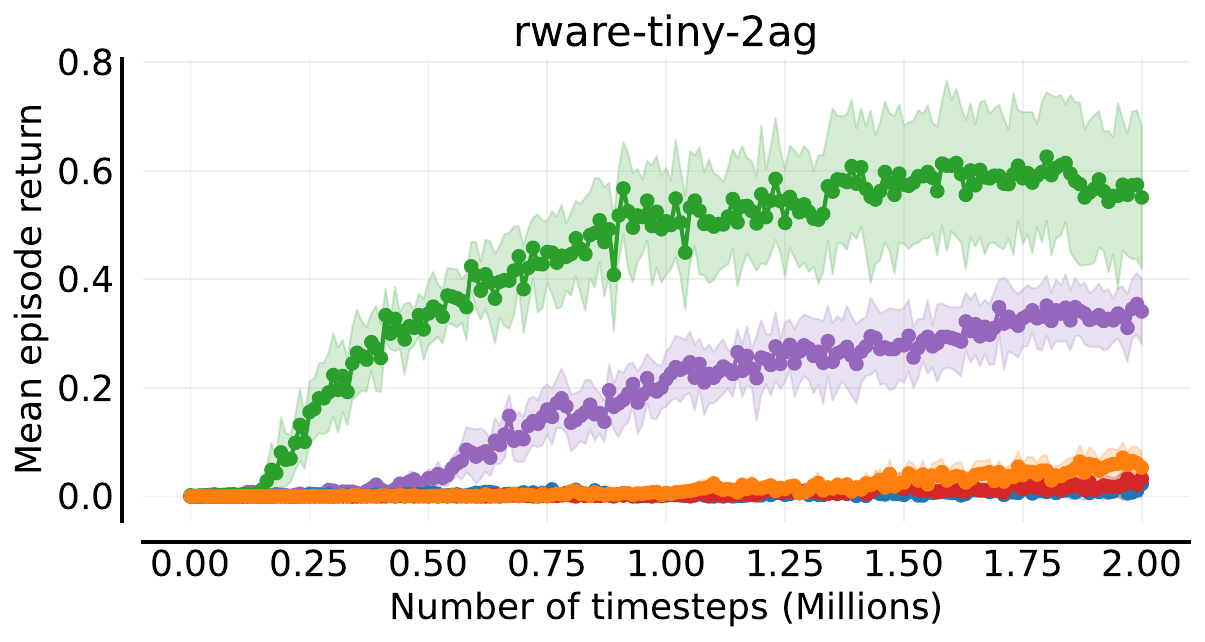}
    \end{subfigure}
  \caption{Mean episode returns for all algorithms without parameter sharing in seven LBF scenarios and three RWARE scenarios, with the mean and 95\% confidence intervals over 10 distinct seeds.}
  \label{fig: non_shared_all_tasks}
\end{figure}

In addition, in Figure \ref{fig: non_shared_all_tasks} we also evaluated the performance of various algorithms without parameter sharing across a range of scenarios. For the LBF environment, we considered seven different scenarios, each presenting unique challenges and variations in agent and food levels. Similarly, for the RWARE environment, we explored three distinct scenarios, encompassing different warehouse sizes and numbers of agents.

Similarly to the parameter sharing case discussed in Section \ref{sec: param_share_exp}, the results of algorithm performance in the LBF and RWARE environments are presented in Tables \ref{tab: stat_lbf_s} and \ref{tab: stat_rware_s}. 

\begin{table}
  \centering
  \caption{Normalized Episode Return: Aggregated Scores with 95\% Confidence Intervals in the LBF Environment without Parameter Sharing}
  \label{tab: stat_lbf_ns}
\resizebox{0.5\textwidth}{!}
  {\begin{tabular}{lccccc}
    \toprule
      & \textbf{MAPPO} & \textbf{IQL} & \textbf{MAA2C} & \textbf{QMIX} & \textbf{VDN} \\
    \midrule
    Median & 0.91 $\pm 0.03$ & 0.23 $\pm 0.02$ & 0.35 $\pm 0.03$ & 0.78 $\pm 0.03$ & 0.46 $\pm 0.08$ \\
    IQM & 0.93 $\pm 0.01$ & 0.34 $\pm 0.03$ & 0.42 $\pm 0.01$ & 0.77 $\pm 0.01$ & 0.43 $\pm 0.04$ \\
    Mean & 0.89 $\pm 0.01$ & 0.42 $\pm 0.02$ & 0.47 $\pm 0.01$ & 0.74 $\pm 0.02$ & 0.48 $\pm 0.03$ \\
    Optimality Gap & 0.11 $\pm 0.01$ & 0.58 $\pm 0.02$ & 0.53 $\pm 0.01$ & 0.26 $\pm 0.02$ & 0.52 $\pm 0.03$ \\
    \bottomrule
  \end{tabular}}
\end{table}

\begin{table}
  \centering
  \caption{Normalized Episode Return: Aggregated Scores with 95\% Confidence Intervals in the LBF Environment without Parameter Sharing}
  \label{tab: stat_rware_ns}
  \resizebox{0.5\textwidth}{!}
  {\begin{tabular}{lccccc}
    \toprule
      & \textbf{MAPPO} & \textbf{IQL} & \textbf{MAA2C} & \textbf{QMIX} & \textbf{VDN} \\
    \midrule
    Median & 0.45 $\pm 0.08$ & 0.02 $\pm 0.02$ & 0.03 $\pm 0.02$ & 0.48 $\pm 0.14$ & 0.08 $\pm 0.03$ \\
    IQM & 0.41 $\pm 0.05$ & 0.01 $\pm 0.01$ & 0.04 $\pm 0.02$ & 0.47 $\pm 0.1$ & 0.09 $\pm 0.03$ \\
    Mean & 0.44 $\pm 0.06$ & 0.02 $\pm 0.01$ & 0.04 $\pm 0.01$ & 0.47 $\pm 0.09$ & 0.09 $\pm 0.02$ \\
    Optimality Gap & 0.56 $\pm 0.06$ & 0.98 $\pm 0.01$ & 0.96 $\pm 0.01$ & 0.53 $\pm 0.09$ & 0.91 $\pm 0.02$ \\
    \bottomrule
  \end{tabular}}
\end{table}

\subsection{Additional results on RWARE} \label{additional_RWARE}
\begin{figure}
  \centering
  \begin{subfigure}[t]{0.4\linewidth}
    \includegraphics[width=\linewidth, valign=t]{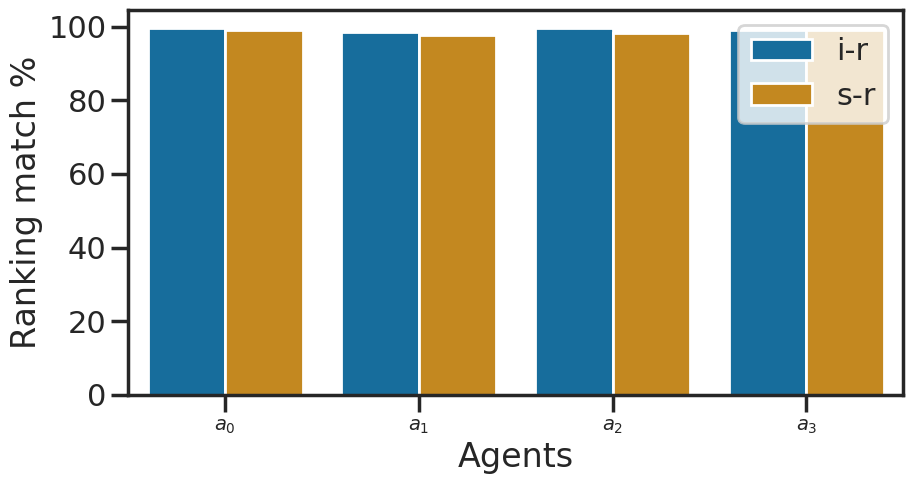}
  \end{subfigure}
  \begin{subfigure}[t]{0.4\linewidth}
    \includegraphics[width=\linewidth, valign=t]{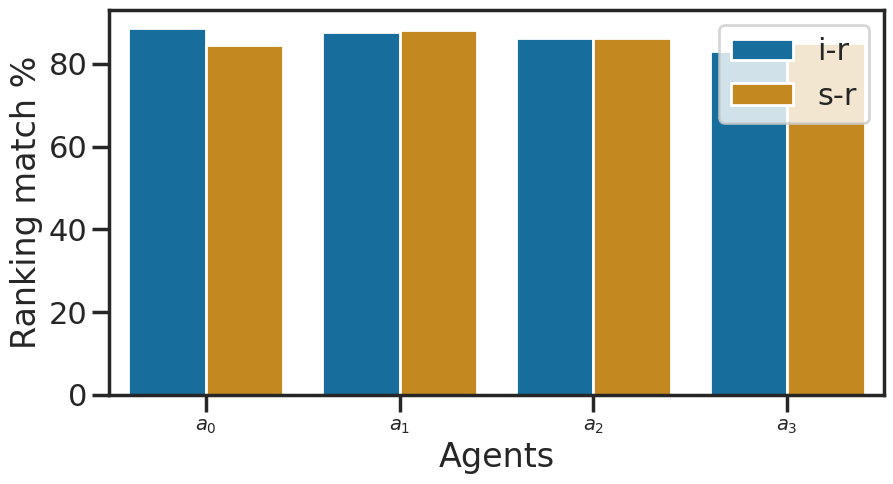}
  \end{subfigure}
  \begin{subfigure}[t]{0.3\linewidth}
    \includegraphics[width=\linewidth, valign=t]{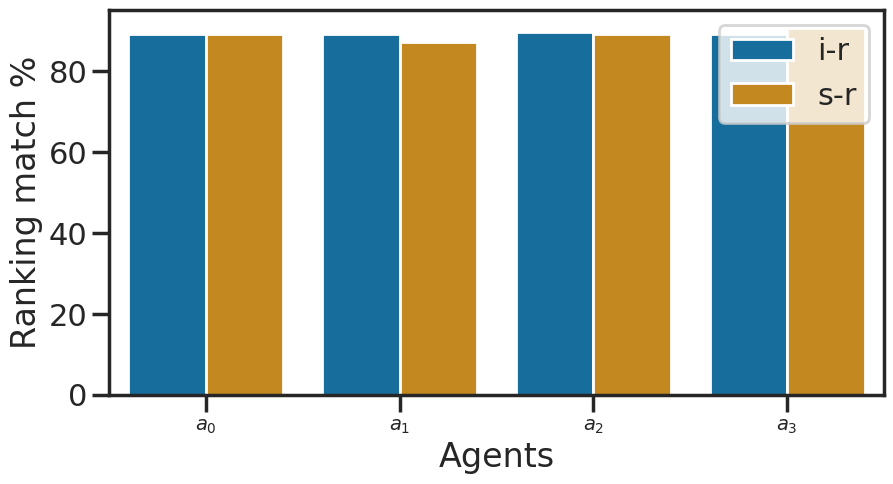}
  \end{subfigure}
  \begin{subfigure}[t]{0.3\linewidth}
    \includegraphics[width=\linewidth, valign=t]{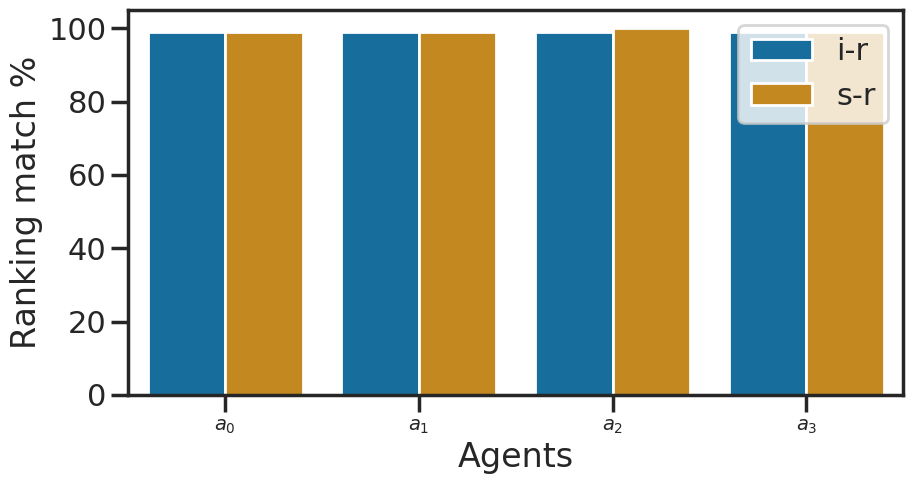}
  \end{subfigure}
   \begin{subfigure}[t]{0.35\linewidth}
    \includegraphics[width=\linewidth, valign=t]{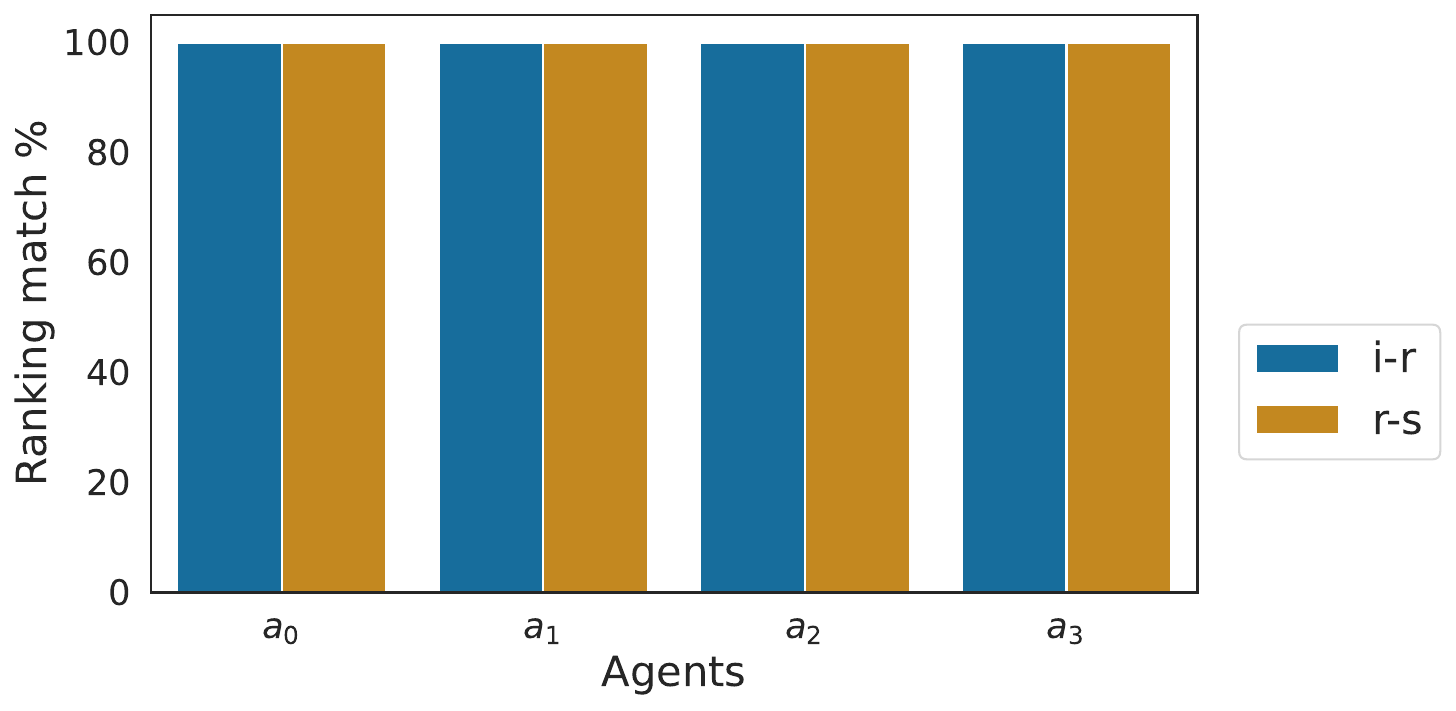}
  \end{subfigure}
   \caption[Fake caption]{Comparison of Shapley value and agent importance rankings to individual rewards for IQL, MAA2C, MAPPO, VDN and QMIX on rware-small-4ag-v1.}
  \label{fig: rware-small-4ag-v1_ranking}
\end{figure}

\begin{figure}
  \centering
  \begin{subfigure}[t]{0.4\linewidth}
    \includegraphics[width=\linewidth, valign=t]{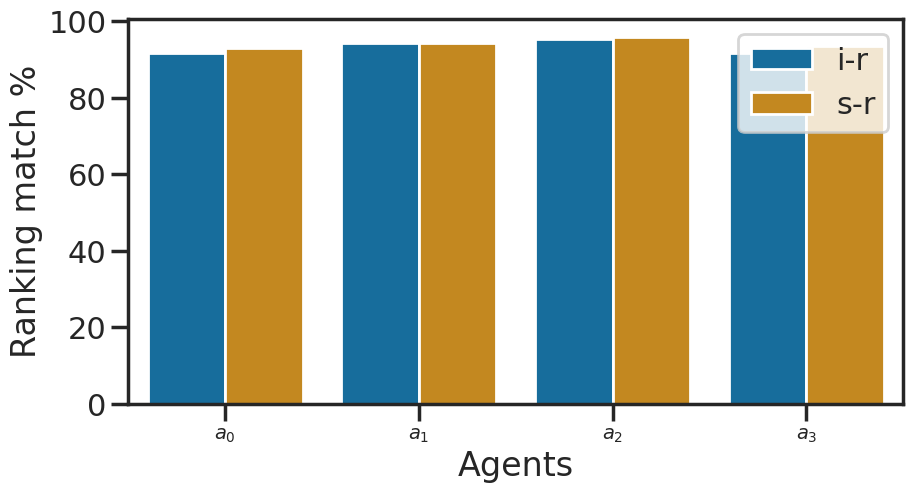}
  \end{subfigure}
  \begin{subfigure}[t]{0.4\linewidth}
    \includegraphics[width=\linewidth, valign=t]{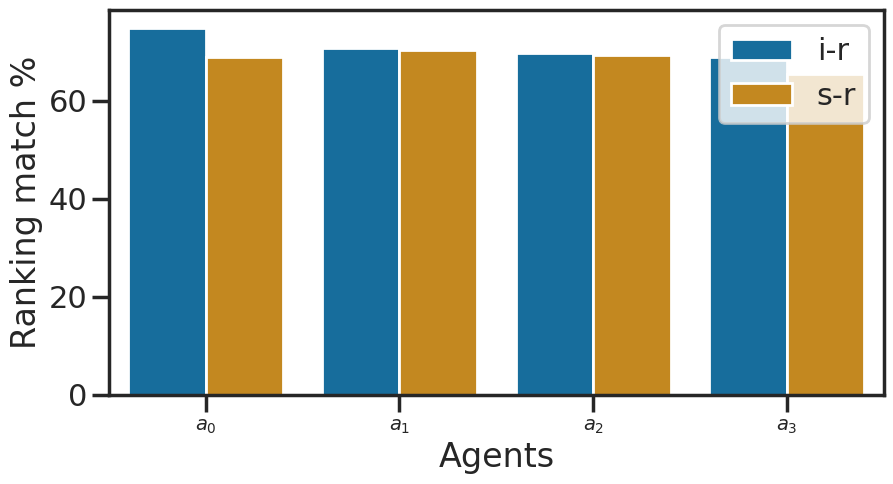}
  \end{subfigure}
  \begin{subfigure}[t]{0.3\linewidth}
    \includegraphics[width=\linewidth, valign=t]{figures/appendix/rware_AI_comparisons/mappo_rware_small_4ag_ranking.png}
  \end{subfigure}
    \begin{subfigure}[t]{0.3\linewidth}
    \includegraphics[width=\linewidth, valign=t]{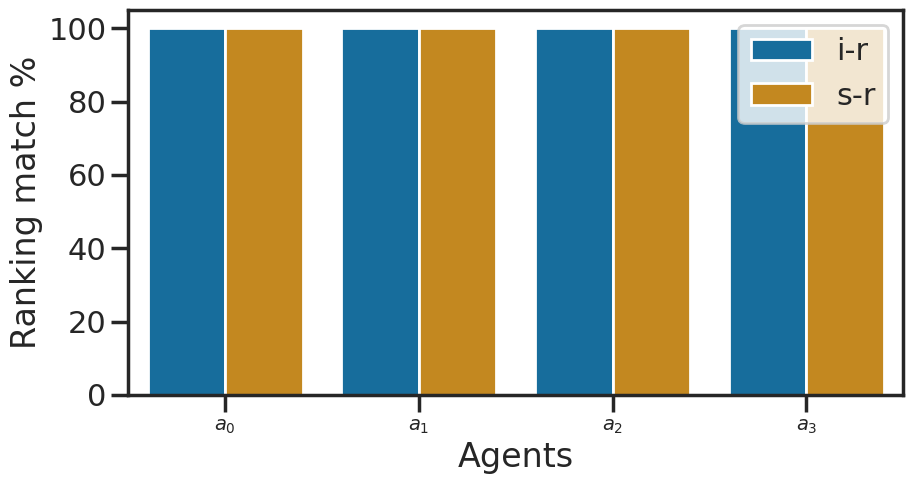}
  \end{subfigure}
  \begin{subfigure}[t]{0.3\linewidth}
    \includegraphics[width=\linewidth, valign=t]{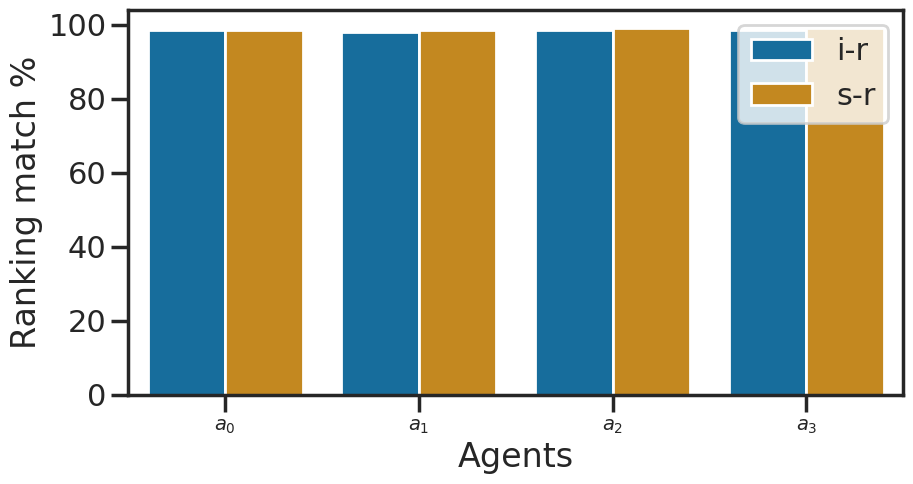}
  \end{subfigure}

   \caption{Comparison of Shapley value and agent importance rankings to individual rewards for IQL, MAA2C, MAPPO, QMIX and VDN on rware-tiny-4ag-v1}
  \label{fig: rware-tiny-4ag-v1_ranking}
\end{figure}

\begin{figure}
  \centering
  \begin{subfigure}[t]{0.3\linewidth}
    \includegraphics[width=\linewidth, valign=t]{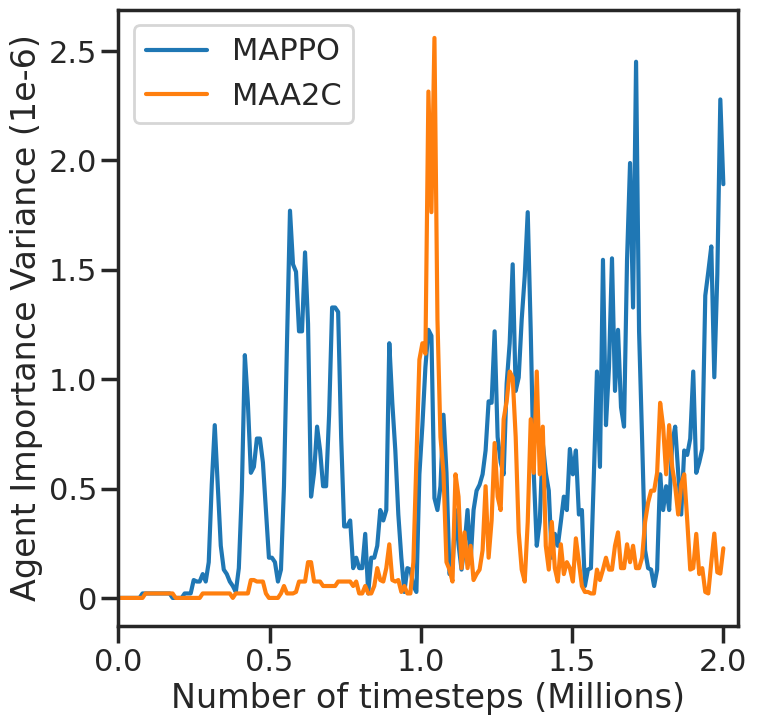}
    \caption{rware-small-4ag-v1}
  \end{subfigure}
  \begin{subfigure}[t]{0.3\linewidth}
    \includegraphics[width=\linewidth, valign=t]{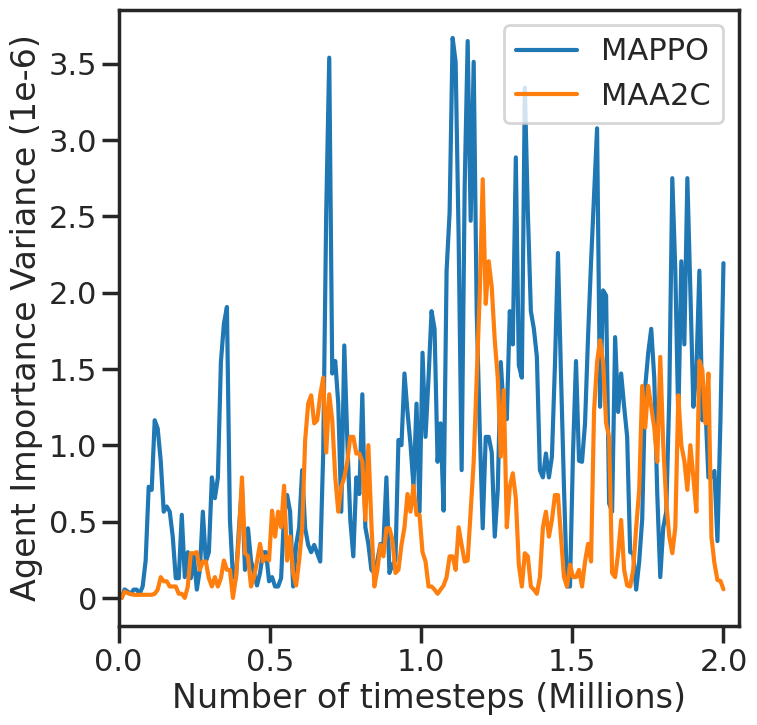}
    \caption{rware-tiny-4ag-v1}
  \end{subfigure}
  \begin{subfigure}[t]{0.3\linewidth}
    \includegraphics[width=\linewidth, valign=t]{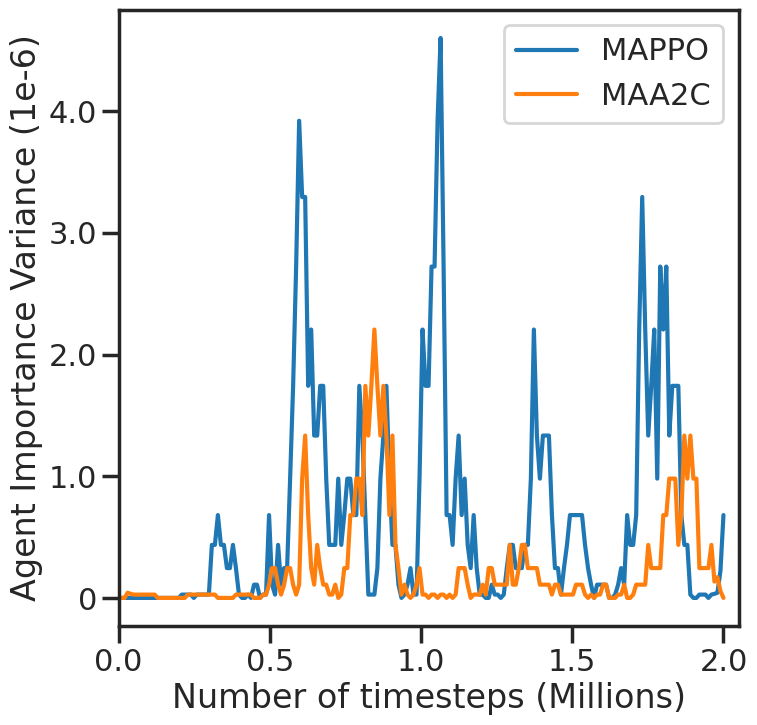}
    \caption{rware-tiny-2ag-v1}
  \end{subfigure}

   \caption{Comparisons of the agent importance variance on RWARE for MAPPO and MAA2C}
  \label{fig: MAPPO_MAA2C_AI_RWARE_VAR}
\end{figure}

\begin{figure}
  \centering
  \begin{subfigure}[t]{0.3\linewidth}
    \includegraphics[width=\linewidth, valign=t]{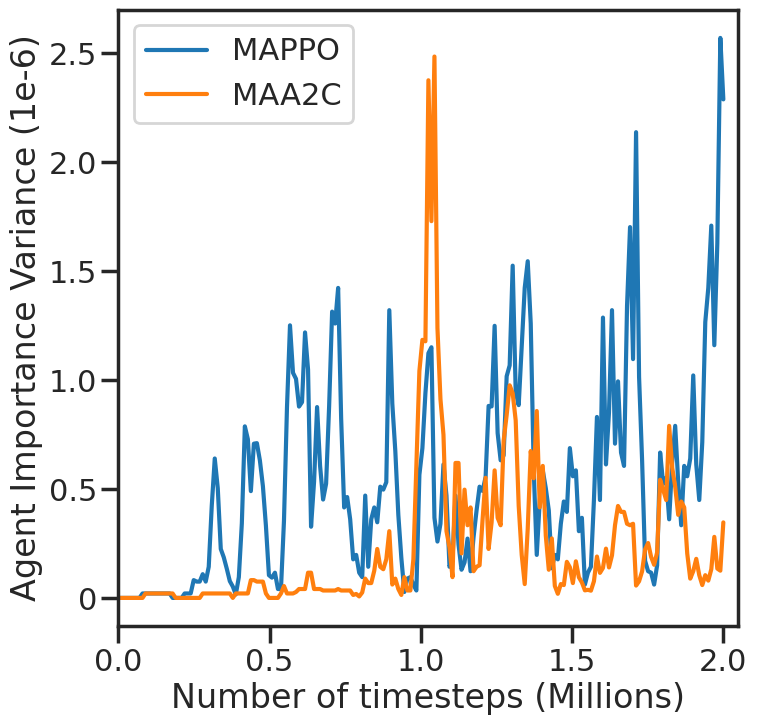}
    \caption{rware-small-4ag-v1}
  \end{subfigure}
  \begin{subfigure}[t]{0.3\linewidth}
    \includegraphics[width=\linewidth, valign=t]{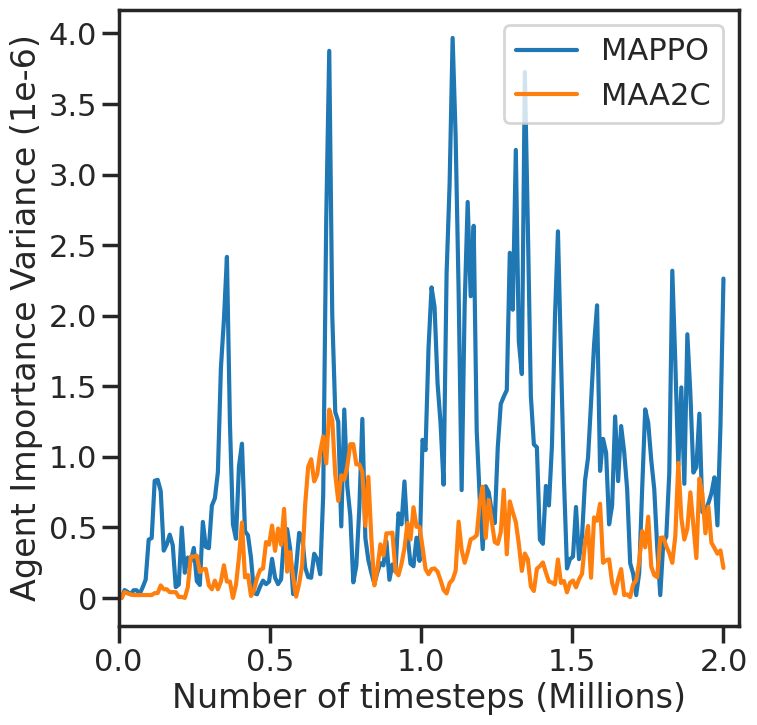}
    \caption{rware-tiny-4ag-v1}
  \end{subfigure}
  \begin{subfigure}[t]{0.3\linewidth}
    \includegraphics[width=\linewidth, valign=t]{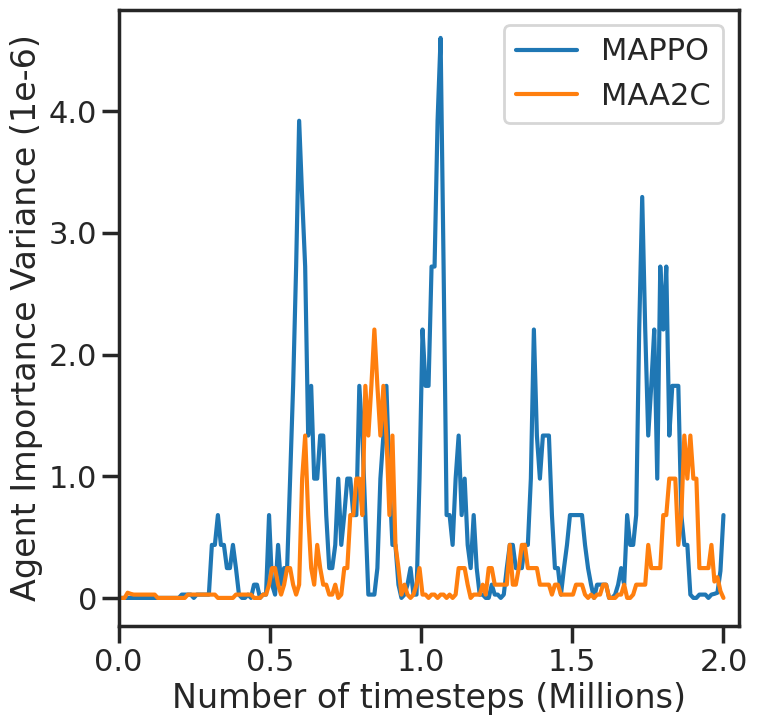}
    \caption{rware-tiny-2ag-v1}
  \end{subfigure}

   \caption{Comparisons of the Shapley value variance RWARE for MAPPO and MAA2C}
  \label{fig: MAPPO_MAA2C_SHAP_RWARE_VAR}
\end{figure}

We perform additional experimentation comparing agent importance and the Shapley values in RWARE which is a sparse setting. From figures \ref{fig: rware-small-4ag-v1_ranking} and \ref{fig: rware-tiny-4ag-v1_ranking} we can see that both method perform similarly in the sparse setting when performance is poor like for IQL, QMIX and VDN. However when the agents are able to achieve some level of success, accuracy drops for both methods. This is likely due to the sparse reward creating many zeros in the data and creating erroneous predictions. This is most noticeable for MAPPO and MAA2C where the Shapley Value estimations can drop below 90\%. We can further verify this from figures \ref{fig: MAPPO_MAA2C_AI_RWARE_VAR} and \ref{fig: MAPPO_MAA2C_SHAP_RWARE_VAR} where the agent importance and Shapey values exhibit similar variance over time.

\begin{figure}
  \centering
  \begin{subfigure}[t]{0.3\linewidth}
    \includegraphics[width=\linewidth, valign=t]{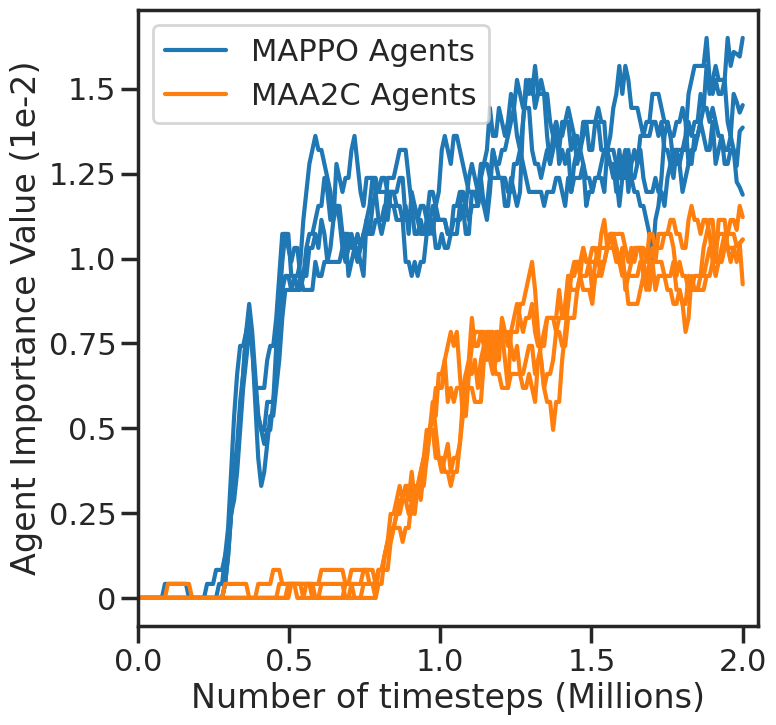}
    \caption{rware-small-4ag-v1}
  \end{subfigure}
  \begin{subfigure}[t]{0.3\linewidth}
    \includegraphics[width=\linewidth, valign=t]{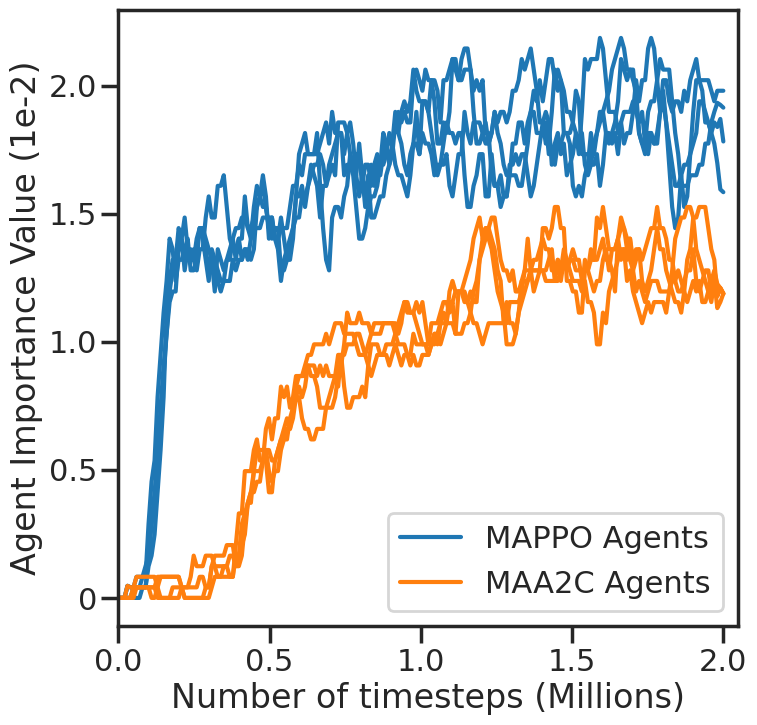}
    \caption{rware-tiny-4ag-v1}
  \end{subfigure}
  \begin{subfigure}[t]{0.3\linewidth}
    \includegraphics[width=\linewidth, valign=t]{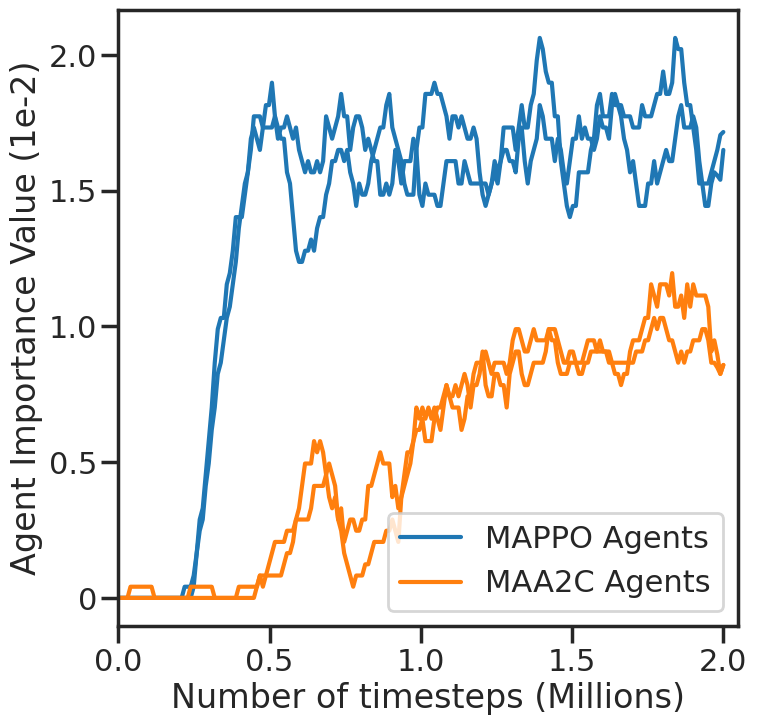}
    \caption{rware-tiny-2ag-v1}
  \end{subfigure}

   \caption{Comparisons of the agent importance on rware-small-4ag-v1 for MAPPO and MAA2C}
  \label{fig: MAPPO_MAA2C_AI_RWARE}
\end{figure}

\begin{figure}
  \centering
  \begin{subfigure}[t]{0.3\linewidth}
    \includegraphics[width=\linewidth, valign=t]{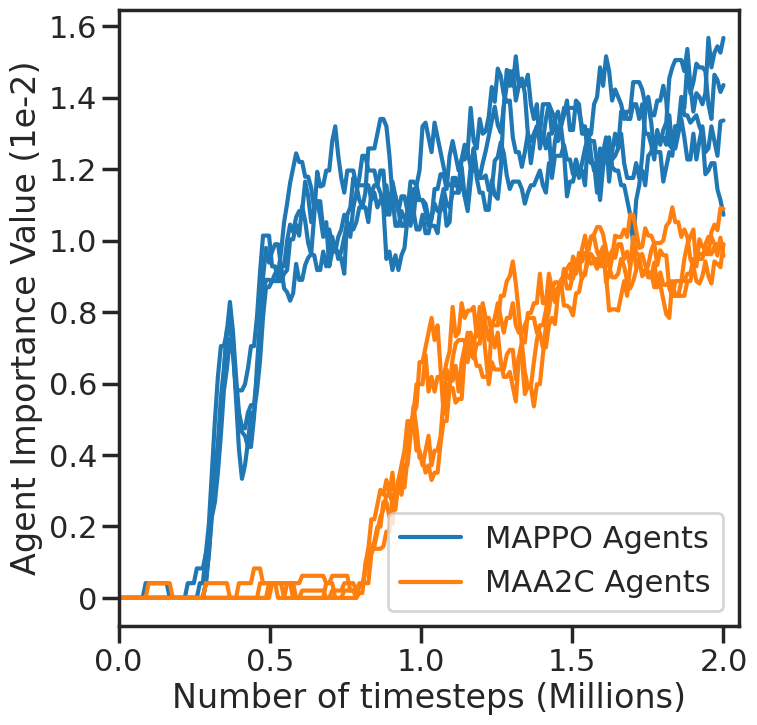}
    \caption{rware-small-4ag-v1}
  \end{subfigure}
  \begin{subfigure}[t]{0.3\linewidth}
    \includegraphics[width=\linewidth, valign=t]{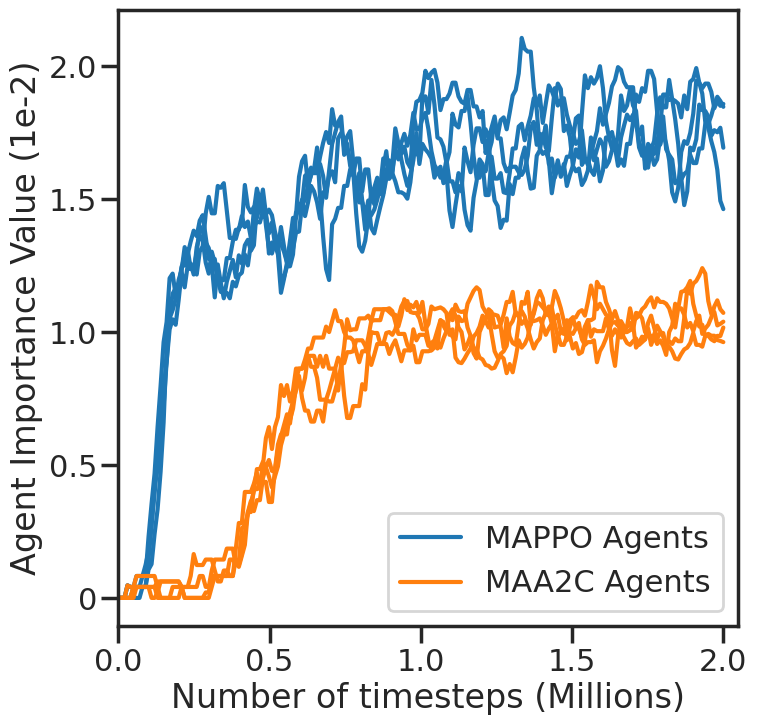}
    \caption{rware-tiny-4ag-v1}
  \end{subfigure}
  \begin{subfigure}[t]{0.3\linewidth}
    \includegraphics[width=\linewidth, valign=t]{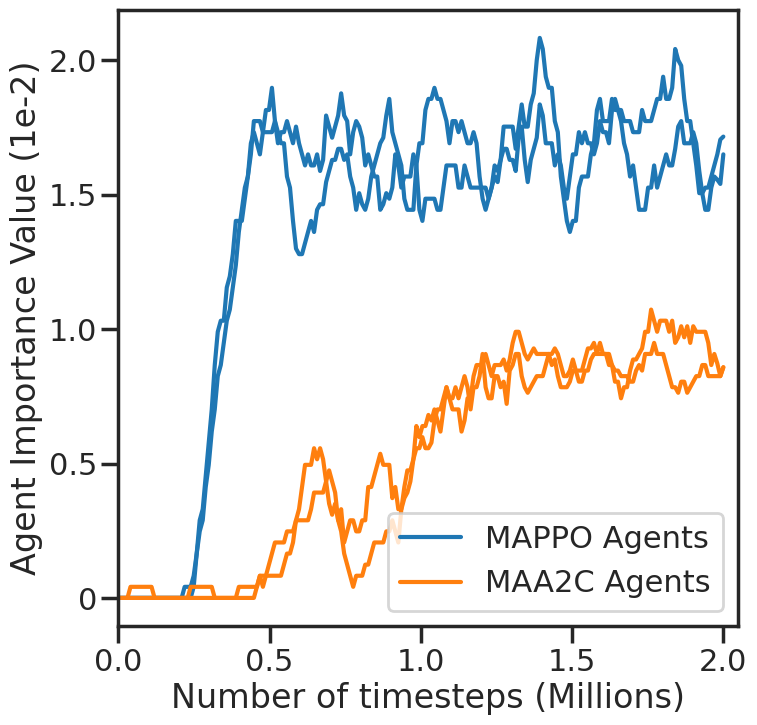}
    \caption{rware-tiny-2ag-v1}
  \end{subfigure}

   \caption{Comparisons of the Shapley values on RWARE for MAPPO and MAA2C}
  \label{fig: MAPPO_MAA2C_SHAP_RWARE}
\end{figure}

\begin{figure}
  \centering
  \begin{subfigure}[t]{0.30\linewidth}
    \includegraphics[width=\linewidth, valign=t]{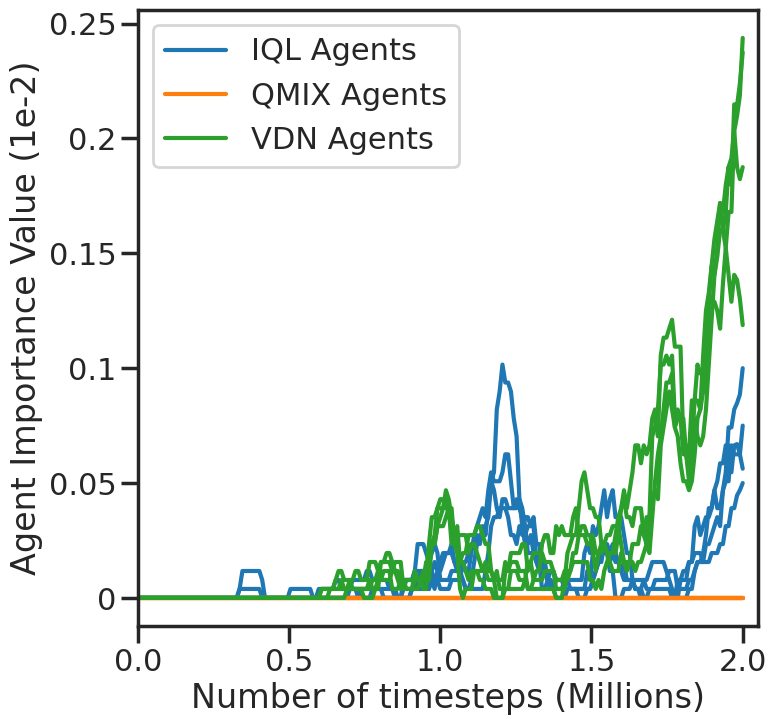}
    \caption{rware-small-4ag-v1}
  \end{subfigure}
  \begin{subfigure}[t]{0.30\linewidth}
    \includegraphics[width=\linewidth, valign=t]{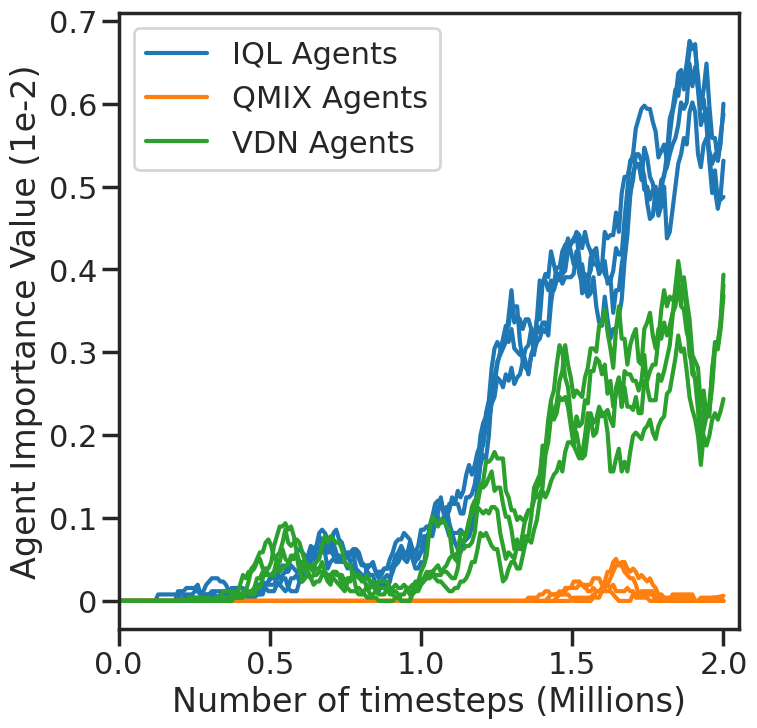}
     \caption{rware-tiny-4ag-v1}
  \end{subfigure}
  \begin{subfigure}[t]{0.30\linewidth}
    \includegraphics[width=\linewidth, valign=t]{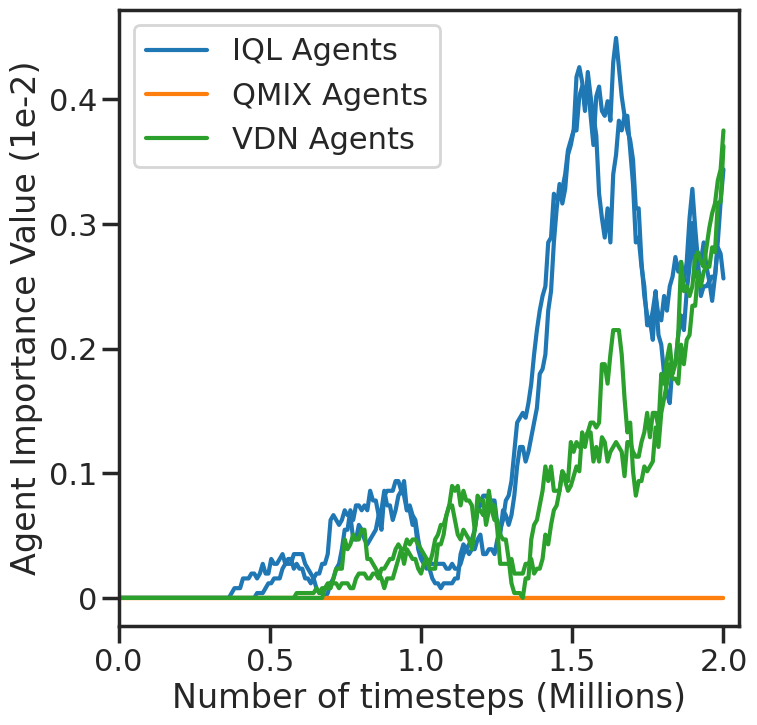}
     \caption{rware-tiny-2ag-v1}
  \end{subfigure}

   \caption{Comparisons of the agent importance on RWARE for QMIX, VDN and IQL}
  \label{fig: QL_AI_RWARE}
\end{figure}

\begin{figure}
  \centering
  \begin{subfigure}[t]{0.30\linewidth}
    \includegraphics[width=\linewidth, valign=t]{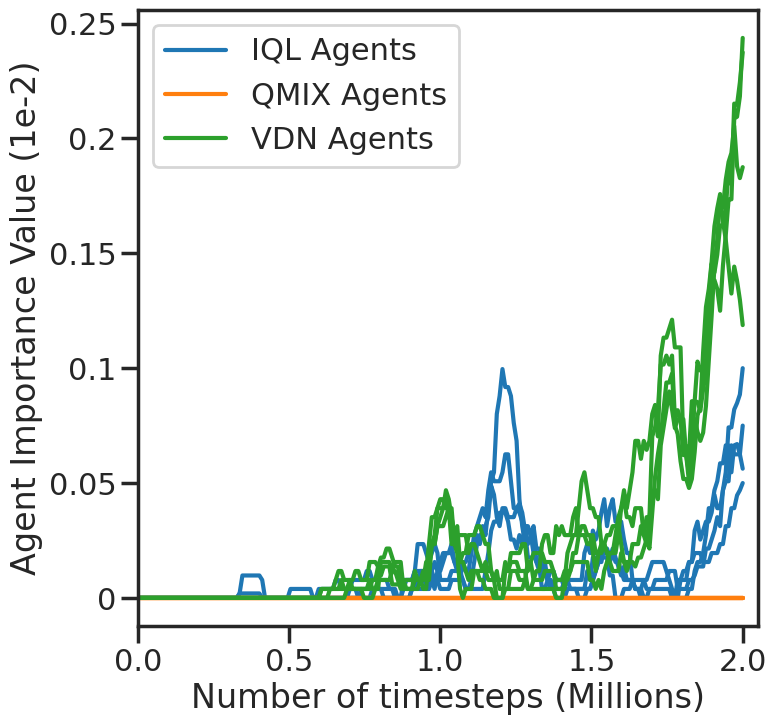}
    \caption{rware-small-4ag-v1}
  \end{subfigure}
  \begin{subfigure}[t]{0.30\linewidth}
    \includegraphics[width=\linewidth, valign=t]{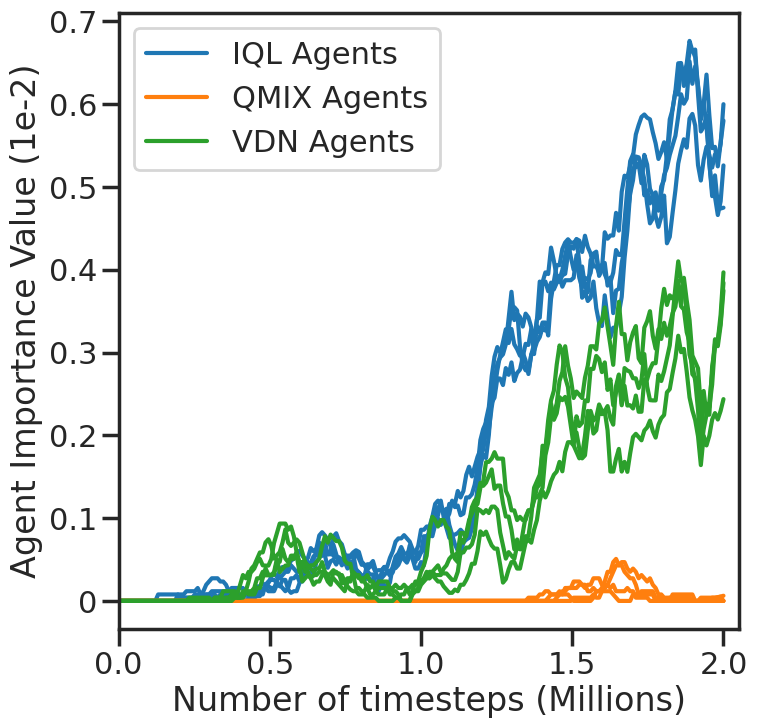}
    \caption{rware-tiny-4ag-v1}
  \end{subfigure}
  \begin{subfigure}[t]{0.30\linewidth}
    \includegraphics[width=\linewidth, valign=t]{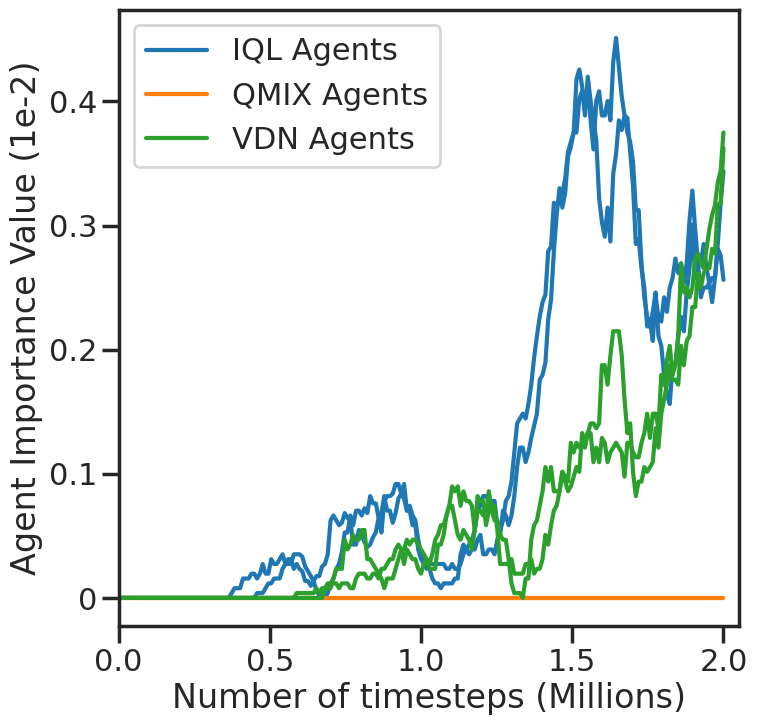}
    \caption{rware-tiny-2ag-v1}
  \end{subfigure}

   \caption{Comparisons of the Shapley value on RWARE for QMIX, VDN and IQL}
  \label{fig: QL_SHAP_RWARE}
\end{figure}

\begin{figure}
  \centering
  \begin{subfigure}[t]{0.30\linewidth}
    \includegraphics[width=\linewidth, valign=t]{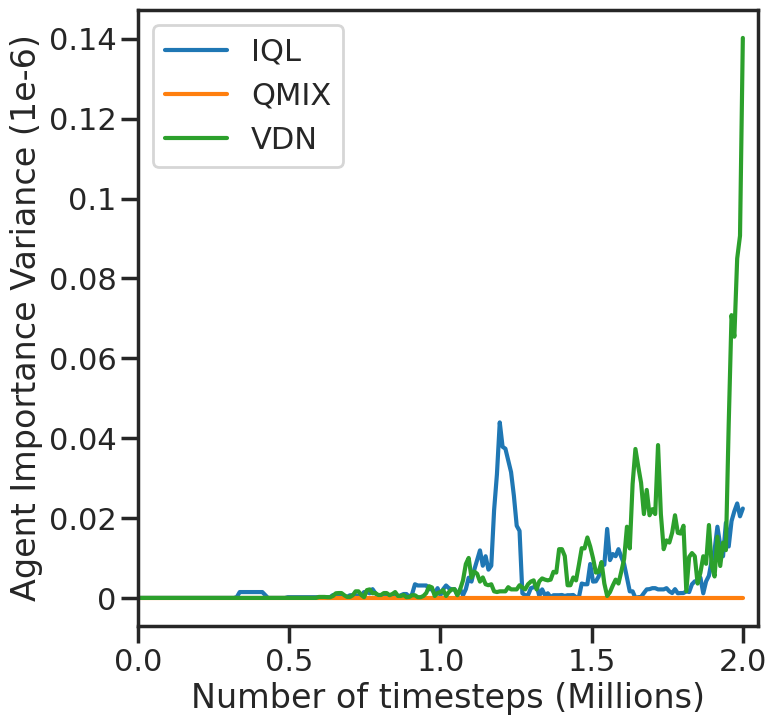}
    \caption{rware-small-4ag-v1}
  \end{subfigure}
  \begin{subfigure}[t]{0.30\linewidth}
    \includegraphics[width=\linewidth, valign=t]{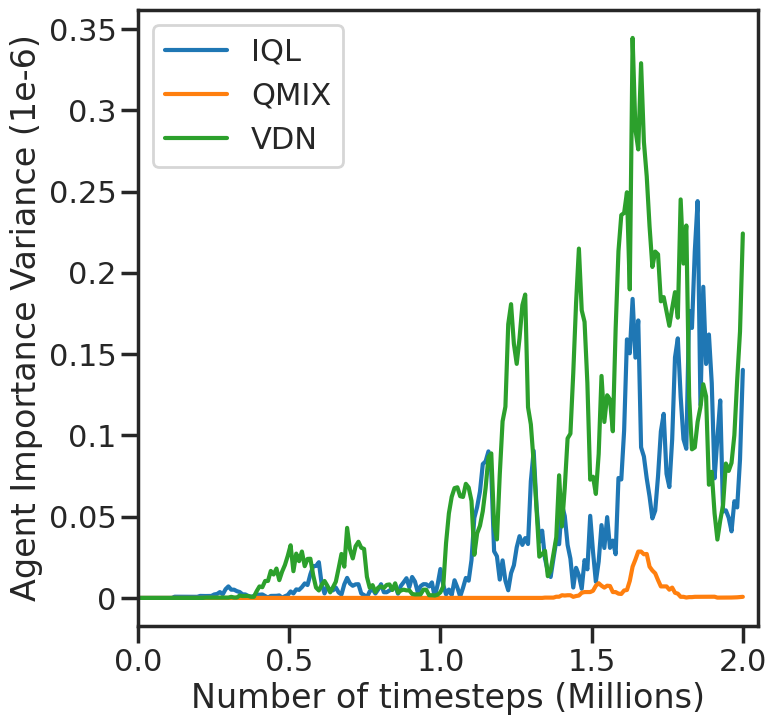}
    \caption{rware-tiny-4ag-v1}
  \end{subfigure}
  \begin{subfigure}[t]{0.30\linewidth}
    \includegraphics[width=\linewidth, valign=t]{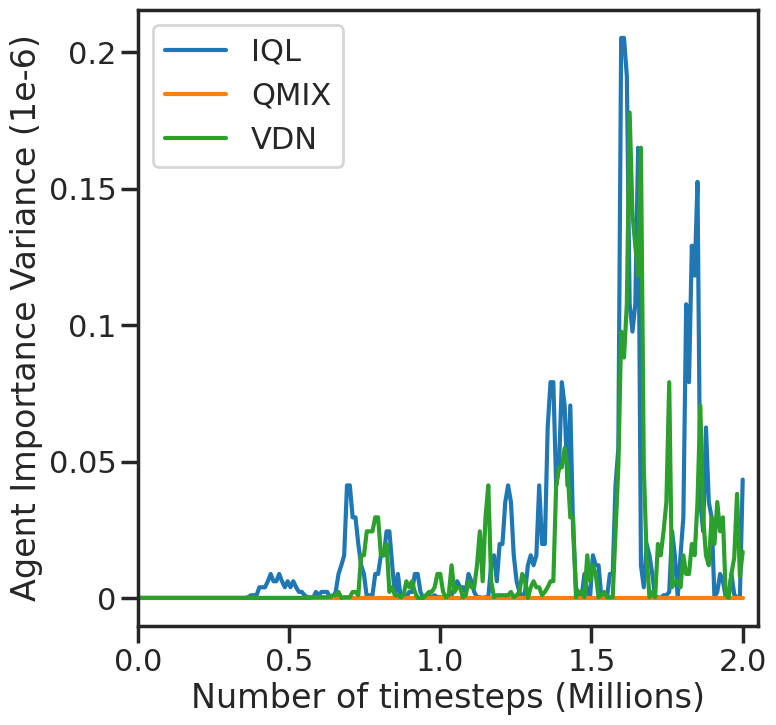}
    \caption{rware-tiny-2ag-v1}
  \end{subfigure}

   \caption{Comparisons of the agent importance variance on RWARE for QMIX, VDN and IQL}
  \label{fig: QL_AI_RWARE_VAR}
\end{figure}

\begin{figure}
  \centering
  \begin{subfigure}[t]{0.30\linewidth}
    \includegraphics[width=\linewidth, valign=t]{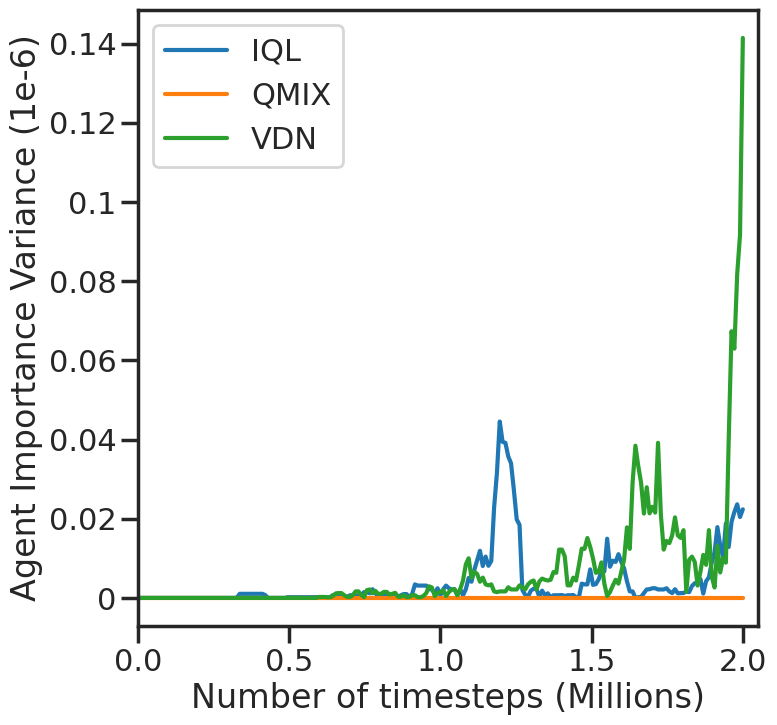}
    \caption{rware-small-4ag-v1}
  \end{subfigure}
  \begin{subfigure}[t]{0.30\linewidth}
    \includegraphics[width=\linewidth, valign=t]{figures/appendix/rware_AI_comparisons/q_learning_rware_tiny_4ag_variance.png}
    \caption{rware-tiny-4ag-v1}
  \end{subfigure}
  \begin{subfigure}[t]{0.30\linewidth}
    \includegraphics[width=\linewidth, valign=t]{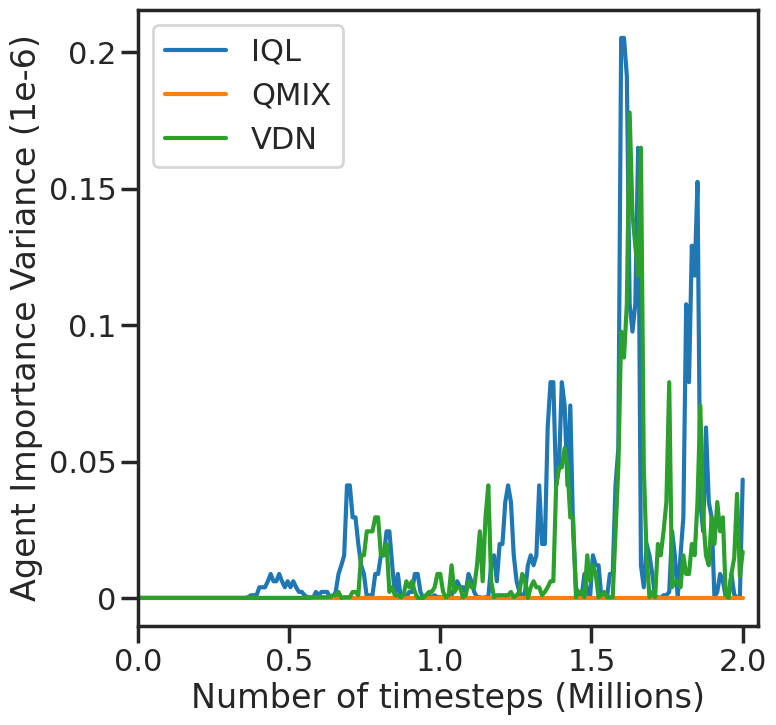}
    \caption{rware-tiny-2ag-v1}
  \end{subfigure}

   \caption{Comparisons of the Shapley value variance on RWARE for QMIX, VDN and IQL}
  \label{fig: QL_SHAP_RWARE_VAR}
\end{figure}

\subsection{Aggregation of agent importance}\label{aggregation}
\begin{figure}
     \centering
     \begin{subfigure}[b]{0.23\textwidth}
     \includegraphics[width=\textwidth]{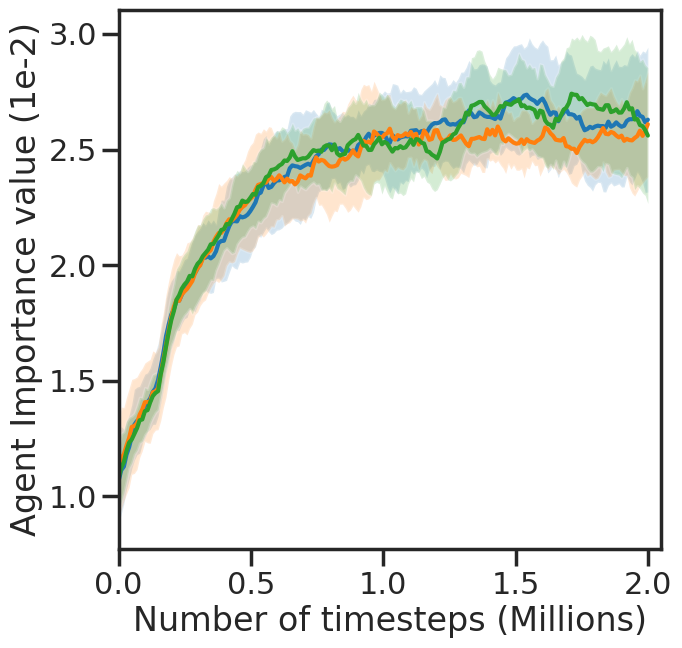}
     \end{subfigure}
     \begin{subfigure}[b]{0.23\textwidth}
         \includegraphics[width=\textwidth]{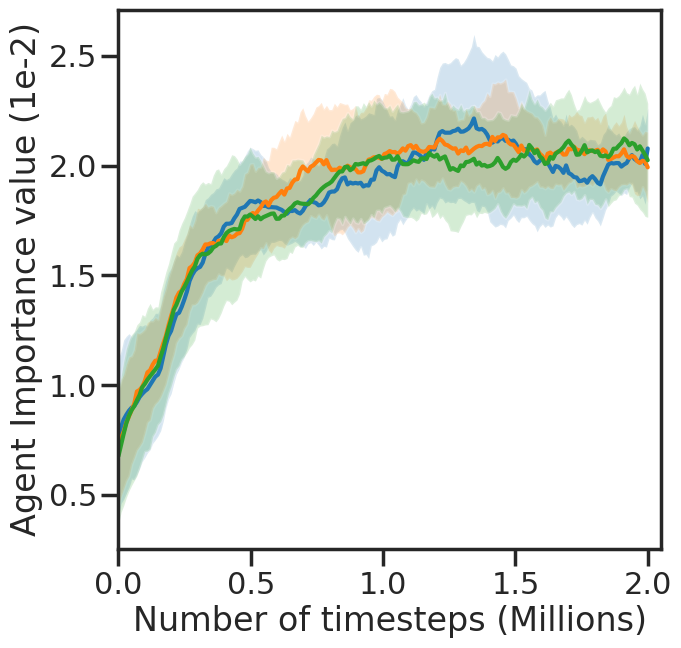}
     \end{subfigure}
     \begin{subfigure}[b]{0.23\textwidth}
     \includegraphics[width=\textwidth]{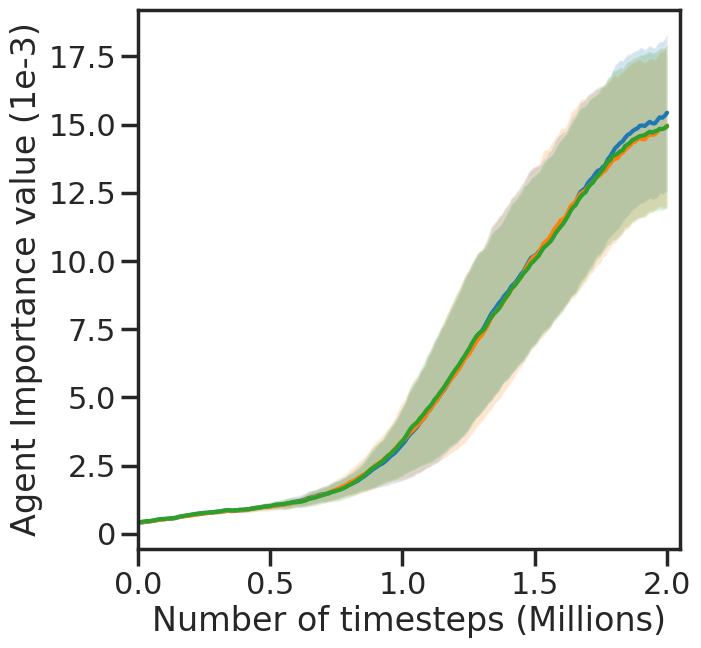}
     \end{subfigure}
     \begin{subfigure}[b]{0.23\textwidth}
         \includegraphics[width=\textwidth]{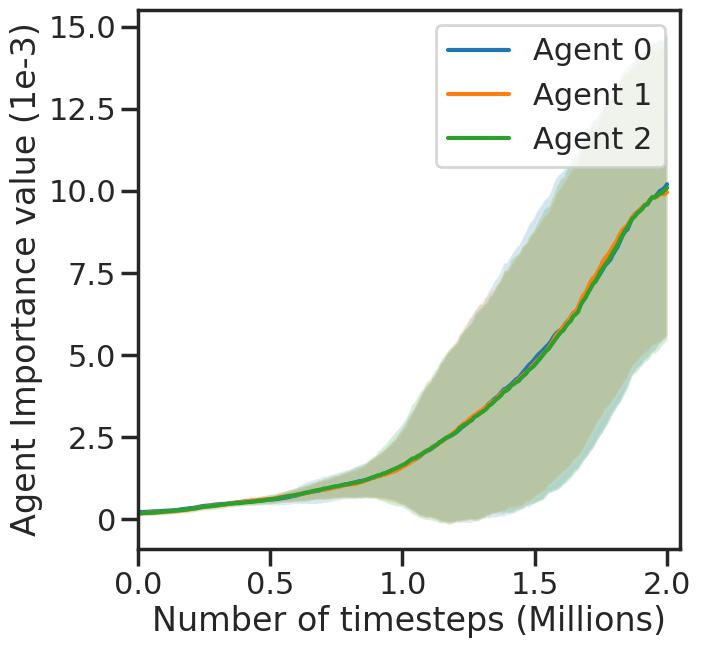}
     \end{subfigure}
     \caption{\textit{Agent importance scores on the stochastic Foraging-10x10-3p-3f-v2 LBF scenario for MAA2C, MAPPO, VDN and QMIX.}} 
     \label{fig:lbf_agg}
 \end{figure}

\begin{figure}
     \centering
     \begin{subfigure}[b]{0.23\textwidth}
     \includegraphics[width=\textwidth]{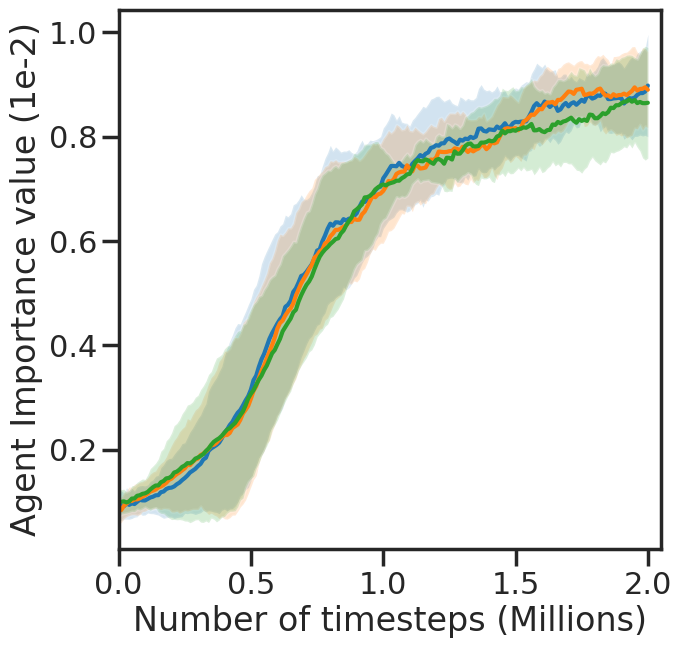}
     \end{subfigure}
     \begin{subfigure}[b]{0.23\textwidth}
         \includegraphics[width=\textwidth]{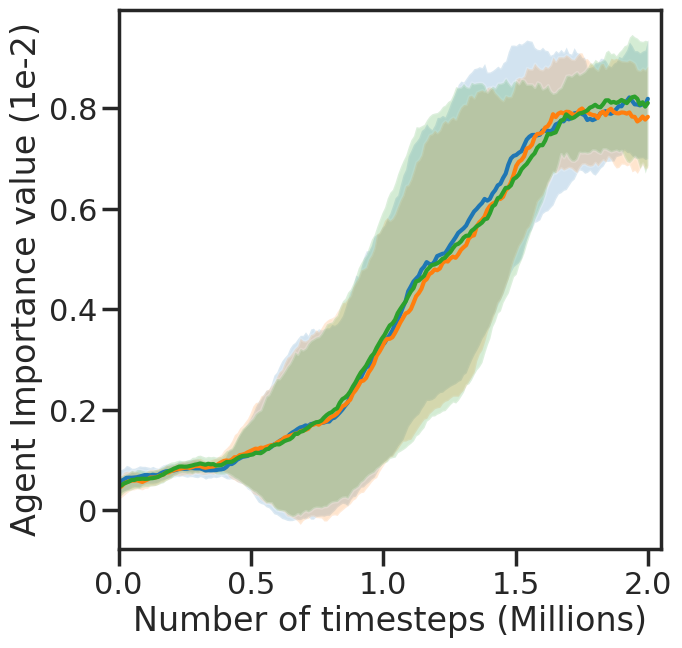}
     \end{subfigure}
     \begin{subfigure}[b]{0.23\textwidth}
     \includegraphics[width=\textwidth]{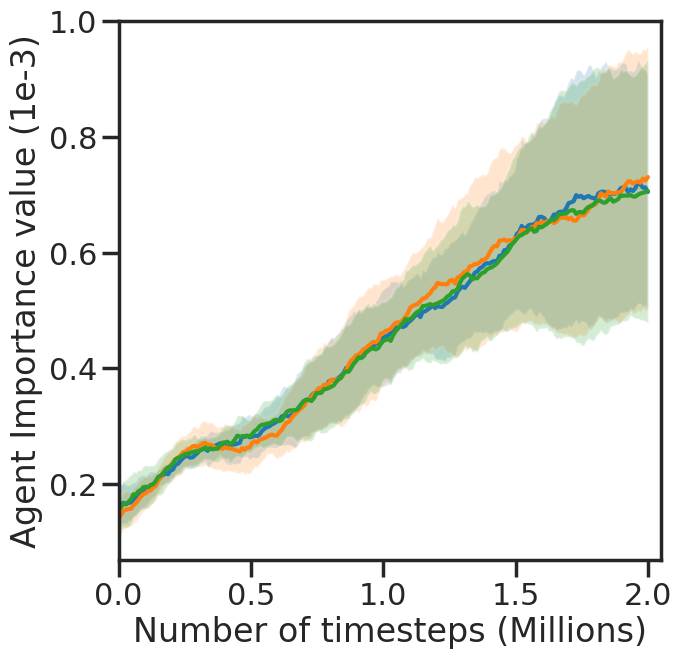}
     \end{subfigure}
     \begin{subfigure}[b]{0.23\textwidth}
         \includegraphics[width=\textwidth]{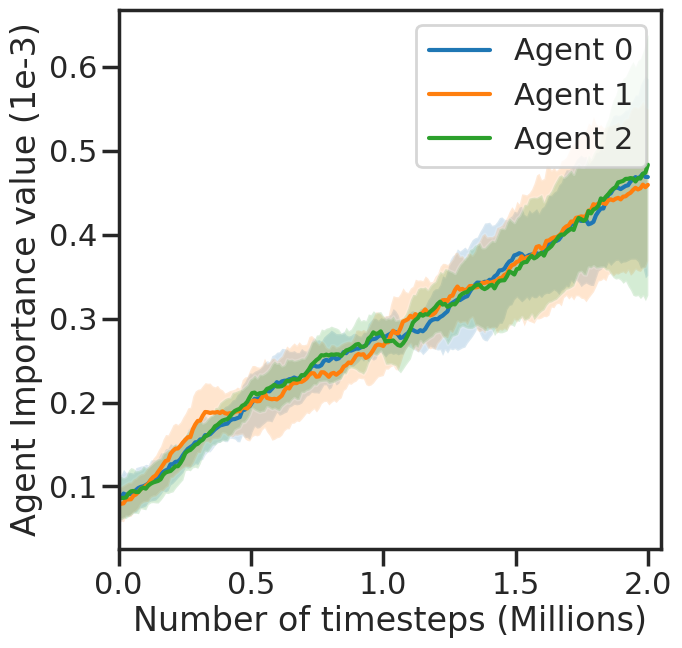}
     \end{subfigure}
     \caption{\textit{Agent importance scores on the stochastic lbforaging:Foraging-15x15-3p-5f-v2 LBF scenario for MAA2C, MAPPO, VDN and QMIX.}} 
     \label{fig:lbf_agg_15}
 \end{figure}

Throughout the paper we use a single seed to display agent importance over time. For homogeneous settings with parameter sharing this is required as agents can take different roles in each seed depending on the training conditions. Essentially agents of a similar type can fulfil multiple different sub-roles during training which makes aggregating agent contributions over multiple seeds inconsistent in the stochastic case as seen in \ref{fig:lbf_agg}. When compared to the parameter sharing explain in figure \ref{fig: ps_vs_non-ps} we can see that determining individual agent contributions becomes difficult. Similar issues can be seen in figure \ref{fig:lbf_agg_15} when compared to figure \ref{fig: maa2c_vs_mappo}.

\section{On heterogeneous settings}\label{smaclite}

For most of our experiments, we make use of the LBF and RWARE settings. RWARE is completely homogeneous as all agents have the same capabilities as each other and their roles and importance within the coalition are developed during training time as the individual policies associated with each agent's IDs are learnt. This means that agent roles are inconsistent across seeds and parameterisation as depending on external factors across runs, different agent IDs can occupy different roles. For LBF, agents are homogeneous in their action space but their importance rankings are essentially preassigned due to their levels determining the extent of their contribution towards collecting food. Given the limitations of these settings, it is important to determine how effectively \textit{agent importance} is able to determine the contributions of agents in complex heterogeneous settings where there are clear agent types with differing capabilities that compose the coalition.

A popular setting with heterogeneous agents in cooperative MARL is the Starcraft Multi-Agent Challenge (SMAC) \citep{SMAC} however, this environment has 2 limitations which make applying agent importance difficult. Firstly, it uses the original game engine for the Starcraft 2 (SC2) video game which is coded in the C++ programming language as a back-end. This makes it unsuitable for creating multiple parallel copies of the setting using the built-in Python copy method which makes applying contribution calculation methods like the Shapley value and agent importance challenging. Secondly, the SC2 engine back-end is computationally expensive to run and the high resource requirements make using contribution calculation methods on top of the existing environment unappealing. Instead, we make use of SMAClite \citep{michalski2023smaclite}, which implements a setting similar to SMAC but in purely Python code which allows it to be copied and reduces computational requirements.

\begin{figure}[h]
  \centering
\begin{subfigure}{0.4\linewidth}
    \includegraphics[width=\linewidth]{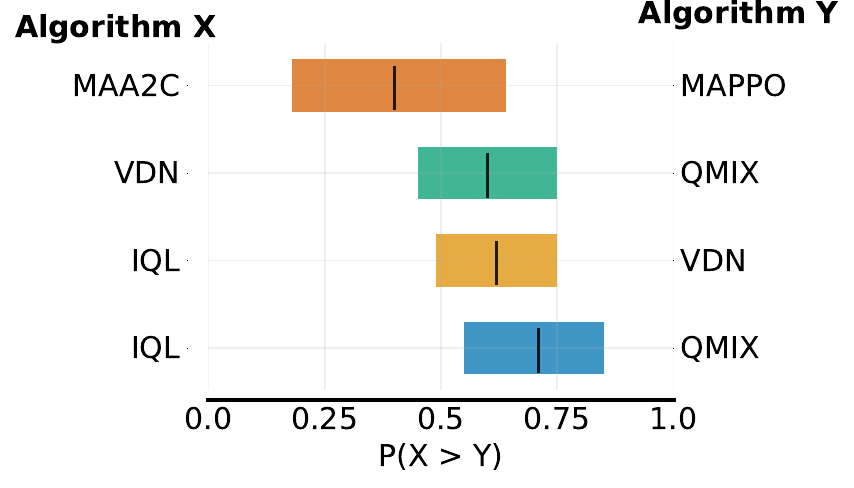}
    \caption{Probability of improvement}
  \end{subfigure}
  \begin{subfigure}{0.5\linewidth}
    \includegraphics[width=\linewidth]{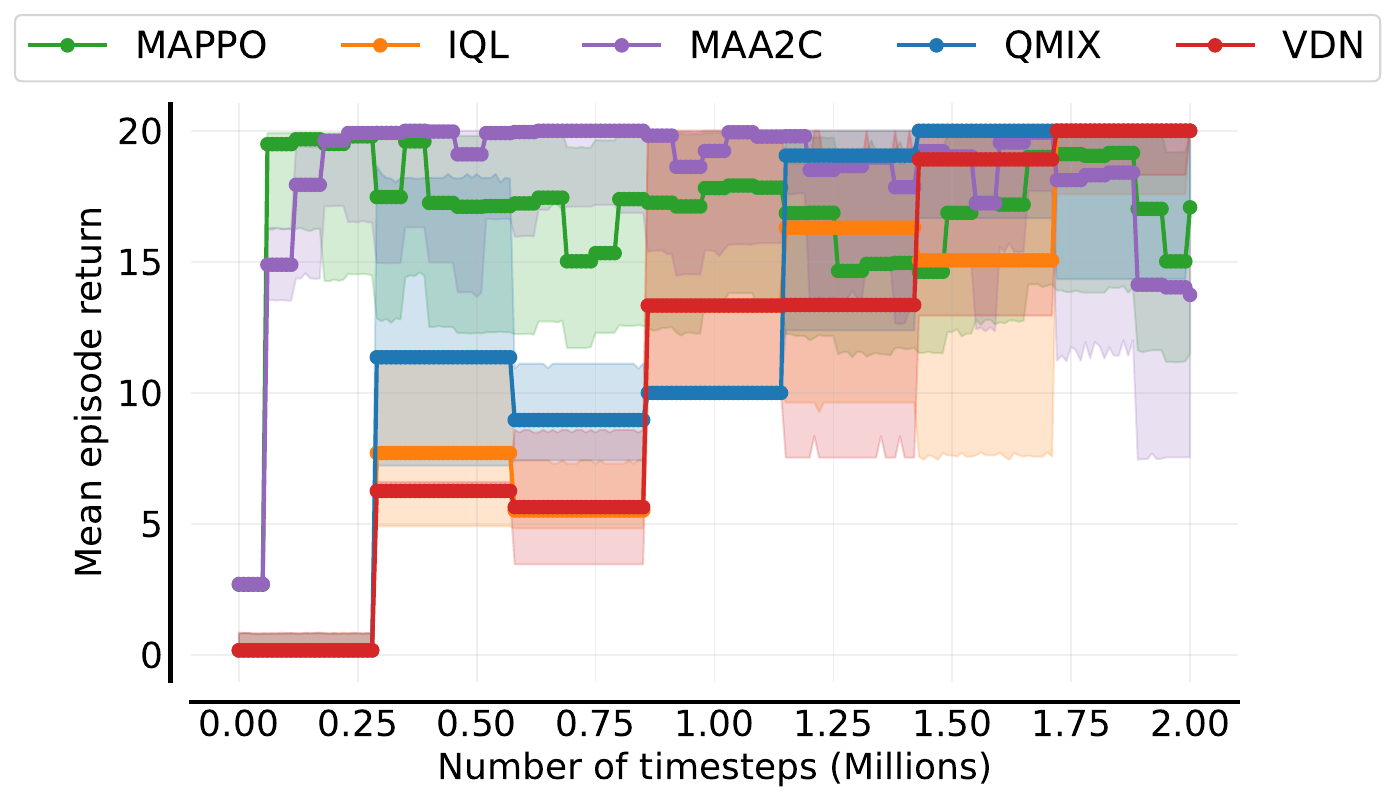}
    \caption{Sample efficiency}
  \end{subfigure}
  \caption{Algorithms performance on SMAClite without parameter sharing}
\end{figure}

\begin{figure}[h]
  \centering
\begin{subfigure}{0.45\linewidth}
    \includegraphics[width=\linewidth]{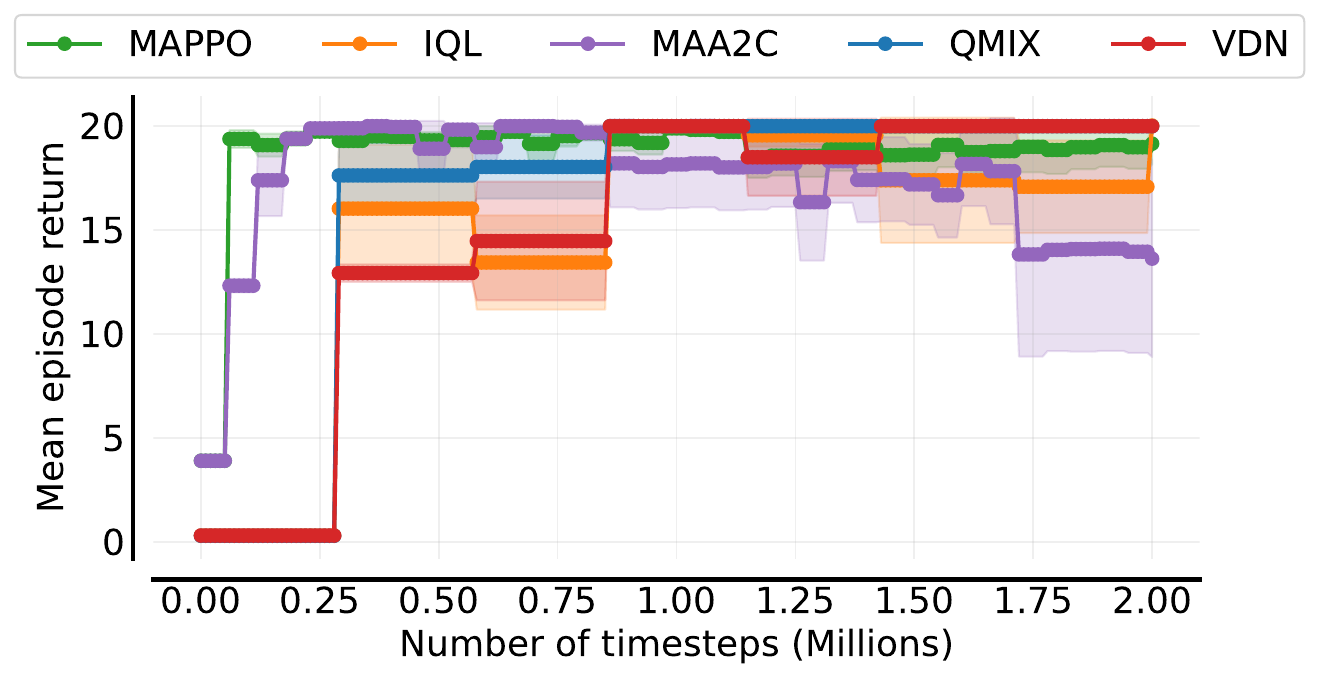}
  \end{subfigure}
  \begin{subfigure}{0.45\linewidth}
    \includegraphics[width=\linewidth]{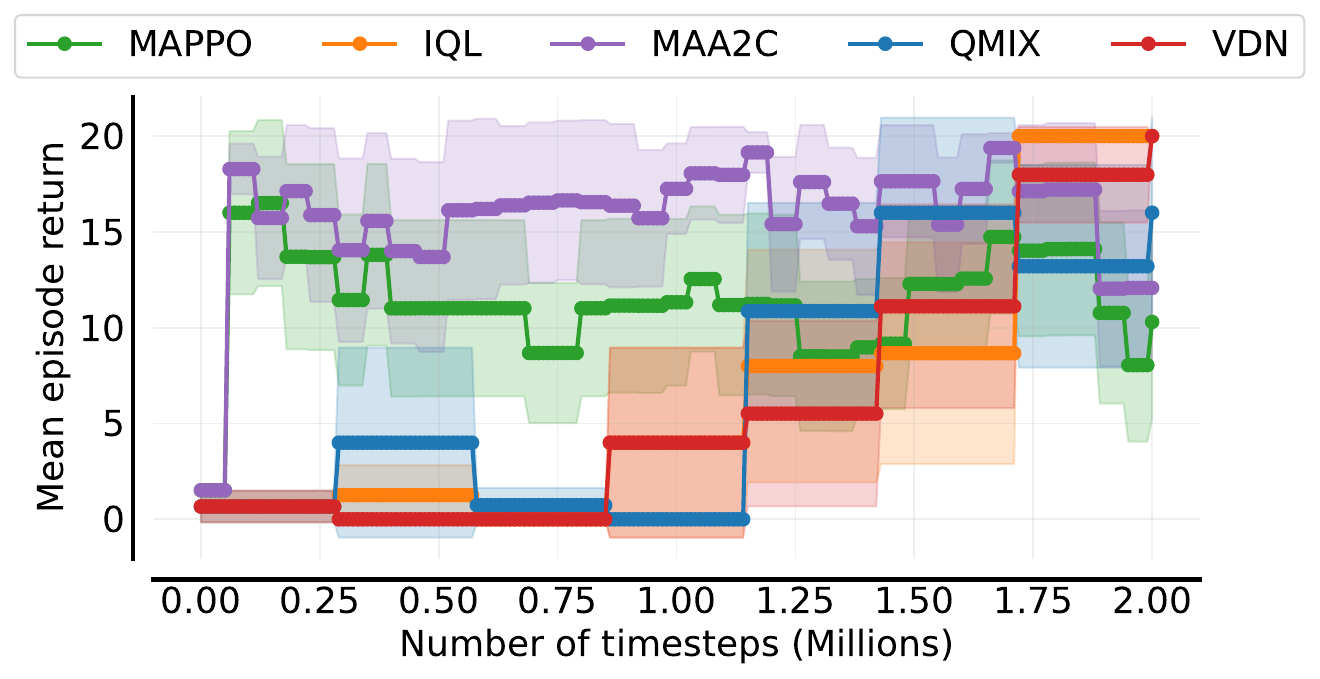}
  \end{subfigure}
  \caption{\textbf{Left:} Mean episode return on 2s3z. \textbf{Right:} Mean episode return on 3m.}
  \label{fig: smac_performance_1}
\end{figure}

We plot performance over training using the absolute metric in figure \ref{fig: smac_performance_1} for the 3m and 2s3z scenarios. We average results over 6 seeds rather than the 5 used in the SMAClite paper and run the policy gradient (PG) methods for 20 million timesteps and the Q-learning methods for 2 million as recommended in the EPymarl benchmark \citep{papoudakis2021benchmarking}. We found there to be high variance in the performance across seeds for the 3m setting especially for the PG methods which often experienced significant performance decay later into training however, for the more complex 2s3z setting performance was fairly stable and algorithms quickly converged to reasonable policies.

\begin{figure}[h]
  \centering
  \begin{subfigure}{0.45\linewidth}
    \includegraphics[width=\linewidth]{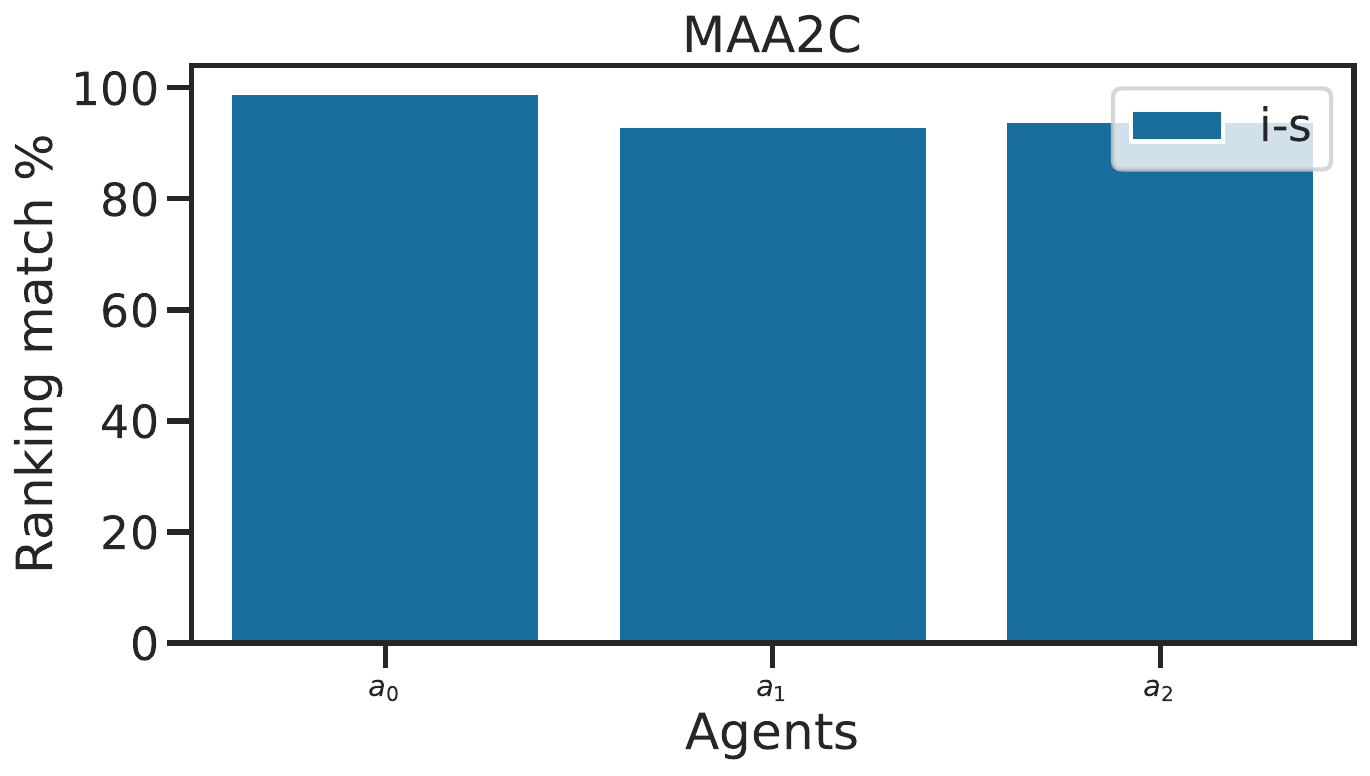}
  \end{subfigure}
  \begin{subfigure}{0.45\linewidth}
    \includegraphics[width=\linewidth]{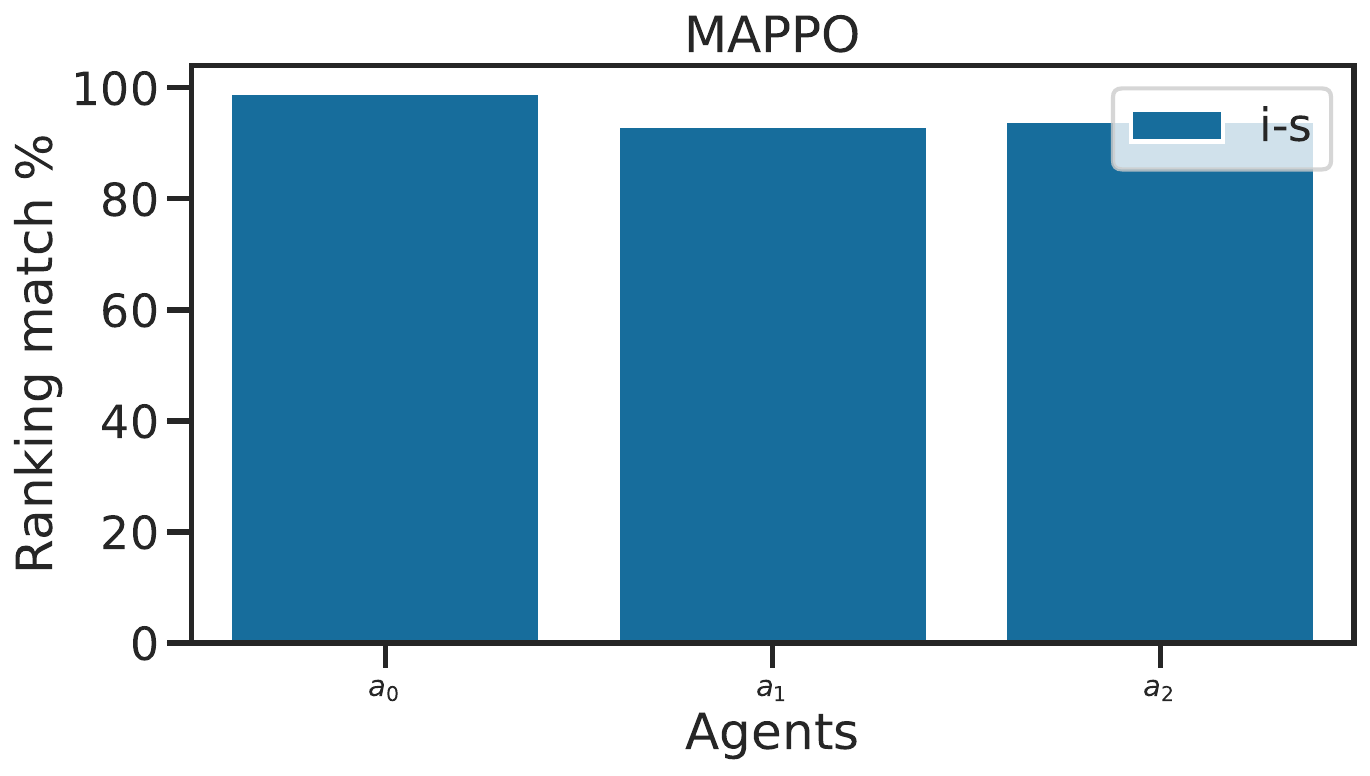}
  \end{subfigure}
  \caption{Ranking agreement percentages for MAPPO and MAA2C for 3m}
  \label{fig: ranking_matching_3m}
\end{figure}

\begin{figure}[h]
  \centering
\begin{subfigure}{0.45\linewidth}
    \includegraphics[width=\linewidth]{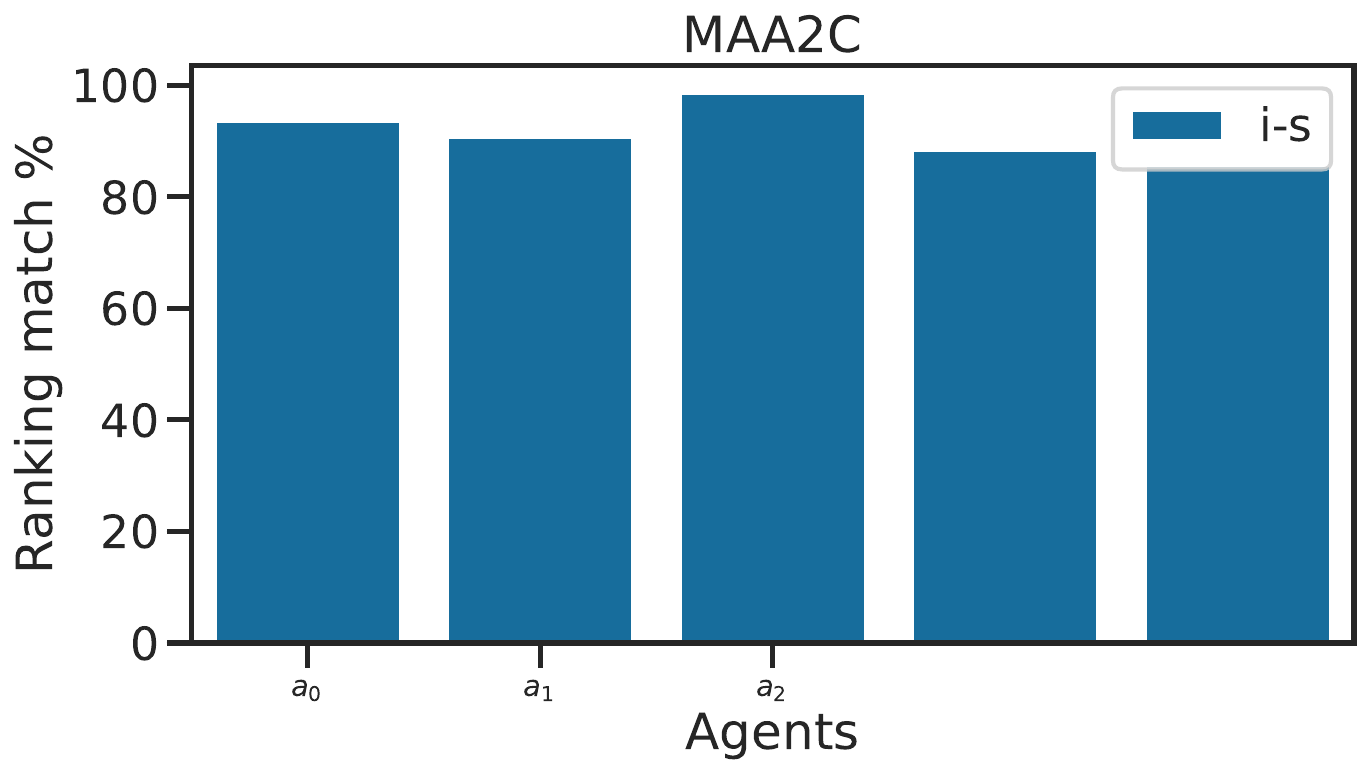}
  \end{subfigure}
  \begin{subfigure}{0.45\linewidth}
    \includegraphics[width=\linewidth]{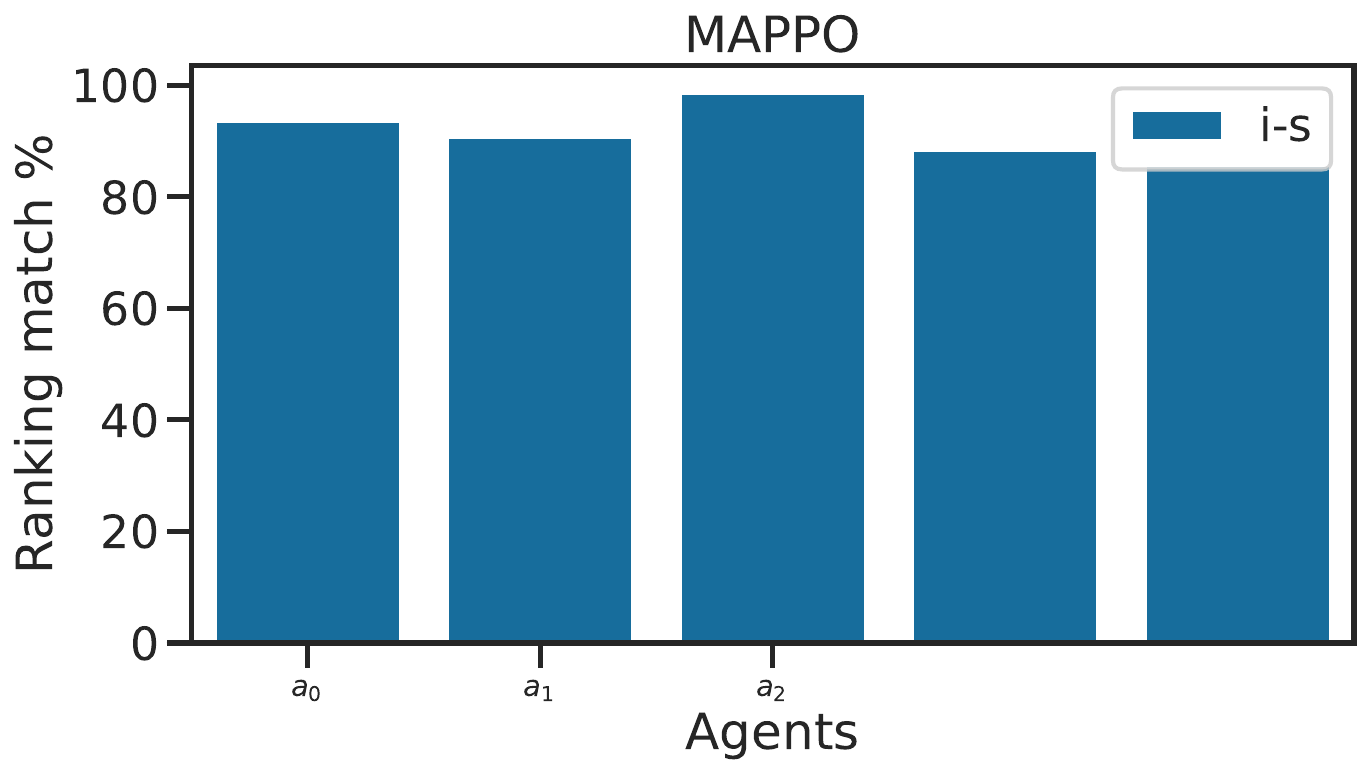}
  \end{subfigure}
  \caption{Ranking agreement percentages for MAPPO and MAA2C for 2s3z}
  \label{fig: ranking_matching_2s3z}
\end{figure}

Firstly, given the high variance between seeds we determine how accurately agent importance is able to capture the inddividual rewards that compose the joint reward in SMAClite. Unlike the RWARE and LBF setting, SMAC and SMAClite do not produce individual rewards which can be used as a ground truth value. Therefore instead of comparing contribution methods to the individual rewards we directly compare agent importance and the Shapley value were we take the shapley value to be an accurate approximation of the ground truth. We note that for 3m, despite the high variance in return across all methods, agent importance and the Shapley value have near 100 percent agreement w.r.t agent rankings. When moving to 2s3z agreement drops to 90 percent for agents 0 and 1 but remains near 100 percent for agents 2 to 4. As agent 0 and 1 are of the same type. Possibly this indicates that agent importance is effective in determining the relative contributions of each agent but as it does not calculate the value of all possible coalitions it can produce erroneous values when multiple agents are closely related but have different importance.

\begin{figure}[h]
  \centering
\begin{subfigure}{0.45\linewidth}
    \includegraphics[width=\linewidth]{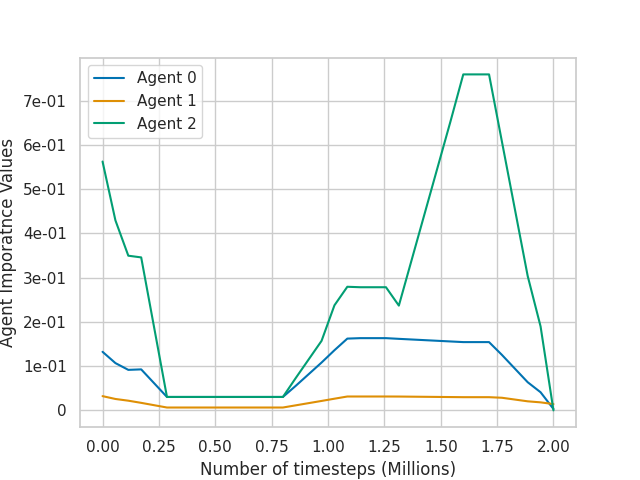}
    \caption{MAPPO agent importance}
  \end{subfigure}
  \begin{subfigure}{0.45\linewidth}
    \includegraphics[width=\linewidth]{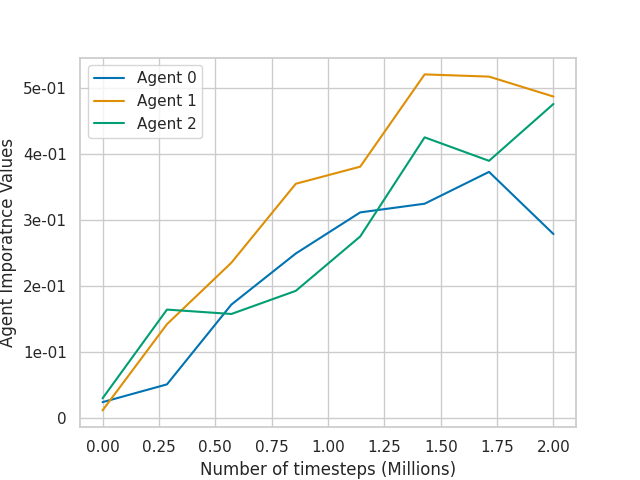}
    \caption{IQL agent importance}
  \end{subfigure}
\begin{subfigure}{0.32\linewidth}
    \includegraphics[width=\linewidth]{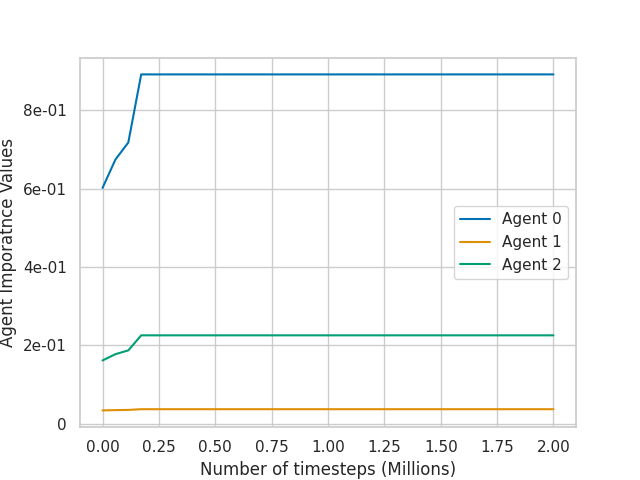}
    \caption{MAA2C agent importance}
  \end{subfigure}
\begin{subfigure}{0.32\linewidth}
    \includegraphics[width=\linewidth]{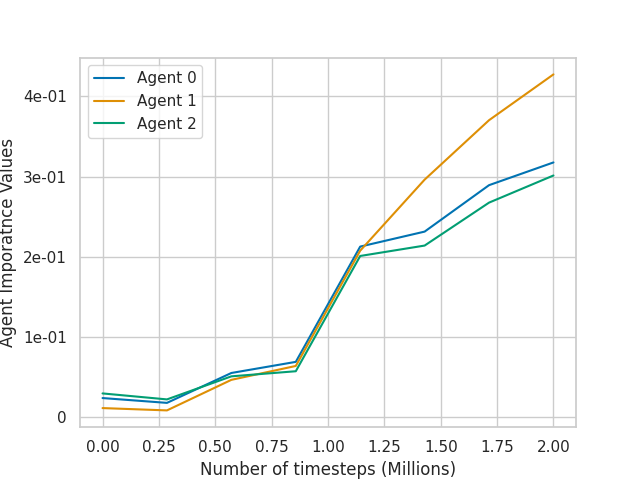}
    \caption{QMIX agent importance}
  \end{subfigure}
  \begin{subfigure}{0.32\linewidth}
    \includegraphics[width=\linewidth]{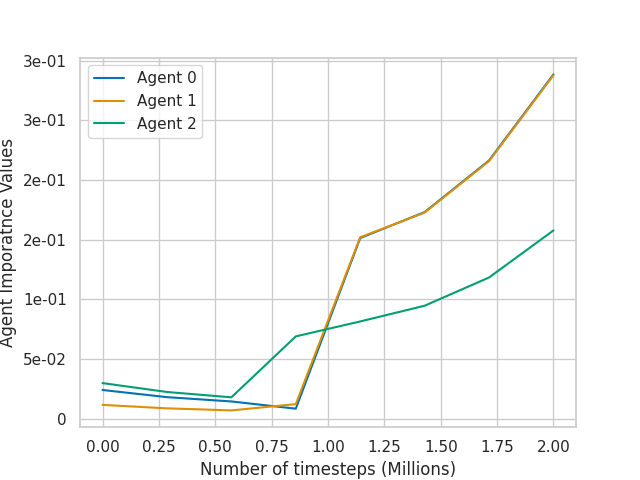}
    \caption{VDN agent importance}
  \end{subfigure}
  \caption{Agent importance plots for 3m from seed 0}
  \label{fig: agent_importance_3m}
\end{figure}

In the homogeneous setting of 3m, we can see from figure \ref{fig: agent_importance_3m} that agent importance follows a similar trend to the RWARE and LBF settings. As agents perform similar functions in the setting, their importance values are closely related. Agent importance also has a high percent ranking match rate with the Shapley values in this case as we can see in figure \ref{fig: ranking_matching_3m}.

\begin{figure}[h]
  \centering
\begin{subfigure}{0.45\linewidth}
    \includegraphics[width=\linewidth]{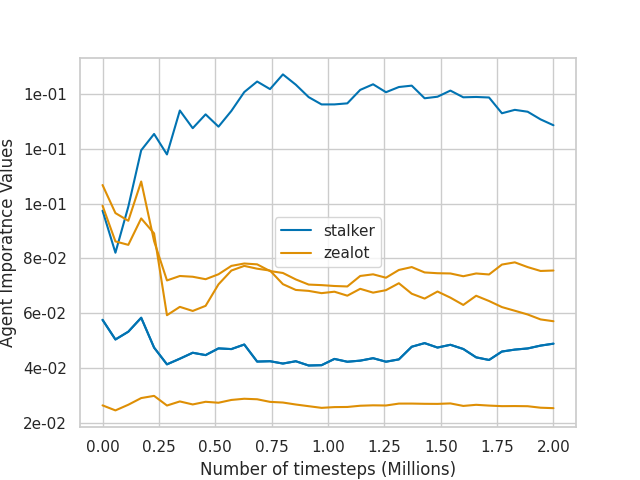}
    \caption{MAPPO agent importance}
  \end{subfigure}
  \begin{subfigure}{0.45\linewidth}
    \includegraphics[width=\linewidth]{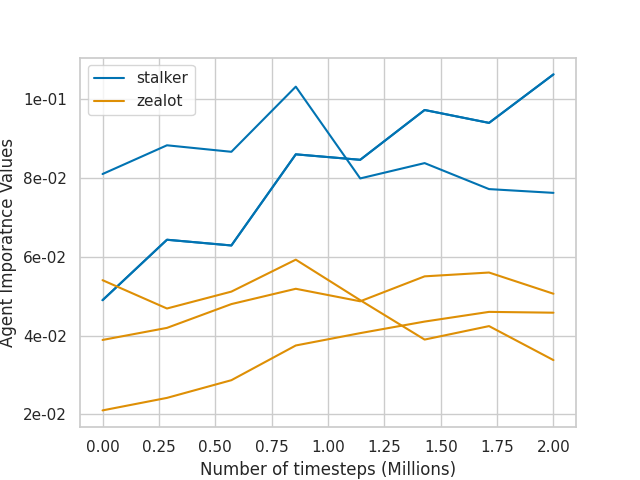}
    \caption{IQL agent importance}
  \end{subfigure}
\begin{subfigure}{0.32\linewidth}
    \includegraphics[width=\linewidth]{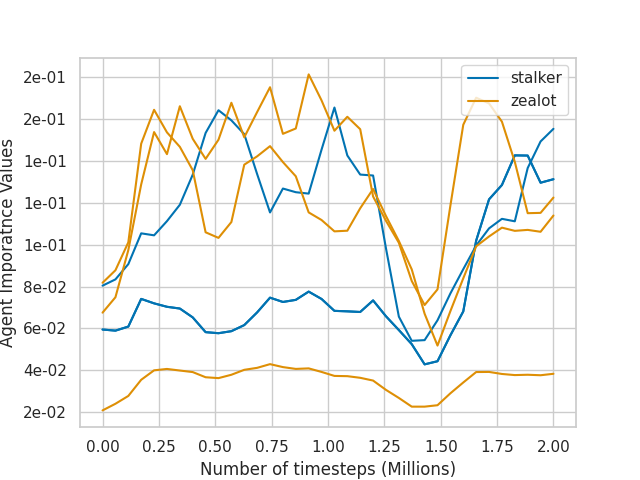}
    \caption{MAA2C agent importance}
  \end{subfigure}
\begin{subfigure}{0.32\linewidth}
    \includegraphics[width=\linewidth]{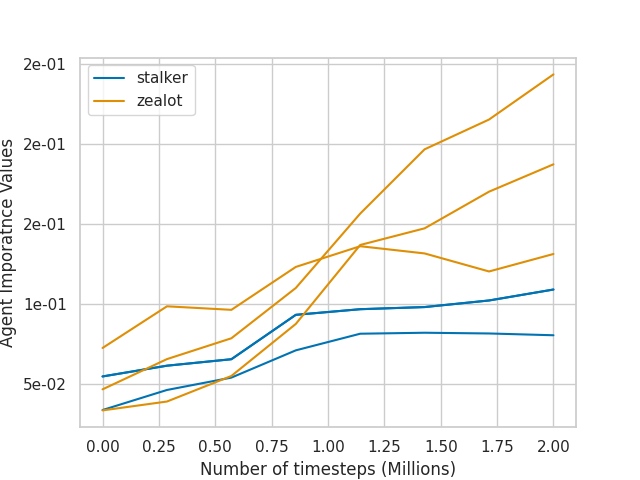}
    \caption{QMIX agent importance}
  \end{subfigure}
  \begin{subfigure}{0.32\linewidth}
    \includegraphics[width=\linewidth]{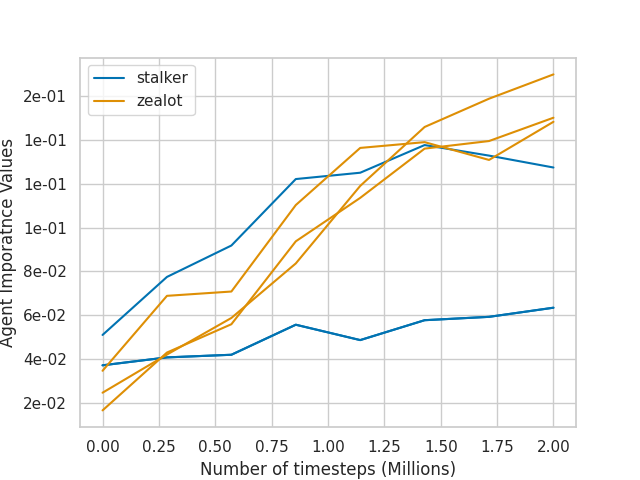}
    \caption{VDN agent importance}
  \end{subfigure}
  \caption{Agent importance plots for 3m from seed 0}
  \label{fig: agent_importance_2s3z}
\end{figure}

In the heterogeneous setting of 2s3z, we can see from figure \ref{fig: agent_importance_2s3z} that the agents tend to naturally separate into very distinct importance ranges. This is more distinct when stable converge has been reached as we can see with MAPPO where after reaching an optimal solution, the agents no longer have highly similar importance values. Comparatively when convergence is unstable like with MAA2C agent importance will oscillate. It is also notable that even agents of the same type can have high variation in contribution score at the end to training. This is inline with existing literature like \citep{yang2020qatten,singh2023challenge} which have show that the importance of different agent agent types varies across the settings of the original SMAC and that importance cannot be assigned uniformly. Additionally as agents in SMAC can die during the episode rollout, the relative importance of the remaining agents increases as dead agents cannot contribute to the coalition which can result in agents of the same type have different assigned weighting based on how long they are able to survive.

\subsection{Parameter Sharing for MMM2} \label{param_sharing_MMM2}

\begin{figure}[h]
  \centering
\begin{subfigure}{0.23\textwidth}
    \includegraphics[width=\textwidth]{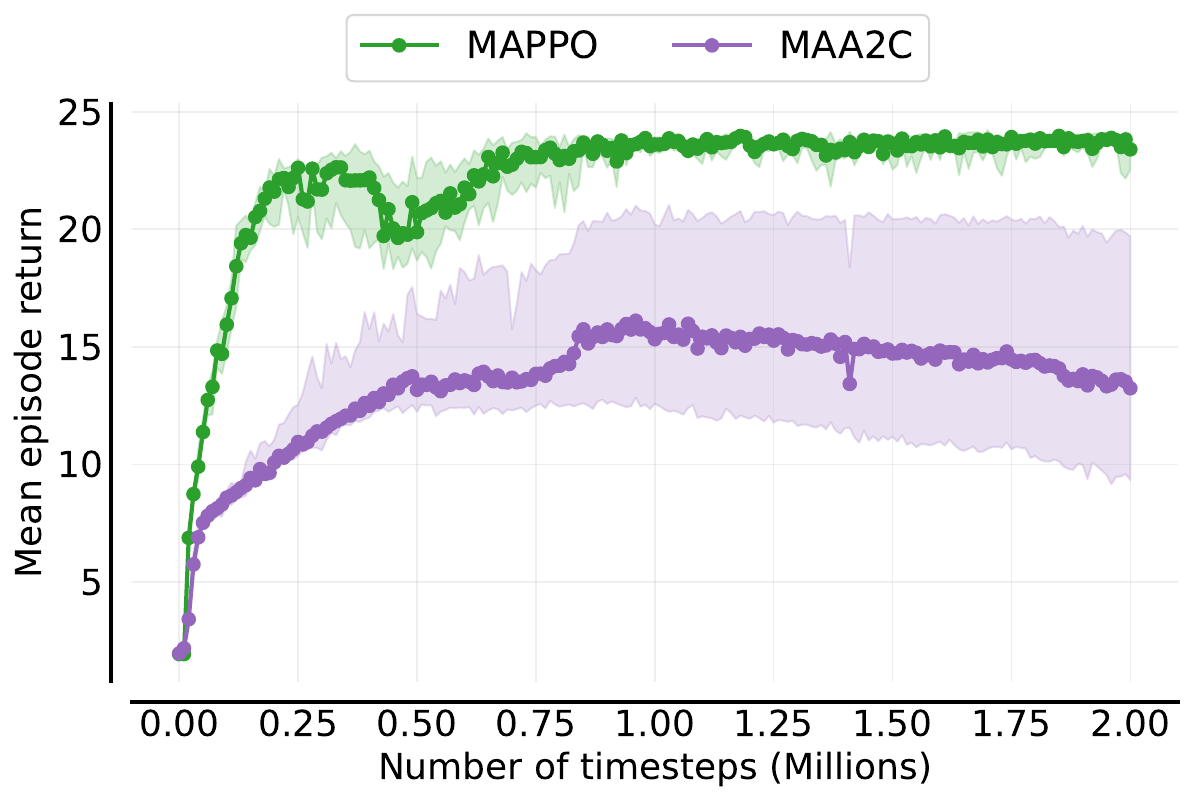}
  \end{subfigure}
  \begin{subfigure}{0.23\textwidth}
    \includegraphics[width=\textwidth]{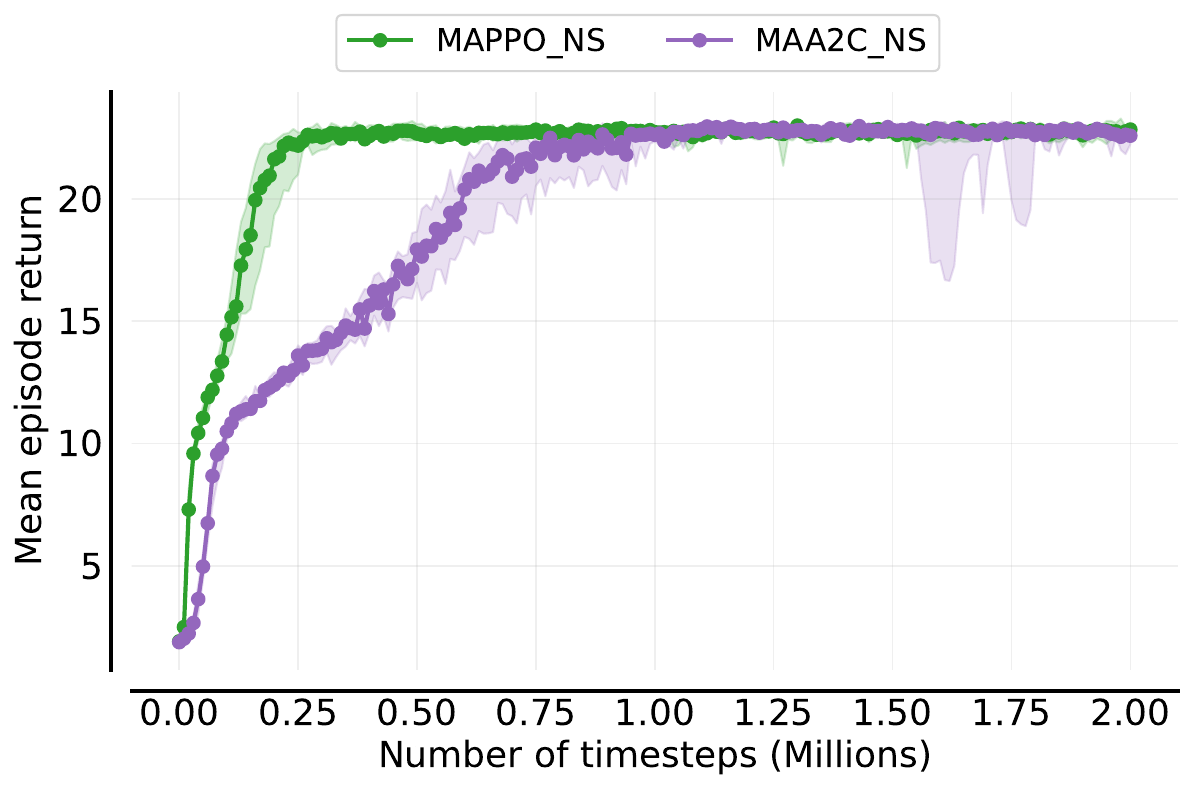}
  \end{subfigure}
  \caption{\textbf{Left:} Mean episode return for MMM2 with parameter sharing. \textbf{Righ:} Mean episode return for MMM2 with without parameter sharing.}
  \label{fig: MMM2_NS_COMPARE}
\end{figure}

We perform additional experimentation on MMM2 to gain insight into how parameter sharing affects agent importance in the heterogeneous case. We can see from figure \ref{fig: MMM2_NS_COMPARE} that unlike in LBF and RWARE, parameter sharing seems to degrade performance. This is most noticeable for MAA2C which is unable to converge to a stable policy across 6 seeds. This is expected as \citep{wen2022multiagent} provide theoretical evidence towards parameter sharing reducing effectiveness in heterogeneous settings. We can also observe that like the findings on SMAC by \citep{wen2022multiagent} showing that it is not a challenging enough to compare parameter vs non-parameter sharing for SOTA algorithms like HATRPO, the MMM2 setting that has been ported to SMAClite does also not show a large change in performance for MAPPO in both the shared and non-shared cases.

\begin{figure}
  \centering
  \begin{subfigure}[t]{0.23\textwidth}
    \includegraphics[width=\textwidth, valign=t]{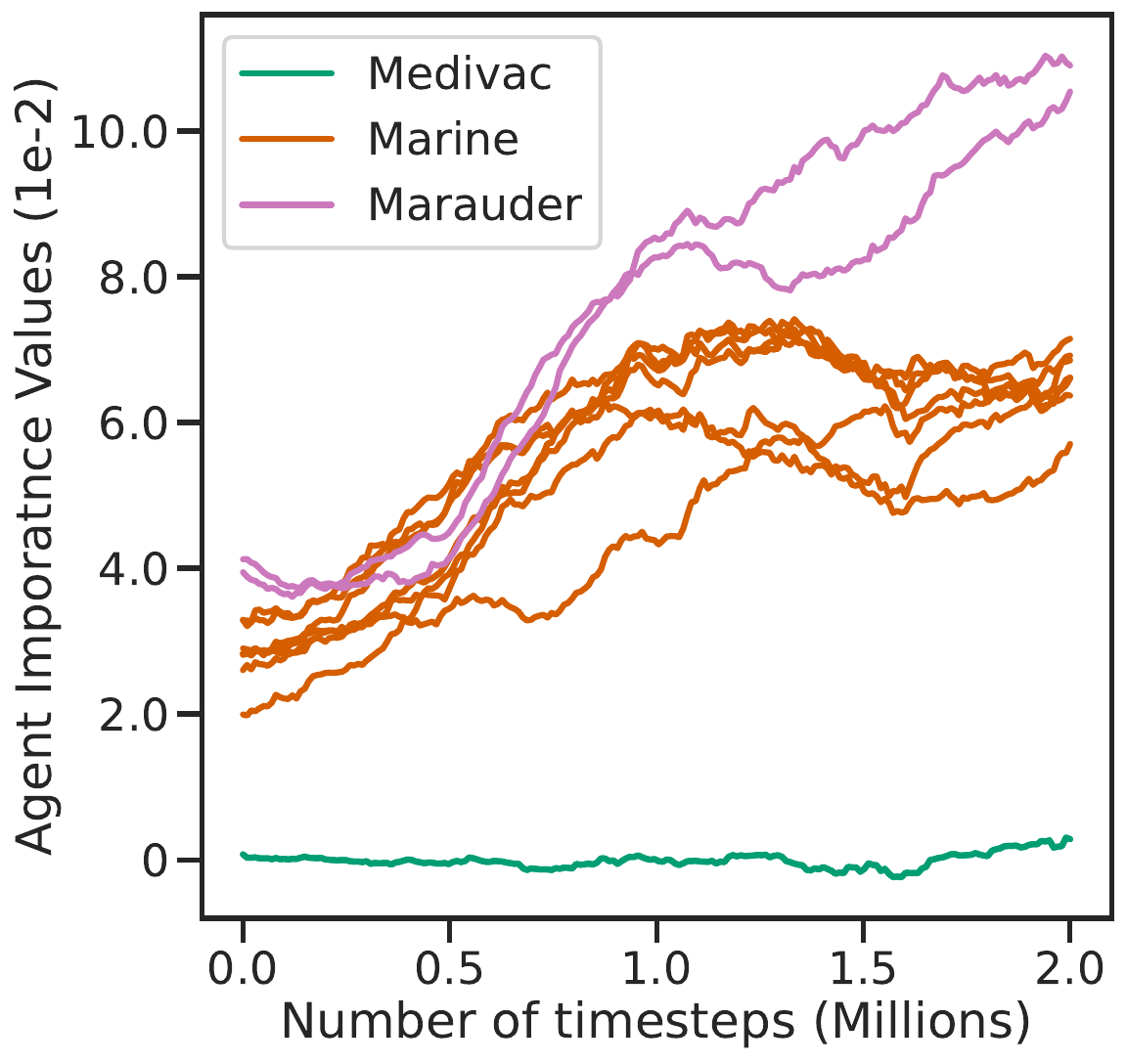}
  \end{subfigure}
  \begin{subfigure}[t]{0.23\textwidth}
    \includegraphics[width=\textwidth, valign=t]{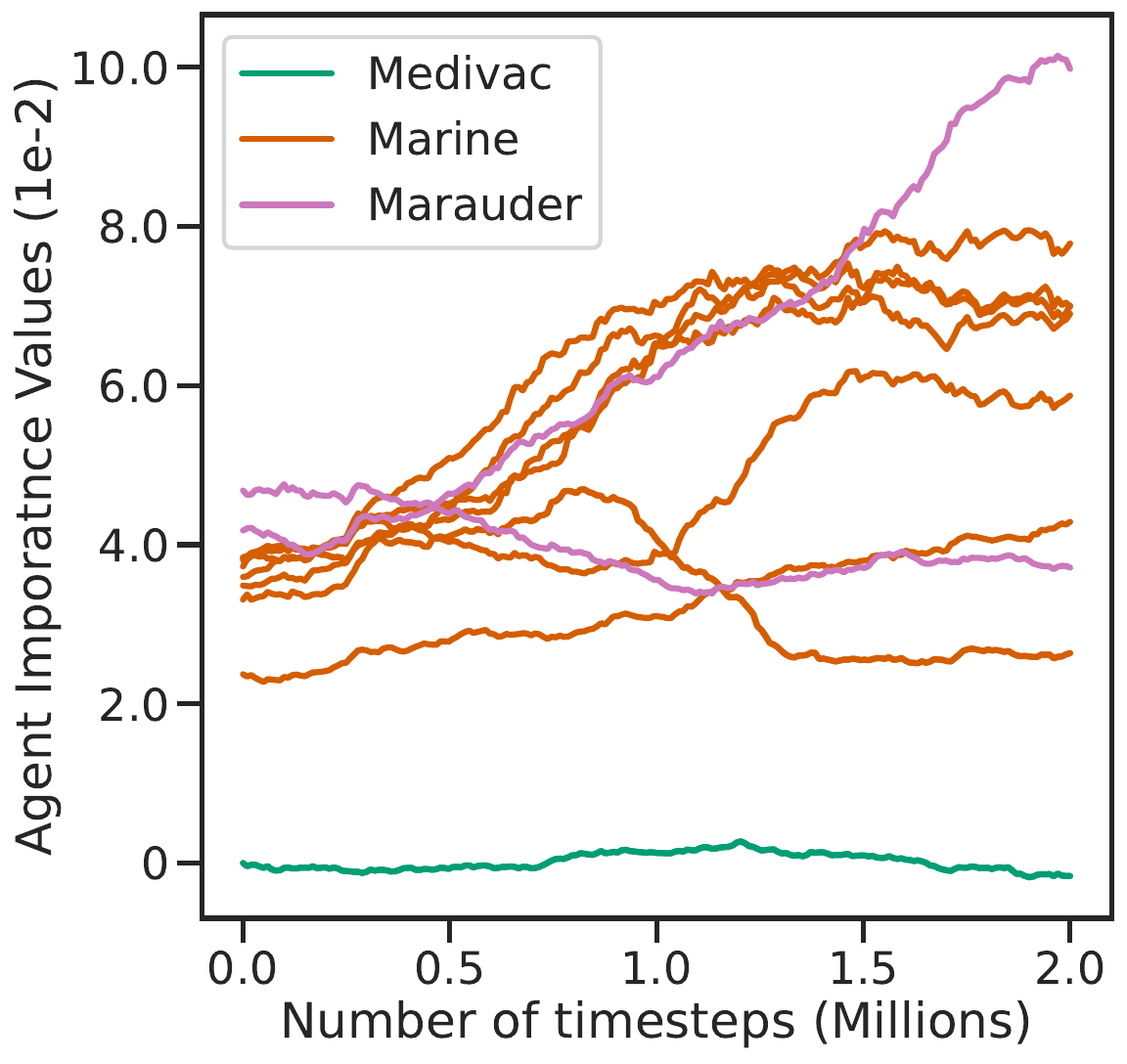}
  \end{subfigure}
  \begin{subfigure}[t]{0.15\textwidth}
    \includegraphics[width=\textwidth, valign=t]{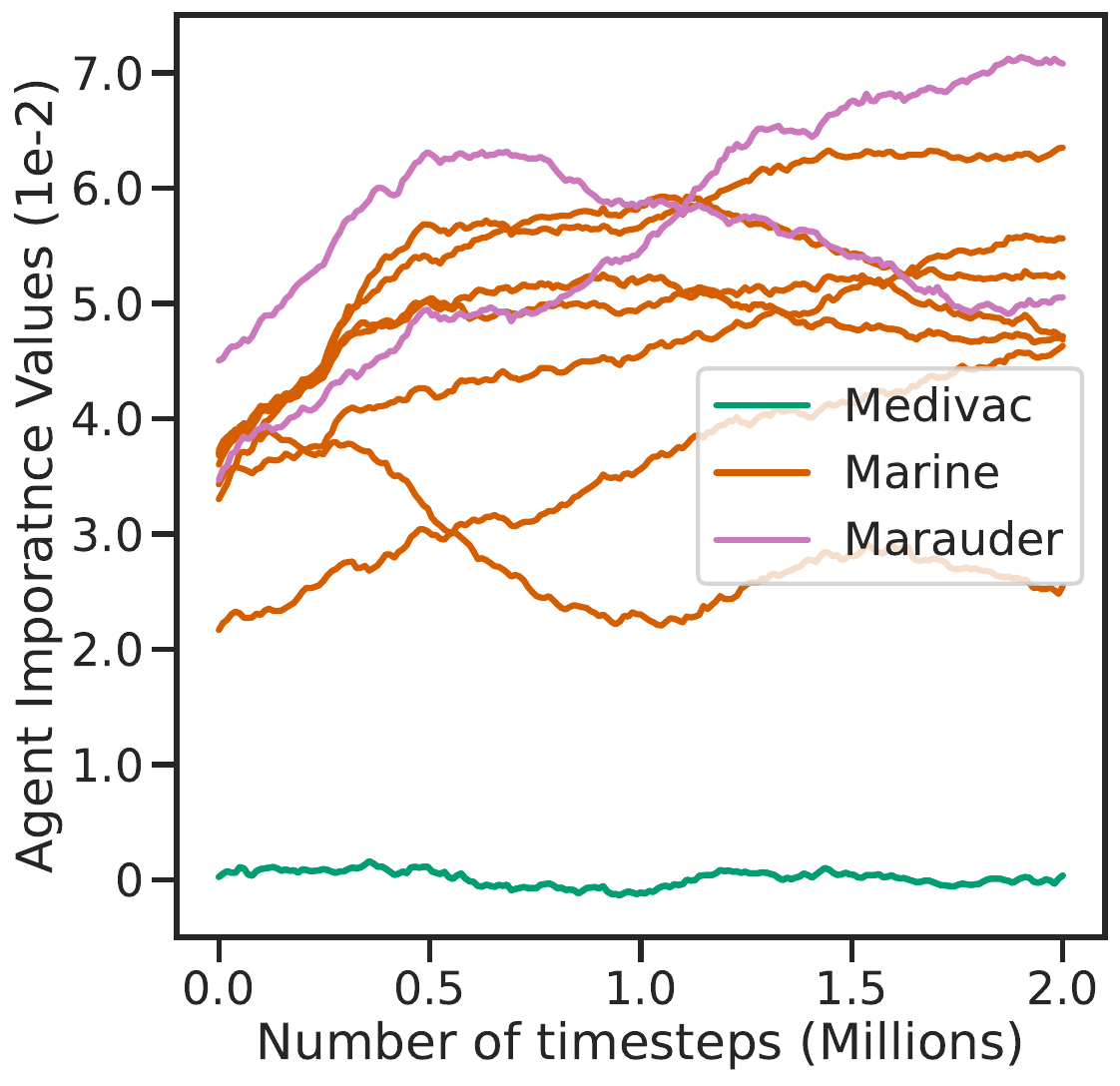}
  \end{subfigure}
    \begin{subfigure}[t]{0.15\textwidth}
    \includegraphics[width=\textwidth, valign=t]{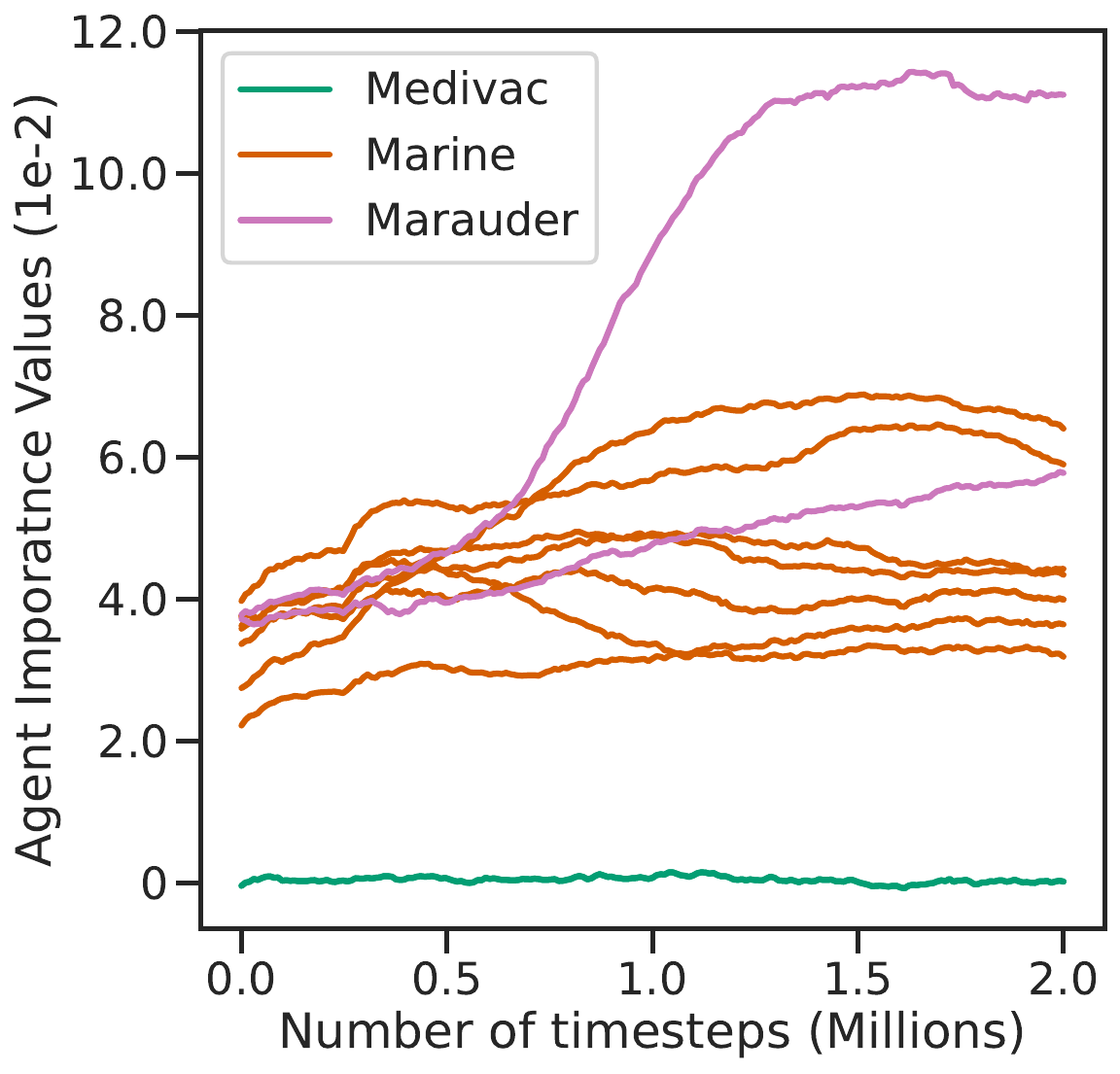}
  \end{subfigure}
  \begin{subfigure}[t]{0.15\textwidth}
    \includegraphics[width=\textwidth, valign=t]{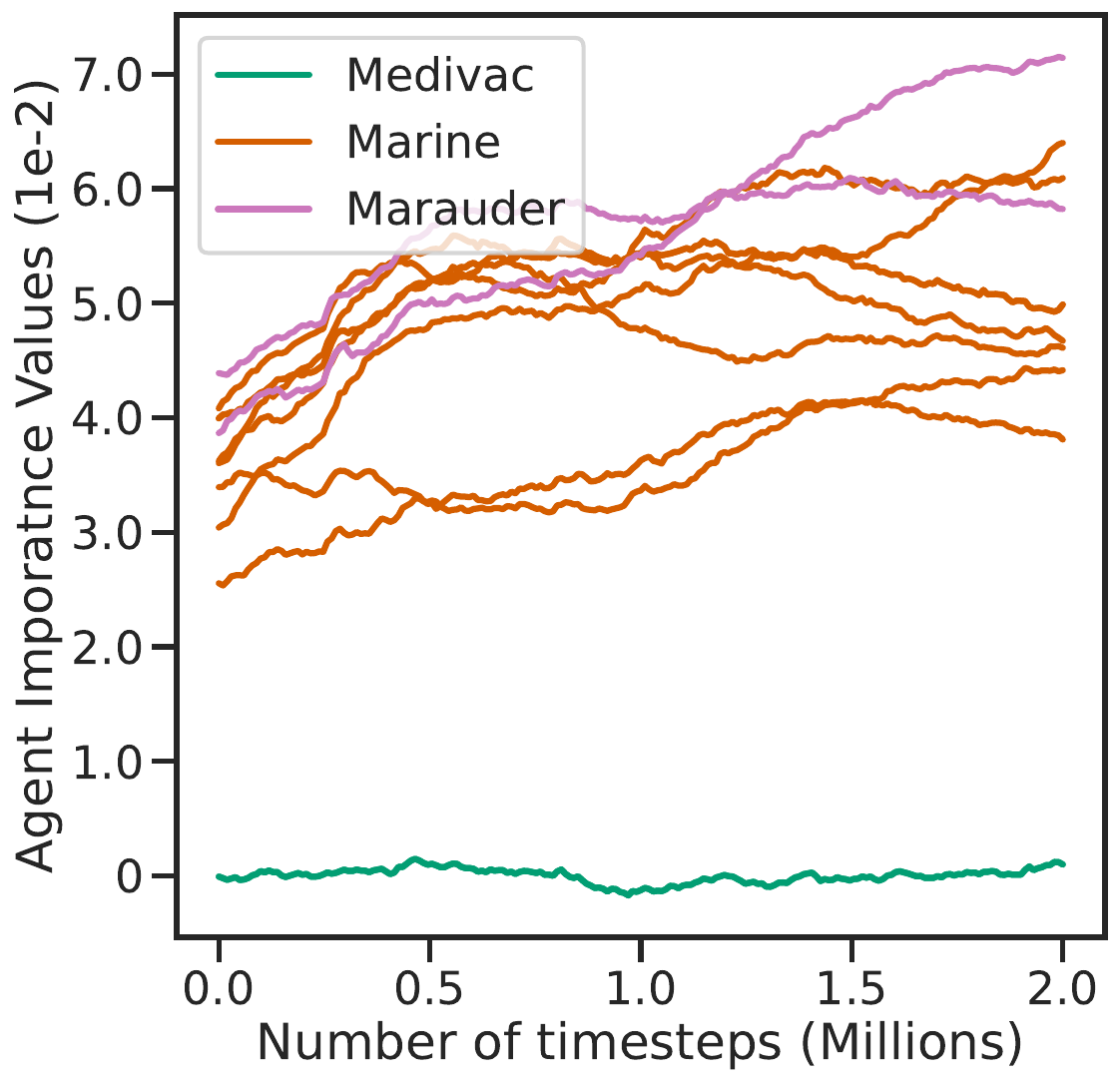}
  \end{subfigure}
   \caption{Agent importance plots for MAA2C with parameter sharing across 5 seeds in MMM2}
  \label{fig: MAA2C_PS_MMM2_ALL}
\end{figure}

\begin{figure}
  \centering
  \begin{subfigure}[t]{0.23\textwidth}
    \includegraphics[width=\textwidth, valign=t]{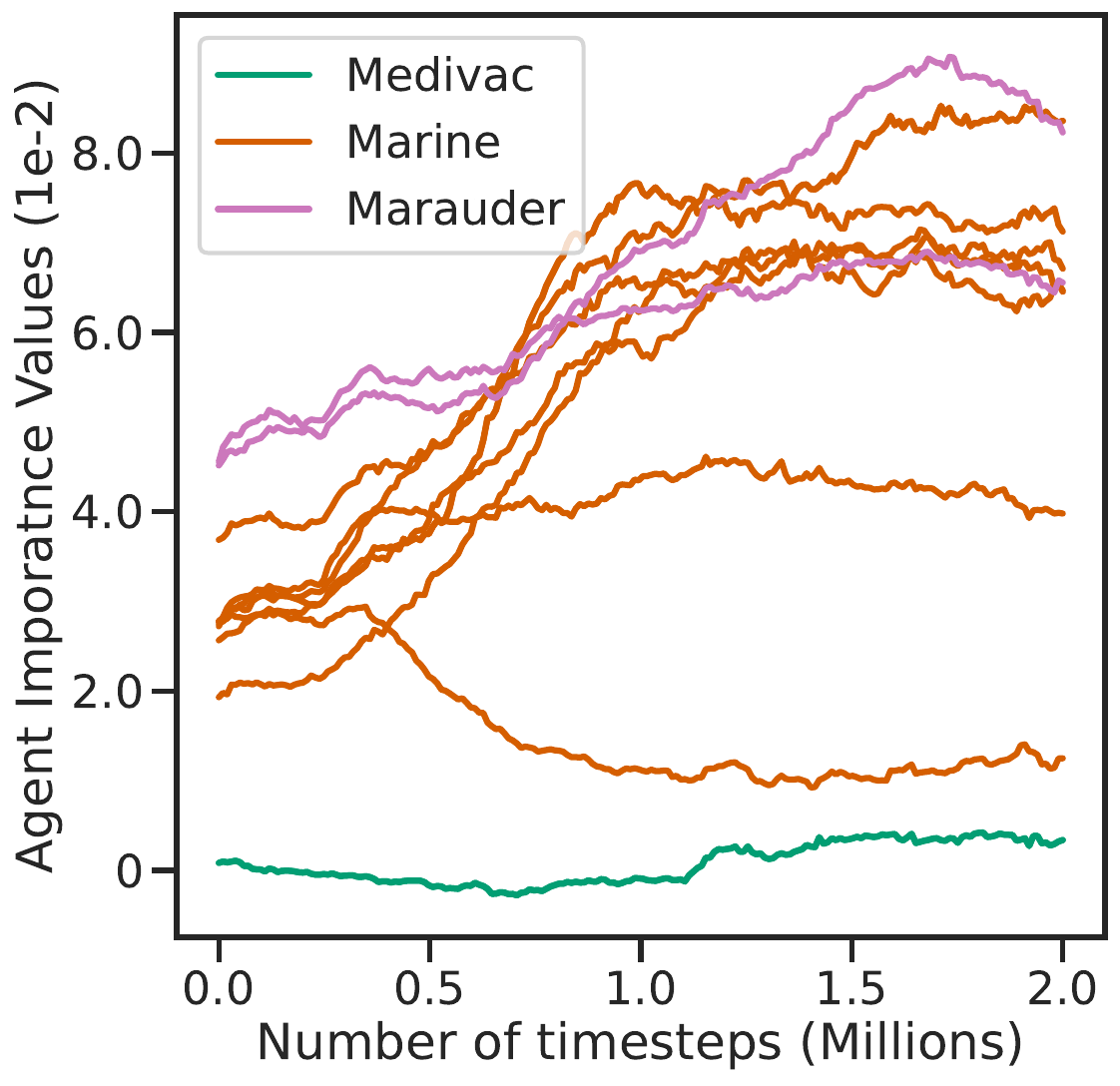}
  \end{subfigure}
  \begin{subfigure}[t]{0.23\textwidth}
    \includegraphics[width=\textwidth, valign=t]{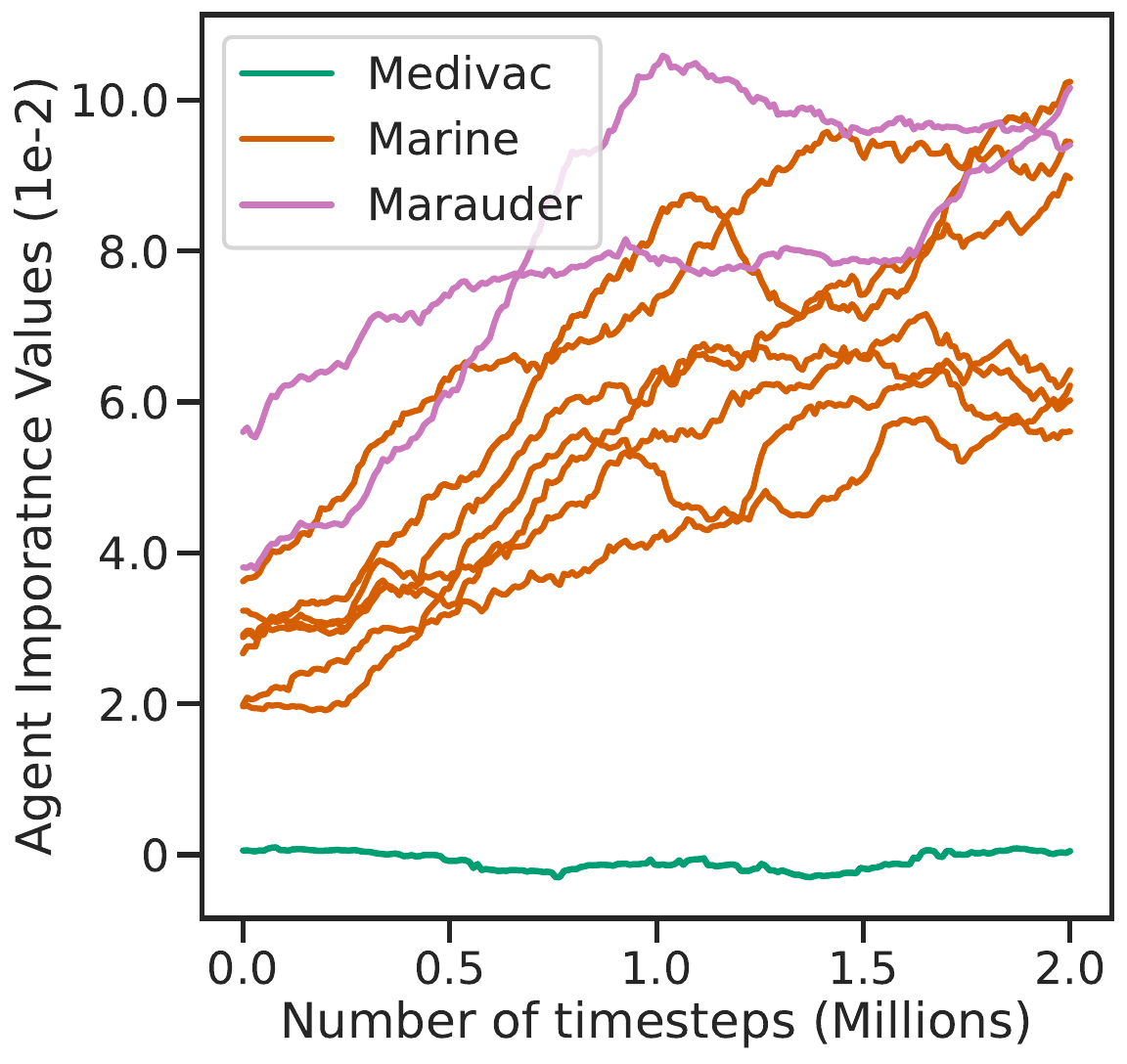}
  \end{subfigure}
  \begin{subfigure}[t]{0.15\textwidth}
    \includegraphics[width=\textwidth, valign=t]{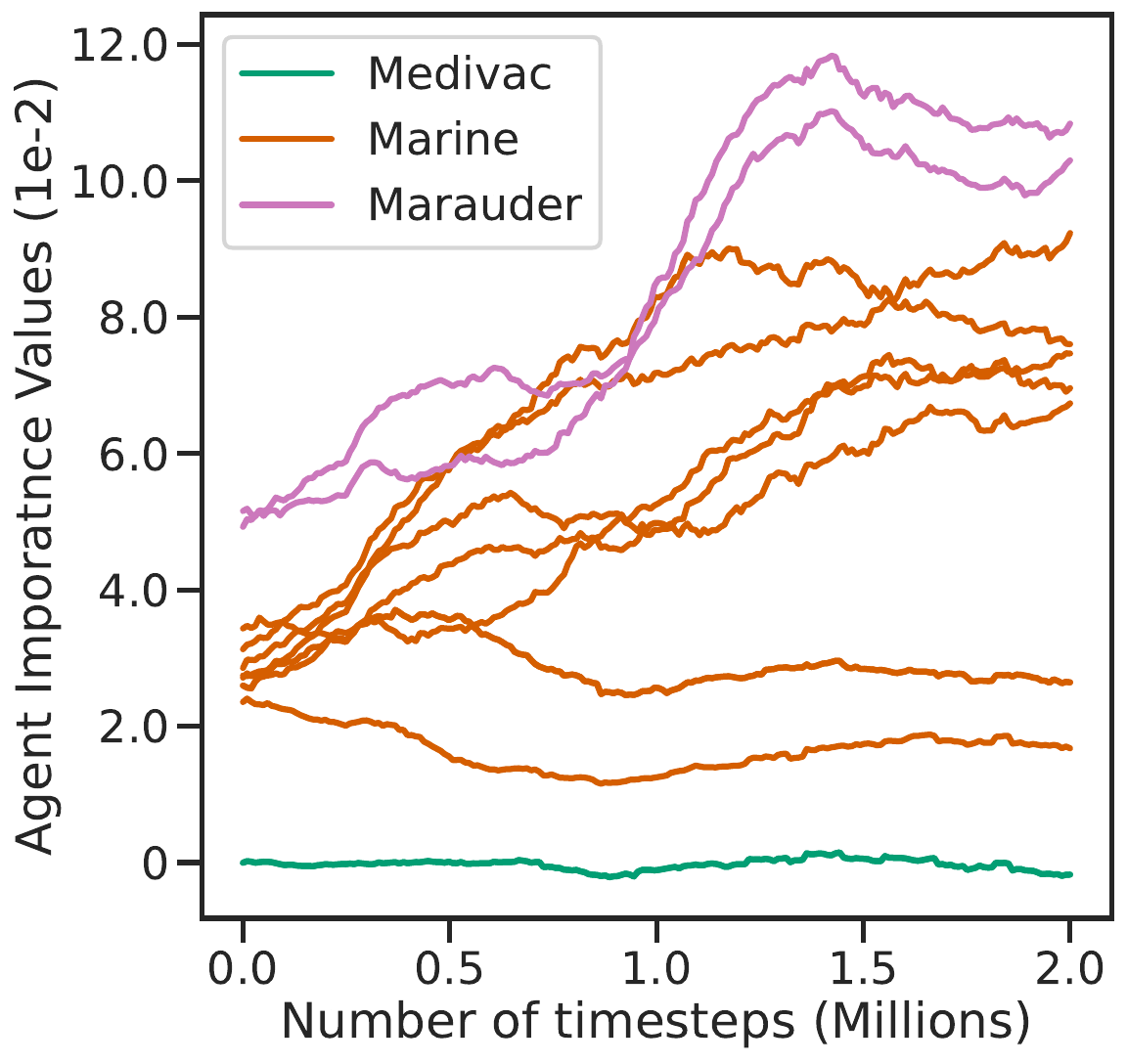}
  \end{subfigure}
    \begin{subfigure}[t]{0.15\textwidth}
    \includegraphics[width=\textwidth, valign=t]{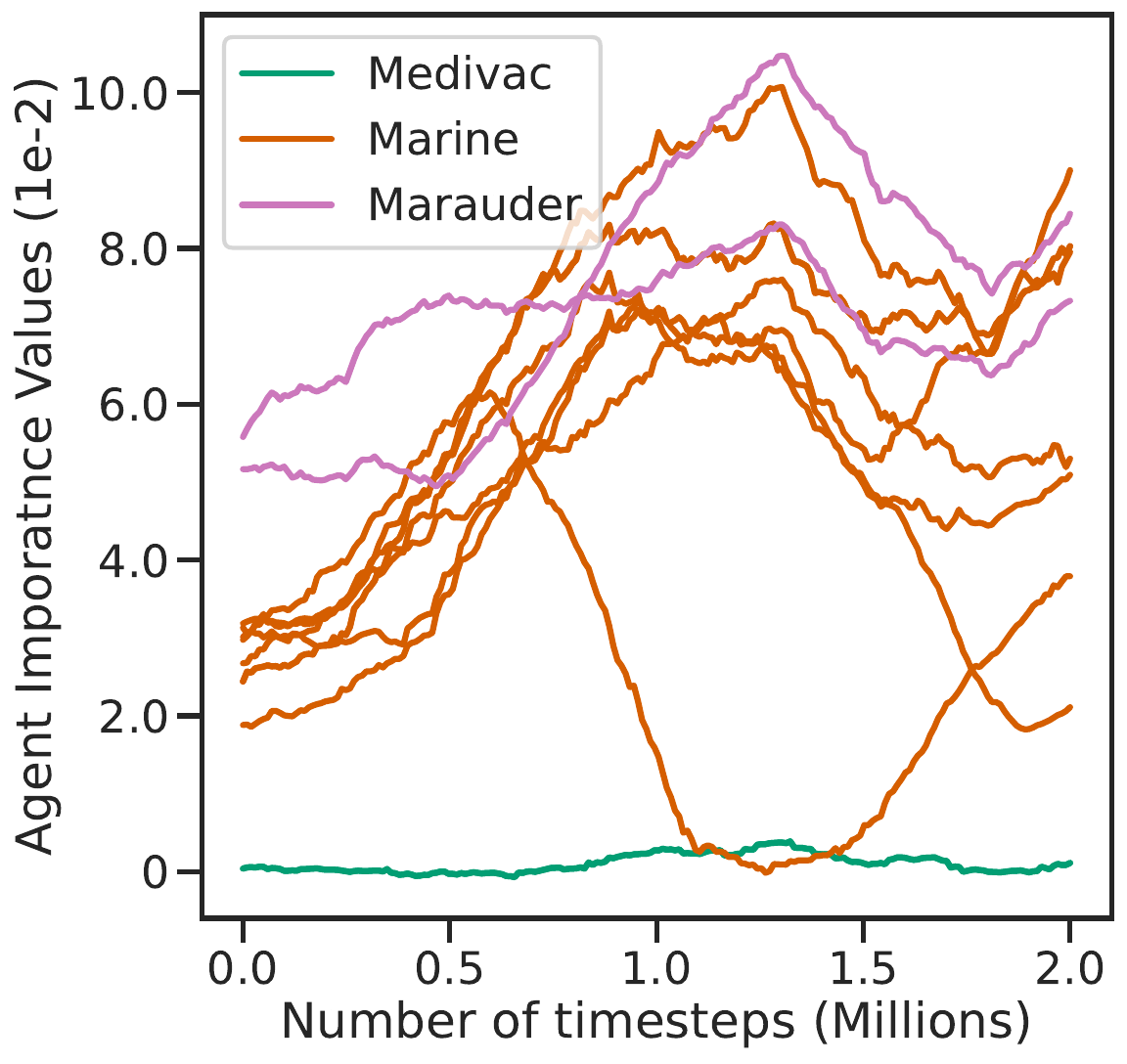}
  \end{subfigure}
  \begin{subfigure}[t]{0.15\textwidth}
    \includegraphics[width=\textwidth, valign=t]{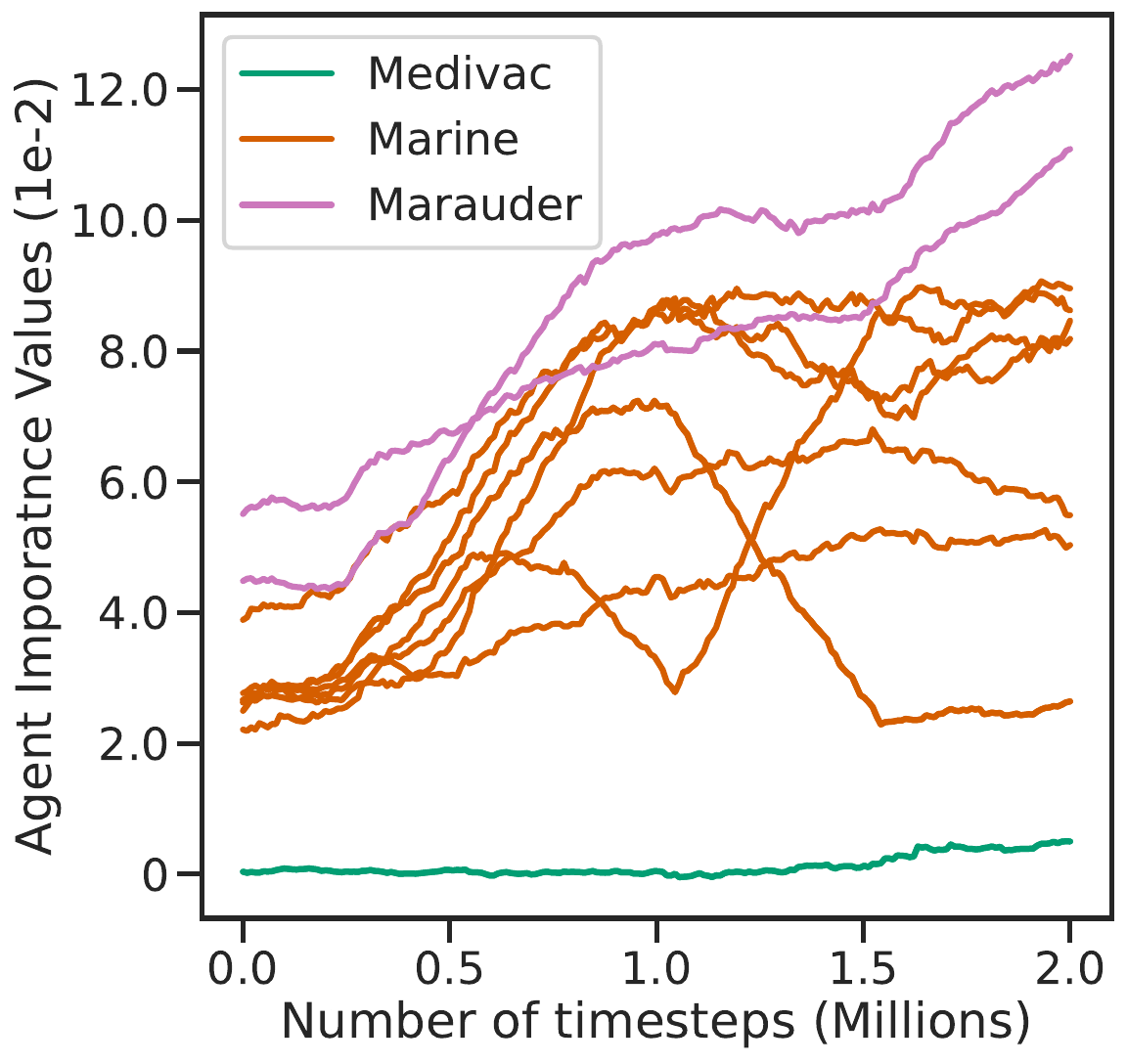}
  \end{subfigure}
   \caption{Agent importance plots for MAA2C without parameter sharing across 5 seeds in MMM2}
  \label{fig: MAA2C_NS_MMM2_ALL}
\end{figure}

From figures \ref{fig: MAA2C_PS_MMM2_ALL} and \ref{fig: MAA2C_NS_MMM2_ALL} we can see that the assigned agent importance rankings vary greatly between the parameter sharing (PS) and no-sharing (NS) results on MMM2. In the NS case the algorithm is able to learn clearer distinctions between the subgroups and doe not consistently over weight the value of the single marauder as it does in the PS case.

\begin{figure}
  \centering
  \begin{subfigure}[t]{0.23\textwidth}
    \includegraphics[width=\textwidth, valign=t]{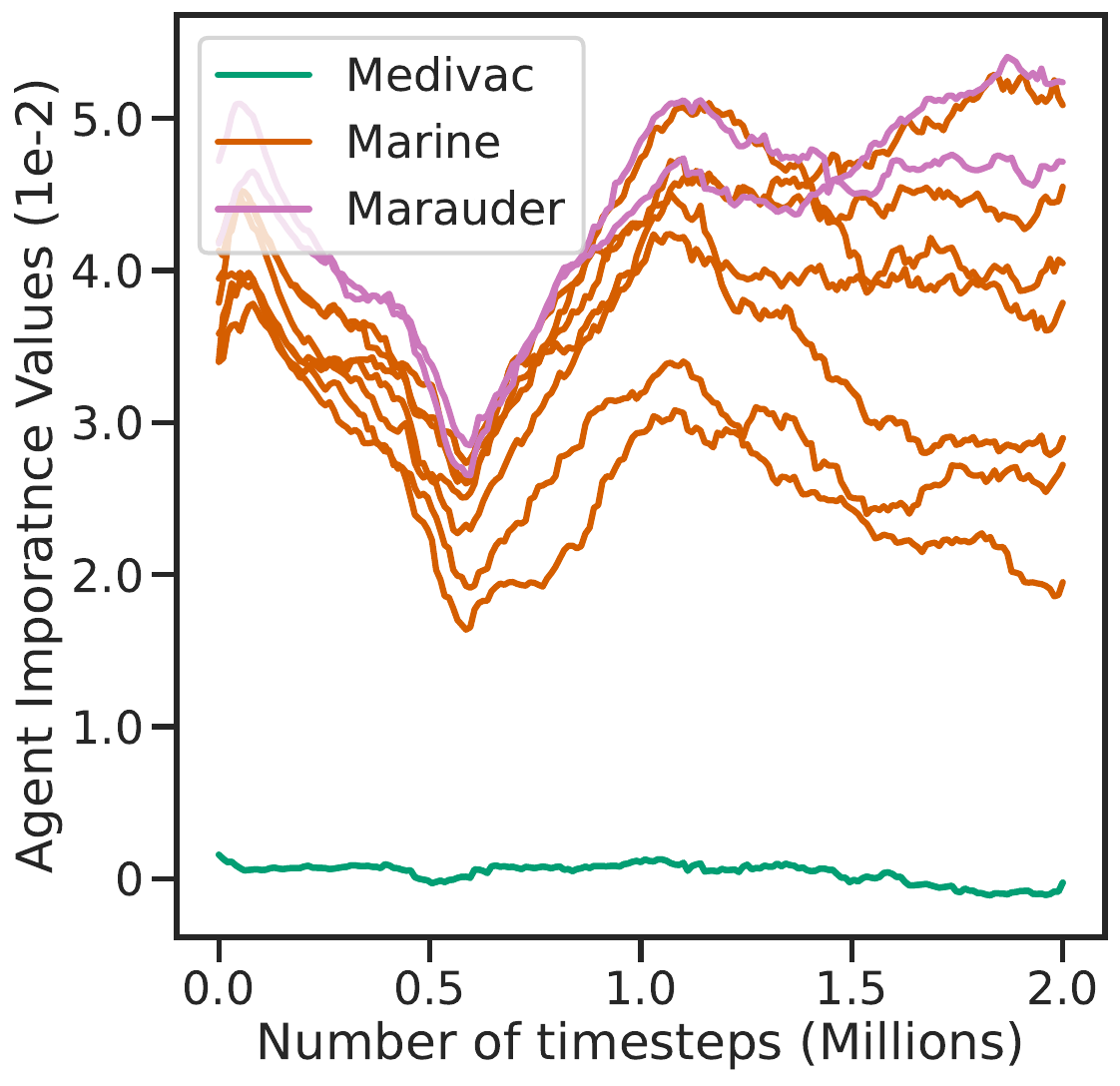}
  \end{subfigure}
  \begin{subfigure}[t]{0.23\textwidth}
    \includegraphics[width=\textwidth, valign=t]{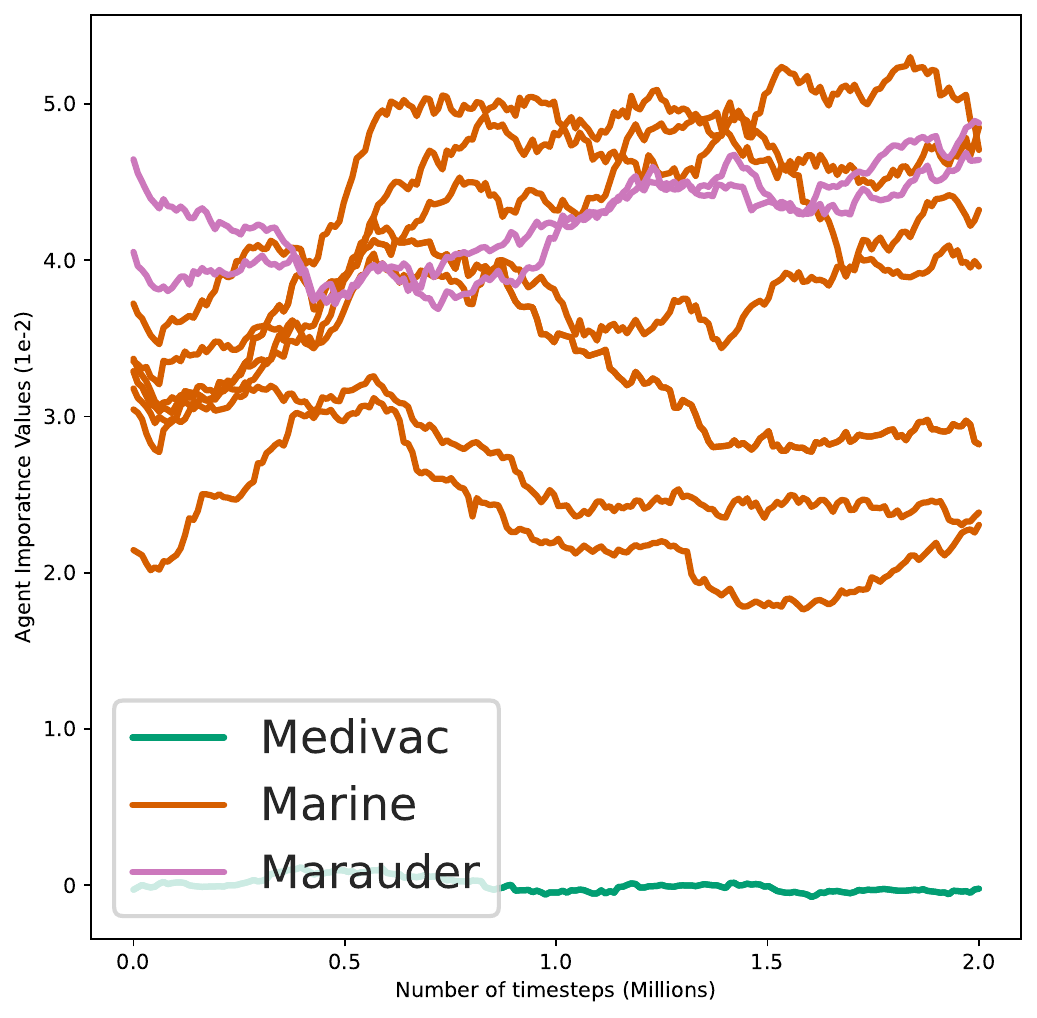}
  \end{subfigure}
  \begin{subfigure}[t]{0.15\textwidth}
    \includegraphics[width=\textwidth, valign=t]{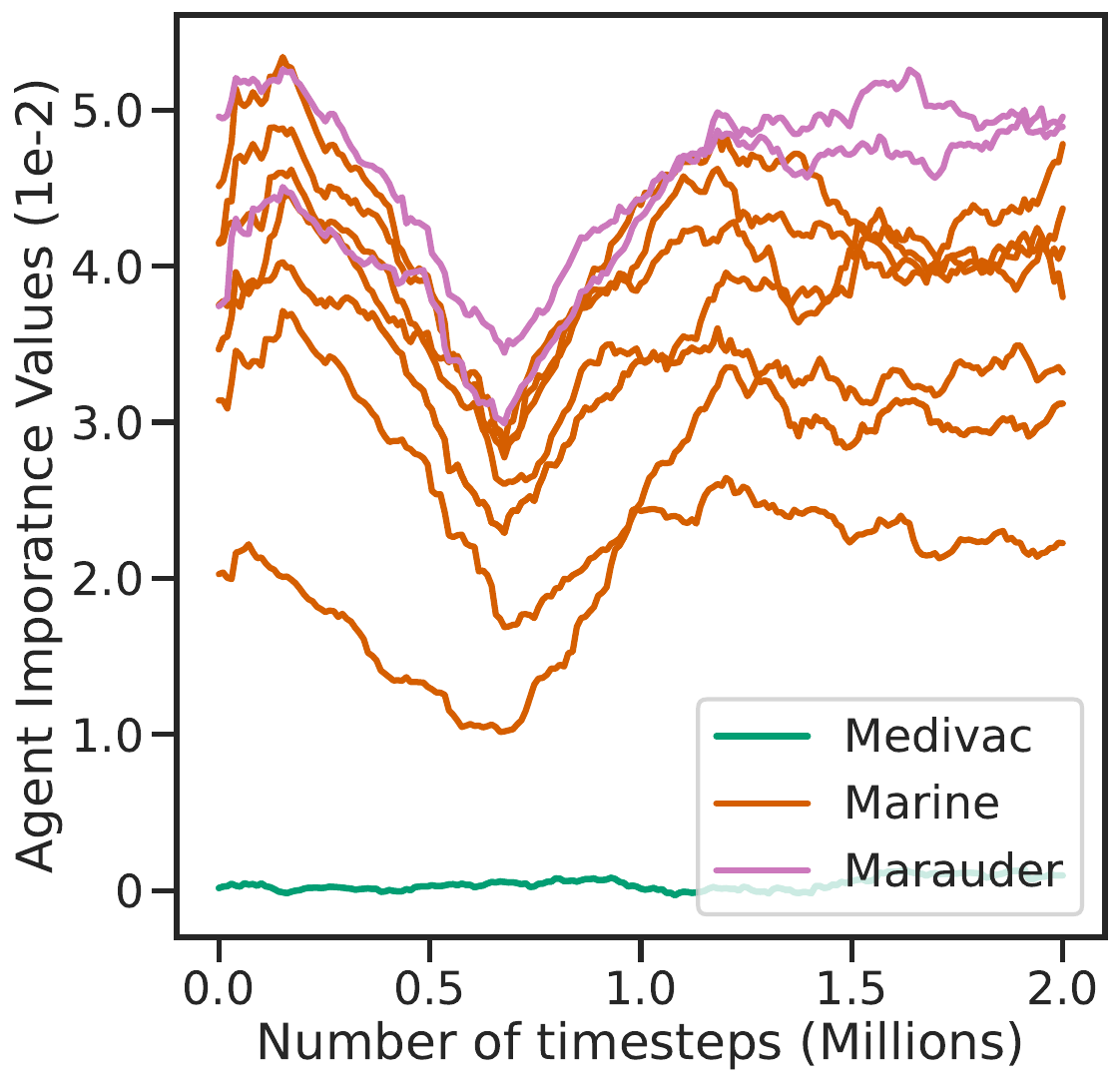}
  \end{subfigure}
    \begin{subfigure}[t]{0.15\textwidth}
    \includegraphics[width=\textwidth, valign=t]{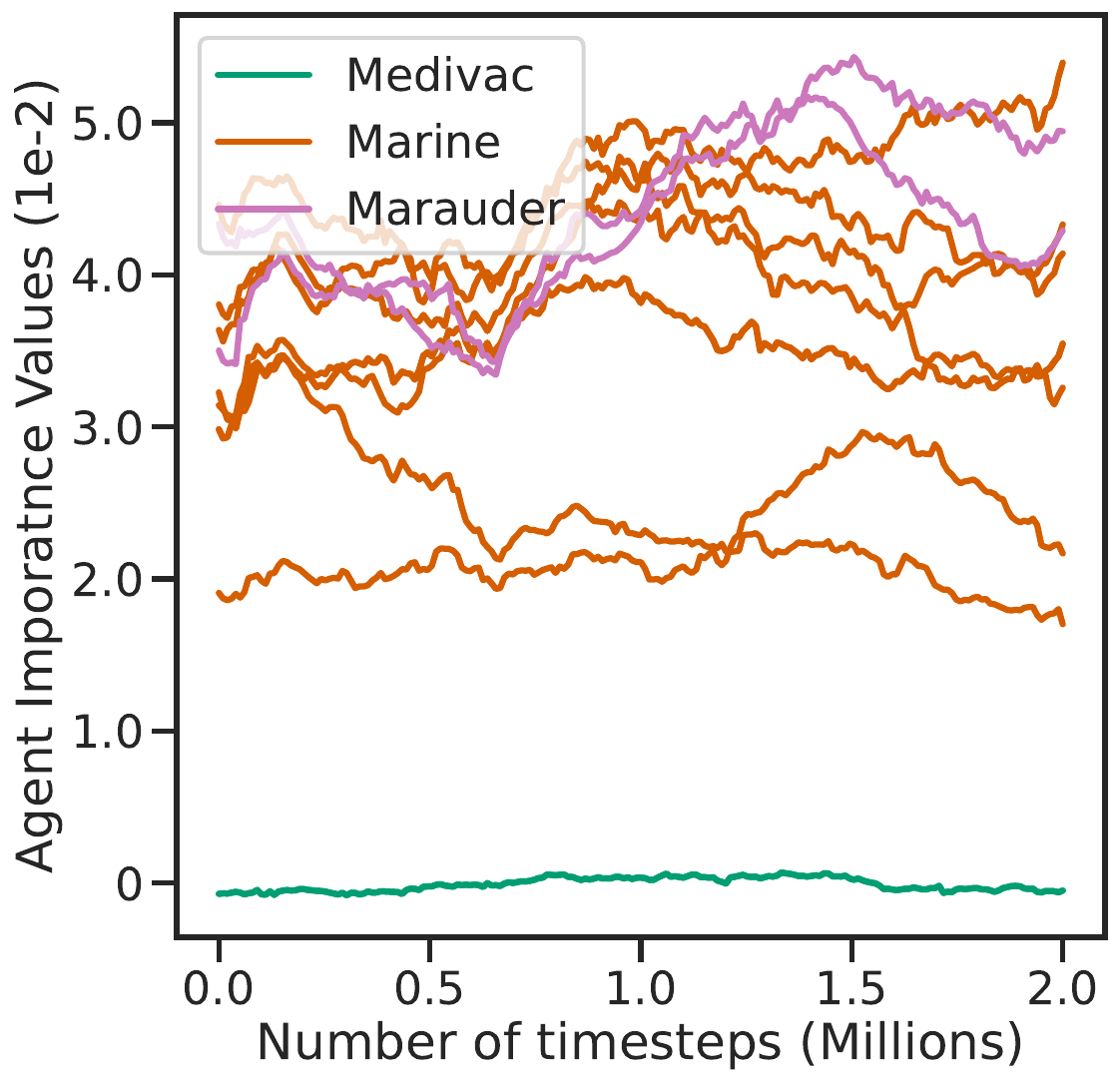}
  \end{subfigure}
  \begin{subfigure}[t]{0.15\textwidth}
    \includegraphics[width=\textwidth, valign=t]{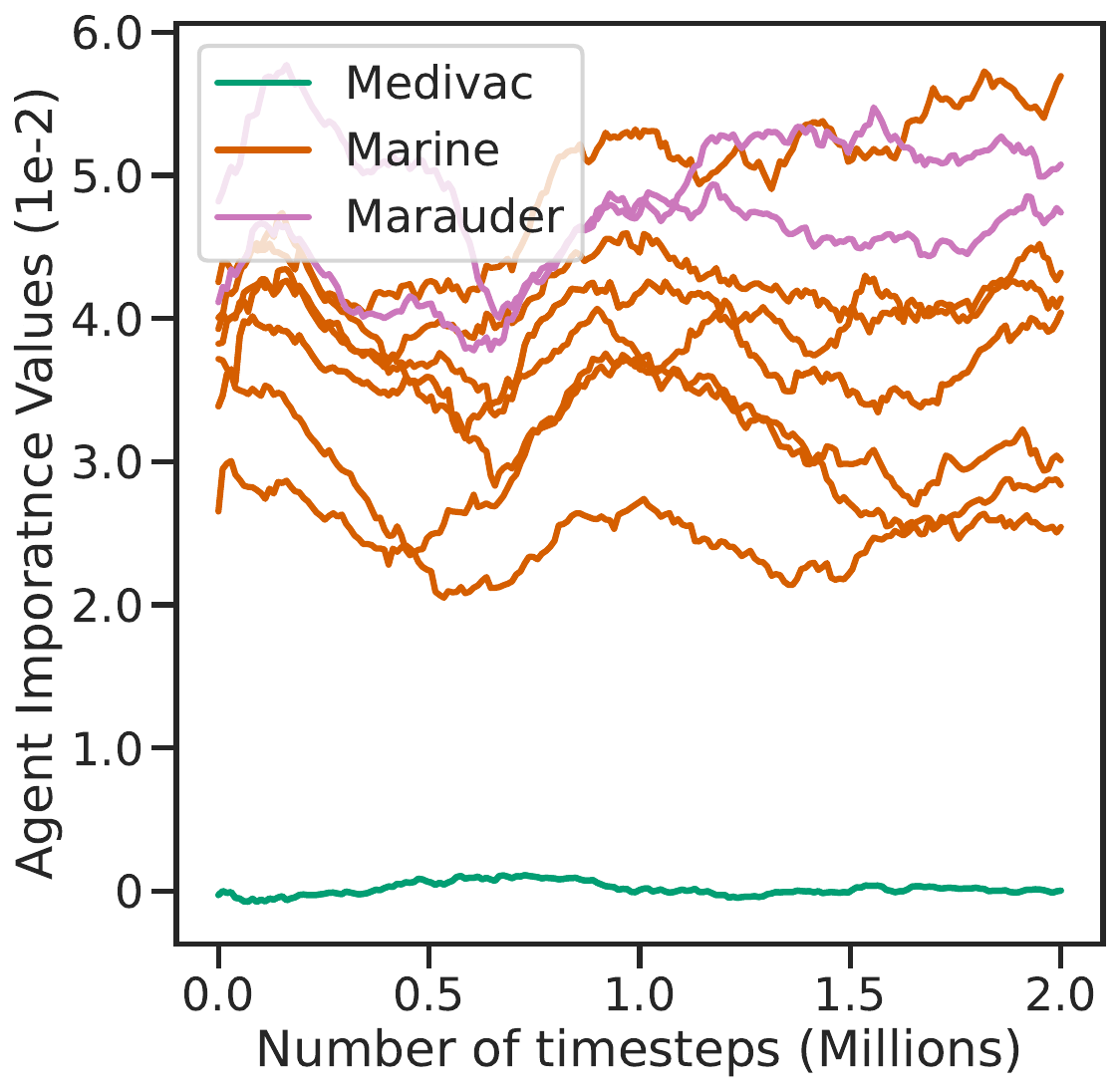}
  \end{subfigure}
   \caption{Agent importance plots for MAPPO with parameter sharing across 5 seeds in MMM2}
  \label{fig: MAPPO_PS_MMM2_ALL}
\end{figure}

\begin{figure}
  \centering
  \begin{subfigure}[t]{0.23\textwidth}
    \includegraphics[width=\textwidth, valign=t]{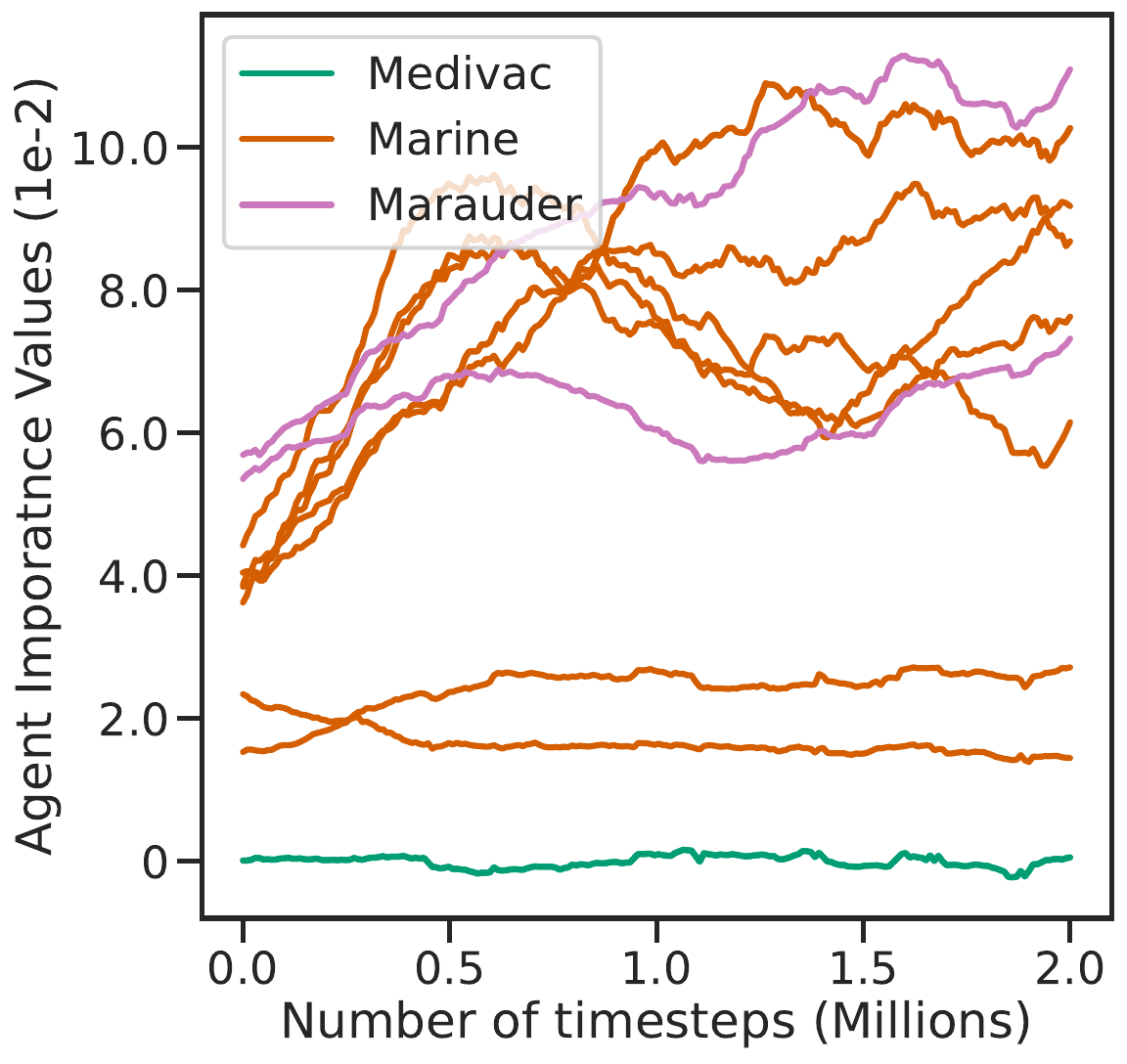}
  \end{subfigure}
  \begin{subfigure}[t]{0.23\textwidth}
    \includegraphics[width=\textwidth, valign=t]{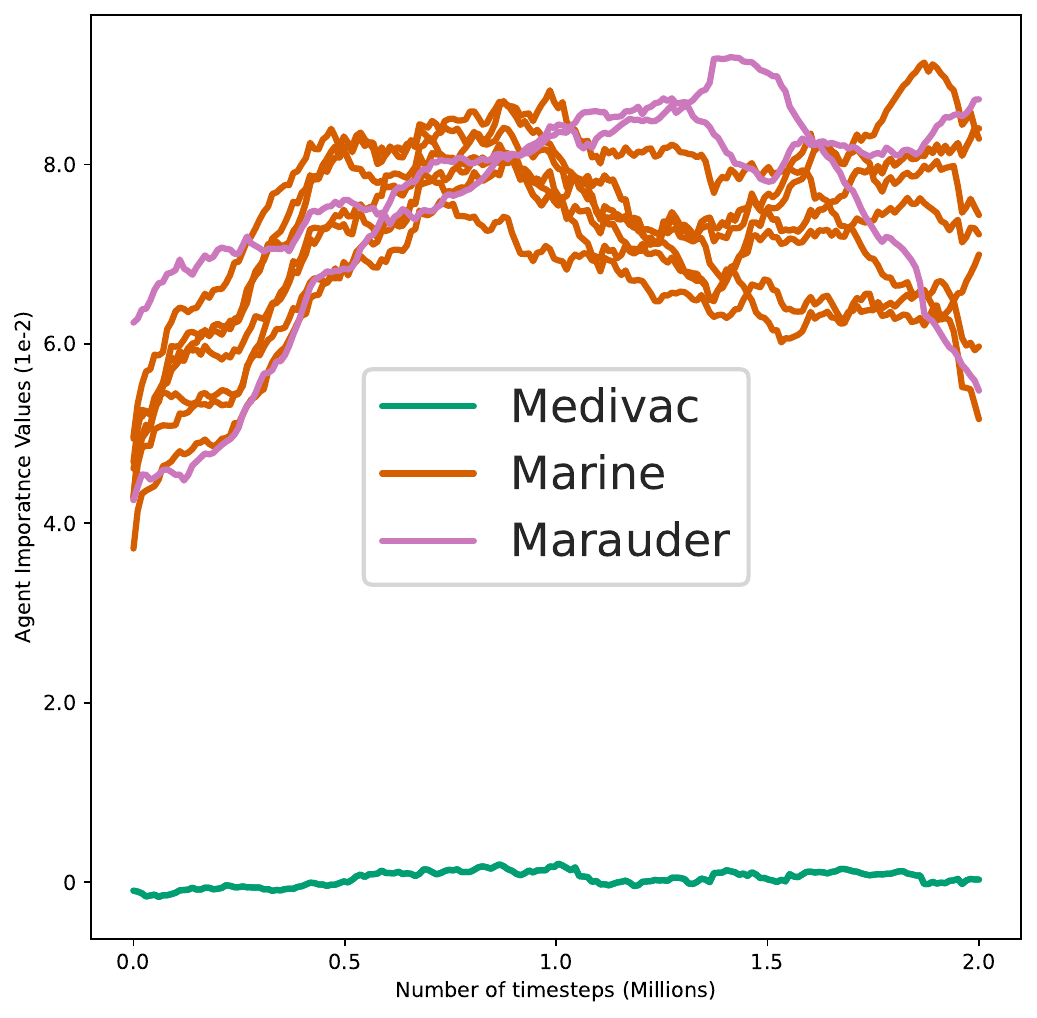}
  \end{subfigure}
  \begin{subfigure}[t]{0.15\textwidth}
    \includegraphics[width=\textwidth, valign=t]{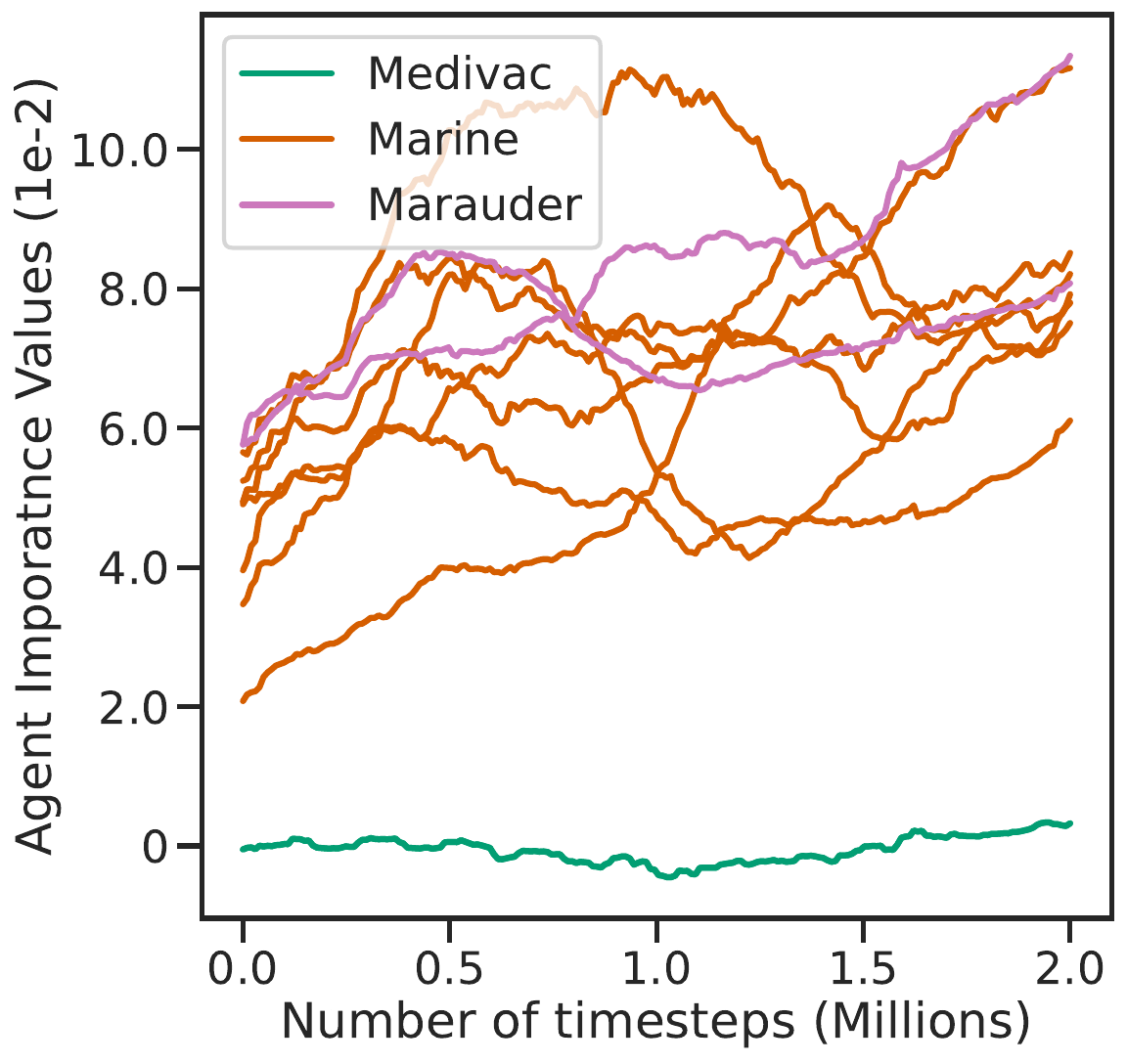}
  \end{subfigure}
    \begin{subfigure}[t]{0.15\textwidth}
    \includegraphics[width=\textwidth, valign=t]{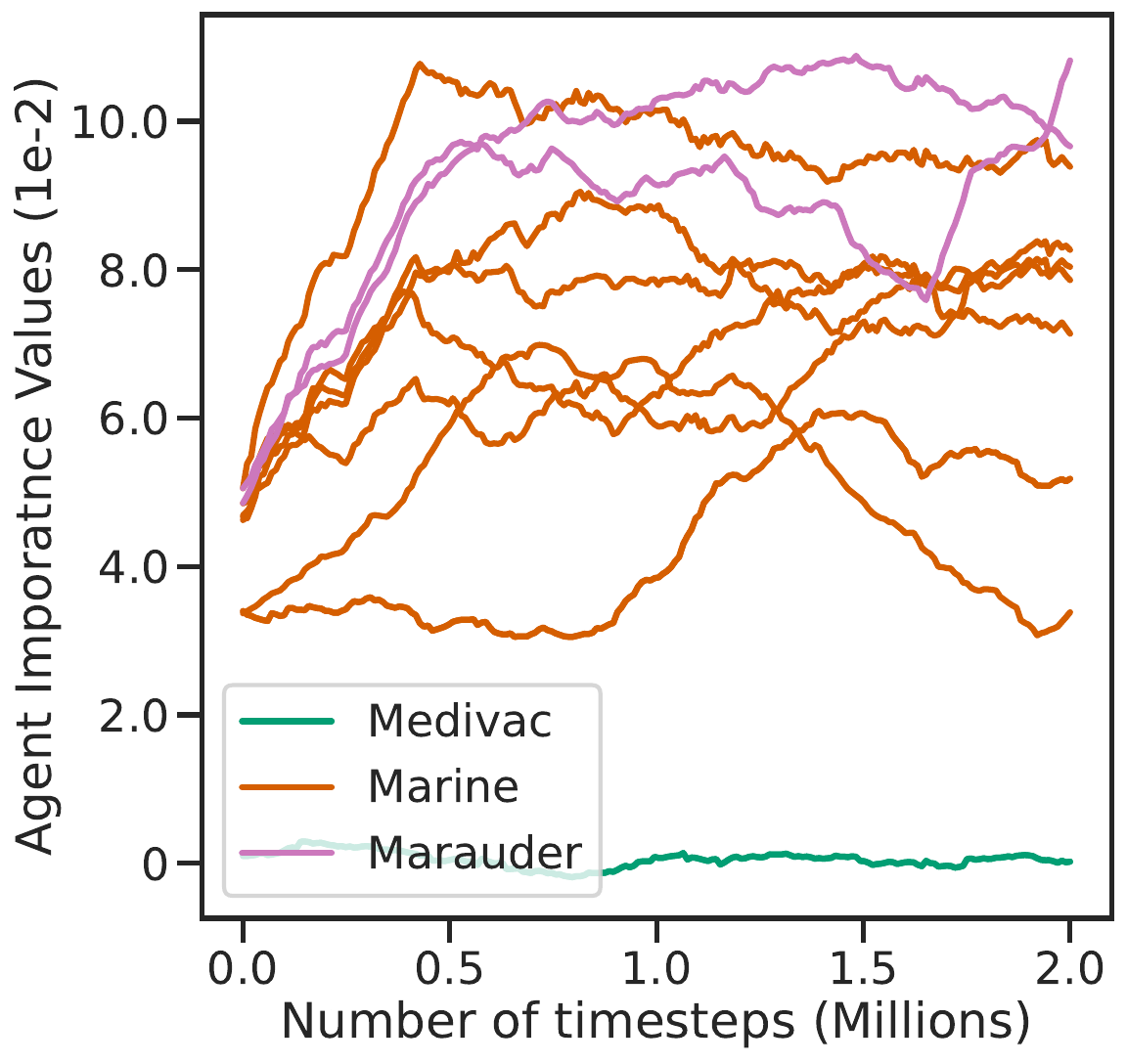}
  \end{subfigure}
  \begin{subfigure}[t]{0.15\textwidth}
    \includegraphics[width=\textwidth, valign=t]{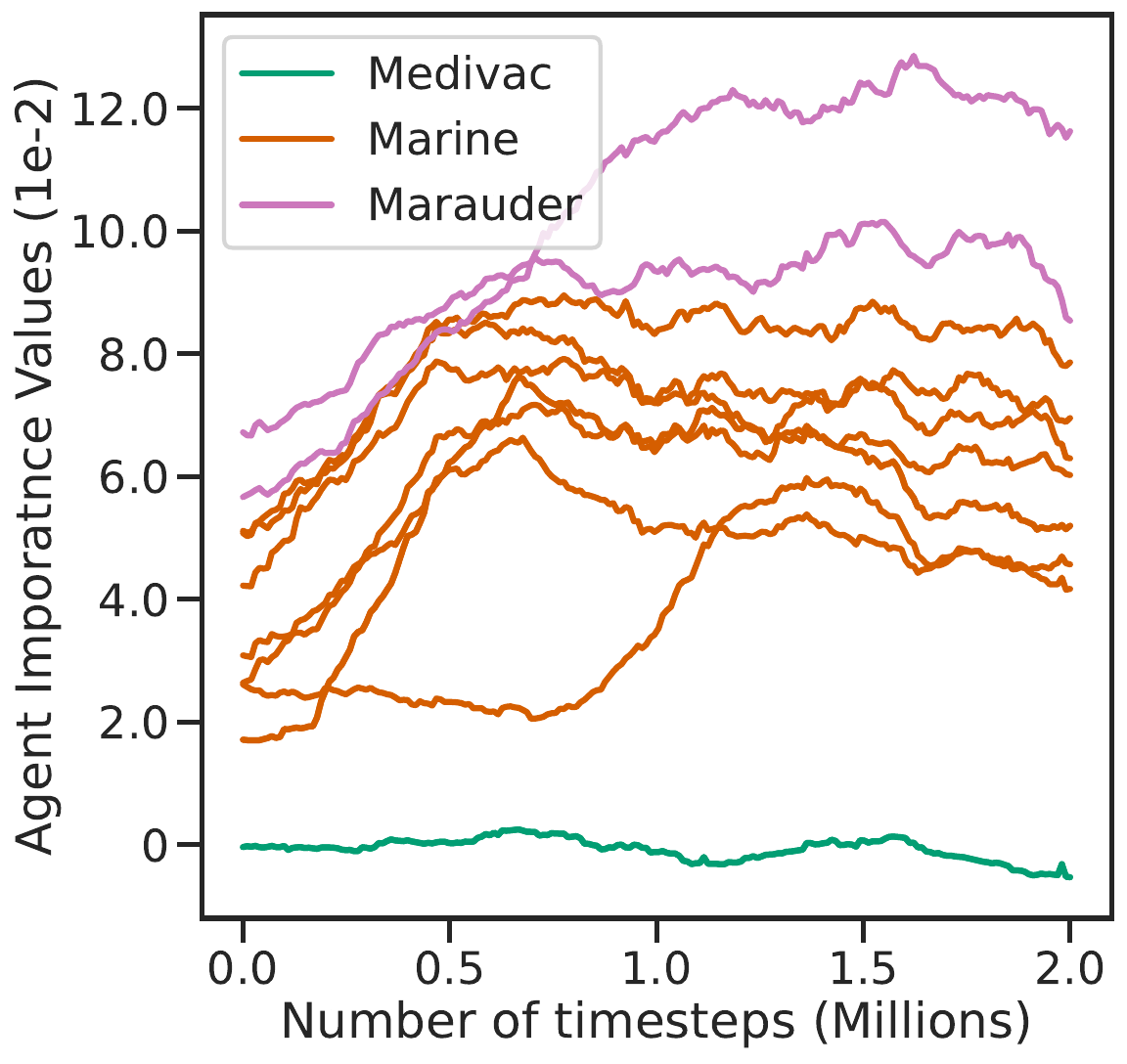}
  \end{subfigure}
   \caption{Agent importance plots for MAPPO without parameter sharing across 5 seeds in MMM2}
  \label{fig: MAPPO_NS_MMM2_ALL}
\end{figure}

From figures \ref{fig: MAPPO_PS_MMM2_ALL} and \ref{fig: MAPPO_NS_MMM2_ALL} we can see that the assigned agent importance rankings are fairly similar for both the PS and NS case. In both cases from figure \ref{fig: MMM2_NS_COMPARE} we can see that MAPPO obtains a similar learning curve. Essentially this supports the claims made by \citep{wen2022multiagent} regarding evaluation of SOTA methods using PS. Although NS does improve performance and correct credit assignment in the heterogeneous case, the improvement is only noticeable if the PS variant of the algorithm is not already able to easily achieve optimal policies.

\section{Further Validation of Agent Importance}
\label{validation_section}
In this section, we present additional results, for further validation of the agent importance metric, thus proving its effectiveness. To do so, we set the tests on a deterministic version of a fixed scenario, within LBF, to assess the reliability of the metric and compare its scalability to the Shapley Value. We provide additional plots to analyze the correlation between the agent importance and the Shapley value.

\begin{figure}
  \centering
  \begin{subfigure}[t]{0.23\textwidth}
    \includegraphics[width=\textwidth, valign=t]{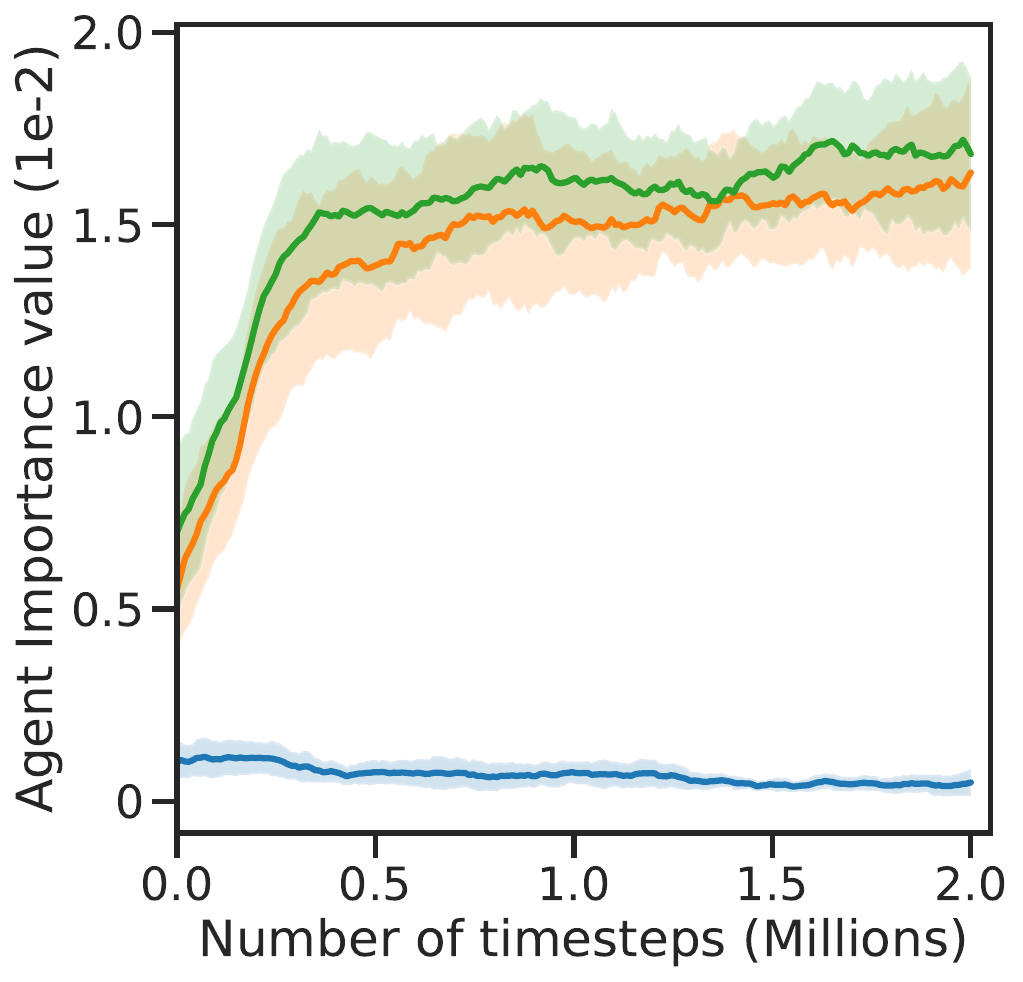}
  \end{subfigure}
  \begin{subfigure}[t]{0.23\textwidth}
    \includegraphics[width=\textwidth, valign=t]{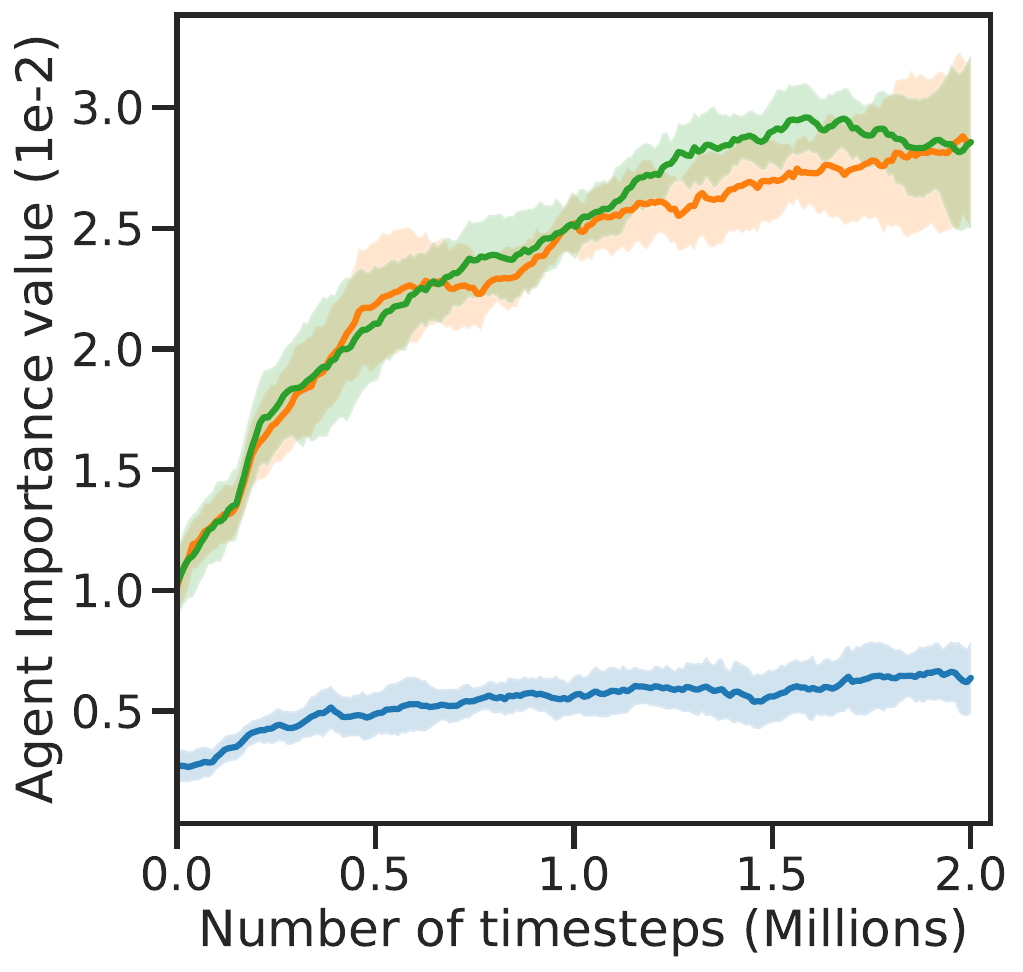}
  \end{subfigure}
  \begin{subfigure}[t]{0.15\textwidth}
    \includegraphics[width=\textwidth, valign=t]{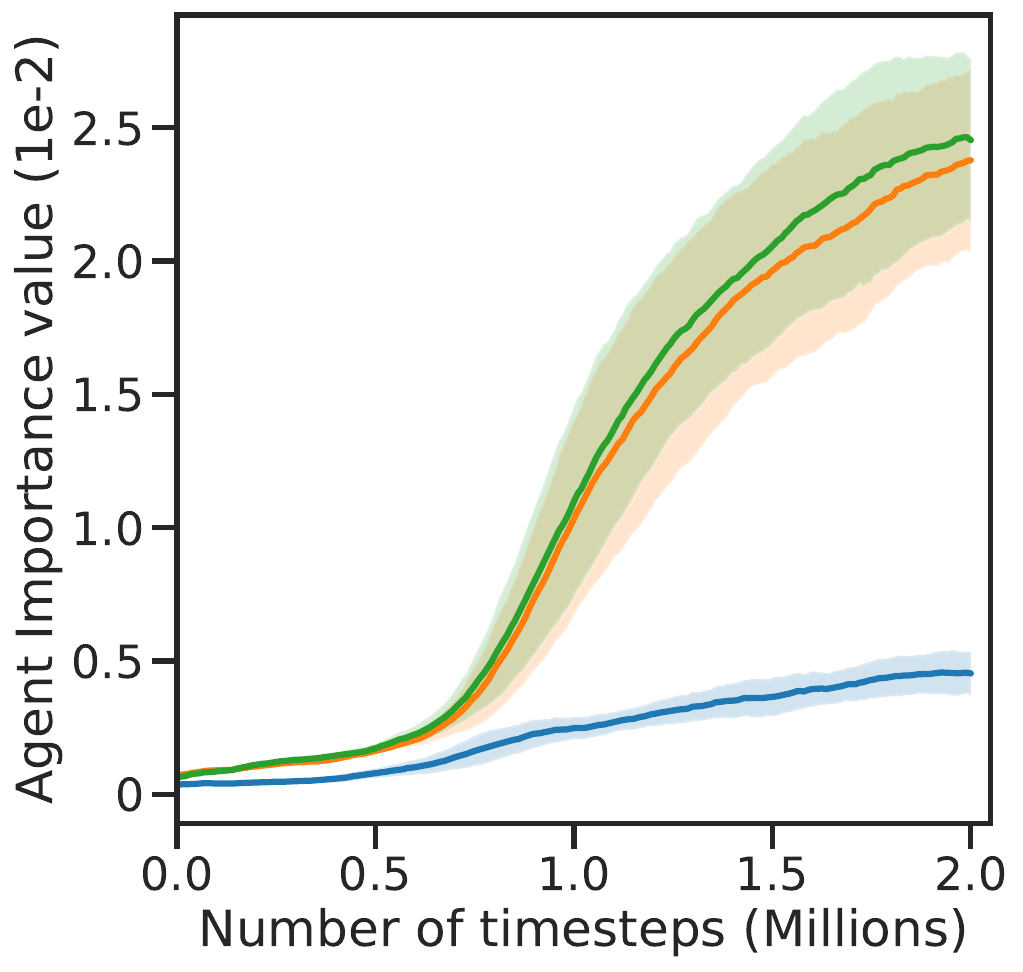}
  \end{subfigure}
    \begin{subfigure}[t]{0.15\textwidth}
    \includegraphics[width=\textwidth, valign=t]{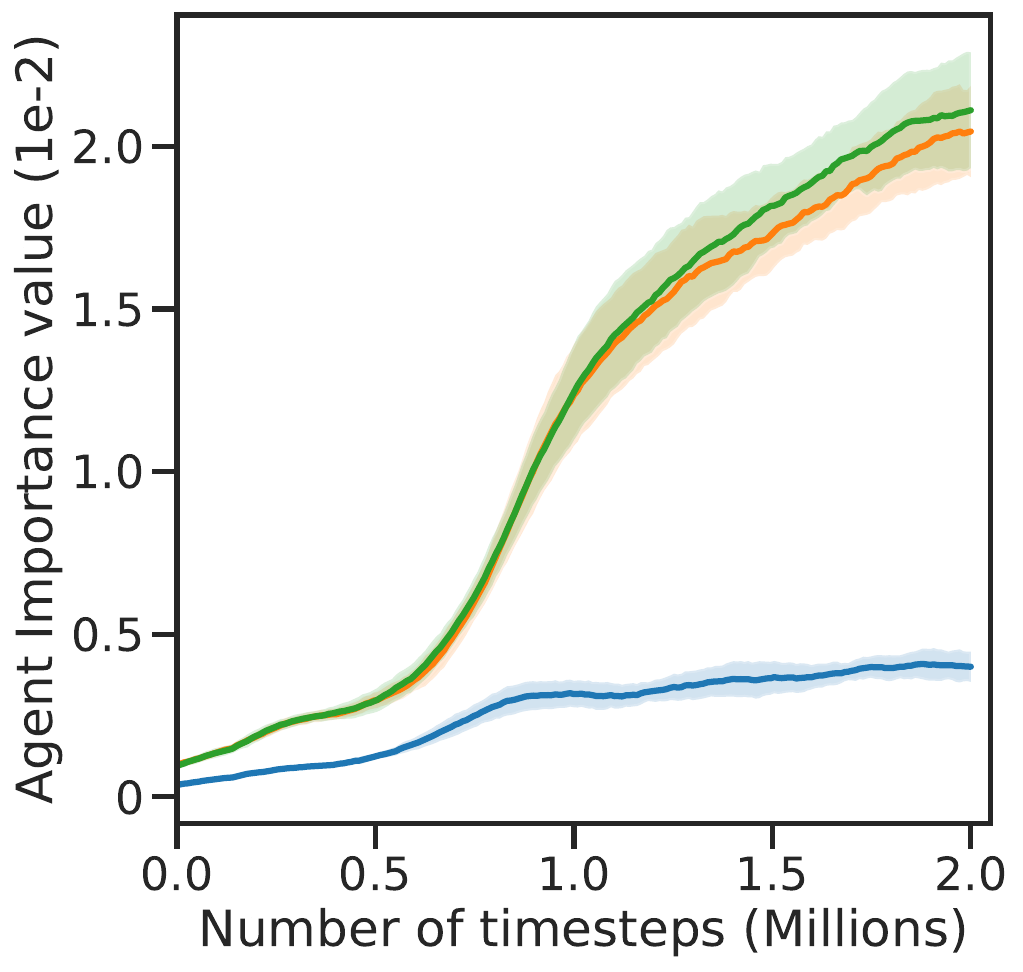}
  \end{subfigure}
  \begin{subfigure}[t]{0.15\textwidth}
    \includegraphics[width=\textwidth, valign=t]{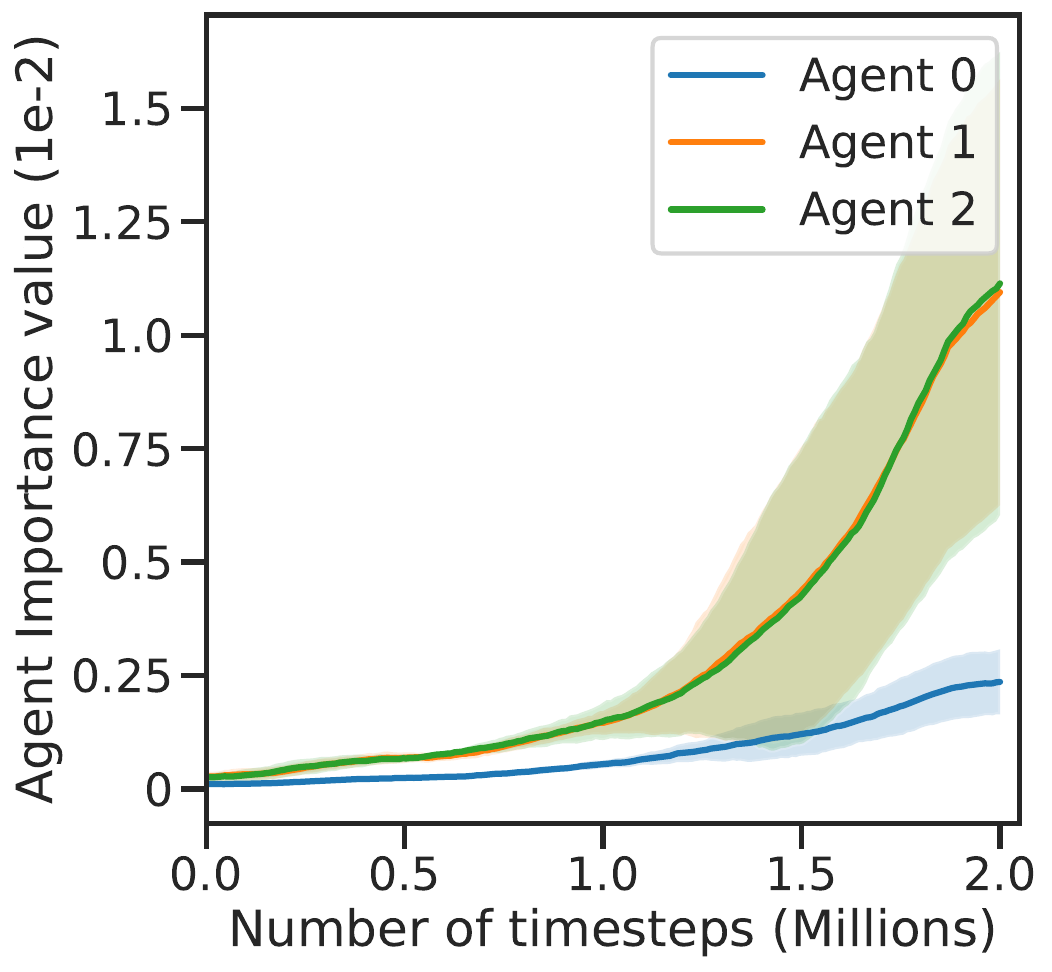}
  \end{subfigure}

   \caption{Agent importance scores on the \textbf{Foraging-15x15-3p-3f-det} scenario with parameter sharing for MAA2C, MAPPO, VDN, IQL and QMIX. Agents 0, 1, and 2 are assigned fixed levels of 1, 2 and 3.}
  \label{fig: lbf_15x15_3p3f_det}
\end{figure}

\begin{figure}
  \centering 
  \begin{subfigure}[t]{0.23\textwidth}
    \includegraphics[width=\textwidth, valign=t]{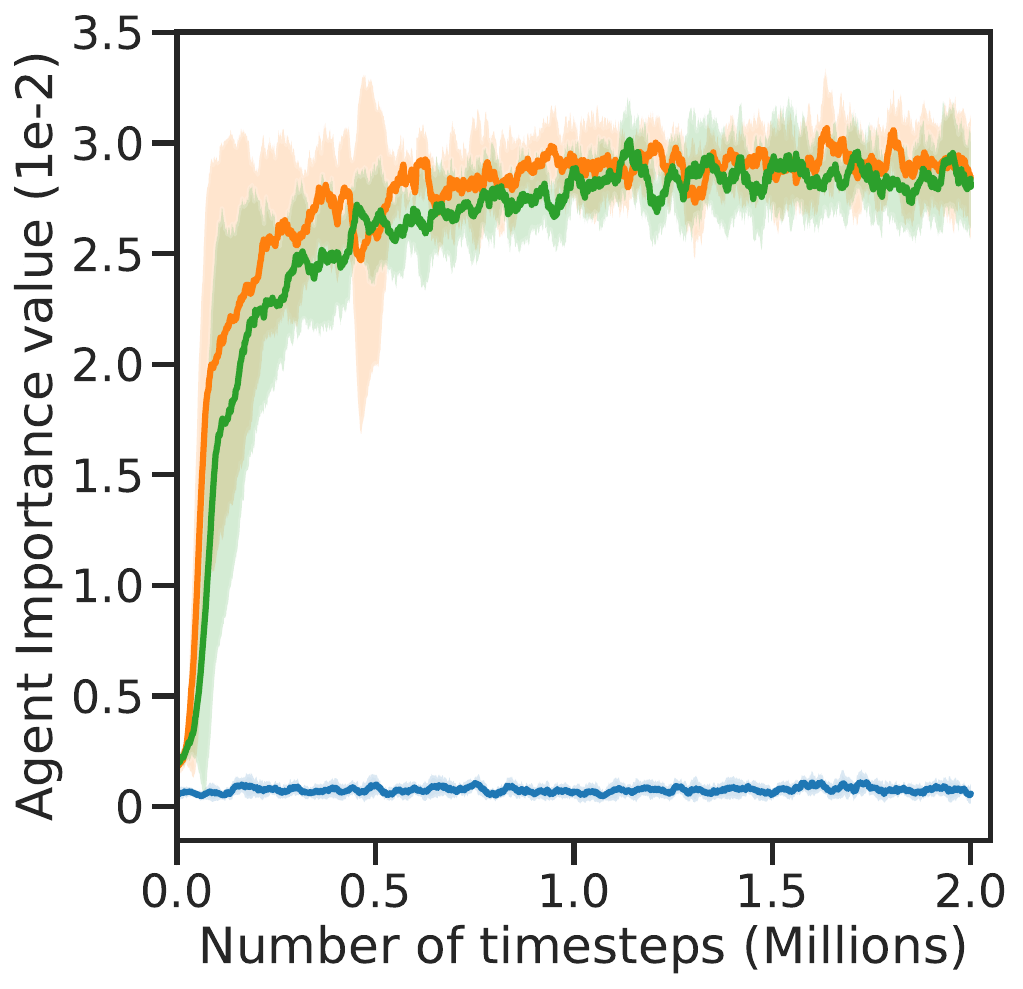}
  \end{subfigure}
  \begin{subfigure}[t]{0.23\textwidth}
    \includegraphics[width=\textwidth, valign=t]{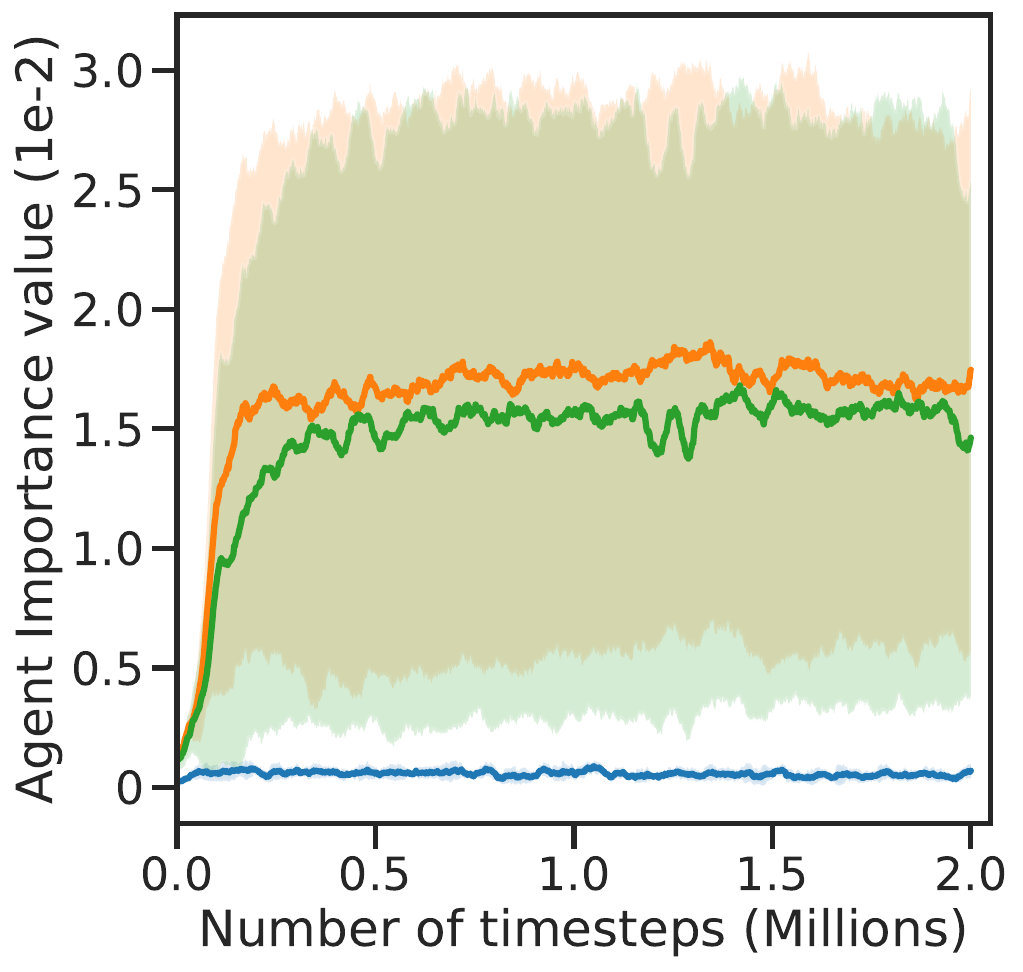}
  \end{subfigure}
  \begin{subfigure}[t]{0.15\textwidth}
    \includegraphics[width=\textwidth, valign=t]{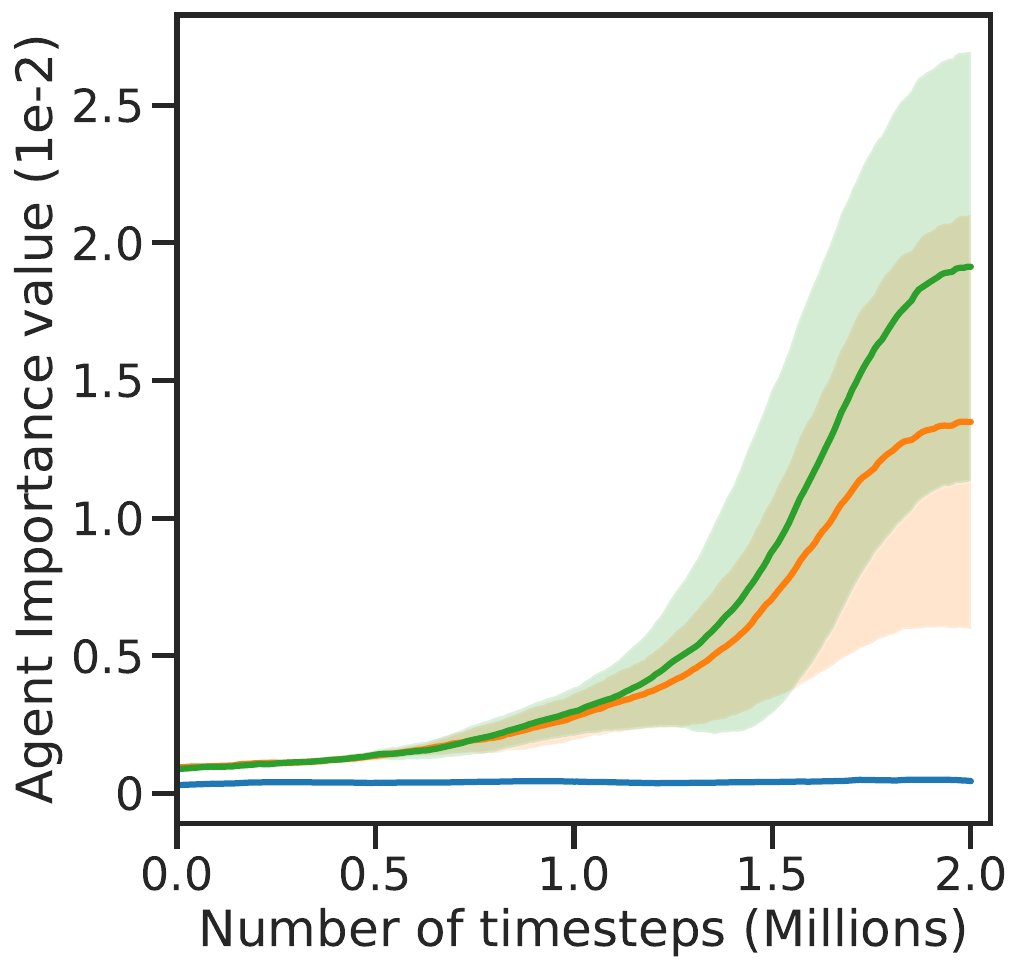}
  \end{subfigure}
      \begin{subfigure}[t]{0.15\textwidth}
    \includegraphics[width=\textwidth, valign=t]{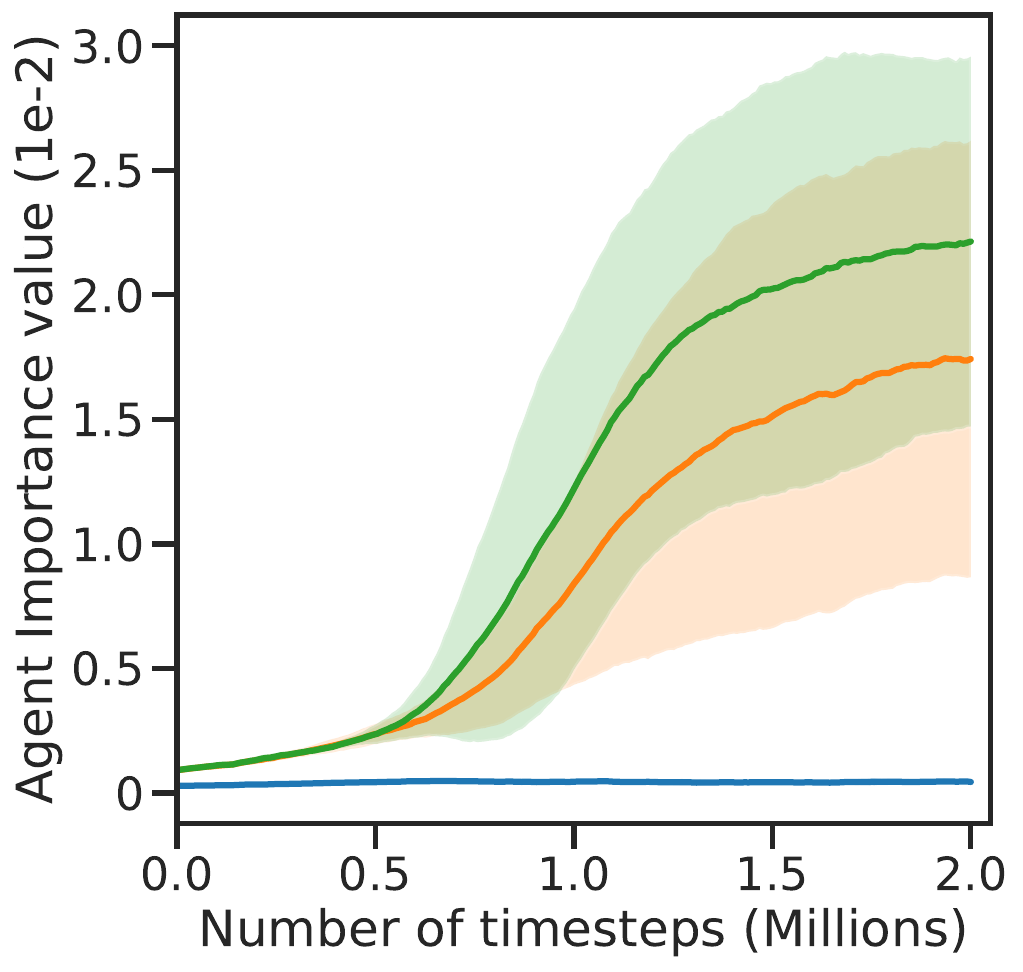}
  \end{subfigure}
  \begin{subfigure}[t]{0.15\textwidth}
    \includegraphics[width=\textwidth, valign=t]{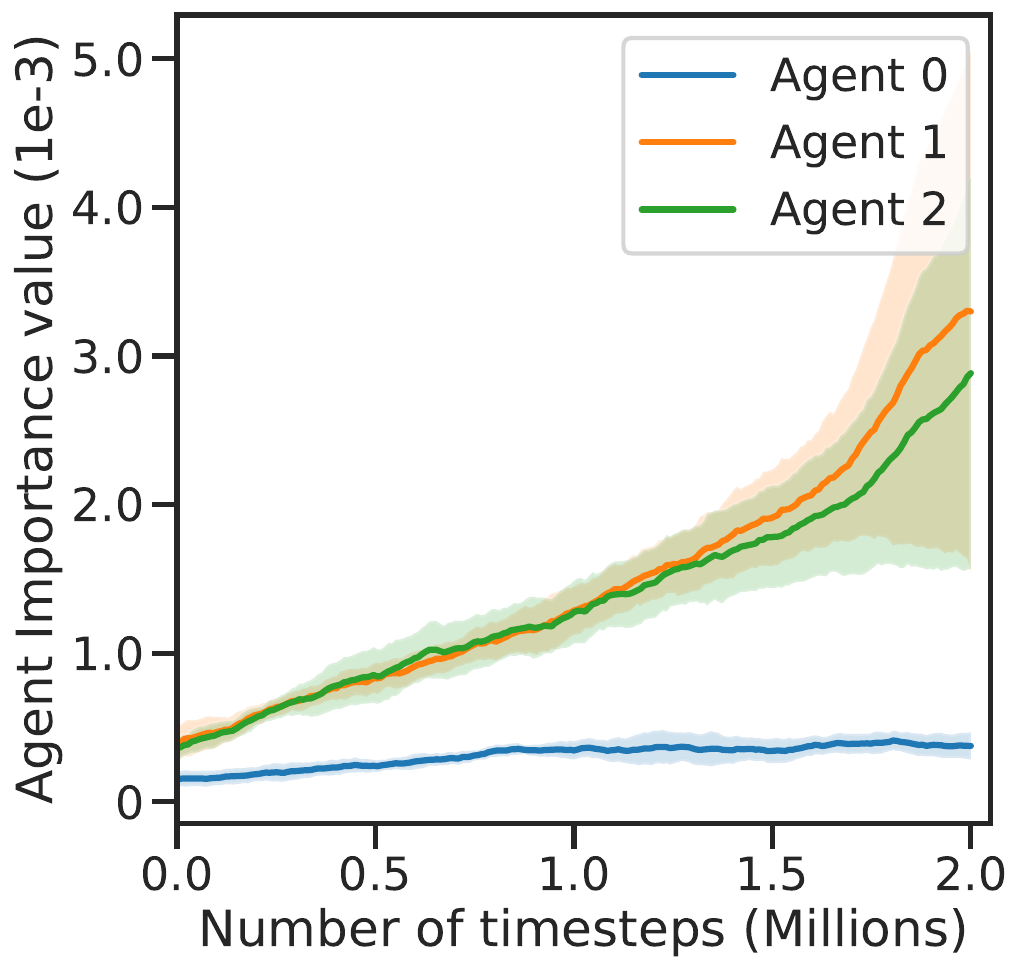}
  \end{subfigure}
  \caption{Agent importance scores on the \textbf{Foraging-15x15-3p-3f-det} scenario without parameter sharing for MAA2C, MAPPO, VDN, IQL and QMIX. Agents 0, 1, and 2 are assigned fixed levels of 1, 2 and 3.}
  \label{fig: lbf_15x15_3p3f_det_ns}
\end{figure}

\subsection{Metric Reliability}
\begin{figure}
  \centering
  \begin{subfigure}[t]{0.23\textwidth}
    \includegraphics[width=\textwidth, valign=t]{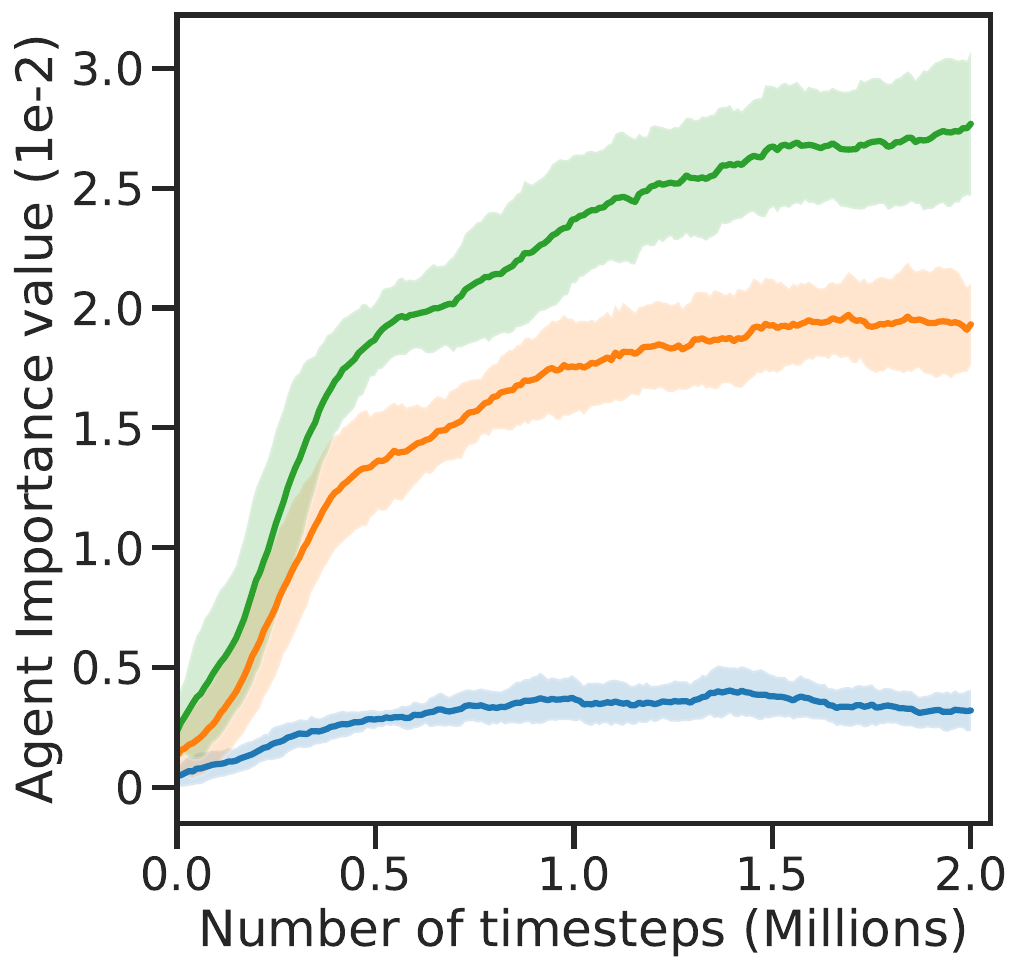}
  \end{subfigure}
  \begin{subfigure}[t]{0.23\textwidth}
    \includegraphics[width=\textwidth, valign=t]{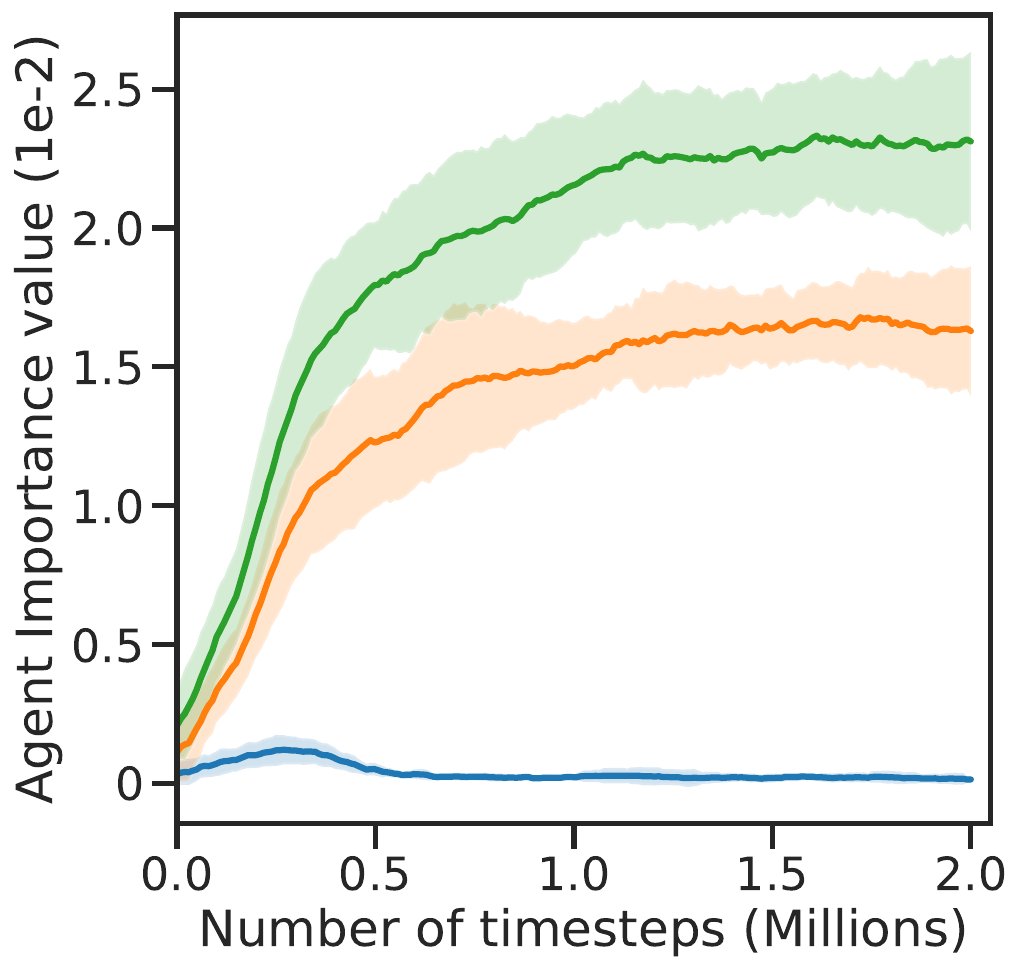}
  \end{subfigure}
  \begin{subfigure}[t]{0.15\textwidth}
    \includegraphics[width=\textwidth, valign=t]{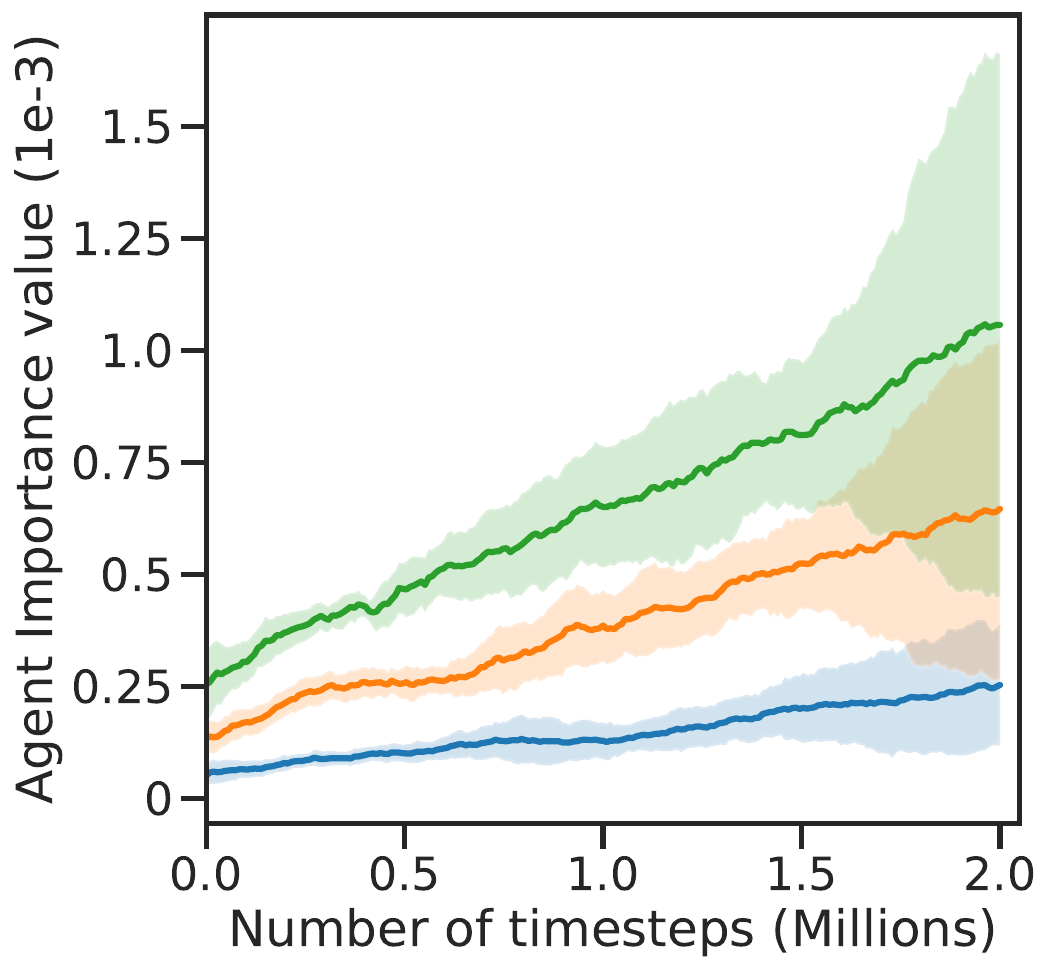}
  \end{subfigure}
    \begin{subfigure}[t]{0.15\textwidth}
    \includegraphics[width=\textwidth, valign=t]{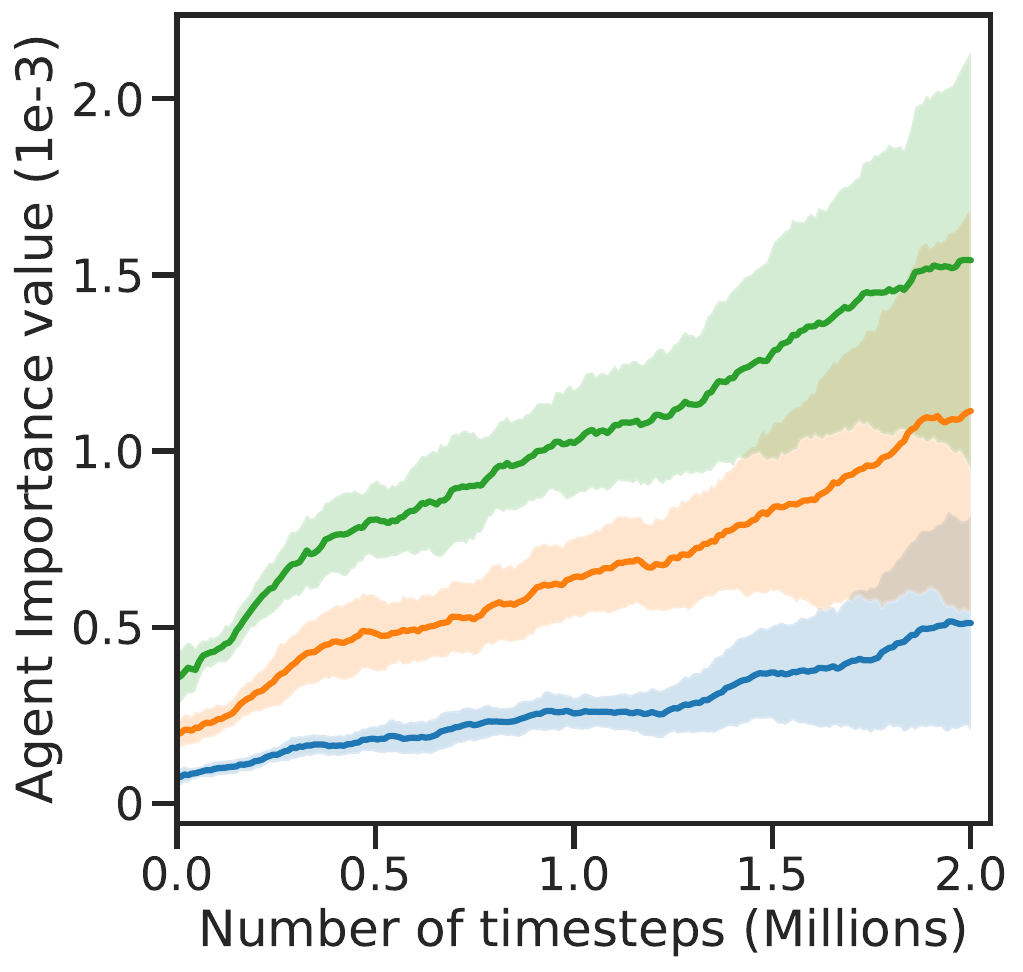}
  \end{subfigure}
  \begin{subfigure}[t]{0.15\textwidth}
    \includegraphics[width=\textwidth, valign=t]{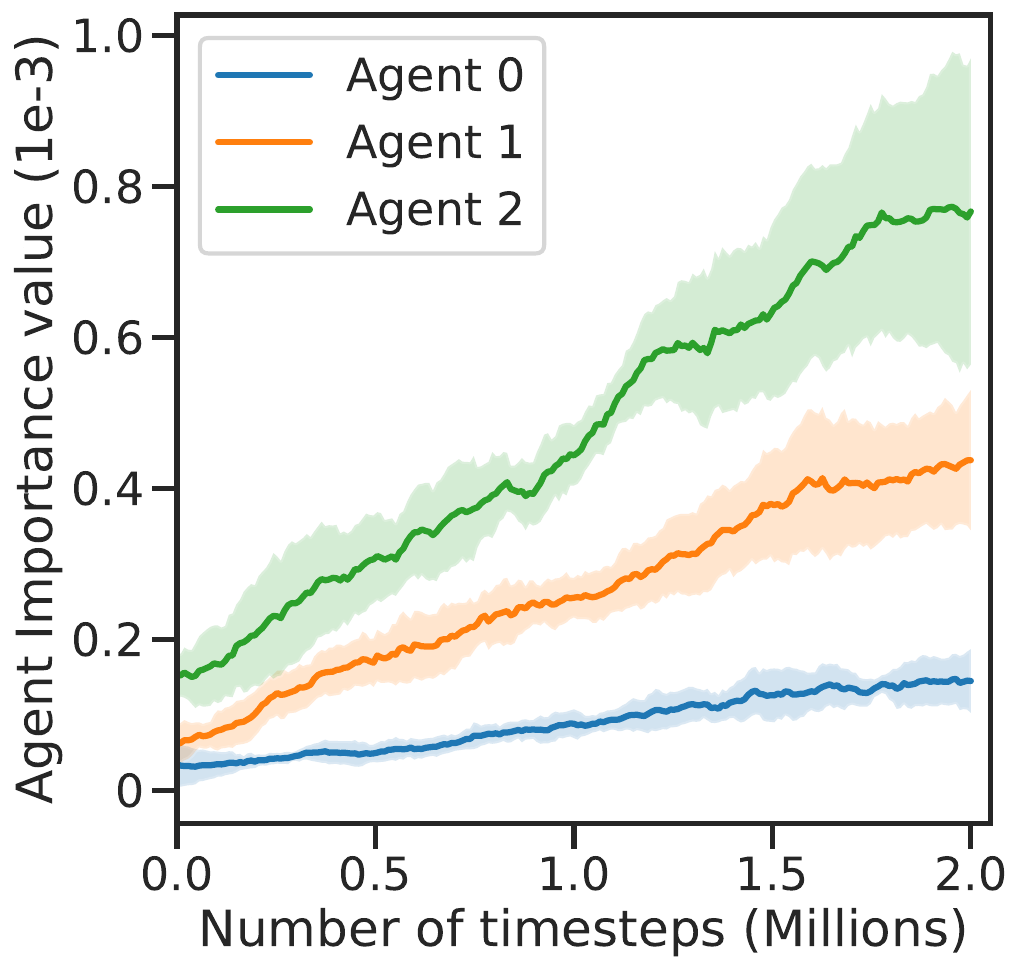}
  \end{subfigure}
  \caption{Agent importance scores on the \textbf{Foraging-15x15-3p-3f-det-max-food-sum} scenario with parameter sharing for MAA2C, MAPPO, VDN, IQL and QMIX. Agents 0, 1, and 2 are assigned fixed levels of 1, 2 and 3.}
  \label{fig: lbf_15x15_3p3f_det_sum}
\end{figure}

\begin{figure}
  \centering
  \begin{subfigure}[t]{0.23\textwidth}
    \includegraphics[width=\textwidth, valign=t]{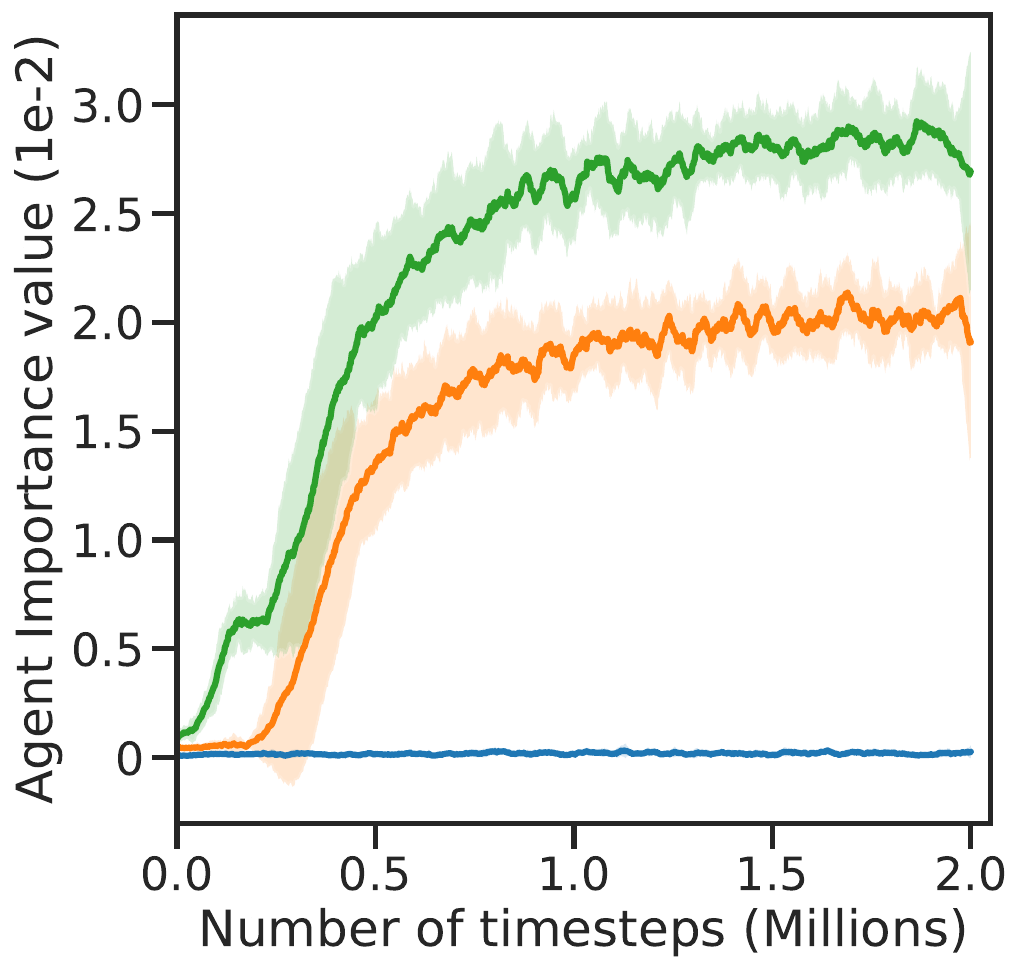}
  \end{subfigure}
 \begin{subfigure}[t]{0.23\textwidth}
    \includegraphics[width=\textwidth, valign=t]{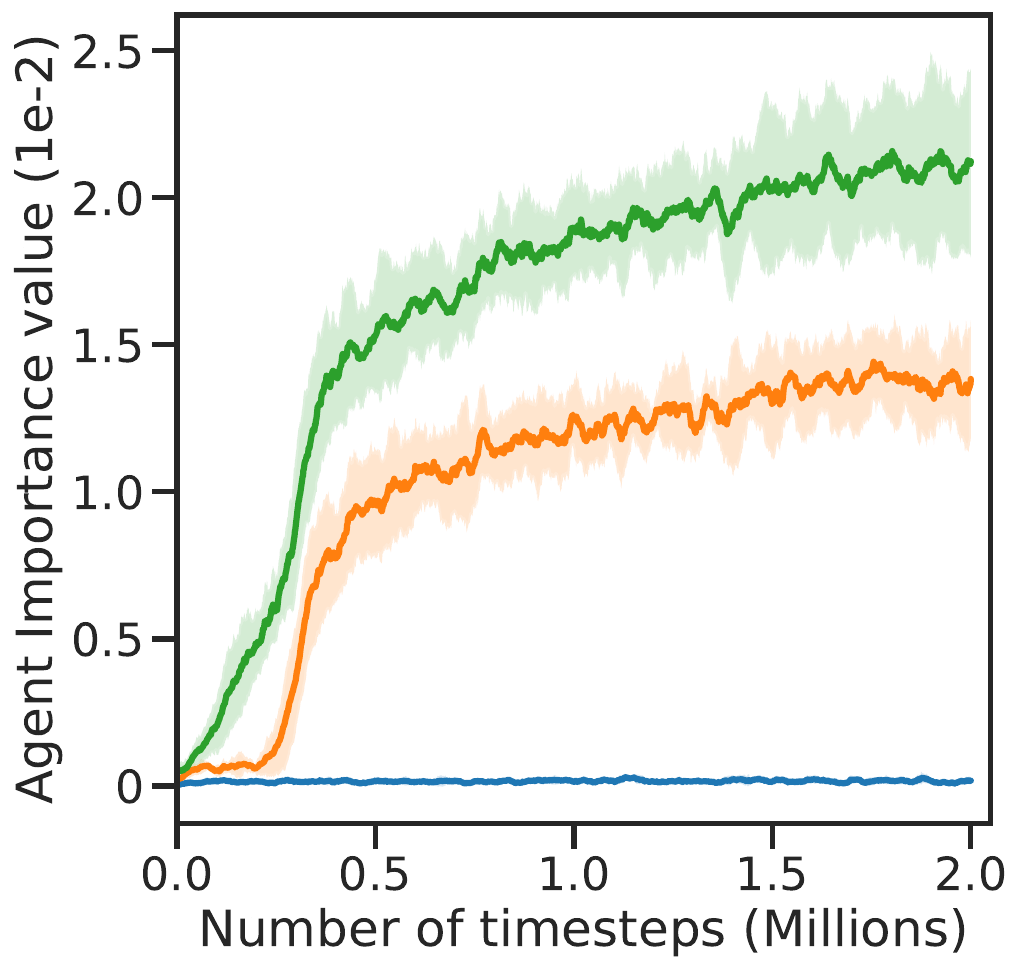}
  \end{subfigure}
    \begin{subfigure}[t]{0.15\textwidth}
    \includegraphics[width=\textwidth, valign=t]{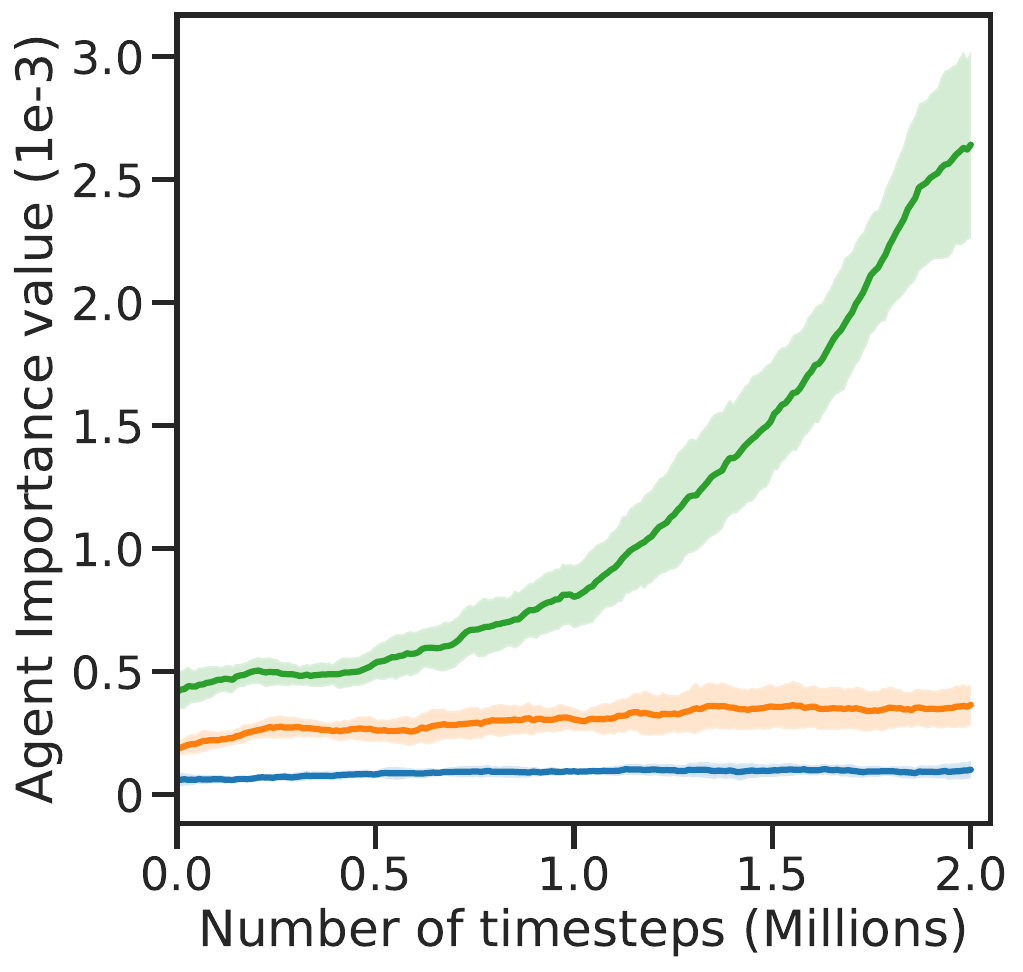}
  \end{subfigure}
      \begin{subfigure}[t]{0.15\textwidth}
    \includegraphics[width=\textwidth, valign=t]{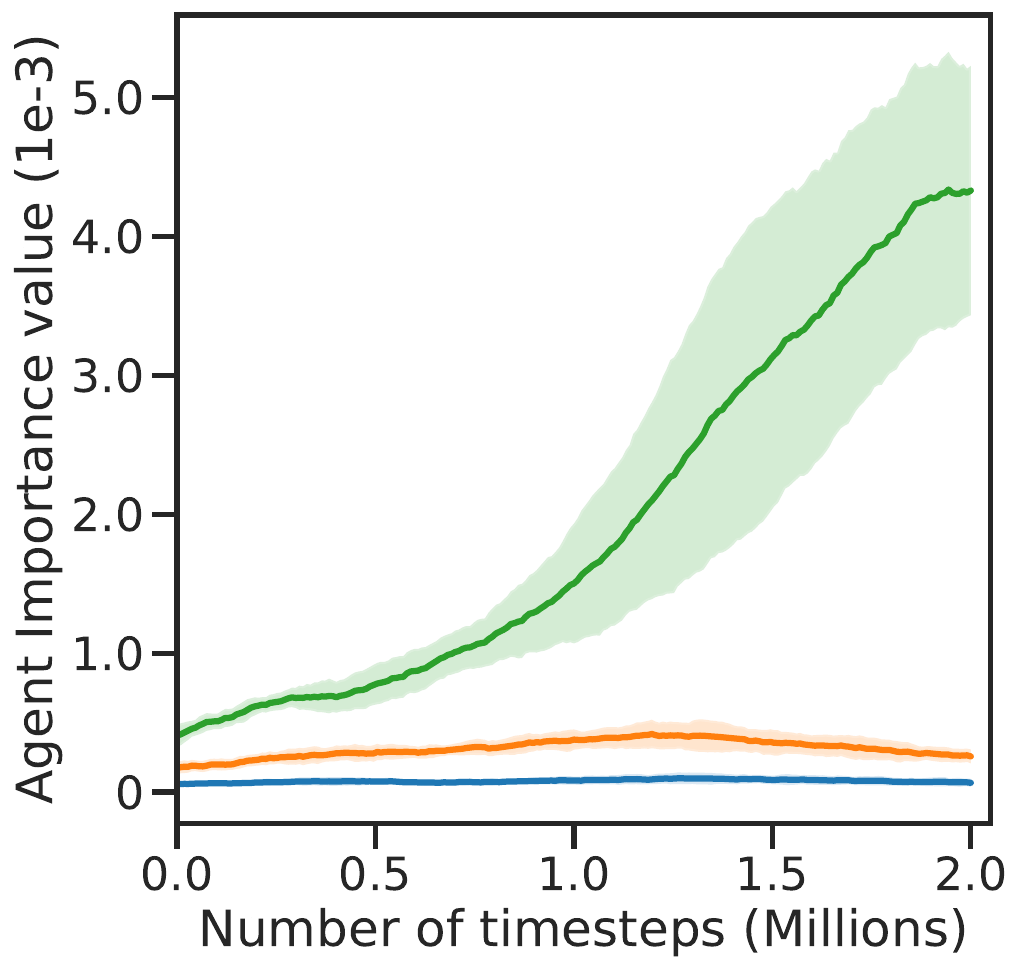}
  \end{subfigure}
    \begin{subfigure}[t]{0.15\textwidth}
    \includegraphics[width=\textwidth, valign=t]{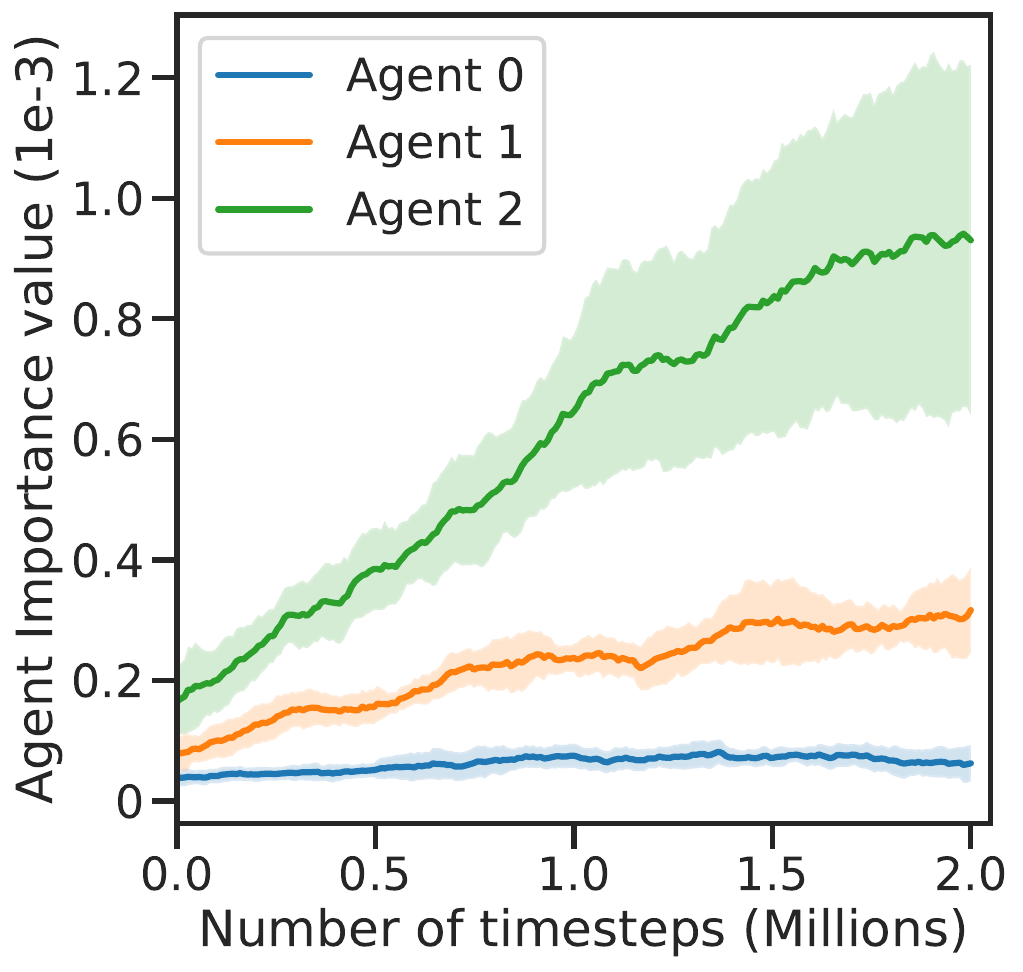}
  \end{subfigure}
 \caption{Agent importance scores on the \textbf{Foraging-15x15-3p-3f-det-max-food-sum} scenario without parameter sharing for MAA2C, MAPPO, VDN, IQL and QMIX. Agents 0, 1, and 2 are assigned fixed levels of 1, 2 and 3.}
  \label{fig: lbf_15x15_3p3f_det_sum_ns}
\end{figure}

Figures \ref{fig: lbf_15x15_3p3f_det} to \ref{fig: lbf_15x15_3p3f_det_sum_ns} showcase the results for the agent importance analysis for all tested algorithms, considering both the parameter-sharing and non-parameter-sharing cases. In the deterministic LBF setting, as outlined in Section 1, agents 0, 1, and 2 are assigned levels of 1, 2, and 3, respectively. The figures demonstrate that agents with higher levels contribute more significantly. These findings are consistent across all algorithms and the reported values resulted from an aggregation over the 10 independent runs.

\subsection{Metric Scalability}

\begin{table}
  \centering
  \caption{Average time required to calculate agent contributions compared to the baseline with standard deviation}
  \label{tab: average_time_consumed}
  \resizebox{0.5\textwidth}{!}
  {\begin{tabular}{cccc}
    \toprule
    \textbf{Number of Agents}& \textbf{Baseline} & \textbf{Agent Importance} & \textbf{Shapley Value} \\
    \midrule
    2 & 0.0018 $\pm 0.0002$ & 0.0074 $\pm 0.0001$ & 0.0079 $\pm 0.0014$ \\
    4 & 0.0023 $\pm 0.0003$ & 0.0118 $\pm 0.0024$ & 0.0313 $\pm 0.0041$ \\
    10 & 0.0038 $\pm 0.0012$ & 0.0392 $\pm 0.0022$ & 3.544 $\pm 0.156$ \\
    20 & 0.0088 $\pm 0.0007$ & 0.1697 $\pm 0.0159$ & 8065.3697 $\pm 832.1829$ \\
    50 & 0.0401 $\pm 0.0019$ & 1.6947 $\pm 0.2295$ & ---\\
    \bottomrule
  \end{tabular}}
\end{table}

\begin{figure}
\centering
\includegraphics[width=0.4\textwidth]{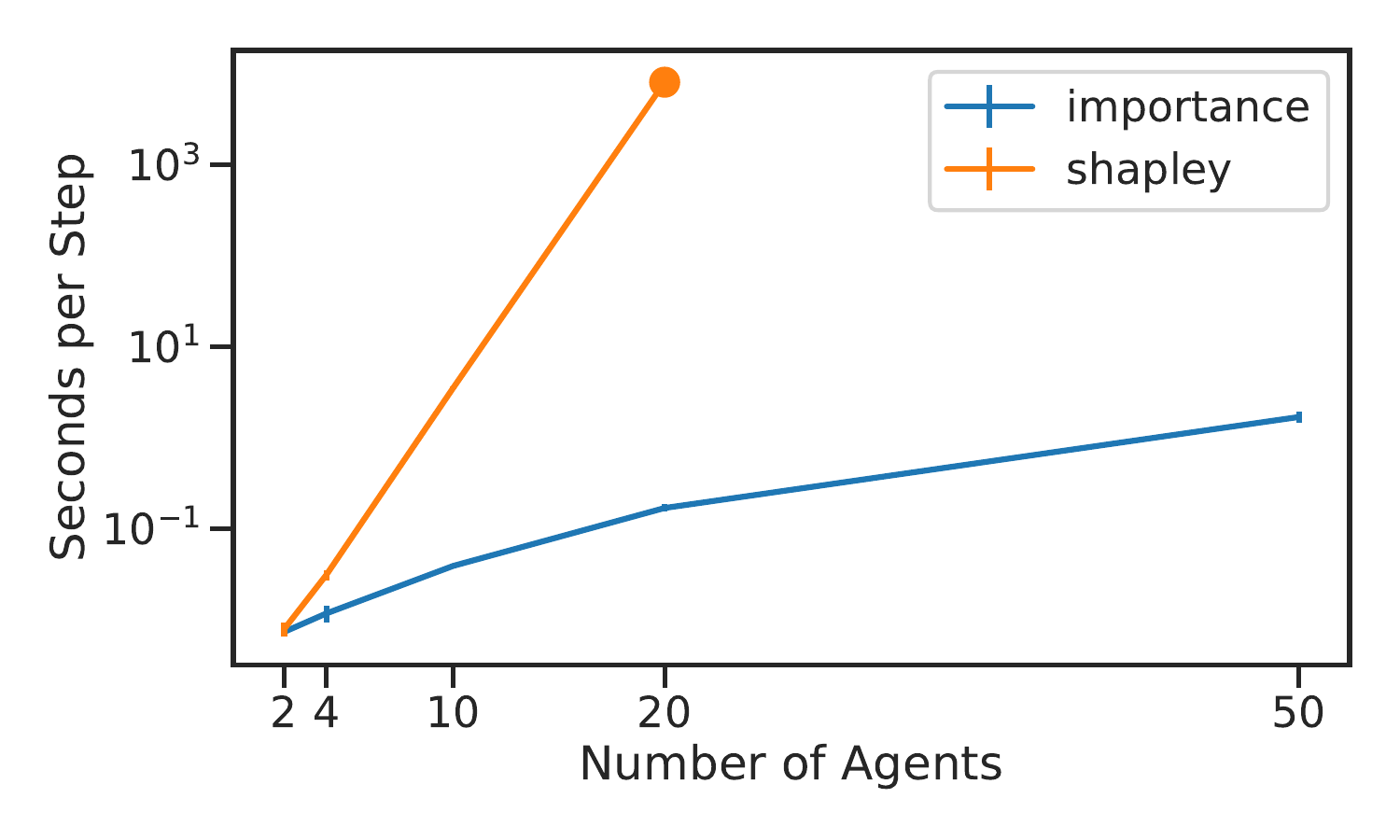}
\caption{Computational cost of computing the Agent Importance and the Shapley value in seconds per environment step (log scale).}
\label{fig: compute_time_log}
\end{figure}

In Table \ref{tab: average_time_consumed}, we present the average time taken per step along with the standard deviation for the baseline algorithm without any metrics,  with agent importance, and with the Shapley Value. These values were calculated over three independent runs for each scenario. The specific scenarios used in these experiments are detailed in section \ref{LBF}. We examined various cases by changing and varying the number of agents from 2 to 50. However, it is important to note that the experiment with 50 agents using the Shapley value took an exceptionally long time to complete and was therefore omitted. Nevertheless, the results obtained from the other experiments demonstrate the scalability w.r.t the number of agents of the Agent Importance metric compared to the Shapley value.
In Figure \ref{fig: compute_time_log}, we present the time consumed by each metric per step, with the values plotted in a logarithmic scale. This scaling helps visualize the significant difference in the required time when using the Shapley Value.

\subsection{Correlation between Agent Importance and the Shapley value}
\begin{figure}
  \centering
\begin{subfigure}{0.4\linewidth}
    \includegraphics[width=\linewidth]{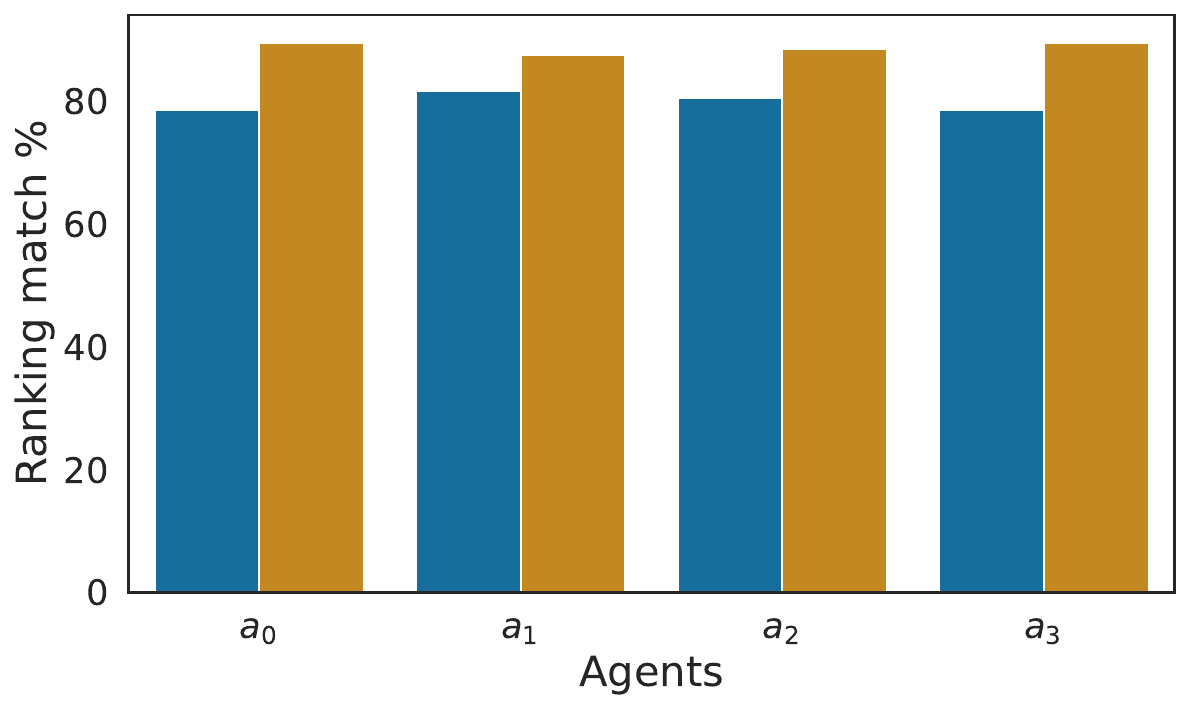}
  \end{subfigure}
  \begin{subfigure}{0.5\linewidth}
    \includegraphics[width=\linewidth]{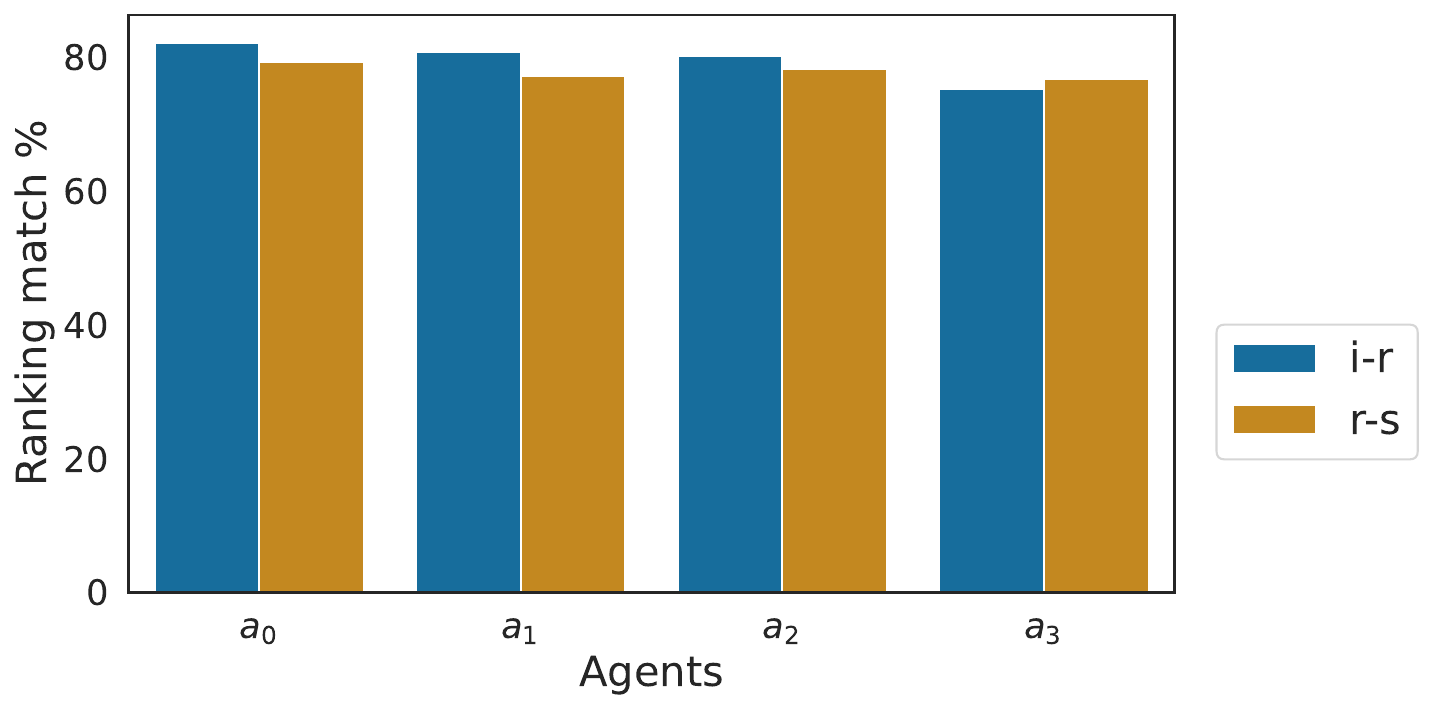}
  \end{subfigure}
  \caption{\textbf{Left:} Matching Rankings Comparison on LBF 15x15-4p-5f. \textbf{Right:} Matching Rankings Comparison on RWARE small-4ag.}
  \label{fig: maappo_rank}
\end{figure}

\begin{figure}[h]
  \centering
\begin{subfigure}{0.4\linewidth}
    \includegraphics[width=\linewidth]{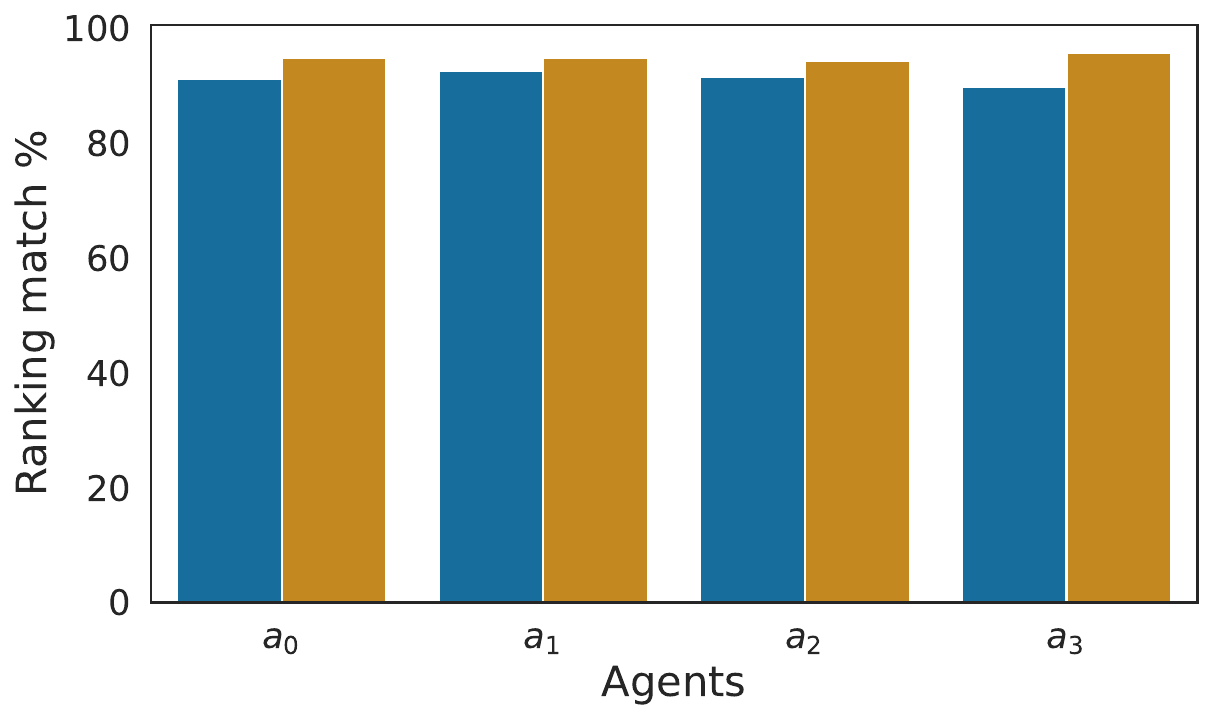}
  \end{subfigure}
  \begin{subfigure}{0.5\linewidth}
    \includegraphics[width=\linewidth]{figures/appendix/ranking/qmix_ranking_rware_rware_small_4ag_v1.pdf}
  \end{subfigure}
  \caption{ \textbf{Left:} Matching Rankings Comparison on LBF 15x15-4p-5f. \textbf{Right:} Matching Rankings Comparison on RWARE small-4ag.}
  \label{fig: qmix_rank}
\end{figure}

\begin{figure}[h]
  \centering
\begin{subfigure}{0.4\linewidth}
    \includegraphics[width=\linewidth]{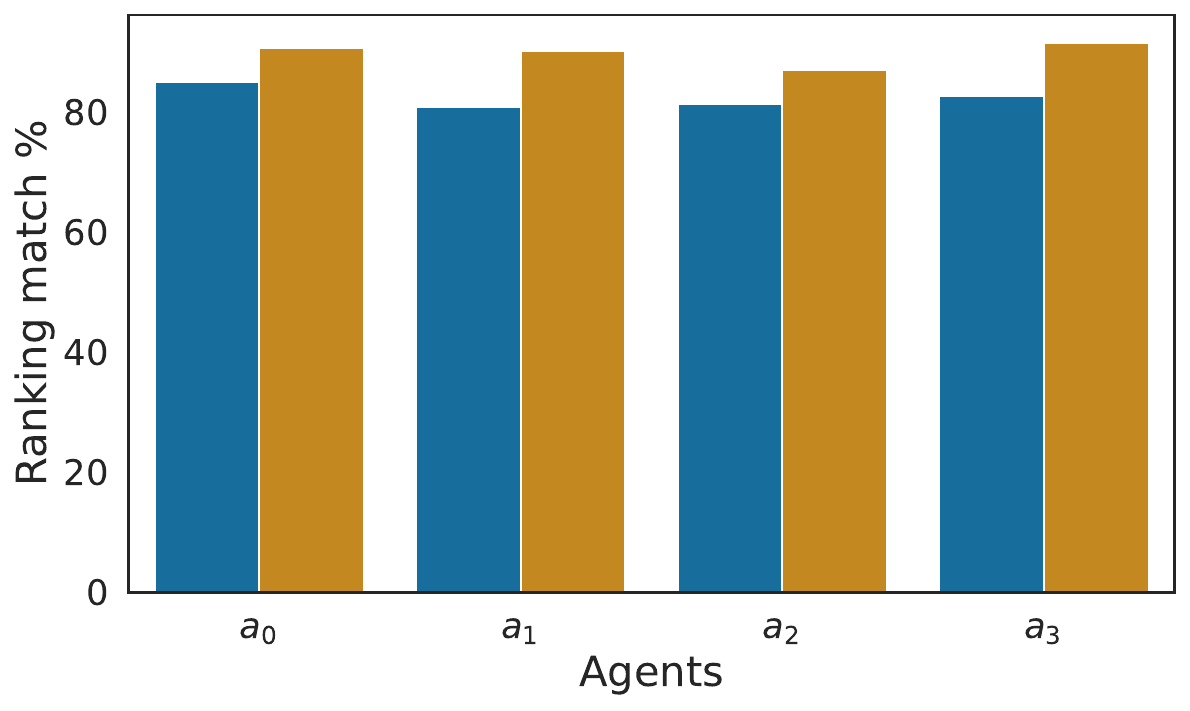}
  \end{subfigure}
  \begin{subfigure}{0.5\linewidth}
    \includegraphics[width=\linewidth]{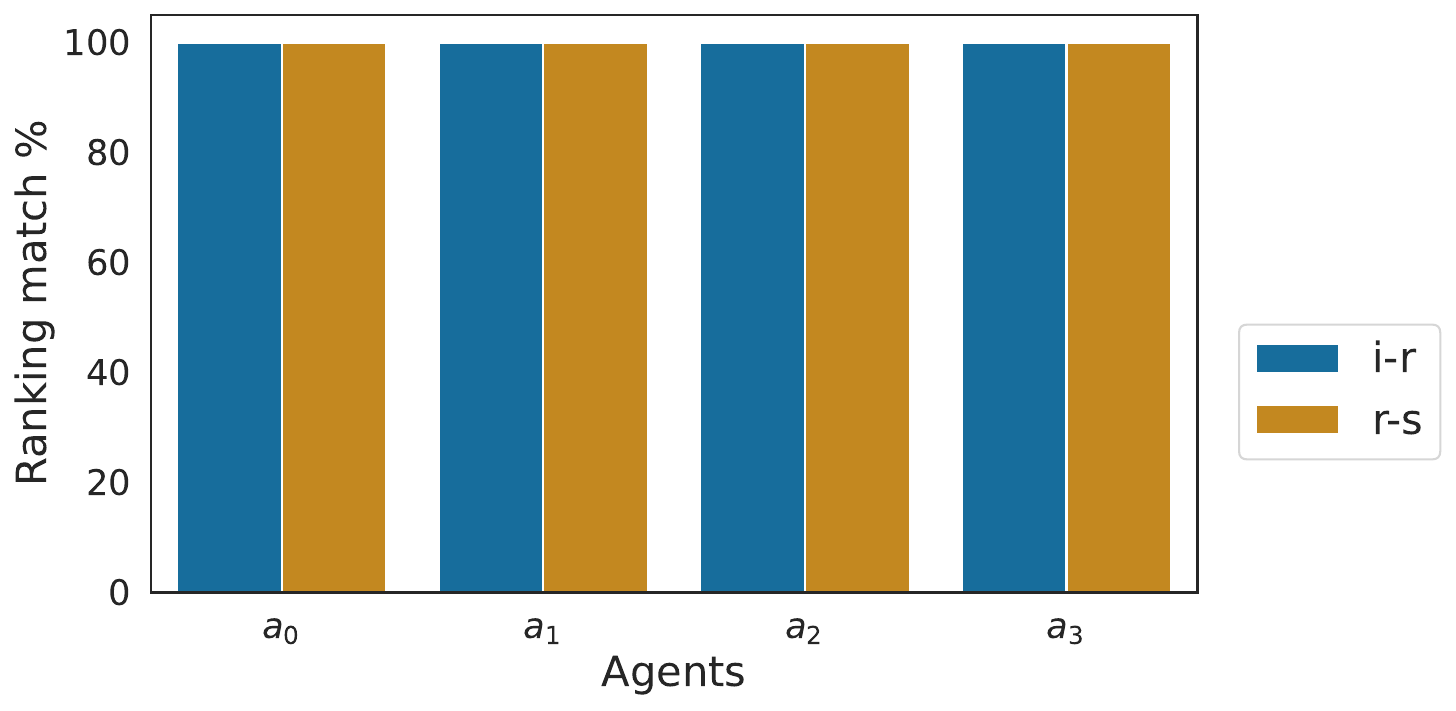}
  \end{subfigure}
  \caption{\textbf{Left:} Matching Rankings Comparison on LBF 15x15-4p-5f. \textbf{Right:} Matching Rankings Comparison on RWARE small-4ag.}
  \label{fig: iql_rank}
\end{figure}

\begin{figure}[H]
\centering
\includegraphics[width=0.7\linewidth]{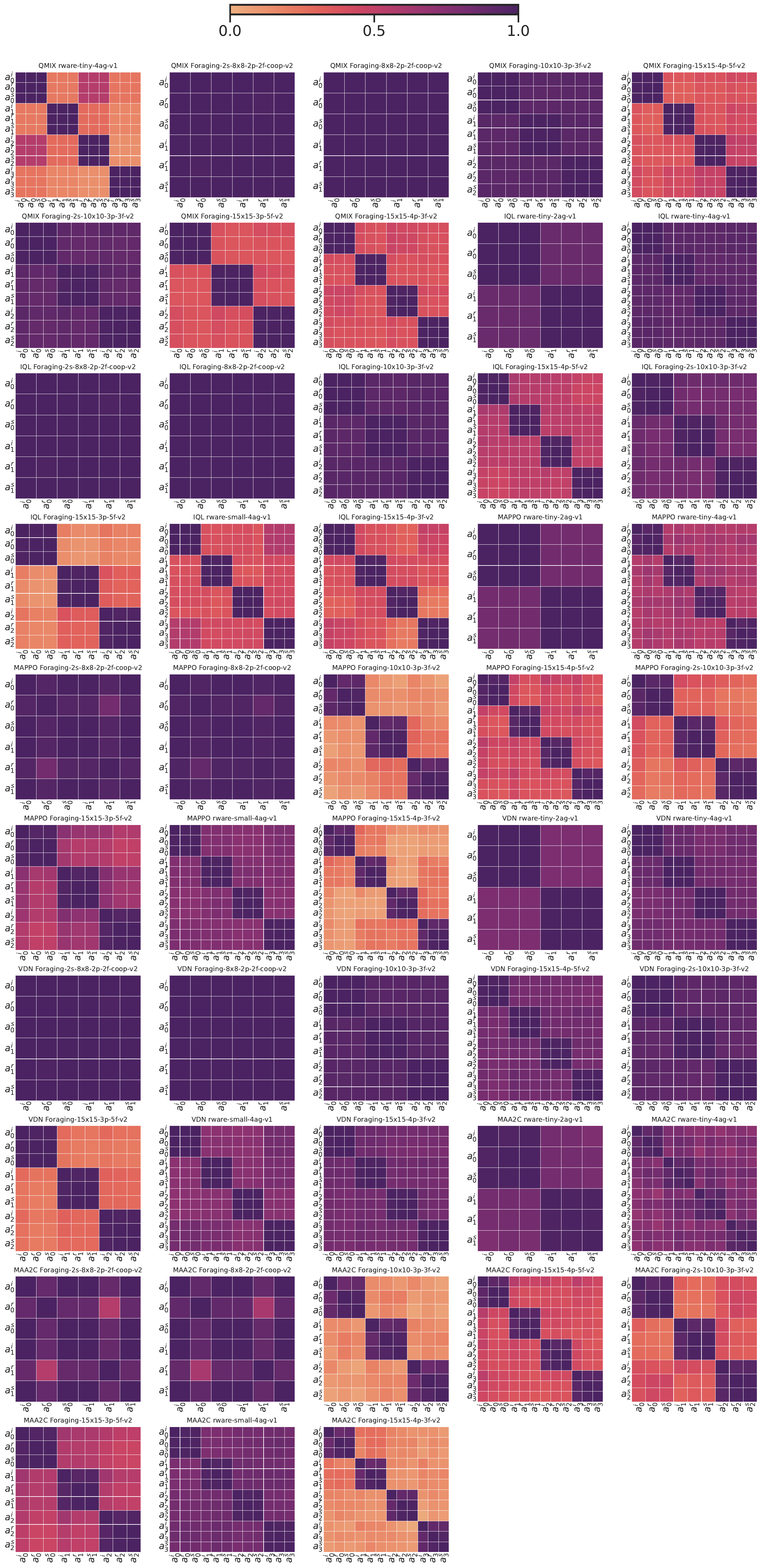}
\caption{\textit{The correlation between Agent Importance, Shapley values, and individual agent rewards is examined for the first independent run.} Each subplot corresponds to a specific algorithm and task, displaying the name of the algorithm followed by the task name. Notably, a strong correlation is observed across all algorithms and tasks.}
\label{fig: correlation_plot_0}
\end{figure}

\begin{figure}[H]
\centering
\includegraphics[width=0.7\linewidth]{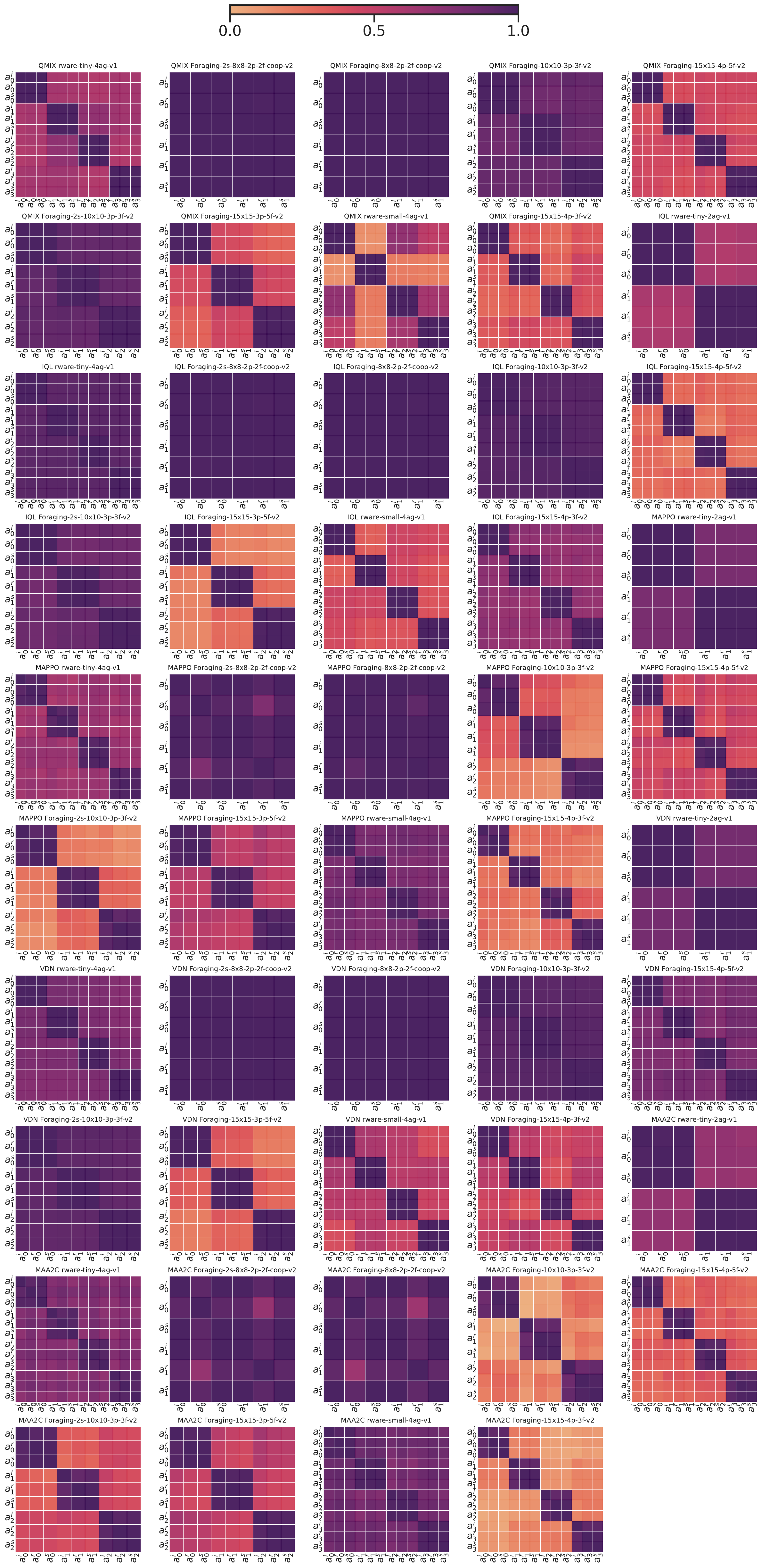}
\caption{\textit{The correlation between Agent Importance, Shapley values, and individual agent rewards is examined for the second independent run}. Each subplot corresponds to a specific algorithm and task, displaying the name of the algorithm followed by the task name. Notably, a strong correlation is observed across all algorithms and tasks.}
\label{fig: correlation_plot_1}
\end{figure}

\begin{figure}[H]
\centering
\includegraphics[width=0.7\linewidth]{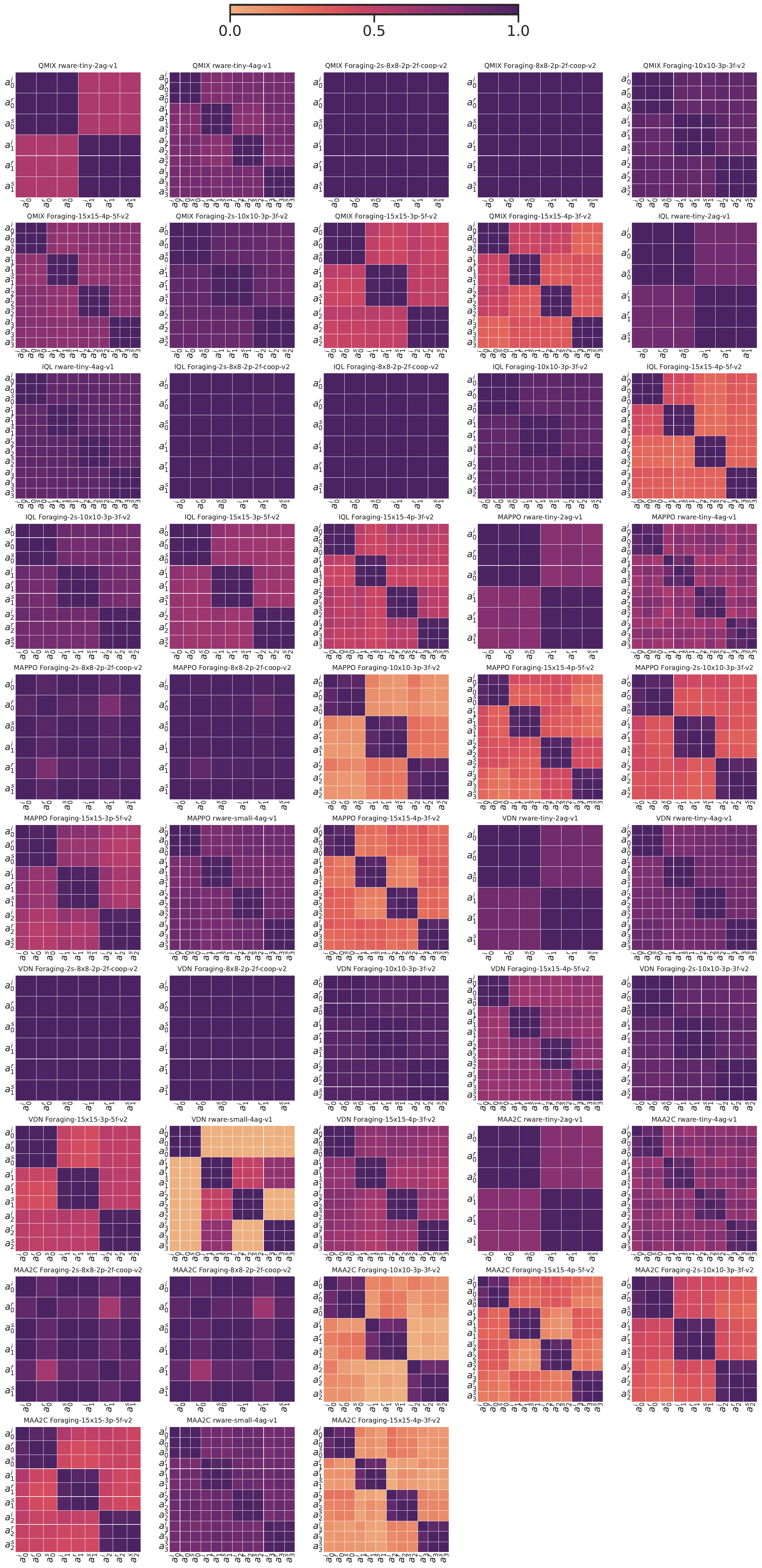}
\caption{\textit{The correlation between Agent Importance, Shapley values, and individual agent rewards is examined for the third independent run.} Each subplot corresponds to a specific algorithm and task, displaying the name of the algorithm followed by the task name. Notably, a strong correlation is observed across all algorithms and tasks.}
\label{fig: correlation_plot_2}
\end{figure}

\begin{figure}
\centering
\includegraphics[width=0.7\linewidth]{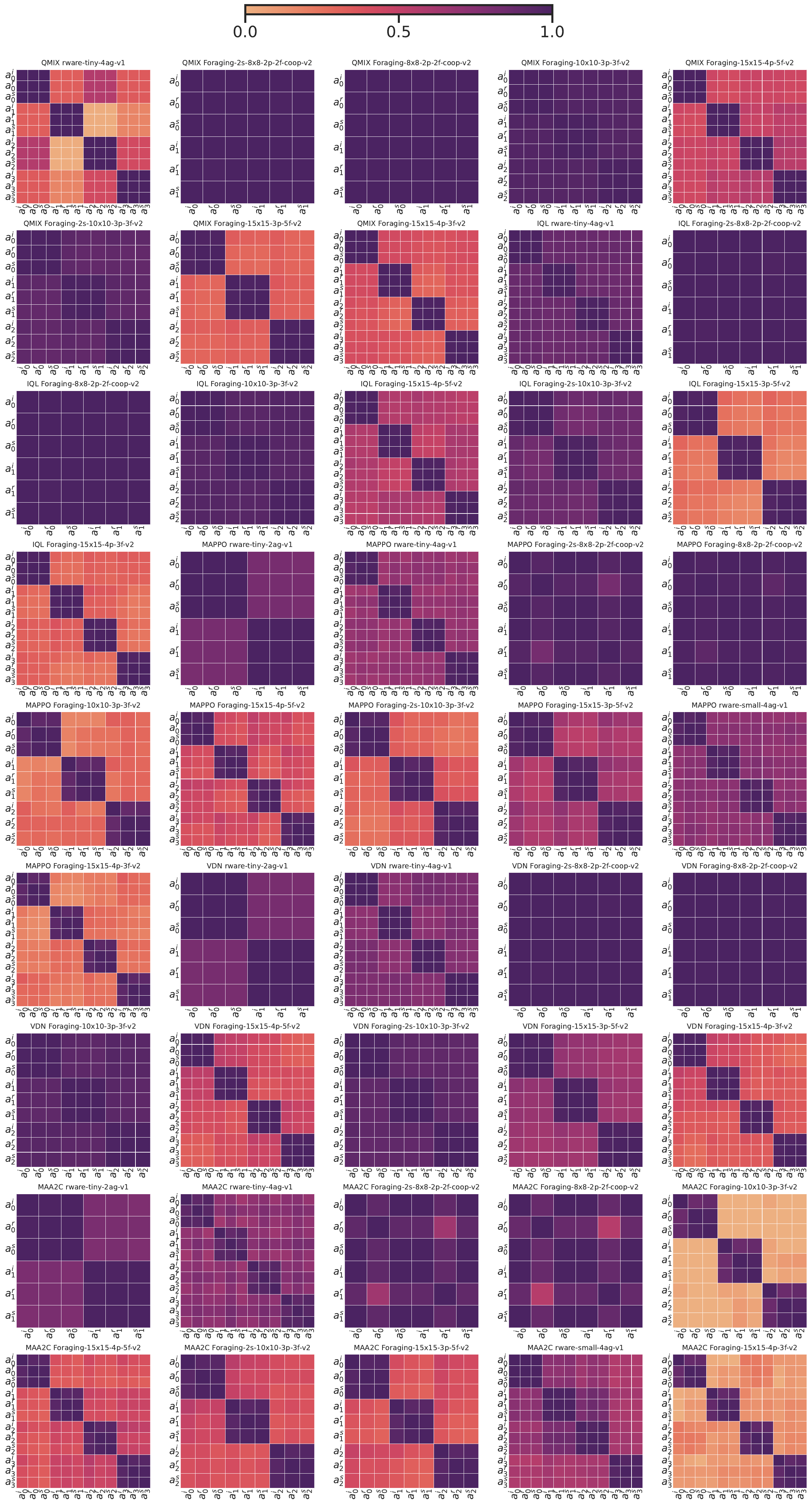}
\caption{\textit{The correlation between Agent Importance, Shapley values, and individual agent rewards is examined for the fourth independent run.} Each subplot corresponds to a specific algorithm and task, displaying the name of the algorithm followed by the task name. Notably, a strong correlation is observed across all algorithms and tasks.}
\label{fig: correlation_plot_3}
\end{figure}

\begin{figure}[H]
\centering
\includegraphics[width=0.7\linewidth]{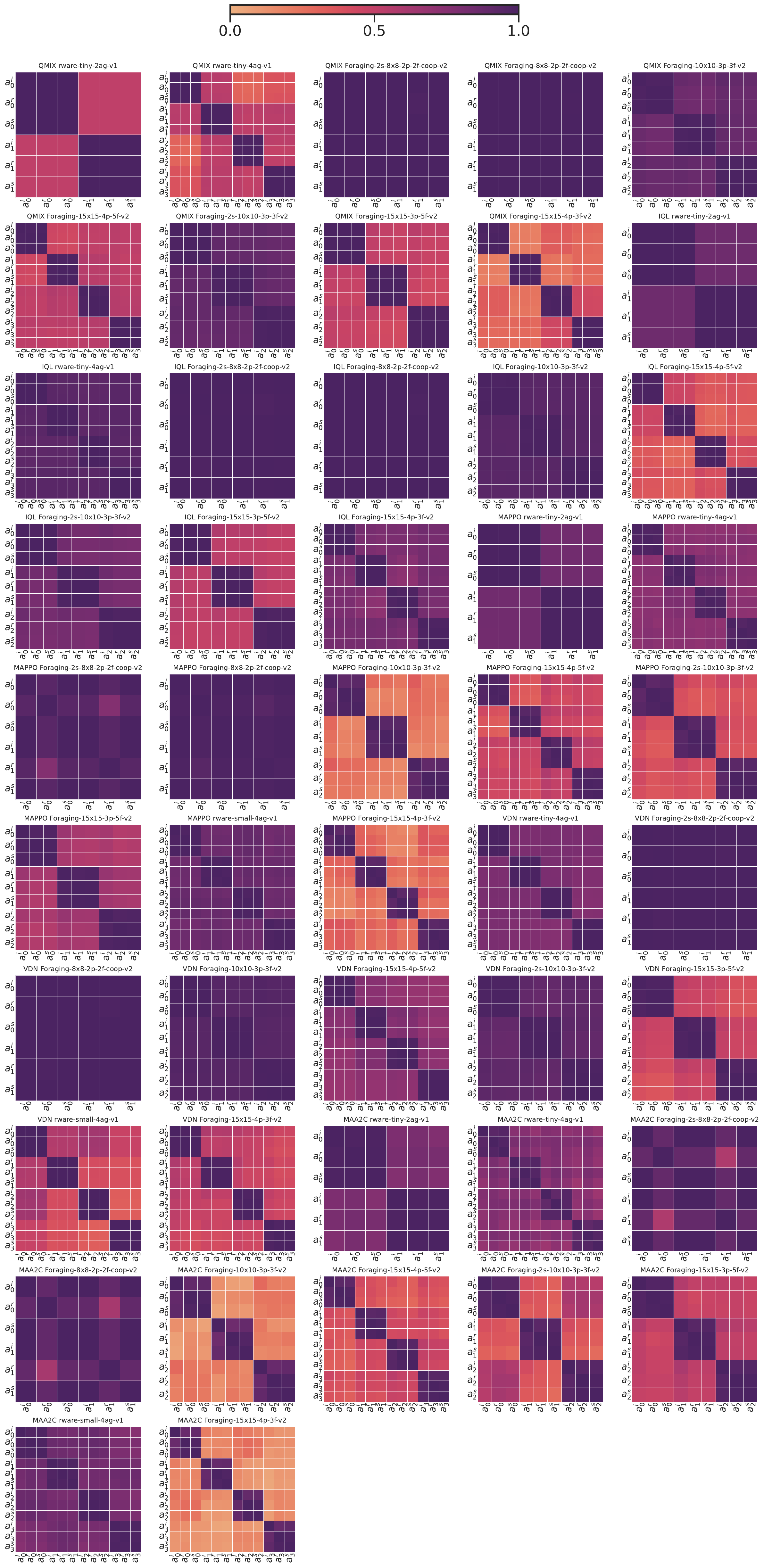}
\caption{\textit{The correlation between Agent Importance, Shapley values, and individual agent rewards is examined for the fifth independent run.} Each subplot corresponds to a specific algorithm and task, displaying the name of the algorithm followed by the task name. Notably, a strong correlation is observed across all algorithms and tasks.}
\label{fig: correlation_plot_4}
\end{figure}

\begin{figure}[H]
\centering
\includegraphics[width=0.7\linewidth]{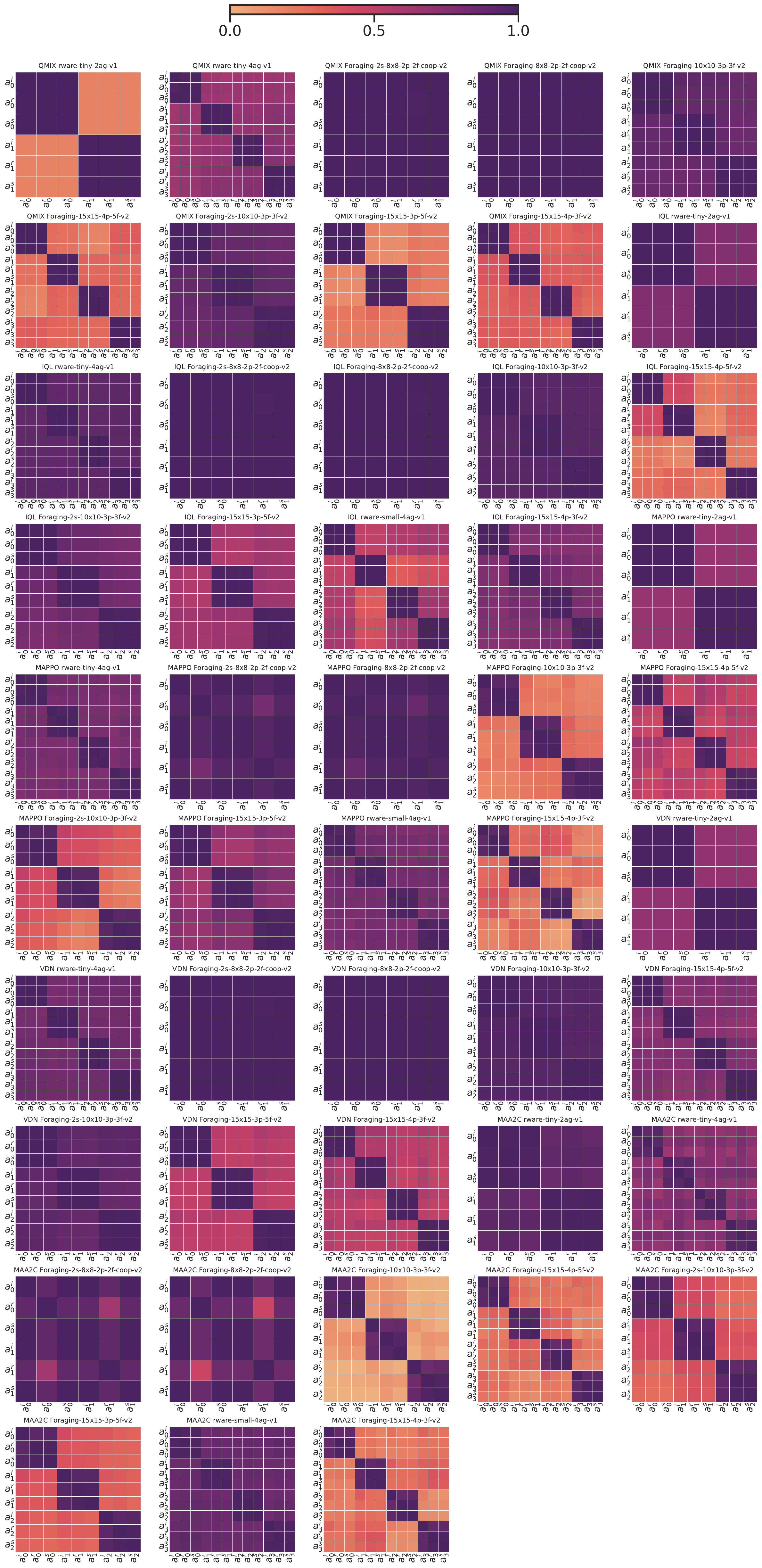}
\caption{\textit{The correlation between Agent Importance, Shapley values, and individual agent rewards is examined for the sixth independent run.} Each subplot corresponds to a specific algorithm and task, displaying the name of the algorithm followed by the task name. Notably, a strong correlation is observed across all algorithms and tasks.}
\label{fig: correlation_plot_5}
\end{figure}

\begin{figure}[H]
\centering
\includegraphics[width=0.7\linewidth]{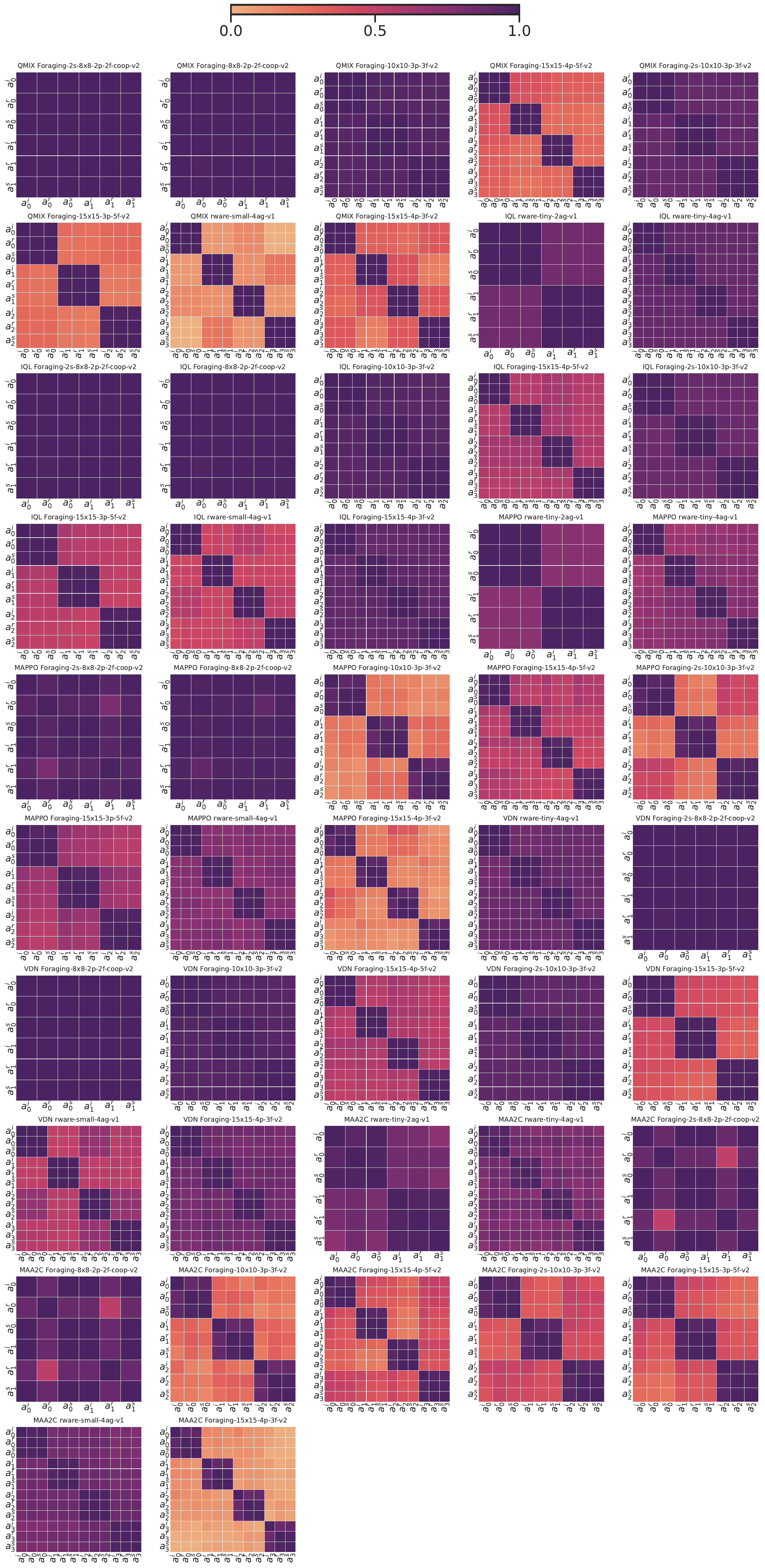}
\caption{\textit{The correlation between Agent Importance, Shapley values, and individual agent rewards is examined for the seventh independent run.} Each subplot corresponds to a specific algorithm and task, displaying the name of the algorithm followed by the task name. Notably, a strong correlation is observed across all algorithms and tasks.}
\label{fig: correlation_plot_6}
\end{figure}

\begin{figure}[H]
\centering
\includegraphics[width=0.7\linewidth]{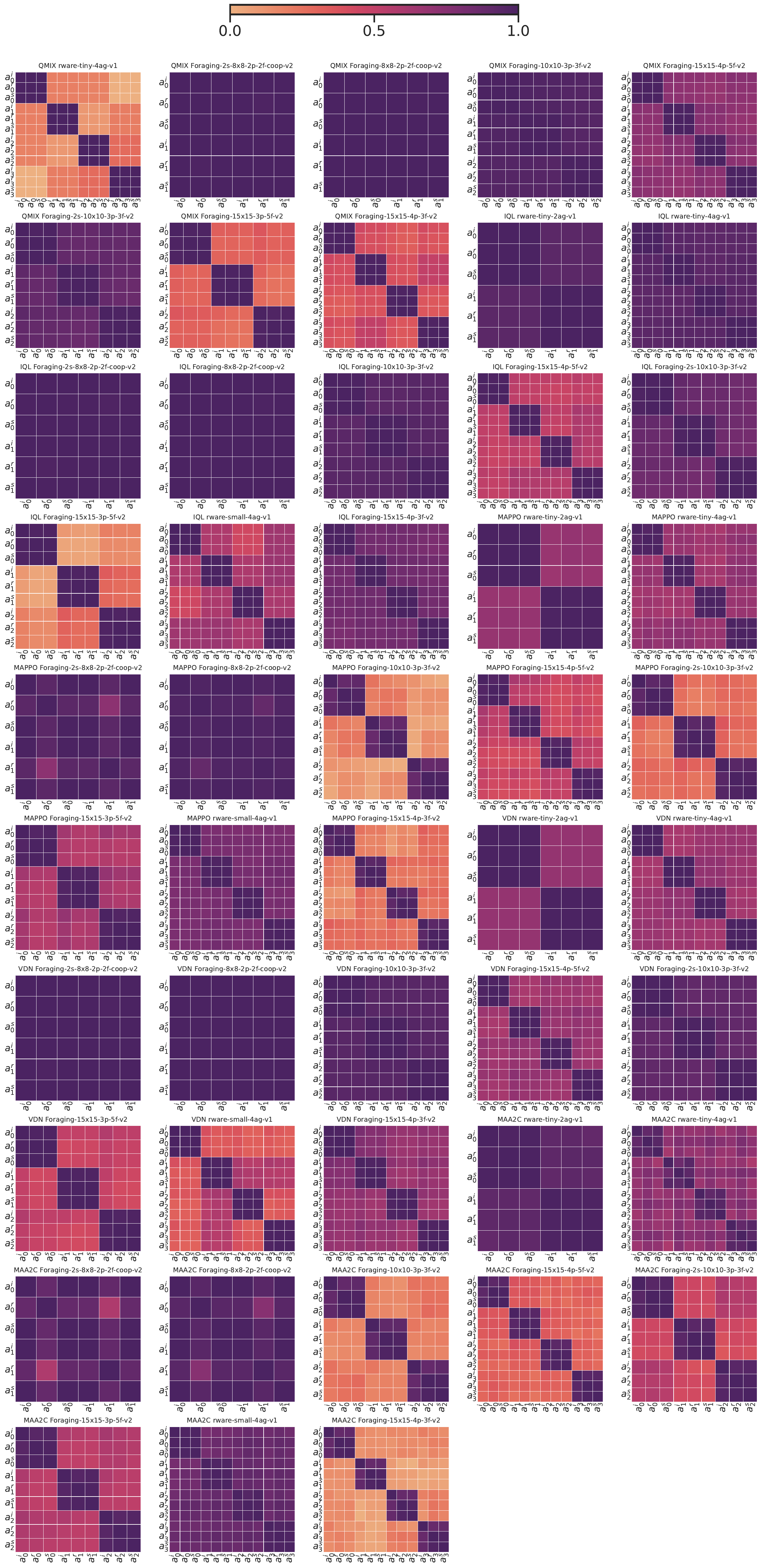}
\caption{\textit{The correlation between Agent Importance, Shapley values, and individual agent rewards is examined for the eighth independent run.} Each subplot corresponds to a specific algorithm and task, displaying the name of the algorithm followed by the task name. Notably, a strong correlation is observed across all algorithms and tasks.}
\label{fig: correlation_plot_7}
\end{figure}

\begin{figure}[H]
\centering
\includegraphics[width=0.7\linewidth]{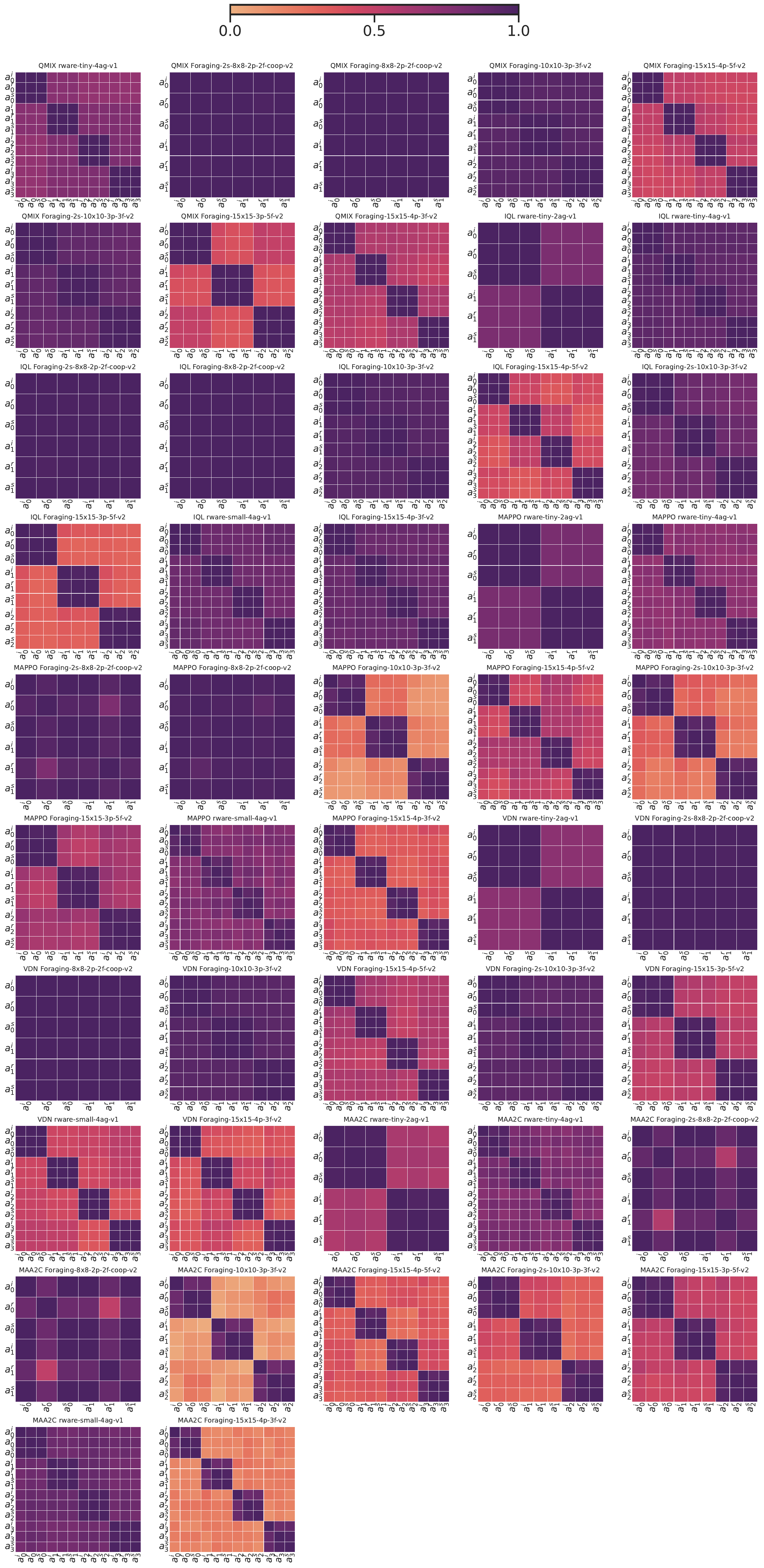}
\caption{\textit{The correlation between Agent Importance, Shapley values, and individual agent rewards is examined for the ninth independent run.} Each subplot corresponds to a specific algorithm and task, displaying the name of the algorithm followed by the task name. Notably, a strong correlation is observed across all algorithms and tasks.}
\label{fig: correlation_plot_8}
\end{figure}

\begin{figure}[H]
\centering
\includegraphics[width=0.7\linewidth]{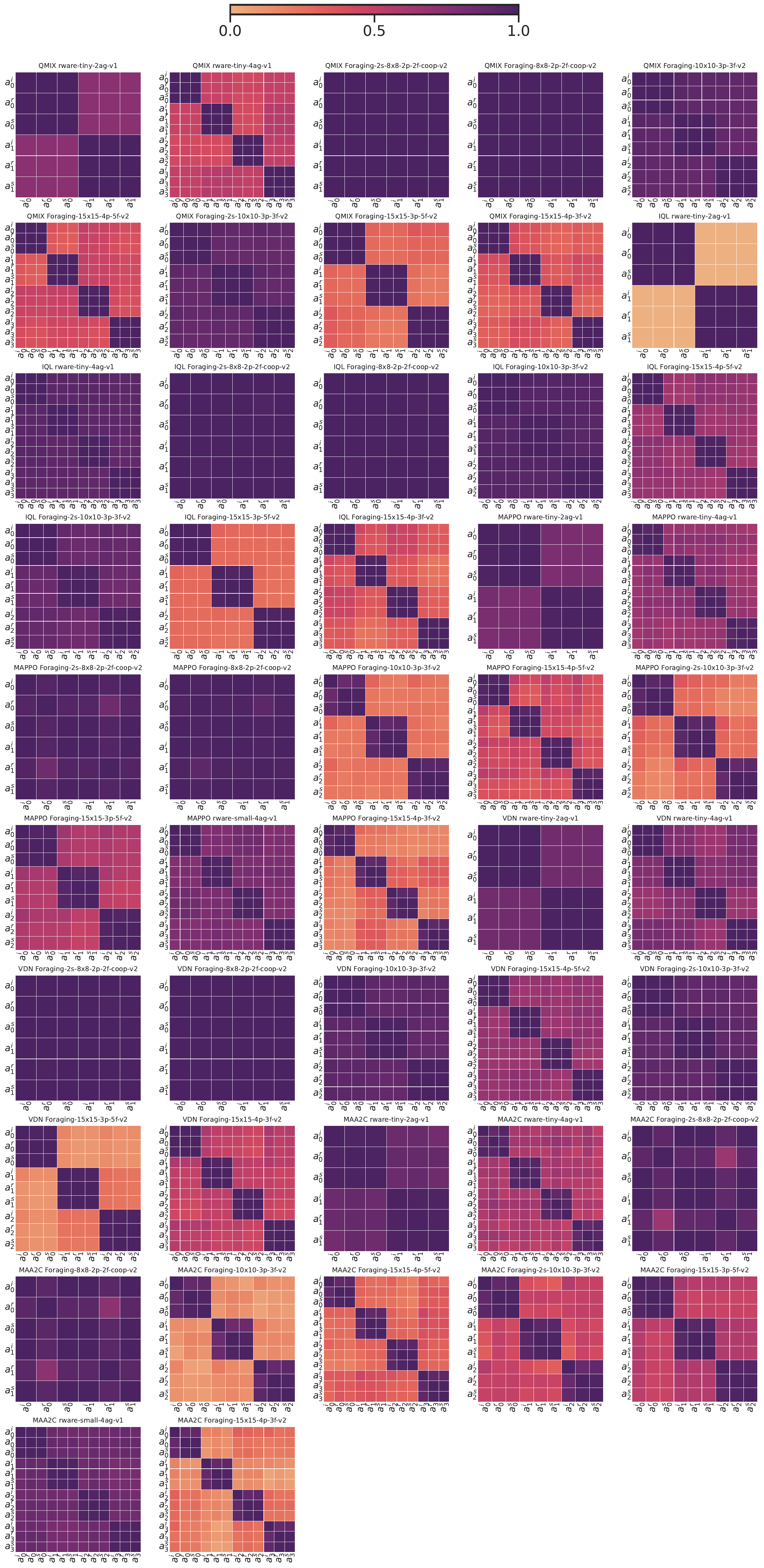}
\caption{\textit{The correlation between Agent Importance, Shapley values, and individual agent rewards is examined for the tenth independent run.} Each subplot corresponds to a specific algorithm and task, displaying the name of the algorithm followed by the task name. Notably, a strong correlation is observed across all algorithms and tasks.}
\label{fig: correlation_plot_9}
\end{figure}

To demonstrate the robust correlation between Agent Importance, the Shapley value, and individual reward, we present a comprehensive set of correlation heatmaps in Figures \ref{fig: correlation_plot_0} to \ref{fig: correlation_plot_9}. Each heatmap corresponds to a specific algorithm-task combination, as indicated by the title of the respective plot. It's important to note that these figures show a symmetrical pattern, unlike the asymmetry seen in Figure \ref{fig:correlation}(a). Furthermore, in Figures \ref{fig: maappo_rank} to \ref{fig: iql_rank}, we have extended the analysis by examining the consistency of rankings among individual rewards, agent importance, and the Shapley value for the scenario depicted in Figure \ref{fig:correlation}. However, it is worth noting that these specific plots are specific to alternative algorithms, namely IQL, QMIX, MAPPO, and MAA2C.

Moreover, we provide an overarching assessment of the correlation between Agent Importance and the Shapley value in Table \ref{tab:average_correlation}. This evaluation entails computing the average correlation coefficient across multiple independent runs, algorithms, tasks, and agents, thereby yielding a consolidated metric. To ensure fairness, tasks involving 2, 3, and 4 agents are aggregated separately in the analysis.

\begin{figure}
  \centering
\begin{subfigure}{0.4\linewidth}
    \includegraphics[width=\linewidth]{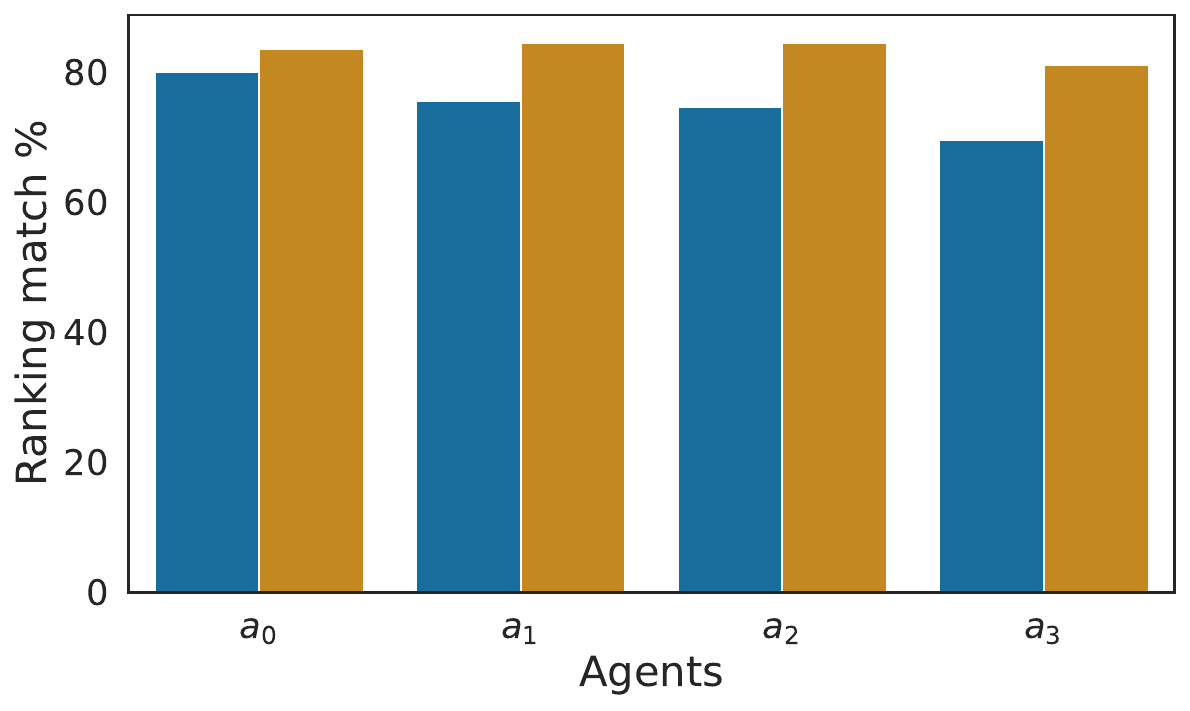}
  \end{subfigure}
  \begin{subfigure}{0.5\linewidth}
    \includegraphics[width=\linewidth]{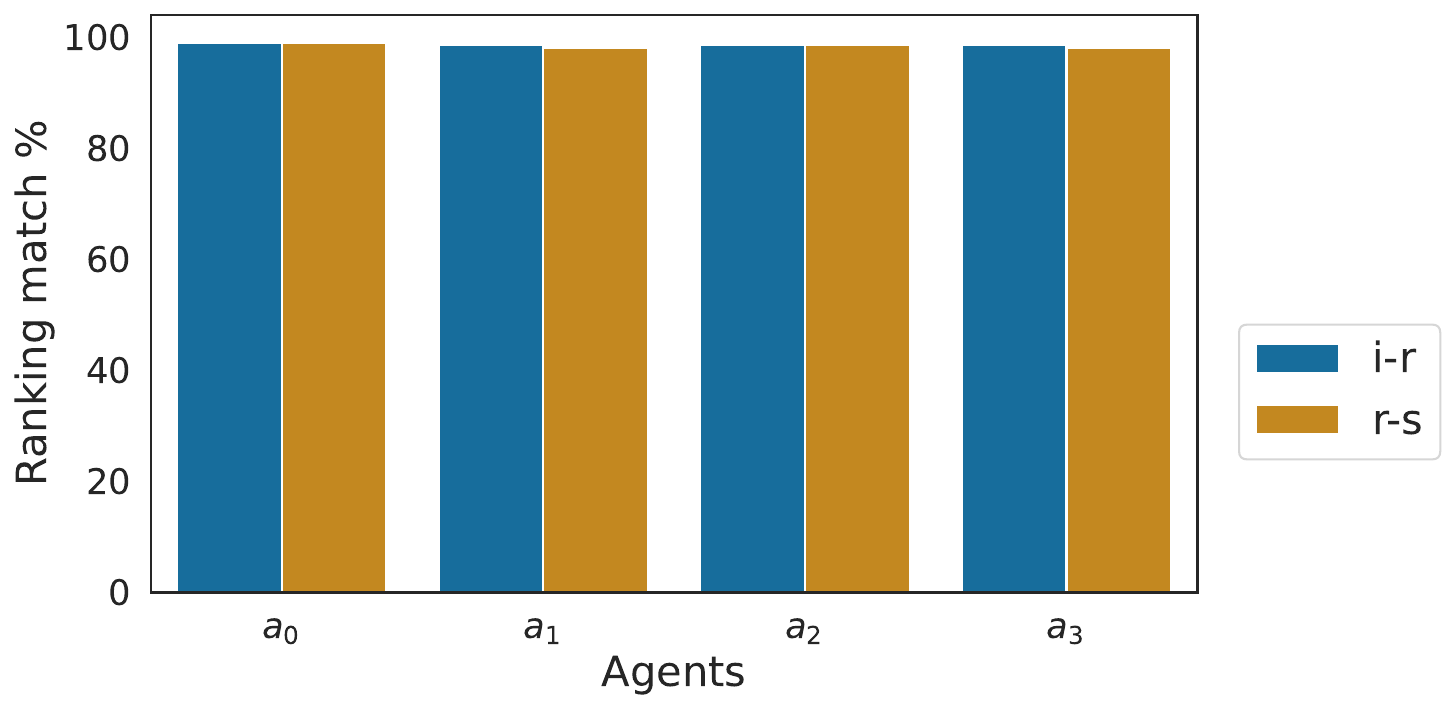}
  \end{subfigure}
  \caption{\textbf{Left:} Matching Rankings Comparison on LBF 15x15-4p-5f. \textbf{Right:} Matching Rankings Comparison on RWARE small-4ag.}
  \label{fig: maa2c_rank}
\end{figure}

\begin{table}
  \centering
  \caption{\textit{Average correlation of Agent Importance and the Shapley value.} Even when aggregating over multiple independent runs, algorithms, tasks, and agents, the strong correlation still holds.}
  \label{tab:average_correlation}
  \begin{tabular}{cc}
    \toprule
    \textbf{Number of Agents}& \textbf{Correlation Value} \\
    \midrule
    2 & 0.97 $\pm 0.01$\\
    3 & 0.96 $\pm 0.01$\\
    4 & 0.96 $\pm 0.01$\\
    \bottomrule
  \end{tabular}
\end{table}

[H]

\end{document}